\documentclass[a4paper,11pt,twoside,openright,notitlepage]{book}
\usepackage[a-1b]{pdfx}

\usepackage[numbers]{natbib}
\PassOptionsToPackage{numbers, compress}{natbib}

\makeatletter
\def\bstctlcite{\@ifnextchar[{\@bstctlcite}{\@bstctlcite[@auxout]}}
\def\@bstctlcite[#1]#2{\@bsphack
 \@for\@citeb:=#2\do{%
   \edef\@citeb{\expandafter\@firstofone\@citeb}%
   \if@filesw\immediate\write\csname #1\endcsname{\string\citation{\@citeb}}\fi}%
 \@esphack}
\makeatother

\newcommand{\chapabstract}[1]{
    \begin{quote}
        \singlespacing\small
        #1
        \vskip-4mm
\end{quote}}

\usepackage{xpatch}
\makeatletter
\xpretocmd{\@endpart}{%
  \ifx\@abstract\@empty\else
    \bigskip
    \begin{quote}\@abstract\end{quote}
    \global\let\@abstract\@empty
  \fi
}{}{}
\newcommand{\partabstract}[1]{%
  \renewcommand{\@abstract}{#1}%
}
\newcommand{\@abstract}{}
\makeatother

\usepackage[utf8]{inputenc}
\usepackage[T1]{fontenc}			
\usepackage{lmodern}

\usepackage{amsfonts}   
\usepackage{amsmath}					
\usepackage{amssymb}

\usepackage{units}				
\usepackage{amstext}        
\usepackage{amsthm}     

\usepackage{array}						 
\usepackage{multicol}					
\usepackage{multirow}
\usepackage{lscape}
\usepackage{xspace}						
\usepackage[pdftex]{graphicx}	
\usepackage{subcaption}
\usepackage{indentfirst} 			
\usepackage{latexsym}					
\usepackage{setspace}					
\usepackage{algorithmic}			
\usepackage{makeidx}					
\usepackage{vmargin}					
\usepackage{float}            
\usepackage{enumerate}        
\usepackage{verbatim}         
\usepackage{floatflt}         
\usepackage{rotating}					
\usepackage{longtable}        
\usepackage{diagbox}        
\usepackage{url}
\usepackage{pdfpages}           
\usepackage{titlesec} 
\usepackage{graphicx}
\usepackage{minitoc}
\usepackage{xcolor,lipsum}
\usepackage[framemethod=tikz]{mdframed}
\usepackage{setspace}
\usepackage{wrapfig}
\usepackage{multicol}
\usepackage{multirow,makecell}
\usepackage{tabularx, booktabs}
\usepackage{nicefrac}       
\usepackage{microtype}      
\usepackage{float}
\usepackage{mathtools,cases}
\usepackage{empheq}
\usepackage{bm}

\usepackage{amsmath,amsfonts,bm}

\def\Figref#1{Figure~\ref{#1}}

\def\eqref#1{Eq.~(\ref{#1})}

\def\1{\bm{1}}

\DeclareMathAlphabet{\mathsfit}{\encodingdefault}{\sfdefault}{m}{sl}
\SetMathAlphabet{\mathsfit}{bold}{\encodingdefault}{\sfdefault}{bx}{n}

\newcommand{\R}{\mathbb{R}}

\usepackage[utf8]{inputenc}		
\usepackage[T1]{fontenc}		
\usepackage{lmodern}
\usepackage{ae,aecompl}
\usepackage[top=2.5cm, bottom=2cm, 
			left=3cm, right=2.5cm,
			headheight=15pt]{geometry}
\usepackage{graphicx}
\usepackage{eso-pic}	
\usepackage{array}	
\usepackage{doi}
\usepackage{adjustbox}
\usepackage{multirow,makecell}
\usepackage{tabularx, booktabs}
\usepackage{arydshln}
\usepackage[inline]{enumitem}
\usepackage{arydshln}
\usepackage{mathtools,cases}
\usepackage{empheq}
\usepackage{bm}
\usepackage{lettrine}
\usepackage[ruled,vlined,noend]{algorithm2e}

\definecolor{redHESAM}{RGB}{210,0,37}
\usepackage{color, colortbl}
\definecolor{Gray}{gray}{0.9}
\usepackage{pifont}

\def\eg{e.g.~}
\def\ie{i.e.~}

\def\etc{\textit{etc}}
\def\etal{\textit{et al.~}}
\newcommand{\x}{\mathbf{x}}
\newcommand{\y}{\mathbf{y}}
\newcommand{\z}{\mathbf{z}}
\def\D{{\mathcal{D}}}
\def\I{{\mathcal{I}}}
\def\F{{\mathcal{F}}}
\def\A{{\mathcal{A}}}
\newcommand*\diff{\mathop{}\!\mathrm{d}}

\usepackage{hyperref}

\usepackage{fancyhdr}
\usepackage{optidef}

\newtheorem{prop}{Proposition}
\newtheorem{definition}{Definition}

\makeindex

\newenvironment{vcenterpage}
{\newpage\vspace*{\fill}}
{\vspace*{\fill}\par\pagebreak}
\makeindex

\mtcsettitle{minitoc}{Content}

\begin{document}


\dominitoc
\thispagestyle{empty}

\newgeometry{
        showframe,
		top=20mm,
		bottom=20mm,
		inner=20mm,
		outer=20mm,
		includehead,
		ignorefoot,
		nomarginpar,
		headsep=10mm,
		footskip=10mm,
	}

\setmarginsrb{5mm}{15mm}{18mm}{5mm}{0mm}{0mm}{0mm}{0mm}

%

\hspace{0.5cm}\includegraphics[width=5.5cm]{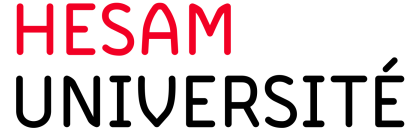} \hspace{6cm} \includegraphics[width=5.5cm]{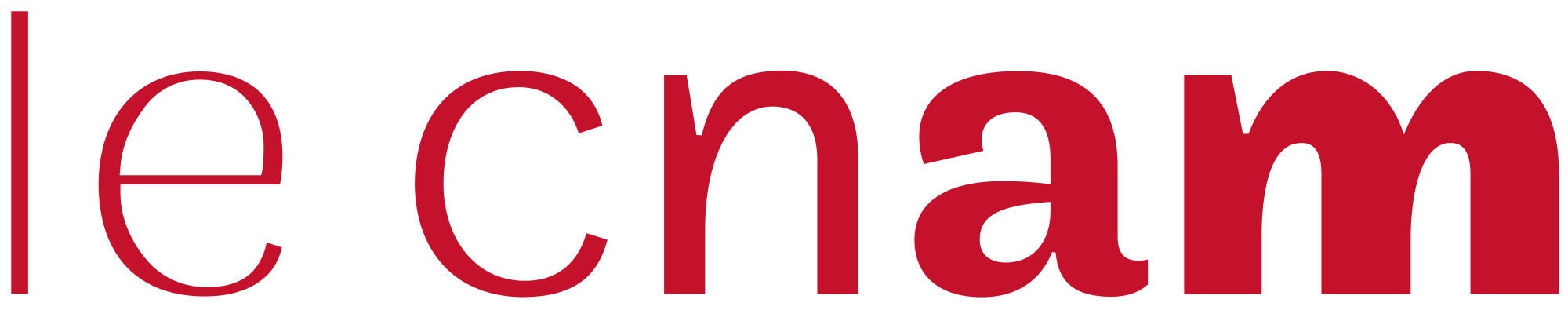}


\vspace{1cm}
\makebox[\textwidth][c]{\centering{\LARGE{\textbf{\'ECOLE DOCTORALE Sciences des Métiers de l'Ingénieur}}}}\\

\vspace{-0.2cm}
\hspace{0.5cm}\makebox[\textwidth][c]{\centering{\LARGE{\textbf{Centre d'études et de recherche en informatique et communications}}}}\\
\vspace{0.2cm}

\makebox[\textwidth][c]{{\Huge{\textbf{TH\`ESE}}}}\\

\vspace{0.1cm}
\makebox[\textwidth][c]{\centering{\large{\textit{présentée par} : }} {\Large{\textbf{Vincent LE GUEN}}}}\\
\vspace{.4cm}
\makebox[\textwidth][c]{\centering{\large{\textit{soutenue le} : }} {\Large{\textbf{30 novembre  2021}}}}\\
\vspace{.4cm}
\makebox[\textwidth][c]{\centering{\large{\textit{pour obtenir le grade de} : }} {\Large{\textbf{Docteur d'HESAM Université}}}}\\
\vspace{.2cm}
\makebox[\textwidth][c]{\centering{\large{\textit{préparée au} : }} {\Large{\textbf{Conservatoire national des arts et métiers}}}}\\
\vspace{.2cm}
\makebox[\textwidth][c]{\centering{\large{\textit{Discipline} :}} {\large{\textbf{Informatique}}}}
\makebox[\textwidth][c]{\centering{\large{\textit{Spécialité} :}} {\large{\textbf{Informatique}}}}
\vspace{0.2cm}


\resizebox{\textwidth}{!}{
\fcolorbox{redHESAM}{white}{
\parbox{\dimexpr \linewidth-2\fboxsep-2\fboxrule}{%
\begin{tabular}{>{\centering\arraybackslash}p{16.5cm}}
\begin{minipage}{16.5cm}
\centering
\vspace{0.1cm}
{
~\\
\LARGE{\textbf{{Deep learning for spatio-temporal forecasting - application to solar energy \\}}}
~\\}
\vspace{0.1cm}
\end{minipage}	
\end{tabular}
}}}~\\

\vspace{0.2cm}

\makebox[\textwidth][c]{\centering{\textbf{\textsc{\large TH\`{E}SE dirig\'{e}e par :}}}}\\
\vspace{0.5cm}
\makebox[\textwidth][c]{\centering{\large{\textbf{M. Nicolas THOME}}} {\large{, Professeur, Conservatoire national des arts et métiers}}}\\

\vfill

\vspace{-1cm}

\begin{minipage}[c]{0.95\linewidth}
   \centering
   \begin{tabular}{p{12cm} p{3cm}}
\textbf{Jury} & \\

  \textbf{M. Greg \textsc{MORI}} & \\ 
 Professeur, Simon Fraser University et directeur scientifique, Borealis AI   &  Rapporteur  \\
 
   \textbf{M. Patrick \textsc{PEREZ}} & \\ 
Directeur scientifique Valeo AI  &  Rapporteur  \\

\textbf{M. Patrick GALLINARI}  & \\ 
Professeur, Sorbonne Université et chercheur senior, Criteo AI  &  Président du jury  \\   

   \textbf{M. Philippe \textsc{BLANC}} & \\ 
Professeur, Mines ParisTech  &  Examinateur  \\
 
    \textbf{Mme Stéphanie \textsc{DUBOST}} & \\ 
Docteure, EDF R\&D  &  Examinatrice  \\
 
    \textbf{Mme Elisa \textsc{FROMONT}} & \\ 
Professeur, Université Rennes 1, IRISA  &  Examinatrice  \\

   \textbf{M. Etienne \textsc{MEMIN}} & \\ 
Directeur de recherche INRIA  &  Examinateur  \\

   \textbf{M. Nicolas \textsc{THOME}} & \\ 
Professeur, Conservatoire national des arts et métiers &  Directeur de thèse  \\
\end{tabular}

  \end{minipage}
\hfill
  \begin{minipage}[c]{0.05\linewidth}
   \centering
   \vspace{0.5cm}
   \includegraphics[height=7.4cm]{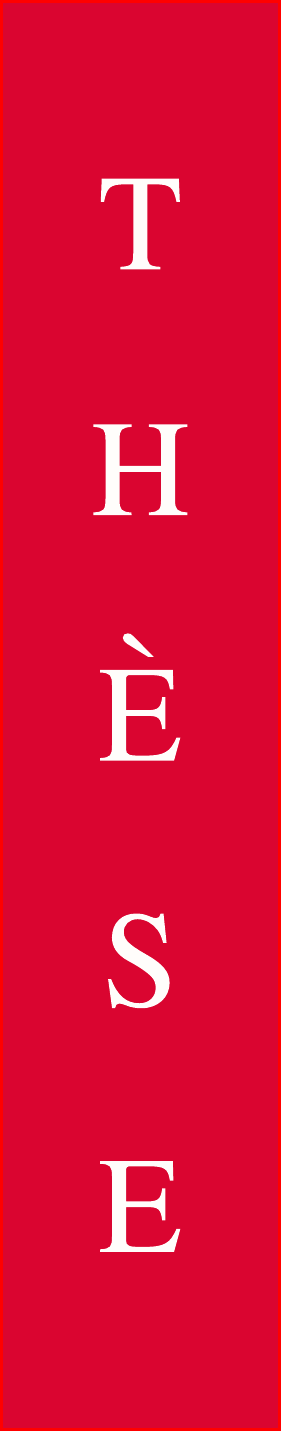}     
  \end{minipage}

\newpage
\thispagestyle{empty}

\vspace{0.5cm}\hspace{0.5cm}\includegraphics[width=4cm]{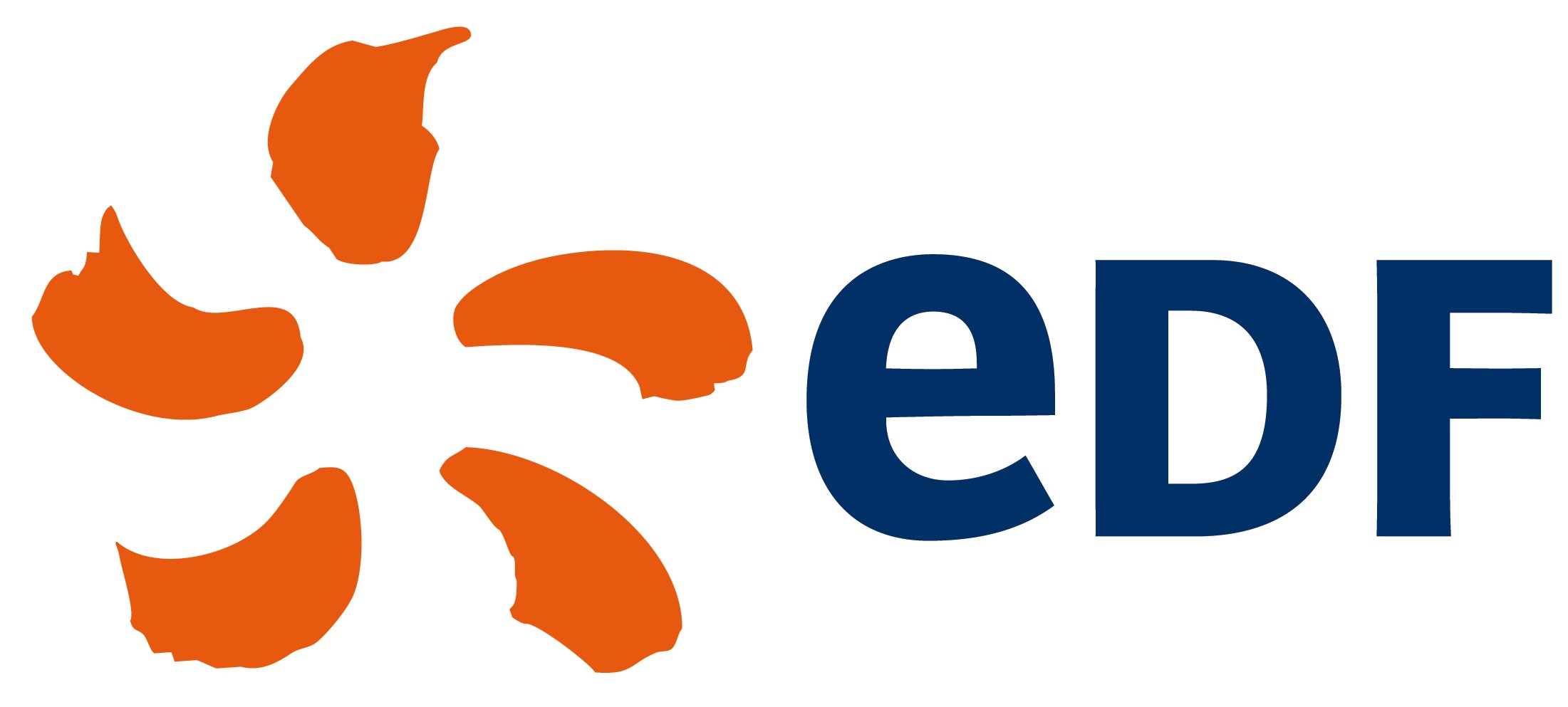} \hspace{6cm} \includegraphics[width=5.5cm]{images/Cnam.jpg}

\vspace{23cm}
Vincent Le Guen: \textit{Deep learning for spatio-temporal forecasting, application to solar energy}, \copyright ~ 2021

\frontmatter

\textbf{Affidavit}\\

Je soussigné, Vincent Le Guen, déclare par la présente que le travail présenté dans ce manuscrit est mon propre travail, réalisé sous la direction scientifique de Nicolas Thome (directeur de thèse), dans le respect des principes d’honnêteté, d'intégrité et de responsabilité inhérents à la mission de recherche. Les travaux de recherche et la rédaction de ce manuscrit ont été réalisés dans le respect de la charte nationale de déontologie des métiers de la recherche.
Ce travail n'a pas été précédemment soumis en France ou à l'étranger dans une version identique ou similaire à un organisme examinateur.\\

Fait à Paris, novembre 2021\\

\begin{figure}[H]
    \centering
    \includegraphics[width=3cm]{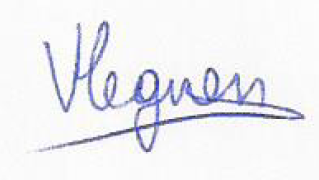}
\end{figure}

\textbf{Affidavit}\\

I, undersigned, Vincent Le Guen, hereby declare that the work presented in this manuscript is my own work, carried out under the scientific direction of Nicolas Thome (thesis director), in accordance with the principles of honesty, integrity and responsibility inherent to the research mission. The research work and the writing of this manuscript have been carried out in compliance with the French charter for Research Integrity.\\

Paris, November 2021\\

\begin{figure}[H]
    \centering
    \includegraphics[width=3cm]{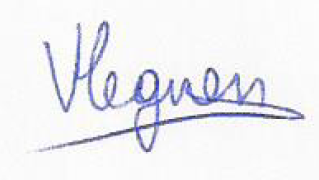}
\end{figure}

\thispagestyle{empty}




\setlength{\baselineskip}{1.05\baselineskip} 
\setlength{\voffset}{0pt} 				
\setlength{\topmargin}{0pt}				
\setlength{\headheight}{30pt}			
\setlength{\headsep}{30pt}			
\setlength{\textheight}{620pt}		
\setlength{\footskip}{30pt}			
\setlength{\hoffset}{-15pt} 		
\setlength{\oddsidemargin}{0pt}	    
\setlength{\evensidemargin}{0pt}	
\setlength{\textwidth}{300pt}		
\setlength{\marginparsep}{15pt}		
\setlength{\marginparwidth}{40pt}	
\parskip=4pt	                    


\pagestyle{fancy}




\let\cleardoublepage\clearpage


\begin{vcenterpage}

\thispagestyle{empty}
\chapter*{Remerciements}
\addcontentsline{toc}{chapter}{Remerciements}
\adjustmtc
\markright{\MakeUppercase{Remerciements}}

Je tiens à remercier ici toutes les personnes qui ont concouru à l'achèvement du travail présenté dans ce manuscrit.

Pour revenir chronologiquement sur le montage de ce projet, commencer une thèse de doctorat 5 ans après la sortie d’école d’ingénieur a été en soi un premier défi. Je remercie chaleureusement mes collègues à EDF pour leur soutien, en particulier Nicolas Paul, Bruno Charbonnier, Loic Vallance. J’ai aussi une grande reconnaissance pour Stéphanie Dubost pour toutes nos discussions utiles sur la prévision d'énergies et pour
avoir défendu dès le début cette idée. Stéphanie a grandement oeuvré pour assurer le financement de ma thèse, réparti sur 4 projets et 4 programmes de recherche différents, ce qui était assez inédit ! J'ai aussi pu compter sur l'appui décisif de ma hiérarchie, Nicolas Roche, Julien Berland, dans cette démarche.
Je remercie aussi Dominique Demengel pour son travail crucial depuis de nombreuses années sur l'instrumentation des caméras au sol et pyranomètres et la bonne qualité des données, sans quoi ce travail n'aurait pas été possible.

Ensuite, mes remerciements les plus sincères s’adressent à Nicolas Thome qui a dirigé ma thèse au conservatoire national des Arts et Métiers. Quand je suis venu la première fois dans son bureau à l'été 2018 avec mon sujet de thèse déjà écrit, Nicolas a accueilli l'idée avec grand intérêt. Ces 3 années de travail commun se sont révélées être intenses et très stimulantes. Nicolas a toujours été très actif pour
guider nos réflexions et discuter de nouvelles pistes quand nous arrivions dans une impasse. Nicolas m'a beacoup appris sur comment mener un projet de recherche, et notamment sur la difficile tâche de rédaction d'articles scientifiques. Un grand merci pour tout le temps passé (soir et week-end compris) pour guider, relire, (ré)écrire, ce qui a été d'une importance cruciale pour l'acceptation des soumissions. Je remercie aussi Clément Rambour au CNAM, pour son co-encadrement très important sur mes travaux de dernière année et ses conseils judicieux. 

Au cours de ces 3 années de thèse, j'ai pu passer de très bons moments avec les autres doctorants de l'équipe au CNAM: Olivier Petit, Thuy Le, Laura Calem, Charles Corbière, Rémy Sun, Elias Ramzi, Loïc Thémyr,  Marc Lafon, Perla Doubinsky, Yannis Karmim. Malgré le confinement et le télétravail résultant de la crise sanitaire, nous avons réussi à maintenir des contacts techniques et conviviaux réguliers à distance. J'en retiendrai l'amitié, la solidarité et l'entraide nées de ces périodes de dur labeur.

Je remercie également mes collègues à EDF avec qui les discussions ont toujours été fructueuses, en particulier Charlotte Gauchet, Christophe Chaussin, Lorenzo Audibert, Louis Apffel, Georges Hebrail, Nicolas Bousquet, Benoît Braisaz, Eric Lajoie-Mazenc, Matthieu Chiodetti, Gerald Kwiatkowski. 

Du côté de Sorbonne Université, je remercie Matthieu Cord et tous ses doctorants pour l'organisation des réunions hebdomadaires "cordettes", à la fois studieuses et conviviales. J'ai également particulièrement apprécié les travaux avec Patrick Gallinari et ses doctorants Yuan Yin, Jérémie Dona, Ibrahim Ayed, Emmanuel de Bézenac, qui ont permis d'aboutir à une publication jointe très approfondie. Je remercie aussi Edouard Oyallon, ancien camarade au master MVA et maintenant chercheur CNRS, dont le regard sur nos travaux a été très pertinent. Sur ses conseils, nous avons collaboré de manière fructueuse avec Edouard Leurent, dont je salue la gentillesse et la disponibilité. 

J'exprime ma gratitude pour tous les membres de mon jury de thèse pour avoir accepté d'évaluer mes travaux et pour leurs retours très pertinents: Greg Mori, Patrick Pérez, Patrick Gallinari, Philippe Blanc, Stéphanie Dubost, Elisa Fromont, Etienne Mémin.

Pour finir, je tiens à remercier mes parents, mon épouse pour leur soutien et patience durant ces 3 années très chargées, avec une pensée pour le petit Louis qui est venu au monde 1 mois avant ma soutenance de thèse.

\end{vcenterpage}

\let\cleardoublepage\clearpage


\thispagestyle{empty}
\begin{vcenterpage}

\chapter*{Abstract}
\addcontentsline{toc}{chapter}{Abstract}
\adjustmtc
\markright{\MakeUppercase{Abstract}}

This thesis tackles the subject of spatio-temporal forecasting with deep learning, which is the task of forecasting complex phenomena represented by time series or videos, involving both complex temporal dynamics and strong spatial correlations. This is of crucial importance for many industrial applications, such as climate, healthcare or finance. The motivating application at Electricity de France (EDF) is short-term solar energy forecasting with fisheye images. Despite the great successes of deep learning in computer vision and natural language processing, pure data-driven methods still struggle in the task of physical process extrapolation, especially in data-scarce contexts and for non-stationary time series that can present sharp variations. We explore two main research directions for improving deep forecasting methods by injecting external physical knowledge. The first direction concerns the role of the training loss function. Instead of using the largely dominant mean squared error (MSE), we show that differentiable shape and temporal criteria, typically used as evaluation metrics in applications, can be leveraged to improve the performances of existing models. We address both the deterministic context with the proposed DILATE loss function and the probabilistic context, for which we aim at describing the predictive distribution with a small set of diverse and accurate scenarios, with our proposed STRIPE model. Our second direction is to augment incomplete physical models with deep data-driven networks for accurate forecasting. For video prediction, we introduce the PhyDNet model that disentangles PDE (partial differential equations) dynamics from residual information necessary for prediction, such as texture or details. We further propose a learning framework (APHYNITY) that ensures a principled and unique linear decomposition between physical and data-driven components under mild assumptions, leading to better forecasting performances and parameter identification. We validate our contributions on many synthetic and real-world datasets, and on the solar energy dataset at EDF.

Keywords : deep learning, machine learning, spatio-temporal forecasting, solar energy forecasting.

\end{vcenterpage}

\clearpage{\pagestyle{empty}\cleardoublepage}


\singlespacing
\tableofcontents
\singlespacing
\listoftables 
\singlespacing
\addcontentsline{toc}{chapter}{List of tables}
\adjustmtc
\listoffigures     
\singlespacing
\addcontentsline{toc}{chapter}{List of figures}
\adjustmtc

\newpage
\mbox{}
\thispagestyle{empty}




\mainmatter

\chapter{Introduction}
\label{chap:intro}
µ
\minitoc   
\newpage

\section{Spatio-temporal forecasting}

\begin{figure}[H]
    \centering
    \includegraphics[width=17cm]{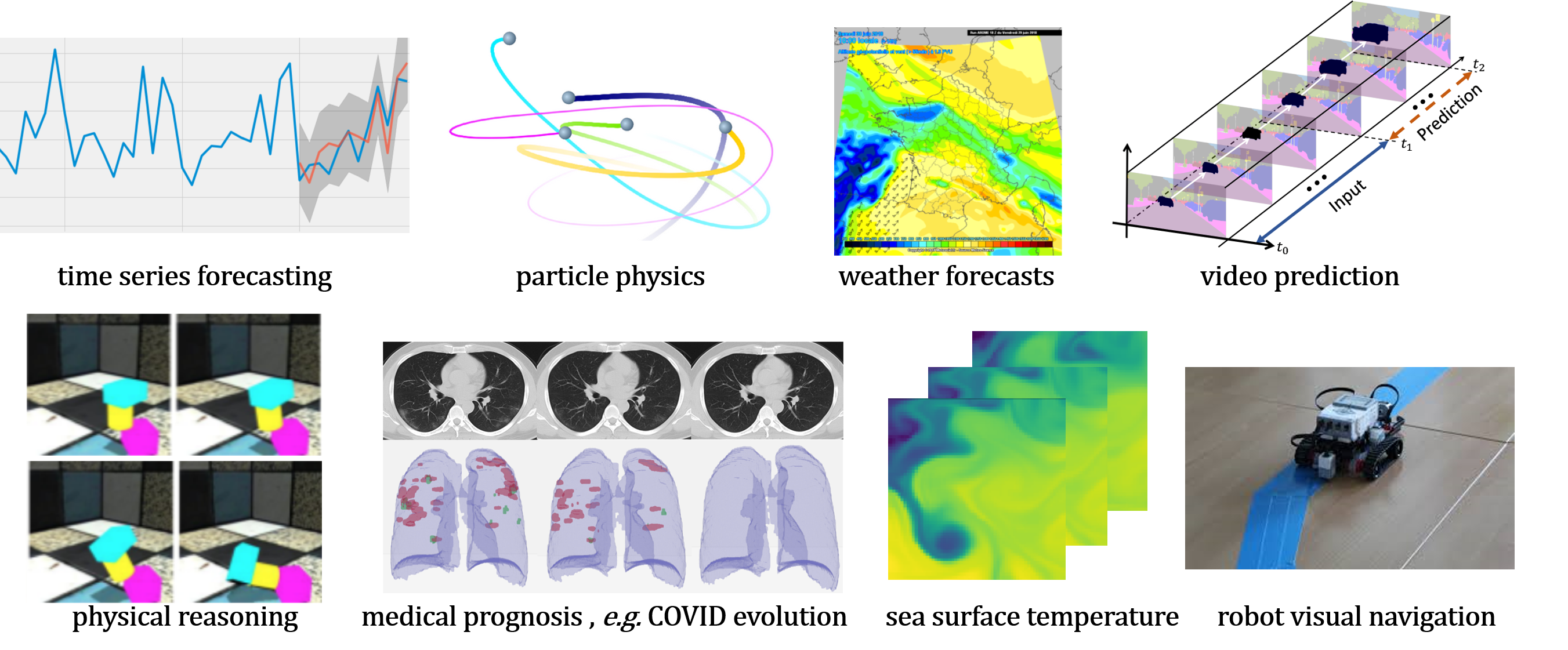}
    \caption[Spatio-temporal forecasting applications.]{Spatio-temporal forecasting applications include time series forecasting, physical systems extrapolation, forecasting phenomena with visual data, generic video prediction, \etc.}
    \label{fig:forecasting}
\end{figure}

\subsection{General context: perception vs extrapolation}

\lettrine[lines=3]{I}n this thesis, we tackle the problem of spatio-temporal forecasting, which is the task of forecasting complex phenomena represented by time series or videos, involving both complex temporal dynamics and strong spatial correlations. Advances in this field could lead to immediate and possibly large impacts in the society. A wide range of sensitive applications heavily rely on accurate forecasts of uncertain events with potentially sharp variations for making  decisions (see Figure \ref{fig:forecasting}). In weather and climate science, better anticipating floods, hurricanes, earthquakes or other extreme events could help taking emergency measures on time and save lives. In medicine, predicting the evolution of a disease is a particularly actual topic. In retail and business, accurately predicting the demand for a product is fundamental for stock management and profit maximization. For industrial applications, failure prediction is an important issue for maintenance.
 
\begin{figure}[H]
    \centering
    \includegraphics[width=17cm]{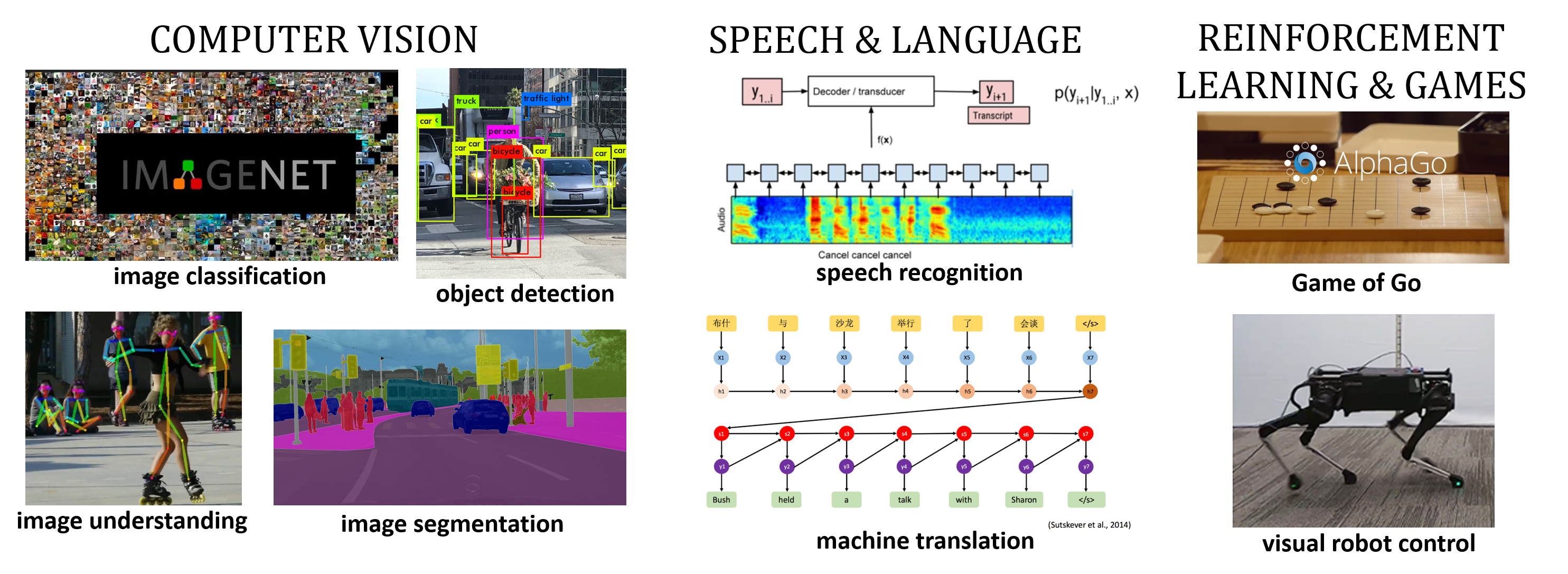}
    \caption[Successes of Artificial Intelligence and Deep Learning.]{The main Artificial Intelligence and Deep learning successful applications include tasks linked to perception, such as computer vision, speech, language, reinforcement learning and games.}
    \label{fig:ai-success}
\end{figure}

We address spatio-temporal forecasting from a machine learning point of view, \ie by leveraging training data for solving the task. Machine Learning (ML) is a subfield of Artificial Intelligence (AI) that is appealing for solving complex problems. Bolstered by the recent advances in computer hardware and the exponential growth of available data, ML has witnessed a renewed interest in the last decade from both academic and industrial actors. At the ImageNet competition in 2012, which consists in classifying images between 1000 categories, the deep neural network of Krizhevsky et al. \cite{krizhevsky2012imagenet} has for the first time outperformed traditional methods by a large margin. Given enough training data, Deep Learning (DL) can automatically learn  meaningful representations useful for downstream tasks, replacing the manual feature extraction necessary in traditional ML algorithms. Since, Deep Learning has shown impressive results in many practical applications (see Figure \ref{fig:ai-success}), such as object detection \cite{carion2020end}, image segmentation \cite{minaee2021image}, natural language understanding \cite{devlin2018bert}, or human speech recognition \cite{amodei2016deep}. Combined with reinforcement learning, DL has led to super-human performance on many board games, \eg at the game of Go with alphaGo \cite{silver2017mastering}.

However, the successes of AI in these tasks are essentially linked to perception and not directly transferable to spatio-temporal forecasting. Modelling and extrapolating complex physical dynamics, such those arising in climate sciences, seems still beyond the scope of pure ML methods. The extrapolation task we address is quite different by nature from perception: future is inherently stochastic and multimodal, \ie multiple outcomes may happen from the same context situation. 
Moreover, the volume of available data for learning complex dynamical systems such as in climate is by several orders of magnitude not sufficient still nowadays \cite{schultz2021can}. Many extreme events appear very scarcely in datasets and are thus highly challenging to learn from data.

\subsection{Incorporating prior knowledge in machine learning models}

\begin{figure}
    \centering
    \includegraphics[width=\linewidth]{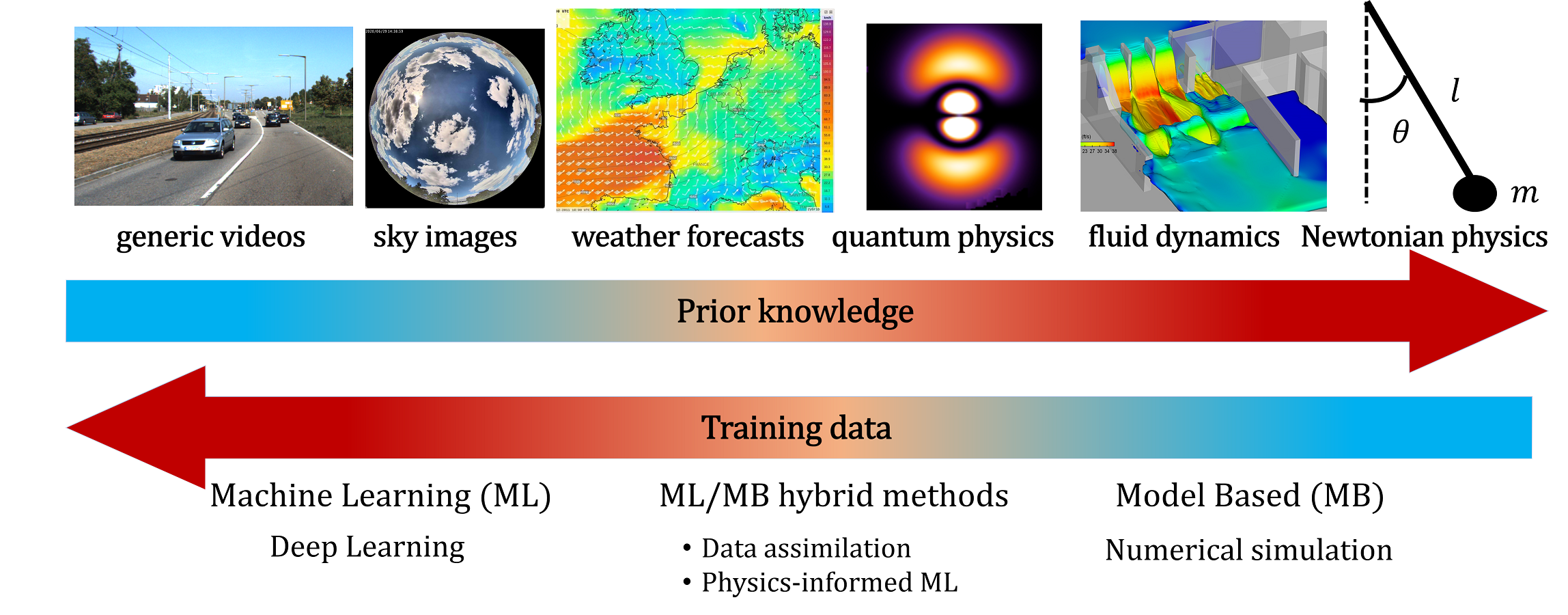}
    \caption[Data \textit{vs} prior knowledge contexts.]{\textbf{Data \textit{vs.} prior knowledge contexts.} On the left, Machine Learning (ML) and particularly Deep Learning can extrapolate dynamical systems with no prior information after training on a large dataset. On the right, traditional Model-Based (MB) approaches assume a full physical knowledge of the system and predict the future with numerical simulation from a set of initial and boundary conditions. In-between, with some data and a possibly incomplete knowledge, the ML/MB coupling is a very active and promising research direction that we explore in this thesis.}
    \label{fig:physics_data}
\end{figure}

To overcome these issues, injecting prior physical knowledge about the system is a key aspect for accurate extrapolation. This is an old question in machine learning that yet remains widely open. 
We illustrate in Figure \ref{fig:physics_data} the main classes of methods for spatio-temporal forecasting. 

On the right side of Figure \ref{fig:physics_data}, the traditional Model-Based (MB) approaches require a
deep mathematical or physical understanding of the underlying phenomena. For time series, classical state space models (SSMs) \cite{hyndman2008forecasting,box2015time} explicitly exploit the trend and seasonality patterns. For physical processes, physicists attempt to model the dynamics with first principles, conservation laws or other empirical behaviours. This physical knowledge can often be formulated through ordinary or partial differential equations (ODE/PDE) with known coefficients. With data available for the initial and boundary conditions, forecasting is performed with numerical simulation solvers. This is the classical setting in many engineering fields, such as in mechanics (where systems are described by Newtonian mechanics) or in computational fluid dynamics (with the Navier-Stockes equations), and the numerical analysis solvers are well theoretically grounded. 

However, this class of methods is limited in the case of \textit{incomplete} physical models. Models can be considered  incomplete in two situations. In the first case, the complexity of the phenomenon prevents from deriving an exhaustive analytical description of the system.  For example when modelling climate change, many complex interactions governing the state of the atmosphere are not modelled. The complete set of input variables of the system may also be unknown, \eg when forecasting financial markets or human interactions. In the second case, certain approximations are made to make the complete equations numerically tractable. For example, the Schrödinger equation that governs the wave function of a quantum-mechanical system is not exactly solvable in many non-trivial situations. Solutions are typically computed by approximate numerical schemes and with several simplifying assumptions, \eg the Born-Oppenheimer approximation. For computational time issues, the equations can also be solved on rather coarse meshes, which can prevent from capturing certain phenomena, \eg the turbulence behaviour in computational fluid dynamics.

On the other side of the spectrum, Machine Learning (ML) represents a more  prior-agnostic approach. Given a large amount of training data, deep learning has encountered impressive successes in automatically learning complex relationships without any prior knowledge, and has become state-of-the-art for many forecasting tasks, such as generic video prediction \cite{wu2021motionrnn}. However, as discussed above, deep learning is still limited for modelling highly complex dynamics of natural phenomena such as climate; although more and more data is collected about the atmosphere with in-situ or remote sensing, it is still largely largely insufficient for matching the complexity of the task. Moreover, deep neural networks lack the physical plausibility required in several domains and cannot properly extrapolate to new conditions. 


In-between, there exists a category of hybrid methods that combine MB approaches and data. Historically, data assimilation techniques \cite{corpetti2009pressure,bocquet2019data} leverage data to correct the predictions of physical models in presence of noisy observations. This includes the popular Kalman filter \cite{kalman1960new},  particle filter  \cite{perez2004data} or 4D-Var \cite{courtier1994strategy} that have achieved great successes for many smoothing/filtering/forecasting applications, for example for tracking objects in videos \cite{perez2002color}. Data assimilation still constitutes the state-of-the-art paradigm for weather forecasting.

Revisiting the ML/MB cooperation with modern deep learning is an emerging research topic motivating a great interest in many communities, attested by the soaring number of publications and workshops in top ML conferences\footnote{For example, the two workshops "Machine learning and the physical sciences" and "Tackling climate change with machine learning" at NeurIPS 2019 gathered together more than 200 papers, and even more at NeurIPS 2020.}. Physics can  be leveraged in the training process of ML models, either as a soft constraint in the loss function \cite{raissi2017physics,sirignano2018dgm} or as hard constraints in the neural network topology \cite{daw2020physics,mohan2020embedding}. From the ML point of view, these physical constraints lead to more interpretable ML models compliant to physical laws that remain robust in case of noisy data. This typically results in an increased data efficiency and better extrapolation performances beyond the training domain. Another particularly appealing direction concerns identifying and discovering physical systems: data-driven models can learn the unknown coefficients or parts in parameterized PDEs \cite{rudy2017data,long2018pde}, and discover new physical connections from data \cite{cranmer2020discovering}.  

In this thesis, we explore this category of hybrid methods and our contributions are targetted towards the following question:

\begin{center}
    \textit{How to properly exploit prior physical knowledge to improve Machine Learning forecasting models?}
\end{center}

We focus on two particular directions: injecting prior knowledge in the training objective (part \ref{part:part1}) and designing augmented MB/ML neural architectures in the case of incomplete physical models (part \ref{part:part2}).

\subsection{Industrial application at EDF: solar energy forecasting with fisheye images}

\begin{figure}
    \centering
    \includegraphics[width=16cm]{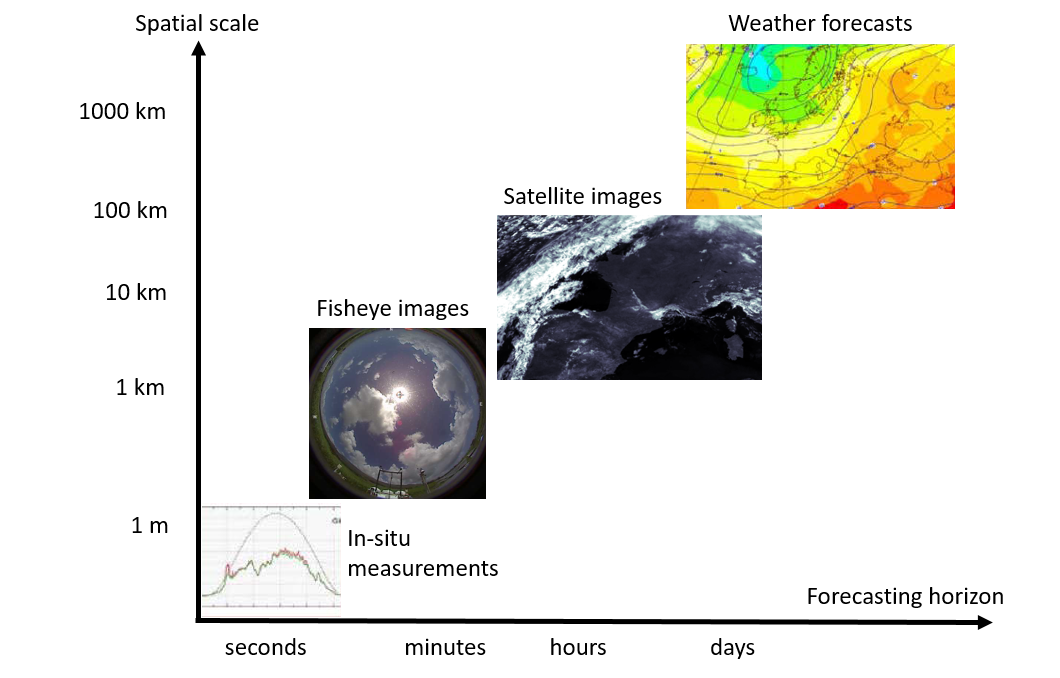}
    \caption{The different data sources for forecasting solar energy.}
    \label{fig:types-observations}
\end{figure}

At Electricité de France (EDF), the industrial use-case motivating this thesis is solar irradiance forecasting. With the increasing share of intermittent renewable energy sources such as solar or wind, accurately forecasting the electricity production and its possibly sharp variations is of great importance since the the consumption-production balance must be satisfied at every timestep. The possible data sources for this task are illustrated in Figure \ref{fig:types-observations}. 
Numerical weather forecasts are commonly used for predicting solar energy for long-term horizons up to a few days, with a typical temporal scale of 1 hour and a spatial scale of approximately 10 km. For shorter term horizons, satellite images offer forecasts up to a few hours, at a 15 min temporal granularity and a 1 to 5km spatial scale. However the spatial and temporal granularity of these two techniques are too coarse to precisely forecast the photovoltaic (PV) energy production of a given plant for very short horizons (< 20min).

To this end, images of the sky from ground-based fisheye cameras have been increasingly investigated in recent years \cite{gauchet2012surface,chu2013hybrid,chu2016sun,marquez2013intra,schmidt2016evaluating}.
Coupled with ground truth solar irradiance measurements from pyranometers, fisheye images offer an hemispheric view of the sky enabling to anticipate the evolution of the cloud cover responsible for the electric production variations. A database of several million annotated fisheye images has been collected by EDF R\&D. Estimating the irradiance corresponding to a given fisheye image is a favorable perception task for the application of deep learning. We have confirmed at the beginning of this thesis \cite{leguen-gretsi} that deep learning indeed provides a large improvement gap over traditional machine learning methods for this estimation task.

On the contrary, predicting future fisheye images for anticipating the PV production is a much more challenging extrapolation task: clouds are deformable objects with complex stochastic behaviour (that can appear or evaporate), several layers with different speeds and directions may be simultaneously present, and the fisheye camera distortion exacerbates the difficulty. In this context, even recent state-of-the-art deep learning algorihms struggle to properly extrapolate the cloud motion. We describe this use-case with more details in Chapter \ref{chap:overview_fisheye}.

\section{Scientific challenges}

We present here the main scientific challenges, highlighted by our industrial application, that we address in this thesis. 

\subsection{Multistep forecasting of non-stationary dynamics}

\begin{figure}[H]
    \centering
    \includegraphics[width=17cm]{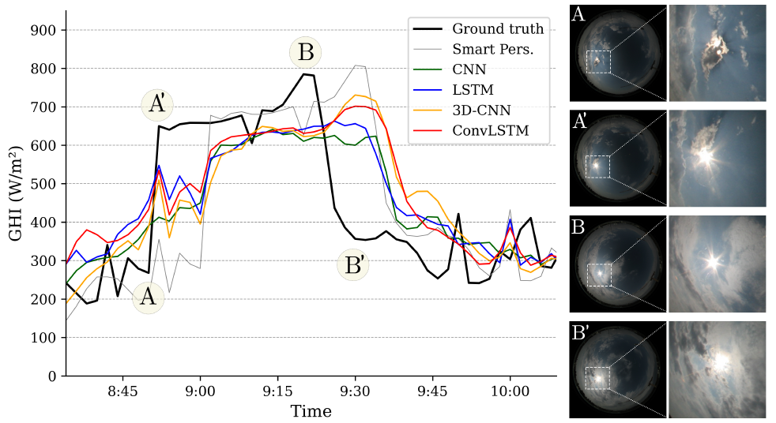}
    \caption[Limitations of standard deep learning model for solar irradiance forecasting.]{Limitations of standard Deep Learning models for 10-min ahead solar irradiance forecasting with fisheye images. Prior-agnostic Deep Learning models trained with the mean squared error do not capture the correct shape of the ground truth nor its exact temporal localization (they are temporally aligned with the smart persistence). Figure taken from Paletta \etal \cite{paletta2021benchmarking}.}
    \label{fig:paletta}
\end{figure}

We address the problem of forecasting complex dynamical systems with non-stationary dynamics, \ie with possible sharp variations. We are interested in describing the distribution of possible futures with a small set of predicted trajectories. In this context, pure data-driven methods are still limited. Paletta \etal \cite{paletta2021benchmarking} compared the performances of mainstream convolutional and recurrent neural networks for solar irradiance forecasting at a 10 minutes horizon. They show (see Figure \ref{fig:paletta}) that Deep Learning (DL) predictions struggle to match the ground truth (black curve). Two main drawbacks can be observed: (1) DL predictions smoothen the shape of the sharp drop of solar irradiance in B, and (2) the predictions are late, for example do not anticipate the drop in B\footnote{Predictions are temporally aligned with the smart persistence, which corresponds to copying the current value for the future time horizon.}. 

This solar energy forecasting problem illustrates a non-stationary forecasting context, with possible abrupt variations that need to be anticipated on time. This also occurs in many other important applications, \eg predicting future traffic flows, stocks markets, \etc. Traditional time series forecasting methods, often relying on stationarity assumptions, are not adapted for this context, and pure data-driven models struggle as well. One of the reasons is the mismatch existing between the evaluation metrics typically used to assess predictions in practice (that take into account shape and temporal errors) and the dominantly used training loss for deep models (the mean squared error). 

The main scientific challenges raised by this use-case are the following:
\begin{itemize}
    \item How to design differentiable metrics for assessing the correctness of shape and the temporal localization of future trajectories?
    \item How to efficiently describe the uncertainty by providing to the decision makers a small set of possible scenarios reflecting the shape and temporal diversity of future trajectories? In particular, how to structure the diversity of future trajectories according to shape and temporal criteria?
\end{itemize}




\subsection{Exploiting incomplete prior physical knowledge in machine learning models}

The majority of existing works for combining machine learning and physics assume a \textit{complete} physical knowledge about the system in the training process \cite{de2017deep,raissi2017physics}. In contrast to this mainstream direction, we investigate in this thesis how to leverage \textit{incomplete} physical models, \ie models that are insufficient for totally describing the dynamics. We have seen that physical models are coarse representations of the reality in many situations, \eg in physics, climate, robotics, finance, \etc.

In the solar forecasting energy example, the dynamics of clouds can be described from fluid mechanics principles. However, an exhaustive physical description is mainly out of reach since the dynamics of atmosphere is governed by many complex and interacting physical phenomena (\eg formation, evaporation of clouds, turbulence). Moreover, even a complete physical model becomes insufficient in case of missing input information, \ie  when the true state of the system (appearing in the dynamical equations) is not fully observed. In our case, we do not have a full observation about the state of the atmosphere above the PV station: we only dispose of fisheye images and we do not use information about the wind speed, the altitude of clouds and we cannot resolve if there exists several cloud layers that mask one another. 

Another exacerbating difficulty is the \textit{non-observability of the prior dynamical model}, \ie when the physical model does not apply directly in the input space. For example common laws of motion for tracking clouds in fisheye images, \eg a simple advection model, suppose that the clouds have been correctly identified and segmented and that a linear translation of clouds translates in a linear translation in the image, which is not the case because of the circular distortion of the fisheye objective.

So far, exploiting incomplete physical models has been explored by very few works \cite{long2018hybridnet,saha2020phicnet,neural20}. This problem poses many technical challenges from several points of view:
\begin{itemize}
\item Neural network architecture: how to design deep architectures with hard or soft physical constraints?
\item Training: how to efficiently train these models? From a theoretical point of view, can we provide guarantees on the quality of the ML/MB decomposition (existence, uniqueness)?
\end{itemize}


\section{Contributions and outline}

In this thesis, we address the two aforementioned scientific challenges for spatio-temporal forecasting. For multistep and non-stationary time series forecasting in deterministic and probabilistic contexts, we propose to incorporate differentiable shape and temporal features in the training scheme of deep forecasting models (part \ref{part:part1} of the thesis). For exploiting physical knowledge in deep architectures in incomplete-knowledge settings, we introduce a disentangling architecture and explore the theoretical properties of the resulting ML/MB decomposition (thesis part \ref{part:part2}). Finally, we apply our proposed ideas to the solar irradiance forecasting problem (thesis part \ref{part:part3}).

\subsection*{Part \ref{part:part1}: Differentiable shape and time criteria for deterministic and probabilistic forecasting}
\label{sec:first_deadlock}

In non-stationary contexts occurring in many industrial applications, current deep learning forecasting methods are often inadequate to properly predict sharp variations. The literature is mainly focused on new neural network architectures to improve forecasts. In contrast, the choice of the training loss function is rarely questioned. The large majority of methods are trained with the proxy Mean Squared Error (MSE) or variants that lead to non-sharp predictions. Besides, current state-of-the-art probabilistic forecasting methods are also ill-adapted for representing the shape and temporal variability of future scenarios. In this part, we propose to design training objectives that account for the shape and temporal localization of predictions. 

Our contributions to tackle the first scientific challenge are the following:
\begin{itemize}
    \item For training deep forecasting models, we introduce in Chapter \ref{chap:criteria} differentiable shape and temporal criteria inspired by evaluation metrics commonly used in applications. We propose an unifying view of these criteria both in terms of dissimilarities (loss functions) and similarities (positive semi-definite kernels). We insist on their efficient computation and differentiability, which allows to use them in deep learning pipelines.
    \item  For deterministic forecasting, we introduce in Chapter \ref{chap:dilate} the DILATE training loss function that combines a shape and a temporal dissimilarity to accurately predict sharp events with precise temporal localization. We show that training with DILATE loss instead of the MSE leads to better results at test time on several non-stationary benchmarks for generic and state-of-the-art architectures. 

    \item  For probabilistic forecasting, we present in Chapter \ref{chap:stripe} the STRIPE model that provides a set of diverse and accurate possible future trajectories. The diversity is structured with shape and temporal positive semi-definite kernels embedded in a determinantal point process (DPP) mechanism. We show that our method leads to predictions with a better quality/diversity tradeoff than competing diversifying mechanisms. 
\end{itemize}

\subsection*{Part \ref{part:part2}: Physically-informed forecasting with incomplete knowledge}
\label{sec:second_deadlock}

To advance towards the exploitation of incomplete physical knowledge in deep forecasting models, we first introduce in this part a new ML/MB deep architecture dedicated to video prediction, for which the physical laws are often not directly applicable at the pixel level. We further delve deeper into the ML/MB decomposition and we propose a new learning framework with uniqueness guarantees.

Our contributions to tackle the second scientific challenge are the following:
\begin{itemize}
    \item In Chapter \ref{chap:phydnet}, we propose a new deep architecture called PhyDNet dedicated to video prediction in non-observable prior contexts. PhyDNet learns physical dynamics parameterized by a general class of PDEs. Since the physical laws may not directly apply at the pixel level in videos, we complement the physical model with a data-driven model in charge of learning the residual information necessary for accurate prediction, such as appearance, texture, details. We show that PhyDNet reaches very good performances on several video prediction benchmarks, from a strong (linear translation for the Moving MNist dataset) to a weak prior physical knowledge (modelling general human motion for Human 3.6 dataset).
    
     \item  In Chapter \ref{chap:aphynity}, we concentrate on the ML/MB decomposition problem and the optimal cooperation between physical and data-driven models. We introduce a principled learning framework, called APHYNITY, for forecasting complex physical systems with incomplete knowledge. Inspired by the least-action principle, APHYNITY minimizes the norm of the data-driven complement under the constraint of perfect prediction of the augmented model, which leads to a unique decomposition under mild assumptions (Chebychev set). We show on several challenging physical dynamics that APHYNITY ensures better forecasting and parameter identification performances than MB or ML models alone, and that competing ML/MB hybrid methods.
\end{itemize}

\subsection*{Part \ref{part:part3}: Application to solar irradiance forecasting}
\label{sec:deadlock_application}

Finally, we apply the methodological contributions of this thesis to the solar irradiance forecasting problem at EDF.
\begin{itemize}
    \item In Chapter \ref{chap:overview_fisheye}, we present the industrial solar irradiance forecasting problem in more details and review the existing literature for solving it. We also propose a first deep learning model for estimating and forecasting solar irradiance.
    \item In Chapter \ref{chap:phydnet_fisheye}, we apply the methodological contributions of this thesis to this problem. We propose an adaptation of the introduced PhyDNet architecture to perform physically-constrained prediction. We also evaluate the DILATE loss and the APHYNITY framework on this problem and discuss future improvement directions.
\end{itemize}

Before delving in the core of the thesis, we present in Chapter \ref{chap:related_work} on overview of the basics of machine learning and the related works on spatio-temporal forecasting  and physically-constrained machine learning. Finally, in Chapter \ref{chap:conclusion}, we summarize our work and propose appealing perspectives for future works.\\
\newpage

This thesis is based on the following list of publications:

\begin{tabular}{p{14cm}|c}
\toprule
    Publication & Chapter  \\
    \midrule
  Vincent Le Guen and Nicolas Thome. "Deep Time Series Forecasting with Shape and Temporal Criteria". IEEE Transactions on Pattern Analysis and Machine Intelligence, 2022.  & \ref{chap:criteria} \\    
   \midrule  
 Vincent Le Guen and Nicolas Thome. "Shape and Time Distortion Loss for Training Deep Time Series Forecasting Models". In Advances in Neural Information Processing Systems (NeurIPS 2019).    & \ref{chap:dilate} \\
 \midrule
 Vincent Le Guen and Nicolas Thome. "Probabilistic Time Series Forecasting with Shape and Temporal Diversity". In Advances in Neural Information Processing Systems (NeurIPS 2020). & \ref{chap:stripe} \\
  \midrule
  Vincent Le Guen and Nicolas Thome. "Disentangling Physical Dynamics from Unknown Factors for Unsupervised Video Prediction". In Proceedings of the IEEE/CVF Conference on Computer Vision and Pattern Recognition (CVPR 2020). & \ref{chap:phydnet} \\
  \midrule
  Yuan Yin$^*$, Vincent Le Guen$^*$, Jeremie Dona$^*$, Ibrahim Ayed$^*$, Emmanuel de Bézenac$^*$, Nicolas Thome and Patrick Gallinari. "Augmenting Physical Models with Deep Networks for Complex Dynamics Forecasting", In International Conference on Learning Representations (ICLR 2021, oral presentation), Journal of Statistical Mechanics: Theory and Experiments (JSTAT 2021). & \ref{chap:aphynity} \\
  \midrule
  Vincent Le Guen and Nicolas Thome. "Prévision de l’irradiance solaire par réseaux de neurones profonds à l’aide de caméras au sol". In: GRETSI 2019. & \ref{chap:overview_fisheye} \\
  \midrule
  Vincent Le Guen and Nicolas Thome. "A Deep Physical Model for Solar Irradiance Forecasting With Fisheye Images". In: Proceedings of the IEEE/CVF Conference on Computer Vision and Pattern Recognition Workshops 2020 (OmniCV 2020 workshop) & \ref{chap:phydnet_fisheye}\\
\bottomrule
\end{tabular}

\clearpage{\pagestyle{empty}\cleardoublepage}


\mbox{}
\thispagestyle{empty}

\chapter{State-of-the-art on spatio-temporal forecasting}
\label{chap:related_work}
\chapabstract{

\minitoc

\begin{center}
   \textsc{Chapter abstract}
\end{center}
\textit{
In this Chapter, we first present the basic concepts of machine learning and deep learning targeted to the problem of time series forecasting (Section \ref{sec:dl-background}). Then we present an historical view of spatio-temporal forecasting, from the traditional to the more recent deep approaches for deterministic and probabilistic forecasting (Section \ref{sec:spatiotemp-forecasting}). We make a focus on the training and testing metrics, on the specific challenges of video prediction and on the question of diversity in probabilistic forecasting. Finally, we introduce the concepts of physics-based machine learning (Section \ref{sec:physicsbased-ml}). We comment the existing strategies for regularizing machine learning with physical knowledge, at the training loss and at the architectural level. We also review the question of physical system identification with machine learning and discuss the few recent works for augmenting incomplete physical models.
}
}

\section{Machine Learning}
\label{sec:dl-background}

\subsection{Background}

\begin{figure}[H]
    \centering
    \includegraphics[width=16cm]{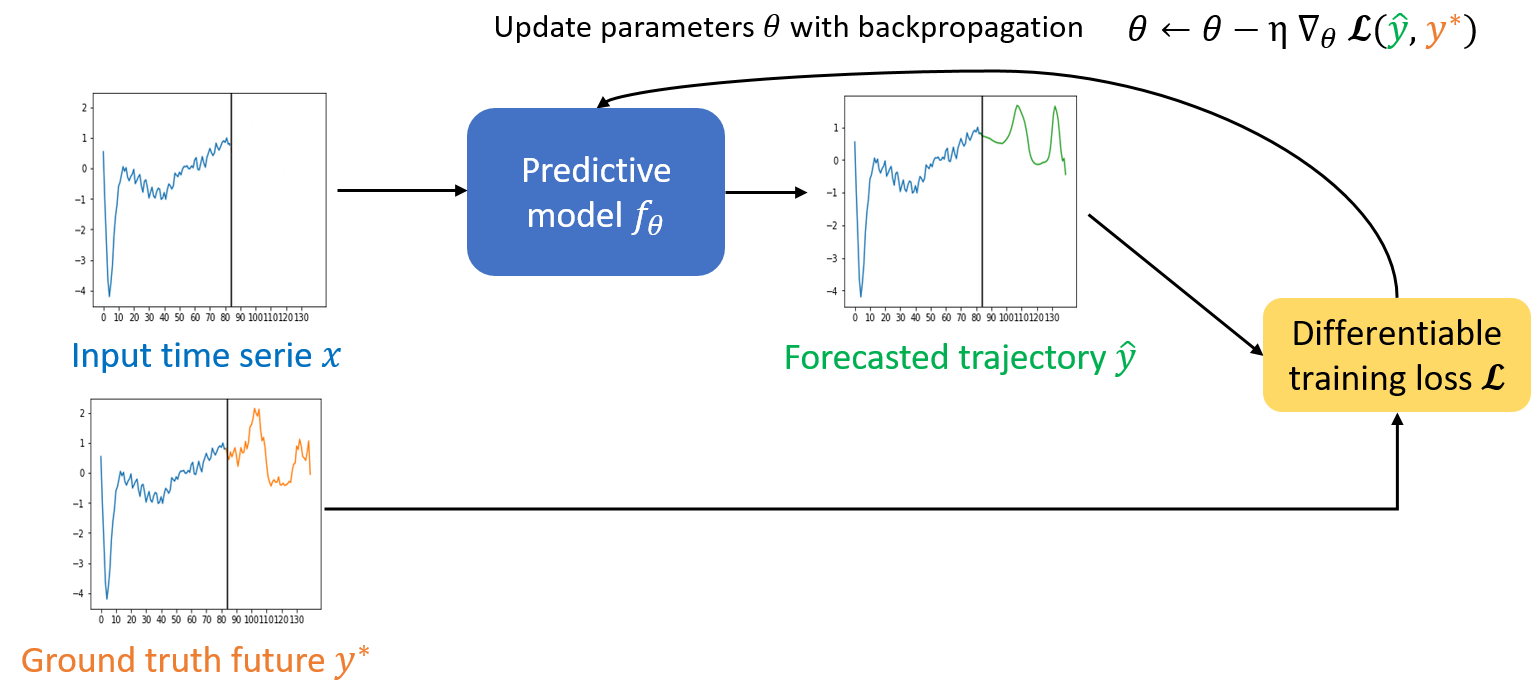}
    \caption{Supervised machine learning framework for time series forecasting.}
    \label{fig:ml_framework}
\end{figure}

\lettrine[lines=3]{D}eep Learning belongs to the broader category of statistical machine learning. In the \textit{supervised learning} context, the goal is to estimate the optimal mapping $Y= f(X)$ between inputs X and outputs Y, given a training dataset of $N$ labelled examples $\left\{ (X_i,Y_i) \right\}_{i=1}^N \in (\mathcal{X} \times \mathcal{Y})^N$. The inputs are represented by the attribute (or feature) vectors $X_i \in \mathbb{R}^d$, and the target $Y_i$ can be a categorical variable $Y_i \in \left\{ 0,1,...,K\right\}$ for classification tasks or a real variable $Y_i \in \mathbb{R}^k$ for regression tasks. We illustrate in Figure \ref{fig:ml_framework} the supervised machine learning framework in the case of time series forecasting.

\paragraph{Learning framework} The classifier or regressor function $f$ is optimized over an hypothesis class $\mathcal{H}$ of functions. Examples of classes include the linear models, the kernel methods, or the neural networks. This class should be carefully chosen for the task, guided by the bias-variance tradeoff \cite{bishop:2006:PRML}. The class $\mathcal{H}$ should be sufficiently expressive for modelling the solution of the problem; on the contrary, a too large model capacity reduces the bias but favors the overfitting phenomenon on the training set.

Once the class $\mathcal{H}$ is defined, we want to select the function $f$ that best fits the training data, while generalizing correctly to unseen input data coming from the same distribution. Training the model consists in minimizing the risk $R(f)$ that measures the disagreement between the predictions and the ground truth labels with a loss function $\ell: \mathcal{X} \times \mathcal{Y} \rightarrow \mathbb{R}^+$:
\begin{align}
    R(f) &:= \mathbb{E}_{(X,Y) \sim \mathcal{D}} ~ \ell(f(X),Y) \\
    f^* &= \text{argmin}_{f \in \mathcal{H}} ~ R(f).
\end{align}
In practice, the joint distribution $\mathcal{D}$ over $\mathcal{X} \times \mathcal{Y}$ is unknown, therefore we minimize the empirical risk defined with the training samples:
\begin{equation}
      R_n(f) := \frac{1}{n} \sum_{i=1}^N \ell(f(X_i),Y_i).
\end{equation}

\paragraph{Training loss functions} In the context of binary classification ($\mathcal{Y} = \left\{0,1\right\}$), a common loss function is the binary cross-entropy:
\begin{equation}
     \ell(f(X),Y) = - [ Y \log f(X) + (1-Y)  \log (1-f(X)) ].
\end{equation}
For regression problems such that found in time series or video prediction, the most common loss function is the mean squared error (MSE), corresponding to the L2 loss averaged over input-output pairs:
\begin{equation}
    \ell(f(X),Y) = \Vert f(X)-Y \Vert_2^2.
    \label{eq:mse}
\end{equation}

\paragraph*{Monostep vs. multistep forecasts} For time series forecasting, the loss function $\ell$ can either applied to compare monostep or multistep forecasts. Monostep forecasting methods compute a one-step ahead prediction $\mathbf{\hat{y
}}_{T+1}$ given past values $(\y_1,\cdots, \y_T)$, which is compared to the ground truth future $\y^*_{T+1}$: $\ell( \mathbf{\hat{y
}}_{T+1},\y^*_{T+1})$. In contrast, multistep forecasts compute the loss on multiple predicted timesteps:  $\ell(\mathbf{ (\hat{y}}_{t})_{T+1:T+H} , (\mathbf{y}^*)_{T+1:T+H})$. The mean squared error (MSE), dominantly used in applications, is \textit{separable}, \ie the multistep loss is the sum of the loss for all individual timesteps. In this thesis, we study dedicated loss functions for multistep forecasting, that are non separable, for explicitly imposing a desired behaviour based on the whole predicted dynamic's trajectory.

\paragraph{Regularization} Machine learning models are optimized to predict the labels of the training set. However, a model that perfectly predicts those labels does not necessarily generalize well to unseen data. With high capacity models such as deep neural networks, the risk is to learn the training set by heart and represent a too complex function; this phenomemon is called \textit{overfitting}.

To overcome this issue, a common strategy is to add a \textit{regularization} term $\Omega$ to the training objective for penalizing the complexity of the model:
\begin{equation}
    \underset{f \in \mathcal{H}}{\min} ~~ R_n(f) + \Omega(f).
\end{equation}
From a Bayesian point of view, many regularizers correspond to certain prior distributions over the model parameters. The most popular choices include the L2 and L1 weight normalization. As we will dicuss in Section \ref{sec:physicsbased-ml}, normalization is a possible way to leverage physical priors in a model.

\subsection{Deep neural networks}

Neural networks are based on the simple artificial neuron modelling proposed by MCulloch and Pitts \cite{mcculloch1943logical} and have been explored from the 1980's \cite{lecun1989backpropagation}. Standard feedforward neural networks are composed of a succession of mathematical functions called \textit{layers} that progressively transform the inputs $X$ to the outputs $Y$ through a sequence of intermediate representations $\mathbf{h}_l$ called \textit{hidden states}. A typical \textit{dense} (or \textit{fully-connected}) layer consists in a linear combination of the inputs followed by a nonlinear activation $\phi$: $\mathbf{h}_{l+1} = \phi(\mathbf{W}_l \mathbf{h}_l + \mathbf{b}_l)$ for the $l^{th}$ layer. The typical nonlinearities are traditionally the sigmoid, hyperbolic tangent or the Rectified Linear Unit (ReLU) $x \mapsto \max(0,x)$.



Neural networks are trained using gradient descent algorithms, such as the basic Stochastic Gradient Descent (SGD) \cite{bottou2010large} or variants with momentum like AdaDelta \cite{duchi2011adaptive} or Adam \cite{kingma2014adam}. The gradient of the loss with respect to the model's parameters is computed by the backpropagation method \cite{lecun1989backpropagation}. Thus all applied operations in the model should be differentiable, in particular the loss function. We will see in this thesis that the choice of a differentiable loss function is a key aspect for imposing a desired behaviour. 

 Deep Learning has become popular since the victory of the AlexNet model \cite{krizhevsky2012imagenet} at the ImageNet competition in 2012. The main revolution of Deep Learning relies in the depth of the neural networks. By stacking many layers, the network progressively learns more and more complex feature representations of the input, from the low-level concepts (such as color or contours) to the most semantics concepts (such as the recognition of a particular object) necessary for image classification.

\begin{figure}[H]
    \centering
    \includegraphics[width=16cm]{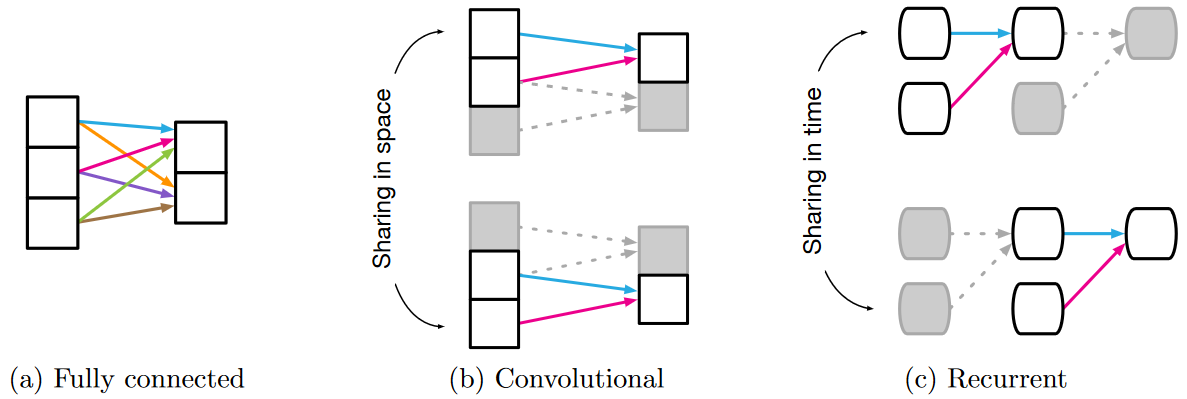}
    \caption[The common layers used in deep learning.]{Common layers used in deep learning models. Shared parameters are shown with the same color. Figure taken from Battaglia \etal \cite{battaglia2018relational}.}
    \label{fig:deep-archis}
\end{figure}

The choice of the neural network architecture is a critical aspect for solving a task. We illustrate in Figure \ref{fig:deep-archis} the three main kinds of layers. The Multi-Layer Perceptron (MLP) \cite{rosenblatt1961principles}, only composed of fully-connected layers, is the most generic architecture but at the expense of a number of parameters exponentially growing with the number of layers, making it not amenable for many applications. Other architectures encode specific inductive biases on data. For example, convolutional neural networks \cite{lecun1989backpropagation} encodes spatial equivariance, \ie the response of a classifier should be independent to the particular location of objects in the image, by sharing a convolutional filter for all spatial positions. Likewise, recurrent neural networks encode translation equivariance for processing sequential data by reusing the same weight in time. More recent architecture also encode other kinds of inductive biases: graph neural networks \cite{battaglia2016interaction} encode permutation invariance among a set of items, and the recent Transformer architecture \cite{vaswani2017attention} implements an attention mechanism over neighbouring positions.

When investigating deeper and deeper architectures, researchers have been faced with training issues like the vanishing gradient problem, \ie the gradient of the loss can become very small after backpropagating through a large number of layers. To overcome this problem, He \etal \cite{he2016deep} has proposed the \textit{residual neural networks} (ResNets) by adding skip connections between a block of standard layers: 
\begin{equation}
    \mathbf{x}_{l+1} = \mathcal{F}(\mathbf{x}_l) + \mathbf{x}_l,
\end{equation}
where $\mathbf{x}_l$ is the hidden state after the $l^{th}$ block and $\mathbf{F}$ denotes a nonlinear function (\eg a series of convolutions and nonlinear activations). These "identity shortcuts" allow a direct flow of the gradient and have significantly improved the training of very deep networks, leading to new state-of-the-art performances on ImageNet. Pursuing this idea, the \textit{densely connected networks} (DenseNets) of Huang \etal \cite{huang2017densely}, connecting all layers together within a block with skip connections, have further improved the performances.

\begin{figure}[H]
    \centering
    \includegraphics[width=17cm]{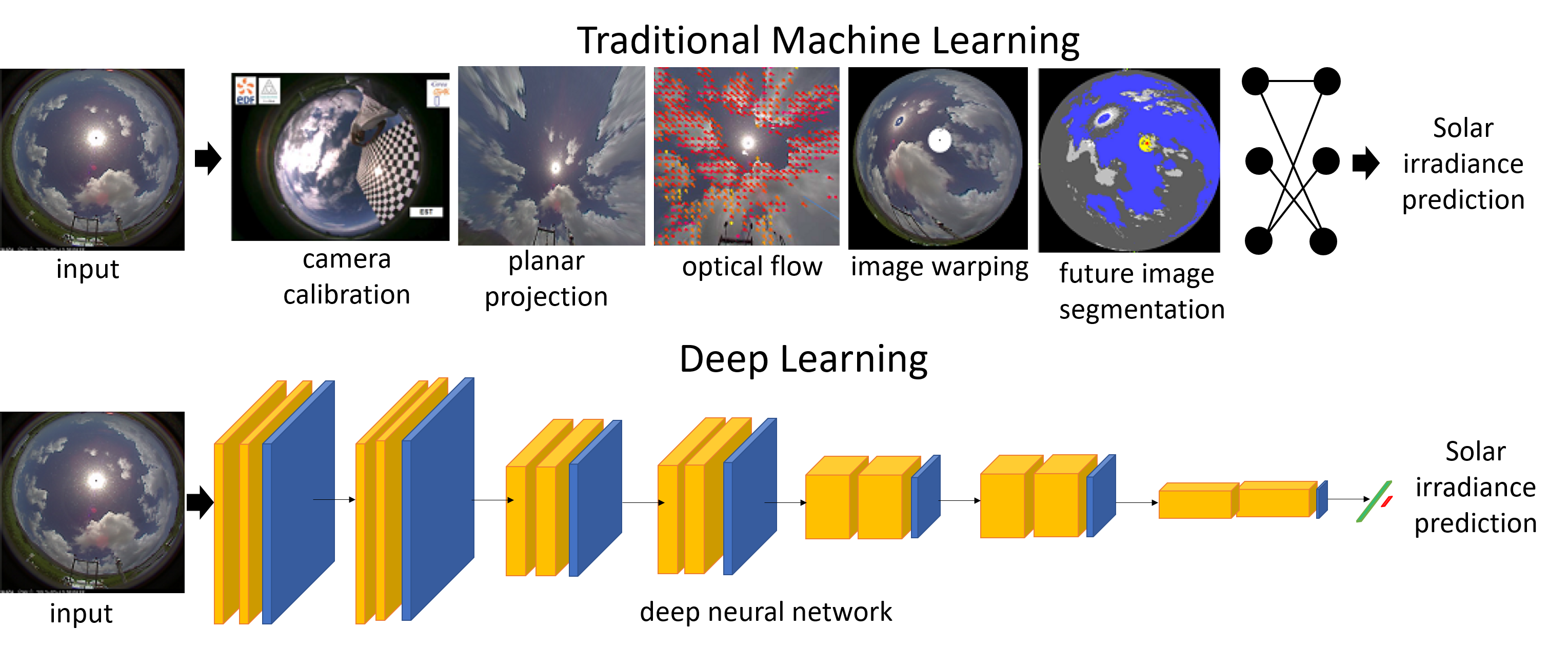}
    \caption{Traditional Machine Learning vs. Deep Learning for forecasting solar irradiance with fisheye images.}
    \label{fig:mldl_diff}
\end{figure}

\paragraph{Difference between traditional ML and DL}
The main differences between traditional Machine Learning (ML) and Deep Learning (DL) are illustrated in Figure \ref{fig:mldl_diff} for the case of solar irradiance forecasting with fisheye images. The traditional ML pipeline (from the existing method at EDF \cite{gauchet2012surface}) is composed of several steps with manual intervention: camera calibration for compensating the fisheye distortion, projection of the input images on a plane at a given altitude, optical flow estimation, image warping for computing the future frame, future image segmentation with handcrafted features and thresholds, and finally prediction of the future irradiance with a traditional regressor (\eg linear regression). Many of these steps require expert manual intervention. On the other side, the Deep Learning approach directly learns the image to irradiance mapping on raw fisheye images and automatically derives the appropriate intermediate concepts.

In fact, the difficulty of the task has shifted from the handcrafted feature engineering of traditional ML methods to the manual neural network architecture design of DL that encodes appropriate inductive biases or behaviours. 

\section{Spatio-temporal forecasting}
\label{sec:spatiotemp-forecasting}

In this Section, we review the main existing machine learning approaches for spatio-temporal forecasting, from the traditional statistical time series forecasting to the most recent deep learning methods.

\subsection{Context and notations}

As discussed in Introduction (Chapter \ref{chap:intro}), we are interested in forecasting spatio-temporal processes driven by some underlying physical phenomenon. We consider dynamical systems formalized through a differential equation of the form:
\begin{equation}
    \frac{\diff X_t}{\diff t} = F(X_t).
\label{eq:ode-relatedwork}
\end{equation}
The \textit{state} of the system $X_t$ represent the variables whose knowledge at time $t_0$ is sufficient, in combination with the evolution function $F$, for describing the phenomenon for each time $t>t_0$. The state $X_t$ can be either be parameterized by:
\begin{itemize}
 \setlength\itemsep{0em}
    \item a  $d$-dimensional vector, \ie we have $X_t\in\R^d$ for every $t$. In that case, equation \ref{eq:ode-relatedwork} is an \textit{ordinary differential equation} (ODE);
    \item a $d$-dimensional vector field over a spatial domain $\Omega\subset\R^k$, with $k\in\{2,3\}$, \ie $X_t(x)\in\R^d$ for every $(t,x)\in[0,T]\times\Omega$. If the description in Eq \ref{eq:ode-relatedwork} involves spatial derivatives of the state, it corresponds to a  \textit{partial differential equation} (PDE).
\end{itemize}

Many phenomena occurring in physics, biology, computer vision, finance follow a general equation of the form \ref{eq:ode-relatedwork}.

To solve the differential equation \ref{eq:ode-relatedwork} numerically, the most common approach is to discretize the phenomenon into a sequence $(\x_1, \x_2, \cdots, \x_T)$ and approximate the time derivative with finite differences. The simplest numerical scheme is the forward Euler method:
\begin{equation}
    \x_{n+1} = \x_n + \Delta t ~ F(\x_n),
\end{equation}
where $\Delta t$ is a fixed step size. We will see that this approximation scheme has strong connections with  residual neural networks (Section \ref{sec:continuous-time-models}). More complex numerical schemes exist with lower truncation errors, \eg Runge-Kutta \cite{butcher2016numerical}.

For predicting a dynamical system of the form \ref{eq:ode-relatedwork}, two main modelling approaches exist:
\begin{itemize}
    \item parameterize the relationship between future time steps and context time steps: $ (\hat{\y}_{T+1},\dots, \hat{\y}_{T+H} ) = g_{\theta}(\x_1,\dots,\x_T)$ with parameters $\theta$. The function $g_{\theta}$ can represent a traditional time series forecasting model like an autoregressive model \cite{box2015time} or a deep neural network.
    \item parameterize the derivative function $F_{\theta}$ and integrate the ODE/PDE with a numerical solver. This is the typical case of numerical simulation with a physical model  $F_{\theta}$. The function  $F_{\theta}$ can also be a deep neural network approximating the dynamics, as done by the Neural ODEs \cite{chen2018neural}  presented in Section \ref{sec:continuous-time-models}.
\end{itemize}


\subsection{Model-Based forecasting methods}

As discussed in Chapter \ref{chap:intro}, the traditional modelling paradigm in physics is to derive analytical laws of motion from first principles and integrate the equations with numerical simulation. These models are often expressed as ordinary or partial differential equations (ODEs/PDEs). This arises in a multitude of scientific fields, such as Newtonian mechanics, fluid dynamics or quantum mechanics. For example, we will consider in this thesis the wave equations:
\begin{equation*}
        \frac{\partial^2 w}{\partial t^2} - c^2\Delta w + k \frac{\partial w}{\partial t}=0 ,
\end{equation*}
where $k$ is the damping coefficient and $c$ the celerity of the wave.

For time series forecasting, traditional Model-Based methods rely on linear state space models (SSMs) \cite{durbin2012time,hyndman2008forecasting}, which provide a principled framework for modelling known temporal patterns. SSMs include the popular integrated autoregressive moving-average model (ARIMA) and Exponential Smoothing. SSMs assume linear dynamics with structural components (\eg level, trend, seasonality), which makes forecasting robust and interpretable. However, the model selection procedure can be tedious and these methods often exploit strong statistical (\eg i.i.d. additive Gaussian noise) and structural assumptions on data (\eg stationarity or stationarity after differentiation) that are not satisfied for many real-world time series that can present abrupt changes of distribution. Moreover, SSMs are fitted independently on each time series, and thus cannot learn patterns between sets of similar series.

Regarding video prediction, traditional methods focus on predicting the motion field with optical flow, rather than predicting future frames at the pixels level. The seminal works of Lucas-Kanade \cite{lucas1981iterative} and Horn-Schunk \cite{horn1981determining} rely on the brightness consistency constraint, which assumes that the intensity value of a pixel remains constant between two frames. In its linearized form, this constraint can be expressed as a PDE:
\begin{equation}
    \frac{\partial I}{\partial t} (t,\mathbf{x}) = - w(t,\mathbf{x}) \cdot \nabla I (t,\mathbf{x}).
    \label{eq:flot}
\end{equation}
Again, this PDE corresponds to an incomplete model, since the  brightness constancy assumption is violated in many situations,  \eg in presence of occlusions, illumination changes, specular reflexions.


\subsection{Deep learning forecasting methods}

Artificial neural networks were first explored in the 1990's for time series forecasting with Multi-Layer Perceptrons (MLPs) \cite{chakraborty1992forecasting,lee1992short,tang1993feedforward}
and Recurrent Neural Networks (RNNs) \cite{connor1994recurrent,kuan1995forecasting}. At that time, most of these architectures were limited to a single hidden layer and trained with one-step targets, restricting their applicability to simple problems.

With the advances in computer hardware and modern training techniques of the deep learning era, neural networks have become appealing for time series forecasting due to their automatic feature extraction, the ability to capture complex nonlinear temporal patterns and the ease to incorporate exogenous variables.

\subsubsection{Recurrent Neural Networks (RNNs)} 

\begin{figure}
    \centering
    \includegraphics[width=12cm]{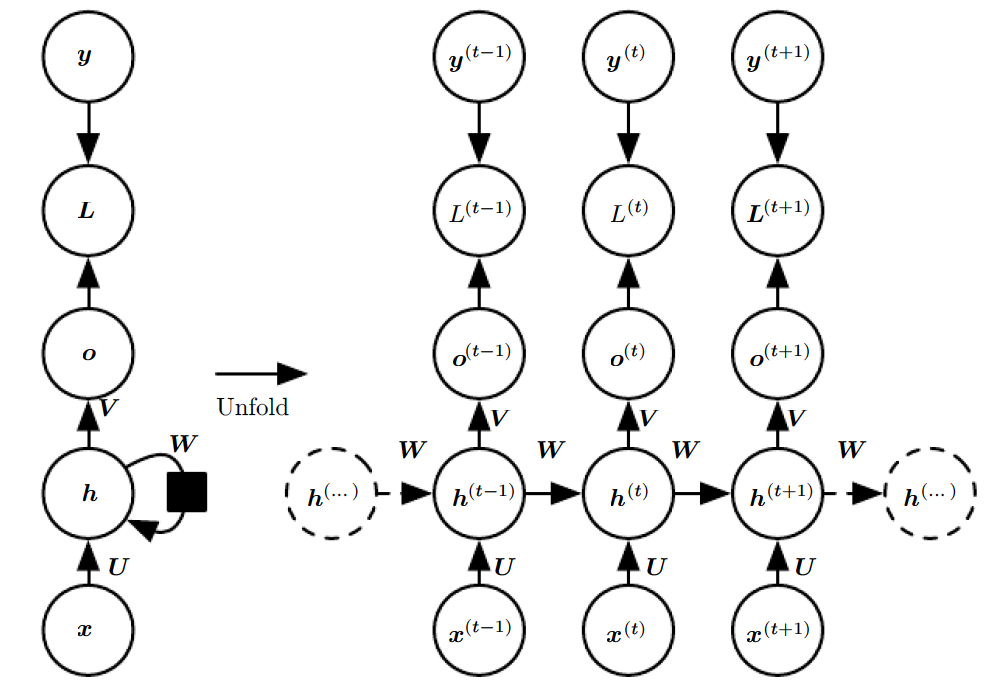}
    \caption[Illustration of a recurrent neural network.]{A recurrent neural network. Figure taken from Goodfellow \cite{goodfellow2016deep}.}
    \label{fig:rnn}
\end{figure}

RNNs denote a family of architectures dedicated to handling sequential data such as text, speech or time series. Illustrated in Figure \ref{fig:rnn}, RNNs implement a discrete time dynamical system, where a hidden variable $\mathbf{h}_t \in \mathbb{R}^d$, serving as a memory of the system, is recurrently updated across time. A basic RNN formulation can be written as:
\begin{align}
    \mathbf{h}_t &= F(\mathbf{W} ~\mathbf{h}_{t-1} + \mathbf{U}~ \mathbf{x}_t + \mathbf{b} )
    \label{eq:rnn}\\
    \mathbf{o}_t &= \mathbf{V} ~ \mathbf{h}_t,
\end{align}
where $\mathbf{U}$ and $\mathbf{W}$ are weight matrices, $\mathbf{b}$ is a bias and $F$ an activation function (\eg $\tanh$). The output $\mathbf{o}_t$ at time $t$, obtained by a projection of the latent state with a weight matrix $\mathbf{V}$, is compared to the ground truth target $\y_t$ with a loss function $L$. Crucially, the weights of the RNN are identical for all timesteps (as shown in Figure \ref{fig:rnn}). Contrary to more general MLPs, weight sharing in RNNs enables to encode time equivariance and to process sequences of arbitrary lengths. Deep recurrent neural networks can be build by stacking  RNN cells.

RNNs are trained by backpropagation through time \cite{mozer1989focused}, \ie by propagating the gradient of the loss function in the unfolded computational graph (see Figure \ref{fig:rnn}). A major drawback of the vanilla formulation in Eq \ref{eq:rnn} is that the vanshing / exploding gradients when processing long sequences \cite{pascanu2013difficulty}. It prevents the network from memorizing long-term information in the current latent state. To address this limitation and model long-term dependencies, Hochreiter \etal \cite{Hochreiter:1997:LSM:1246443.1246450} introduced the Long-Short Term Memory (LSTM) networks which have an additional memory cell $\mathbf{c}_t$ controlled by a learned input gate $\mathbf{i}_t$ and forget gate $\mathbf{f}_t$:
\begin{align*}
    \mathbf{i}_t &= \sigma (\mathbf{W}_{ih} ~ \mathbf{h}_{t-1} + \mathbf{W}_{ix} ~ \mathbf{x}_t + \mathbf{b}_i) \\
    \mathbf{f}_t &= \sigma (\mathbf{W}_{fh}  ~\mathbf{h}_{t-1} + \mathbf{W}_{fx} ~ \mathbf{x}_t + \mathbf{b}_f) \\
    \mathbf{c}_t &= \mathbf{f}_t \odot \mathbf{c}_{t-1} + \mathbf{i}_t \odot    \tanh (\mathbf{W}_{gh} ~ \mathbf{h}_{t-1} + \mathbf{W}_{gx} ~ \mathbf{x}_t + \mathbf{b}_g) \\
    \mathbf{o}_t &= \sigma (\mathbf{W}_{oh} ~ \mathbf{h}_{t-1} + \mathbf{W}_{ox} ~ \mathbf{x}_t + \mathbf{b}_o) \\
    \mathbf{h}_t &= \mathbf{o}_t \odot \tanh(\mathbf{c}_t).
\end{align*}

LSTM networks and their variants such as the Gated Recurrent Unit (GRU) \cite{cho2014learning}, have become a reference for many sequential tasks. Shi \etal \cite{xingjian2015convolutional} proposed the ConvLSTM adaptation for video prediction, by replacing all the full-connected operations of the LSTM by convolutions. The ConvLSTM was adopted in many subsequent studies \cite{finn2016unsupervised,jia2016dynamic,xu2018structure} and is at the basis of the most recent video prediction algorithms such as PredRNN \cite{wang2017predrnn,wang2018predrnn++}, Memory in Memory \cite{wang2019memory} or MotionRNN \cite{wu2021motionrnn}.

\subsubsection{Sequence To Sequence models}

For mapping a variable-length sequence to another variable-length sequence, Cho \etal \cite{cho2014learning} and Sutskever \etal \cite{sutskever2014sequence} proposed the Sequence To sequence (Seq2Seq) architecture. The input sequence $(\x_1,\cdots,\x_{n_x})$ is processed by an encoder RNN that provides a fixed-size context vector $C$ summarizing the sequence, typically defined as the last hidden state of the RNN. This context vector is used for initializing the decoder which is another RNN producing the predictions $(\y_1,\cdots,\y_{n_y})$   one step at a time. In a Seq2Seq model, both RNNs are trained jointly to maximize the likelihood $p(\y_1,\cdots,\y_{n_y} | \x_1,\cdots,\x_{n_x})$ averaged over all the input/output sequences of the training set.

When generating predictions, the RNN decoder is rolled forwards by recursively feeding back its own predictions as inputs for the next timesteps. Seq2Seq models can be trained with \textit{teacher forcing}, consisting in feeding the true targets as inputs to the RNN (that are known at training time) instead of the prediction from the last timestep. A popular curriculum often used in practice to mitigate the train/test discrepancy is \textit{scheduled sampling} \cite{bengio2015scheduled} that randomly chooses to use true values or model predictions as inputs, with a sampling probability to use model predictions increasing over time to gradually converge towards test-time conditions. 

\begin{figure}
    \centering
    \includegraphics[width=16cm]{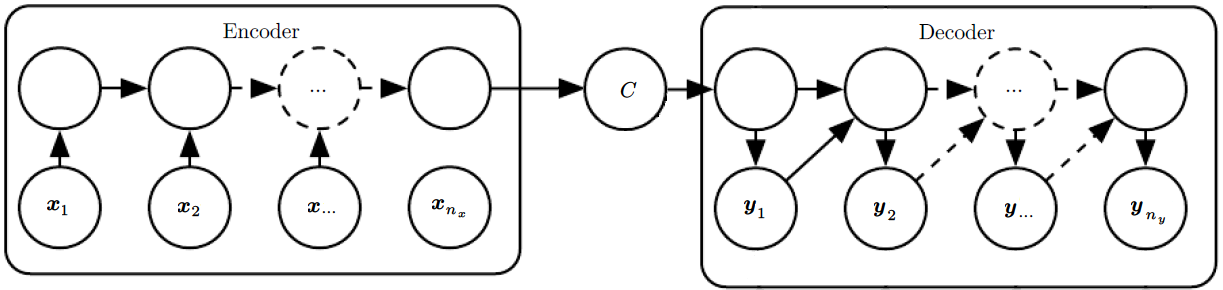}
    \caption[Illustration of a Sequence To Sequence model.]{Sequence To Sequence model. Figure adapted from Goodfellow \cite{goodfellow2016deep}.}
    \label{fig:my_label}
\end{figure}

Seq2Seq architectures with RNNs are at the basis of many successful models \cite{fox2018deep,rangapuram2018deep,kuznetsov2018foundations}. Salinas \etal \cite{salinas2017deepar} proposed DeepAR, a Seq2Seq model which estimates the parameters of a Gaussian distribution for the next timestep. Rangapuram \etal \cite{rangapuram2018deep} revisit the traditional state space models (SSMs) by parameterizing them with deep recurrent networks. To limit error accumulation due to autoregressive predictions, some models directly predict all future values at once, often with a MLP decoder \cite{wen2017multi}.

RNN forecasting can be improved with the attention mechanism, introduced by Bahdanau \etal \cite{Bahdanau2015NeuralMT} for machine translation \cite{qin2017dual,lai2018modeling,fan2019multi}. Attention consists in learning which part of the input sequence is the most relevant for predicting a given timestep. More precisely, the context vector $C$ is replaced with a combination of the hidden states from past timesteps weighted by their learned attention weights.

\subsubsection{Beyond recurrent architectures}

Training RNNs with backpropagation through time is expensive since it requires sequential operations that cannot be parallelized. Researchers have explored alternative architectures than RNNs. Following the success of the Wavenet model for audio processing \cite{van2016wavenet}, temporal convolution networks (TCNs) \cite{borovykh2017conditional,chen2020probabilistic} use causal dilated 1D-convolutions, that exponentially increase the receptive field with additional layers and respect the temporal causality. In addition, TCNs can be easily trained in parallel. 

Recently, a line of works has questioned the convolutional or recurrent layers used in most architectures, showing that fully-connected layers arranged in a careful way can outperform other methods. For example, pure attention-based models have revealed better than LSTMs for capturing long-range relationships. The Transformer architecture of Vaswani \etal \cite{vaswani2017attention}, only composed of self-attention and fully-connected layers, avoids the recurrent structure and provides a direct access to any previous timestep. Several works have proposed adaptations of the Transformer for time series forecasting \cite{li2019enhancing,zhou2020informer}. In particular, the Informer model of Zhou \etal \cite{zhou2020informer} is able to extend the predictions to a long-term horizon with less degradation than competing methods.

\begin{figure}
    \centering
    \includegraphics[width=15cm]{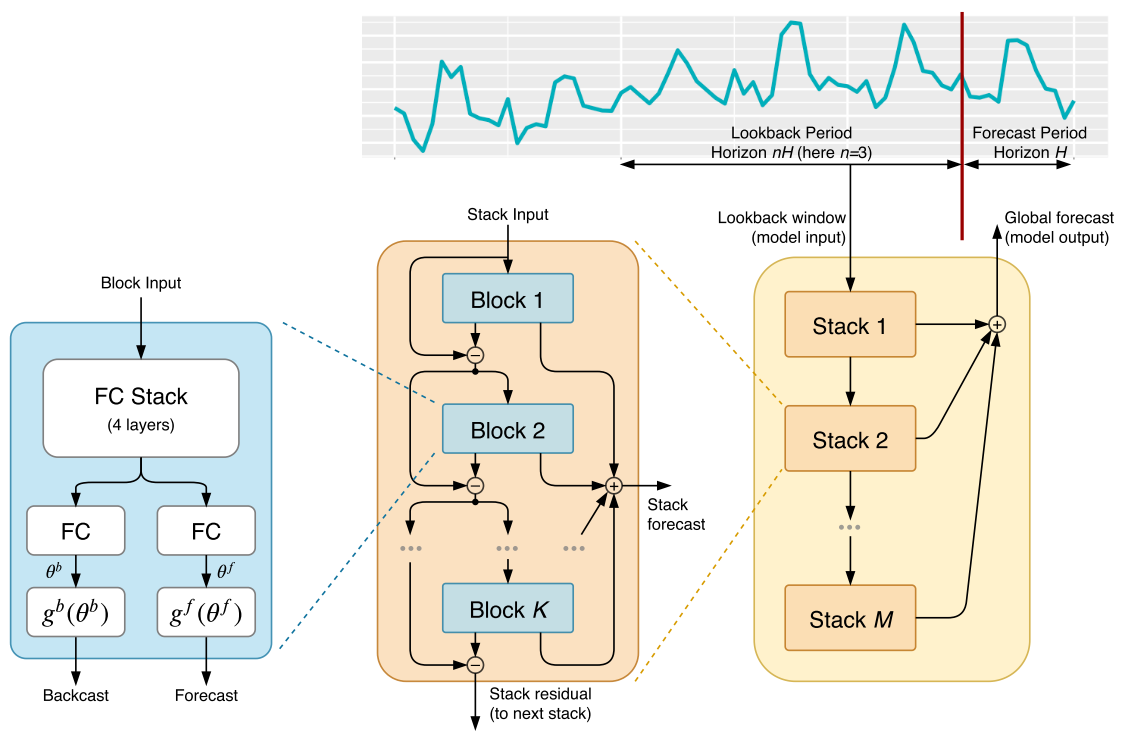}
    \caption[The NBeats model for deterministic forecasting.]{The NBeats model for deterministic forecasting \cite{oreshkin2019n}.}
    \label{fig:nbeats}
\end{figure}

Another example is the NBeats forecasting architecture \cite{oreshkin2019n} shown in Figure \ref{fig:nbeats} that has recently shown state-of-the-art performances for deterministic forecasting. NBeats is composed of stacks of fully-connected layers, each block outputting a forecast for the following block and a backcast that removes the part of the signal that is well-explained by the current block. Partial forecasts from each block are finally combined into the global forecast.

\subsection{Training and evaluation metrics for time series forecasting}

Current research on time series forecasting mainly focuses on new architecture design (the predictive model $f_{\theta}$ in the blue box in Figure \ref{fig:ml_framework}) and the question of the training loss (yellow box in Figure \ref{fig:ml_framework})  is often overlooked. The Mean Squared Error (MSE) in Eq \ref{eq:mse}, Mean Absolute Error (MAE) and its variants (SMAPE, \textit{etc}) are predominantly used as proxies for training models. In practice, forecasts are evaluated with application-specific metrics, often reflecting the shape and temporal localization of future trajectories. However, their non-differentiability makes them unsuitable for training deep models. For characterizing shape, the  Dynamic Time Warping (DTW) algorithm \cite{sakoe1990dynamic,jeong2011weighted,zhang2017dynamic}, originally introduced for speech recognition, computes the similarity between time series after temporal alignment. DTW is particularly popular for time series classification \cite{jeong2011weighted} or clustering \cite{chang2021learning} and has been recently explored for time series forecasting \cite{cuturi2017soft}. Another shape metric is the ramp score  \cite{florita2013identifying,vallance2017towards} that assesses the detection of ramping events in wind and solar energy forecasting. Timing errors can be characterized among other ways by the Temporal Distortion Index (TDI) \cite{frias2017assessing,vallance2017towards}, or by computing detection scores (precision, recall, Hausdorff distance) after change point detection \cite{truong2019supervised}.  

Recently, some attempts have been made to train deep neural networks based on alternatives to MSE, especially based on smooth approximations of DTW \cite{cuturi2017soft, mensch2018differentiable,abid2018learning,vayer2020time,blondel2020differentiable}, in particular the soft DTW \cite{cuturi2017soft} that we will detail in Chapter \ref{chap:dilate}. 

 
In this thesis, we intend to bridge the gap between these common evaluation metrics and the training losses used in practice. We explore how to efficiently combine explicit shape and temporal differentiable criteria at training time, regardless of the training architecture. We will review the most related works in more details in Chapter \ref{chap:criteria}.

\subsection{Particular challenges in video prediction}

Videos are a particular form of multivariate time series, and all the time series forecasting methods presented above could in principle be directly applied to videos by forecasting the dynamics of individual pixels. However this approach neglects the keys properties of images: the spatial coherence between neighboring pixels and the semantics of the scene. Specific architectures dedicated to video prediction were explored \cite{wang2017predrnn,wang2018predrnn++,wang2019memory,wang2018eidetic,wu2021motionrnn}, often based on variants of the seminal ConvLSTM \cite{shi2015convolutional}.

Moreover, extrapolating high-dimensional signals such as images at the pixel level is extremely challenging. To constrain this generation problem, several methods rather use domain-specific knowledge such as predicting geometric transformations between frames \cite{finn2016unsupervised,jia2016dynamic,xue2016visual}, estimating the optical flow \cite{patraucean2015spatio,luo2017unsupervised,liu2017video,liang2017dual,li2018flow} or exploiting the semantics of the scene \cite{bei2021learning}. This is very effective for short-term prediction, but degrades quickly when the video content evolves, where more complex models and memory about dynamics are required. 

\begin{figure}[H]
    \includegraphics[width=14cm]{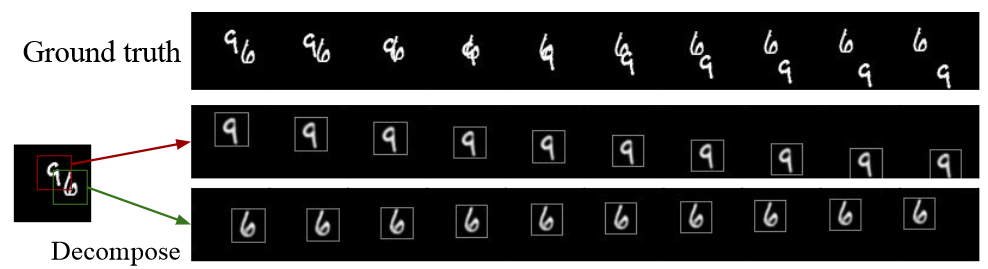}
    \caption[Illustration of disentanglement for video prediction.]{Disentanglement approach for video prediction. In this Moving MNIST example, the DPPAE model \cite{hsieh2018learning} decomposes the two digits and predicts their dynamics separately.}
   \label{fig:dppae}
\end{figure}

\paragraph*{Disentanglement} Another line of work consists in disentangling independent factors of variations in order to apply the prediction model on lower-dimensional representations. The typical decomposition criteria are  as content/motion \cite{villegas2017decomposing,lee2021video} or deterministic/stochastic \cite{denton2017unsupervised}. We illustrate in Figure \ref{fig:dppae} an example of decomposition from the DPPAE model \cite{hsieh2018learning}: the moving objects are extracted and their individual motion estimated separately to provide the final prediction. In specific contexts, the prediction space can be structured using additional information, \eg with human pose \cite{villegas2017learning,walker2017pose} or key points \cite{minderer2019unsupervised}.

We provide a more detailed review of existing deep video prediction methods in Chapter \ref{chap:phydnet}.

\subsection{Diversity in probabilistic forecasting}

Many critical applications require forecasts associated with uncertainty to make relevant decisions. Probabilistic forecasting consists in estimating the predictive distribution of future values given an input sequence. Two main categories of methods exist for probabilistic forecasting. The first class of methods directly characterizes the predictive distribution. This includes estimating the variance of predictions (\eg with Monte Carlo dropout \cite{gal2016dropout}),  estimating the quantiles \cite{wen2017multi,gasthaus2019probabilistic,wen2019deep} or  modelling this distribution by a parametric distribution, \eg a Gaussian for the DeepAR algorithm \cite{salinas2017deepar}).

In this thesis, we focus on a second class of probabilistic methods that propose to describe the predictive distribution with a set of plausible scenarios reflecting the uncertainty of future behaviour. This class includes ensemble methods \cite{smyl2019machine} and generative models, which produce diverse forecasts by sampling multiple latent variables from a prior distribution. The most popular generative models are conditional variational autoencoders (cVAEs) \cite{yuan2019diverse}, conditional generative adversarial networks (cGANs) \cite{koochali2020if}, and normalizing flows \cite{rasul2020multi,de2020normalizing}). For further diversifying forecasts, several repulsive schemes were studied such as the variety loss \cite{gupta2018social,thiede2019analyzing} that consists in optimizing the best sample, or entropy regularization \cite{dieng2019prescribed,wang2019nonlinear} that encourages a uniform distribution.

However the aforementioned methods are limited for representing the diversity of future behaviour with a limited number of scenarios, as discussed in Chapter \ref{chap:intro}. Standard generative models sample points belonging to the dominant mode, \eg by sampling multiple forecasts at test time from a standard Gaussian prior, and do not provide control over the diversity of predictions.

\textbf{Determinantal Point Processes (DPP)} To improve this unstructured mechanism, prior works \cite{yuan2019diverse,yuan2020dlow} introduced proposal neural networks for generating the latent variables that are trained with a diversity objective based on Determinantal Point Processes (DPPs).

 DPPs are appealing probabilistic models for describing the diversity of a set of items $\mathcal{Y}= \left\{\mathbf{y}_1,...,\mathbf{y}_N \right\}$.
A DPP is a probability distribution over all subsets of $\mathcal{Y}$ that assigns the following probability to a random subset $\mathbf{Y}$:
\begin{equation}
    \mathcal{P}_{\mathbf{K}}(\mathbf{Y}=Y) = \frac{\det(\mathbf{K}_Y)}{\sum_{Y' \subseteq \mathcal{Y}}\det(\mathbf{K}_Y')} = \frac{\det(\mathbf{K}_Y)}{\det(\mathbf{K+I})},
\end{equation}{}
where $\mathbf{K}$ is a positive semi-definite (PSD) kernel and $\mathbf{K}_A$  denotes its restriction to the elements indexed by $A$.  

\begin{figure}[H]
    \centering
    \includegraphics[width=14cm]{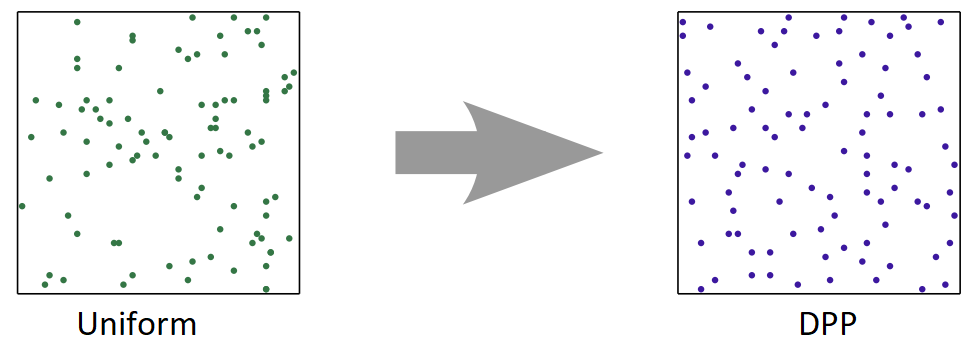}
    \caption[Random points sampled in the plane from a uniform distribution vs a determinantal point process (DPP) distribution.]{Random points sampled in the plane from a uniform distribution vs a determinantal point process (DPP) distribution. Figure taken from Kulesza and Taskar \cite{kulesza2012determinantal}.}
    \label{fig:dpp}
\end{figure}

We illustrate the behaviour of DPPs in Figure \ref{fig:dpp} for sampling random points in the plane. When we draw points randomly according to a uniform distribution, some regions may become more densely populated than other. In contrast, when sampling from a DPP distribution with a Gaussian kernel, points are farther from one another and better spread on to the plane.

DPPs offer efficient algorithms for sequentially sampling diverse items or maximizing the diversity of a set with a given sampling budget. Importantly, the choice of the kernel enables to incorporate prior structural knowledge on the targeted diversity. As such, DPPs have been successfully applied in various contexts, \eg document summarization \cite{gong2014diverse}, recommendation systems \cite{gillenwater2014expectation}, image generation \cite{elfeki2018gdpp} and diverse trajectory forecasting \cite{yuan2019diverse}. 

In this thesis, we design specific shape and temporal PSD kernels for imposing our structured diversity. We further describe the most related works for probabilistic forecasting in Chapter \ref{chap:stripe}.

\section{Physics-informed machine learning}
\label{sec:physicsbased-ml}

As discussed in Chapter \ref{chap:intro}, pure data-driven machine learning methods struggle to extrapolate complex dynamical systems, and often overfit on the training set. Incorporating prior knowledge about the system is an appealing way to regularize the training process. In this Section, we review the main existing approaches for combining machine learning with physical knowledge (called \textit{ML/MB}, \textit{gray-box}, or \textit{hybrid} modelling in the literature).

\subsection{Continuous time models}
\label{sec:continuous-time-models}

Continuous-time models, consisting in modelling the rate of change $F$ of an ODE with a neural network, were first explored from the 1980's \cite{cohen1983absolute,gonzalez1998identification,zhang2014comprehensive}. More recently, researchers have drawn tight connections between dynamical systems and deep (residual) neural networks \cite{weinan2017proposal,lu2018beyond,zhu2018convolutional,chen2018neural}. The residual bloc of a ResNet \cite{he2016deep} 
\begin{equation}
\mathbf{h}_{t+1} = \mathbf{h}_t + \Delta t ~ F(\mathbf{h}_t, \theta)
\end{equation}
 can be interpreted as the forward Euler discretization of the dynamical system 
 \begin{equation}
     \frac{\diff \mathbf{h}(t)}{\diff t} = F(\mathbf{h}(t), \theta).
 \end{equation}
 Mainstream recurrent neural networks also have a continuous-time ODE counterpart. The vanilla RNN $\mathbf{h}_t = F(\mathbf{W} \mathbf{h}_{t-1} + \mathbf{U} \mathbf{x}_t + \mathbf{b} )$ in Eq \ref{eq:rnn} is the Euler discretization of the following ODE:
\begin{equation}
    \frac{\partial \mathbf{h}}{\partial t}(t,\mathbf{x}) = F(\mathbf{W} \mathbf{h}(t) + \mathbf{U} \mathbf{x}(t) + \mathbf{b} ) - \mathbf{h}(t).
\end{equation}
We derive the associated ODE formulation for the LSTM \cite{Hochreiter:1997:LSM:1246443.1246450} and the Gated Recurrent Unit (GRU) \cite{cho2014learning} in Appendix \ref{part:part2}, which makes our ODE assumptions for forecasting (Eq \ref{eq:ode-relatedwork}) quite general.

Since, many other successful deep architectures have been linked to numerical schemes for ODEs \cite{lu2017beyond,fablet2018bilinear} and new architectures were proposed and analyzed with the rich dynamical system theory \cite{haber2017stable,ruthotto2020deep,qin2019data,chang2019antisymmetricrnns,bai2019deep}, \eg with the notions of stability or reversibility.


\begin{figure}
    \centering
    \includegraphics[width=10cm]{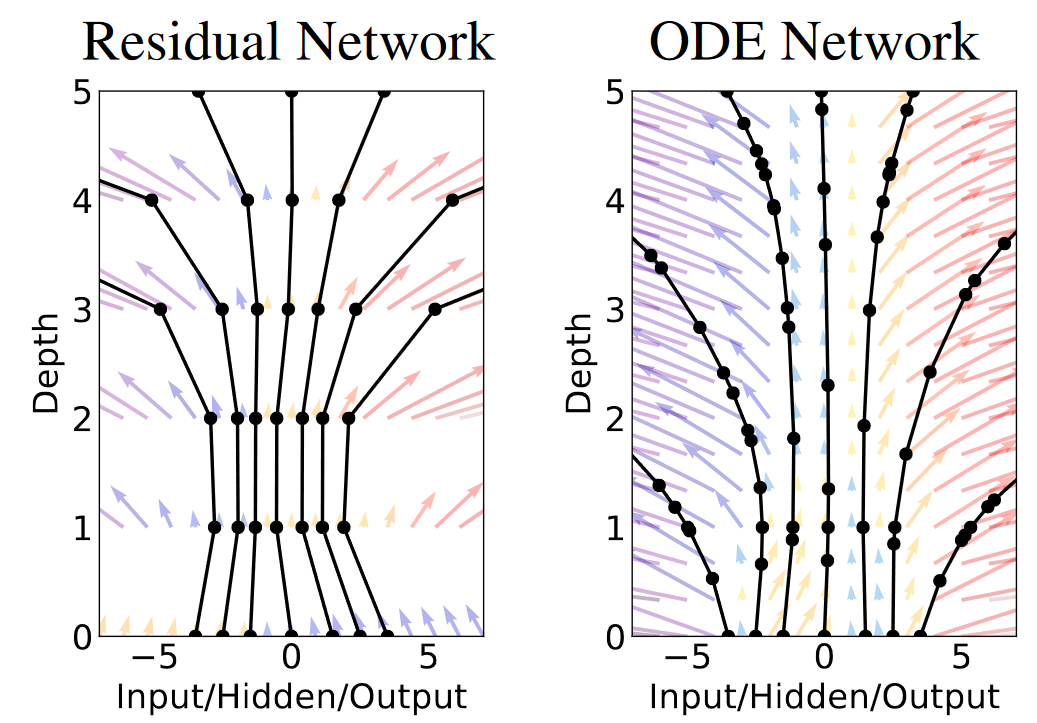}
\caption[Residual neural network vs Neural ODE.]{Left: a residual neural network \cite{he2016deep} defines a discrete sequence of layers from the input to the output. Right: a Neural ODE \cite{chen2018neural} solves an ODE starting from the input for evolving the hidden state. Figure taken from \cite{chen2018neural}.}
    \label{fig:node}
\end{figure}

 The Neural ODEs (or ODE networks) of Chen \etal \cite{chen2018neural} consider the continuous-time limit in residual networks. Instead of a discrete sequence of layers (or timesteps in a RNN), the evolution of the hidden state in the network is supposed to follow an ODE. This leads to a continuous transformation of the hidden state as shown in Figure \ref{fig:node}. Neural ODEs are trained with the adjoint sensitivity method \cite{pontryagin1987mathematical}, which consists in solving a backward ODE instead of backpropagating through the operations of the solver\footnote{This ensures a lower memory footprint for Neural ODEs: intermediate network activations do not need to be stored during the forward pass since they can be recomputed on the fly by solving the backward ODE.}. Many extensions and analyses of Neural ODEs were subsequently proposed \cite{dupont2019augmented,ayed2019learning,massaroli2020dissecting,jia2019neural,zhang2019anodev2,yildiz2019ode2vae} and have shown great successes in several tasks such as generative models with normalizing flows \cite{grathwohl2018ffjord} or modelling continuous-time data \cite{rubanova2019latent,hasani2021liquid}.

For predicting dynamical systems, the advantages of the continuous-time modelling of Neural ODEs are twofold. First, Neural ODEs can accommodate any ODE solver, in particular adaptive solvers that automatically adapt the number of iterations in function of the complexity of the dynamics to reach a given accuracy. Second, Neural ODEs can seamlessly handle irregularly-sampled temporal data, which arises in many applications (\eg medical records) or in case of missing data.

\begin{figure}
    \centering
    \includegraphics[width=16cm]{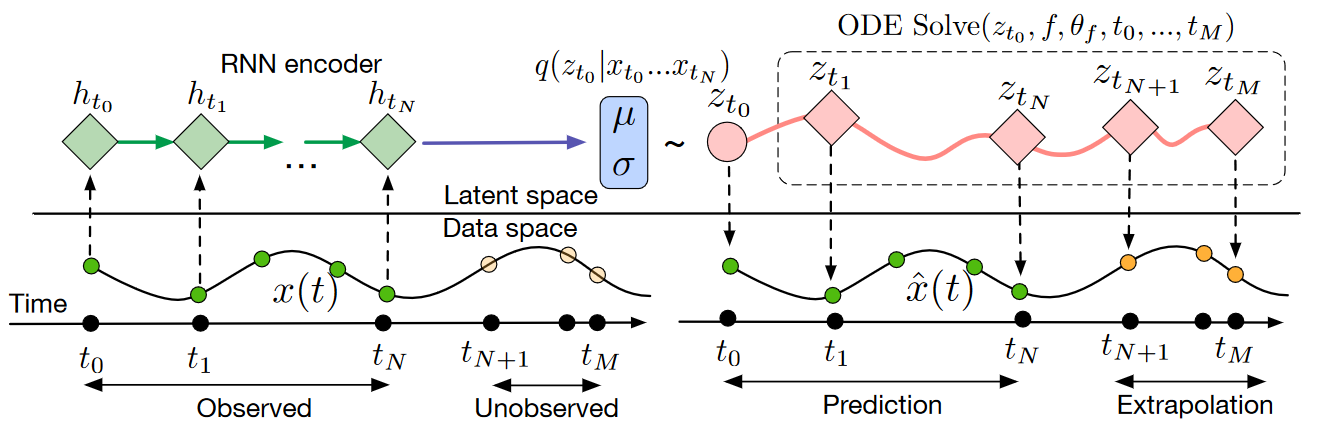}
    \caption[Modelling dynamical systems with Neural ODEs.]{Modelling dynamical systems with Neural ODEs. From an initial condition $z_{t_0}$ inferred by an encoder network, the latent trajectory is computed by solving the dynamical model $F_{\theta}$ (parameterized by a neural network) by a differentiable ODE solver. Figure taken from \cite{chen2018neural}.}
    \label{fig:node_ts}
\end{figure}

Neural ODEs provide a generative approach for modelling dynamical systems. As illustrated in Figure \ref{fig:node_ts}, time series are represented by a latent trajectory $z(t)$ governed by a dynamical function $F_{\theta}$ parameterized by a neural network: $\frac{\partial z(t)}{\partial t}= F_{\theta}(z(t))$. The latent trajectory is computed by solving the ODE with a differentiable ODE solver from an initial condition $z_{t_0}$ (which is known or estimated via an encoder network on an input trajectory). The solution can be evaluated for any time point in the observation range $[t_0,t_N]$ (interpolation) or in the future $[t_N; \infty[$ (extrapolation). The dynamical model $F_{\theta}$ is trained by reconstructing the trajectories of a training dataset.

Although Neural ODEs offer a principled way to model dynamical systems with deep networks in continuous-time, the dynamical model $F_{\theta}$ is still a pure data-driven component and suffers from the main drawbacks as pure ML methods, \ie overfitting in data-scarce contexts and lack of physical plausibility. In this thesis, we explore how to structure the function $F$ with prior physical knowledge.

\subsection{Physically-constrained machine learning}

\begin{figure}
    \centering
    \includegraphics[width=15cm]{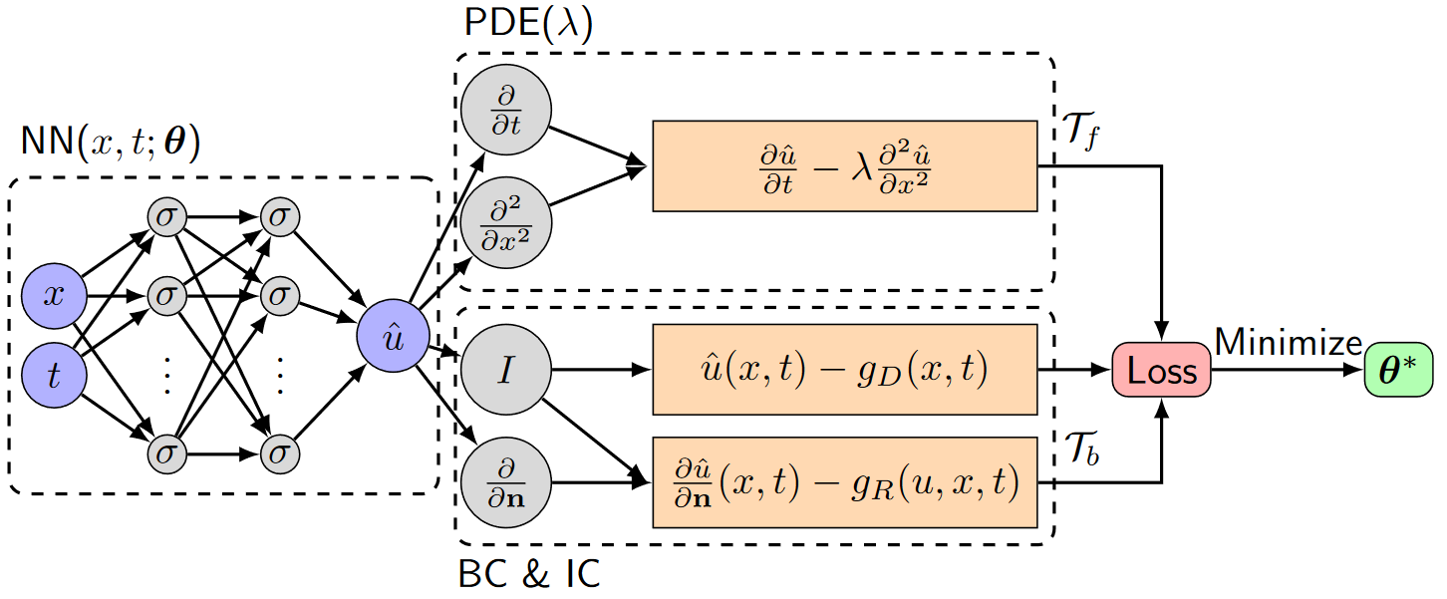}
    \caption[Physics-Informed Neural Networks (PINN)]{Physics-Informed Neural Networks (PINN) for solving the heat equation.}
    \label{fig:pinn}
\end{figure}

In recent years, many researchers have explored how to incorporate physical knowledge into ML models to regularize learning and improve performances. A first solution, made popular by the Physics-Informed Neural Networks (PINNs) of Raissi \etal \cite{Raissi2019}, is to add a physical regularization term in the loss function. Illustrated in Figure \ref{fig:pinn} for solving the heat equation, PINNs are composed of a neural network for predicting the solution $\hat{u}(x,t)$ at a given spatio-temporal location. Partial derivatives are computed during the forward pass by automatic differentiation to form the PDE residual. The total loss function is the sum of the data fidelity term and the adequacy to the PDE constraint and boundary conditions. PINNs are very easy to implement in standard deep learning libraries such as TensorFlow or PyTorch. 

In their initial form, PINNs need to be retrained for each new set of the parameters of the PDE. In order to learn a class of PDEs, Sirignano \etal \cite{sirignano2018dgm} propose to add the PDE parameters as inputs of the physics-informed neural network, and the neural operator approaches propose to directly learn the solution operator of a parametric class of PDEs \cite{li2020neural,lu2019deeponet,li2020fourier,wang2021learning}. However, this class of methods only impose soft constraints, \ie the physical laws are not strictly guaranteed to be respected. 

Other works investigate introducing hard physical constraints in the network architectures. Daw \etal \cite{daw2020physics} propose a monotonicity-preserving architecture for modelling lake temperature along depth, by adapting the LSTM with additional variables playing the role of positive increments. Mohan \etal \cite{mohan2020embedding} impose the divergence-free constraint of incompressible flows by parameterizing the flow as the curl of a learned scalar potential.

For modelling fluids, De Bezenac \etal \cite{de2017deep} propose a hybrid ML/MB architecture that explicitly exploits the advection-diffusion PDE:
\begin{equation}
    \frac{\partial I}{\partial t} + (w . \nabla) I = D \nabla^2 I.
    \label{eq:advection-diffusion}
\end{equation}
Given a sequence of past images, their deep architecture estimates the flow field $w$ and the diffusion coefficient $D$, which are used in a warping scheme implementing the closed-formed solution of the PDE. The model is learned end-to-end for predicting the next frame, without any supervision for the physical parameters. The authors successfully apply this model to predict Sea Surface Temperature (SST) maps. 

\begin{figure}
    \centering
    \includegraphics[width=12cm]{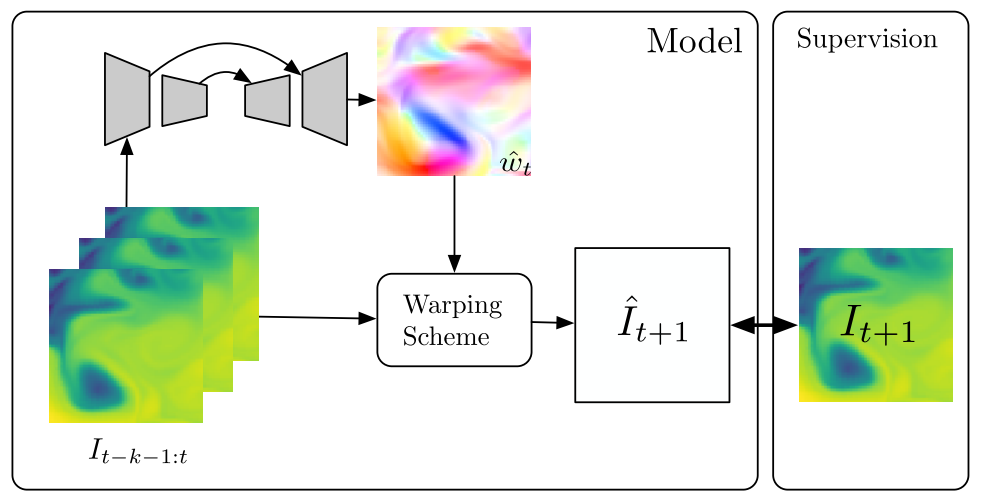}
    \caption{Hybrid ML/MB architecture of De Bezenac \etal \cite{de2017deep} for predicting Sea Surface Temperature with the advection-diffusion PDE.}
    \label{fig:my_label}
\end{figure}

Physical systems are often studied through the conservation of energy, which is encoded in a principled way through Hamiltonian dynamics. Greydanus \etal \cite{greydanus2019hamiltonian} introduce the Hamiltonian Neural Networks to learn physical systems respecting the conservation of energy. With $\mathbf{q}$ the position of a set of particles and $\mathbf{q}$ their momentum, the Hamiltonian $\mathcal{H}(q,p)$ representing the total energy of the systems, obeys the following equations:
\begin{equation}
    \dfrac{d\mathbf{q}}{dt} = \dfrac{\partial \mathcal{H}}{\partial \mathbf{p}}  \;\;\; , \;\;\;    \dfrac{d\mathbf{p}}{dt} = - \dfrac{\partial \mathcal{H}}{\partial \mathbf{q}}.
\end{equation}
HNNs learn the Hamiltonian with a NN and take in-graph gradients to impose the Hamiltonian dynamics. They show in experiments that it better conserves energy than baselines.

Many of the ML/MB approaches described so far are tailored for specific applications, \eg fluid dynamics \cite{de2017deep}, molecular dynamics \cite{chmiela2017machine}, quantum mechanics \cite{schutt2017quantum}, robotics \cite{lutter2019deep}, and are thus not applicable to other domains. Moreover, they often rely on a complete knowledge of the physical equations, and further assume that these equations directly apply in the input space (observed prior as defined in Chapter \ref{chap:intro}). In this thesis, we explore general augmentation strategies that can be applied to all levels of prior knowledge, from the more general prior to the most application-specific equations. We also tackle the case of the unobserved prior by learning representations spaces in which the physical laws apply.

\begin{figure}
    \centering
    \includegraphics[width=15cm]{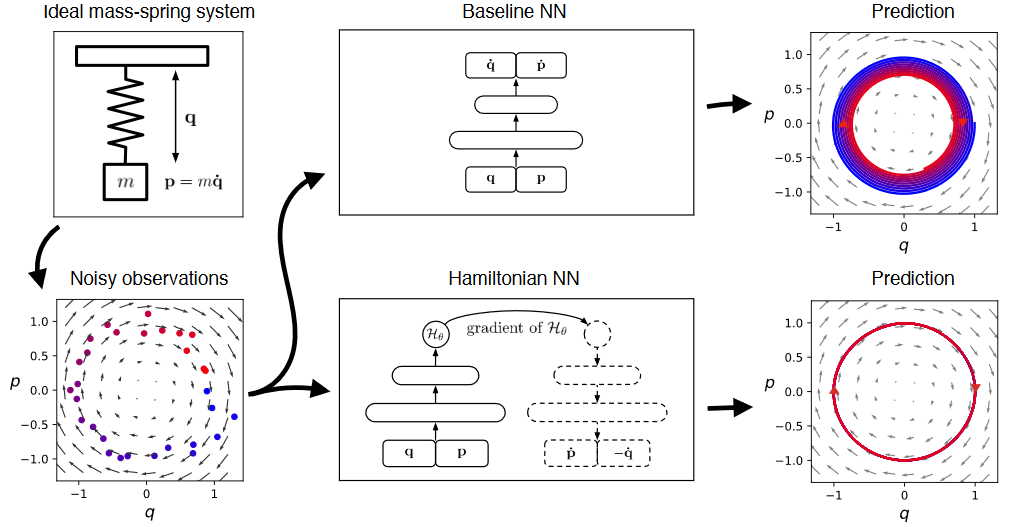}
    \caption{Hamiltonian Neural Networks of Greydanus \etal \cite{greydanus2019hamiltonian}.}
    \label{fig:hnn}
\end{figure}

\subsection{Identifying and discovering physical systems}

Beyond forecasting physical systems, researchers have also explored machine learning for system identification, which consists in estimating the unknown parameters in parameterized physical equations. A basic example is estimating the length of a damped pendulum from observed trajectories. Automatically identifying and discovering physical laws from observations is a long-standing goal for physicists, with many applications in control \cite{kidger2020neural} or robotics \cite{lutter2019deep}. Many approaches
use symbolic regression to search the space of possible mathematical functions, using evolutionary
algorithms \cite{schmidt2009distilling}, sparse regression on dictionaries of potential differentiable terms \cite{brunton2016discovering,rudy2017data,schaeffer2017learning}, or
graph neural networks \cite{cranmer2020discovering}.
 
\begin{figure}[H]
    \centering
    \includegraphics[width=16cm]{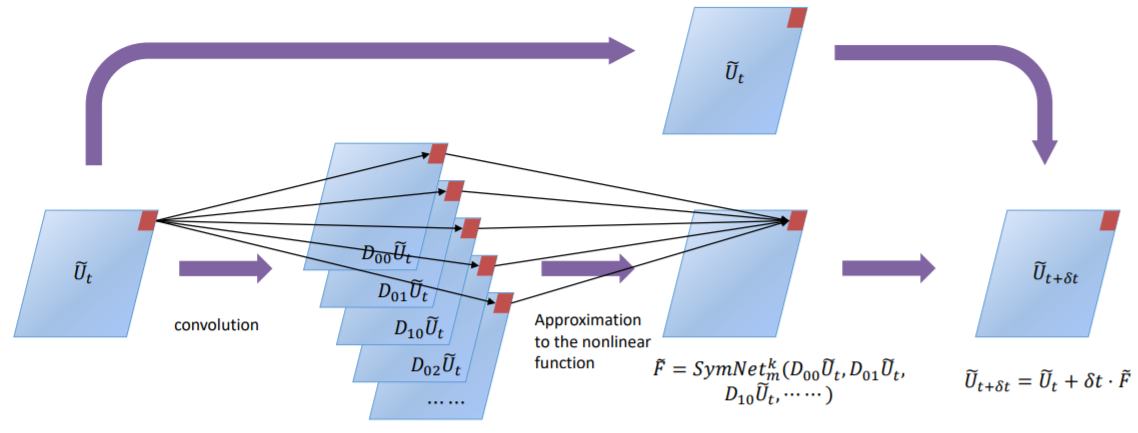}
    \caption[The PDE-Net architecture.]{The basis $\delta t$ block composing the PDE-Net architecture implements on step of forward Euler integration. Constrained convolutional filters estimate each spatial derivative term, that are combined by a symbolic network that estimates the dynamical function $F$. Finally a skip connection provides the solution for the next timestep. Figure taken from \cite{long2019pde}.}
    \label{fig:pdenet}
\end{figure}

Several architectures attempt to predict and identify the PDE governing physical systems \cite{long2018pde,raissi2017physics}, such as the the PDE-Net architecture of Long \etal \cite{long2018pde,long2019pde}. As shown in Figure \ref{fig:pdenet}, the basic bloc composing PDE-Net (the $\delta t$-bloc) is a residual module implementing one forward Euler discretization step. For solving the PDE $\frac{\partial u}{\partial t}=F(u,\frac{\partial u}{\partial x},\frac{\partial u}{\partial y},\cdots)$, the authors use convolutional filters that are constrained to approximate each spatial differential term (we give details about these constrained convolutions in Appendix \ref{app:moment-matrix}\footnote{They show that the flexibility of learned differential filters boost performances compared to handcrafted filters, an observation that has been noted for other discretization schemes learned from data.}. Then a symbolic neural network identifies the nonlinear relationships between the spatial derivatives to form the nonlinear function $F$ of the PDE. A skip connection finally provides the prediction of the next timestep $\hat{u}(t+\delta t)=\hat{u}(t) + \delta t \hat{F}$. The complete PDE-Net architecture is composed of several $\delta t$-blocs concatenated in time for long-term prediction.

In this thesis, we take inspiration from the PDE-Net architecture for imposing physical dynamics, and we take a step further by assuming incomplete physical models and by modelling the residual dynamics for accurate prediction. We also show that a careful training scheme leads to a better identification of the physical parameters than  simplified physical model alone.

\subsection{Augmented physical models}

There exists an abundant literature on statistical methods for calibrating and predicting physical systems in presence of model inadequacy, often expressed in a Bayesian framework; a review of these methods can be found in \cite{pernot2017critical}. In data assimilation techniques, like the Kalman filter \cite{kalman1960new}, the particle filter \cite{perez2004data} or 4D-var \cite{courtier1994strategy}, the predictions errors are modelled probabilistically with random variables reflecting the noise assumption. A correction step using observed data is performed after each prediction step for filtering the noise. Similar residual correction procedures are commonly used in robotics and optimal control \cite{chen2004disturbance,li2014disturbance}. However, these sequential (two-stage) procedures prevent the cooperation between prediction and correction.  Besides, in model-based reinforcement learning, model deficiencies are typically handled by considering only short-term rollouts \cite{janner2019trust} or by model predictive control \cite{nagabandi2018neural} consisting in replanning frequently to mitigate error propagation. 

In this thesis, we take inspiration from data assimilation ideas to augment incomplete physical models with residual terms. However, in contrast to data assimilation, our residual terms are not assumed to correspond to be a stochastic residual, \ie noise, but to a systematic unmodelled part of the dynamics that we learned from data. Moreover, we derive a principled training scheme for making the prediction and correction steps cooperate.

The idea of augmenting physical models with neural networks (\textit{gray-box or \textit{hybrid}} modelling) is not new: in the 1990's, the works \cite{psichogios1992hybrid,thompson1994modeling,rico1994continuous} use neural networks to estimate the unknown parameters of physical models that are difficult to model from first principles, and a classification of the possible augmentation strategies (serial, parallel, modular) was dressed \cite{thompson1994modeling}. The challenge of proper ML/MB cooperation was already raised as a limitation of gray-box approaches but not addressed. Moreover these methods were evaluated on specific applications with a residual targeted to the form of the equation.

In the last few years, there has been a growing interest in deep augmented models that combine physical priors with deep networks \cite{long2018hybridnet,saha2020phicnet,neural20}. Several ML/MB cooperation schemes with deep networks were studied in \cite{wang2019integrating,neural20}. Again, these approaches do not address the issues of uniqueness of the decomposition or of proper cooperation for correct parameter identification. They are also mostly dedicated to the fully-observable case, whereas we also tackle the non-observable prior context in this thesis. We further detail the literature on augmented physical models in Chapter \ref{chap:aphynity}.

\clearpage{\pagestyle{empty}\cleardoublepage}

\partabstract{
\vspace{1cm}
\begin{center}
    \textsc{Abstract}\\
\end{center}
\vspace{1cm}
In this part, we tackle the multistep deep time series forecasting problem, in the challenging context of non-stationary series that can present sharp variations. In deep learning, the mainstream research direction concerns developing new neural forecasting architectures. In contrast, the choice of the training loss function is rarely questioned: the surrogate mean squared error (MSE) is used in the vast majority of cases. We propose here to leverage shape and temporal criteria in the training objective. We introduce differentiable similarities and dissimilarities for characterizing shape accuracy  and temporal localization error (Chapter \ref{chap:criteria}). We leverage these criteria by introducing two approaches dedicated to deterministic and probabilistic forecasting: we introduce the DILATE loss function for deterministic forecasting that ensures both sharp predictions with accurate temporal localization (Chapter \ref{chap:dilate}), and the STRIPE model for probabilistic forecasting with shape and temporal diversity (Chapter \ref{chap:stripe}). We validate our claims with extensive experiments on synthetic and real-world datasets.
}
\part{Differentiable shape and time criteria for deterministic and probabilistic forecasting}
\label{part:part1}

\chapter{Differentiable shape and temporal criteria}
\label{chap:criteria}
\minitoc

\chapabstract{
\begin{center}
   \textsc{Chapter abstract}
\end{center}
\textit{
In this Chapter, we highlight the limitations of the Mean Squared Error (MSE) loss function dominantly used for time series forecasting. As an alternative, we propose to leverage shape and temporal features at training time. We introduce differentiable similarities and dissimilarities for characterizing shape accuracy and temporal localization error. We characterize the shape with the Dynamic Time Warping (DTW) \cite{sakoe1990dynamic} algorithm and the temporal error with the Temporal Distortion Index (TDI) \cite{frias2017assessing}. We provide an unified view of these criteria by formulating them in terms of dissimilarities (loss functions) and similarities (positive semi-definite kernels). We also insist on their differentiability and efficient computation. The work described in this Chapter is based on the following publication:
\begin{itemize}
    \item Vincent Le Guen and Nicolas Thome. "Deep Time Series Forecasting with Shape and Temporal Criteria". IEEE Transactions on Pattern Analysis and Machine Intelligence, 2022. 
\end{itemize} 
}
}

\section{Introduction}\label{sec:introduction}

\begin{figure}[H]
\begin{tabular}{ccc}

\includegraphics[height=4.6cm]{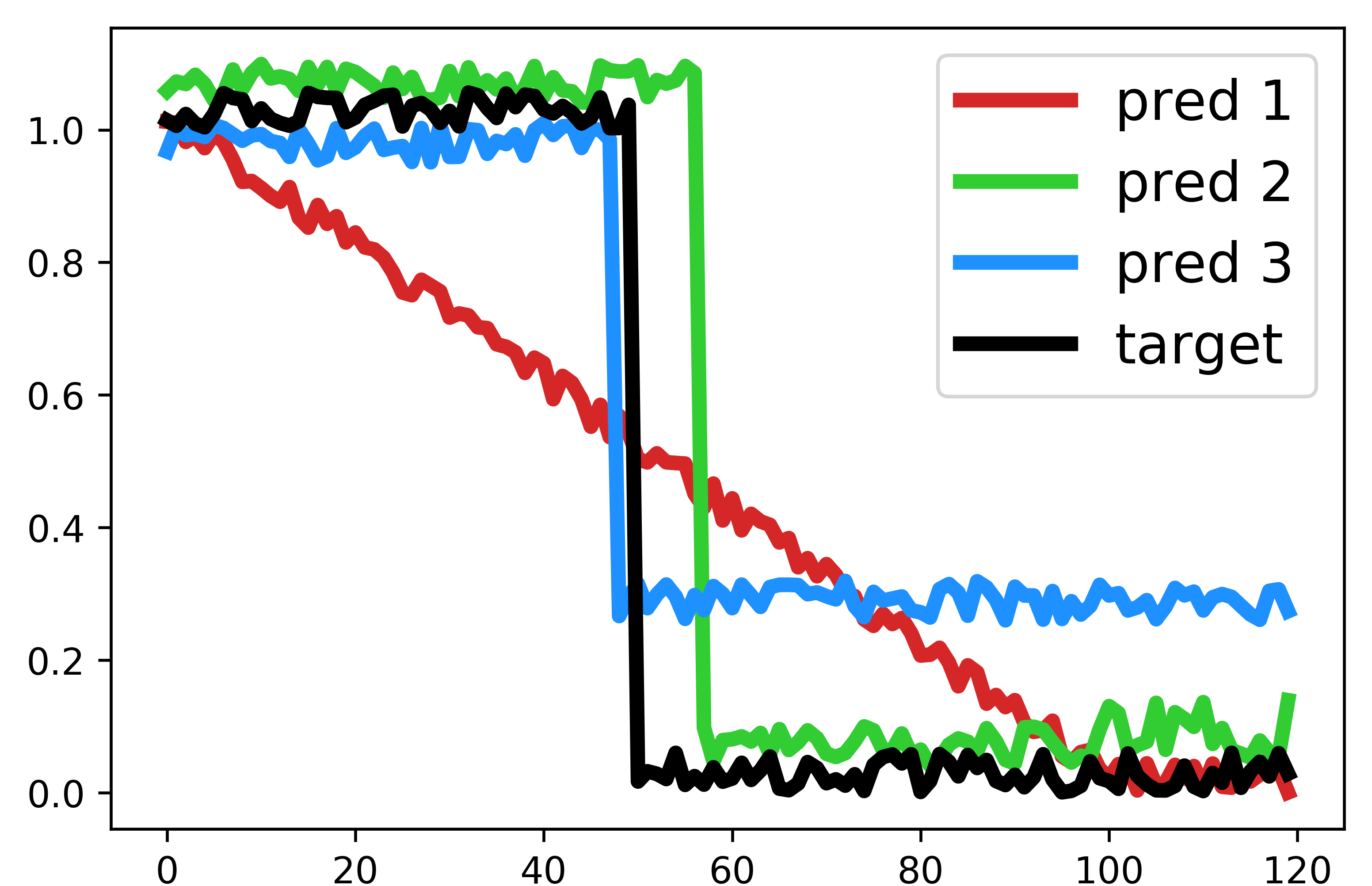}   & \hspace{-0.3cm} 
\includegraphics[height=4.6cm]{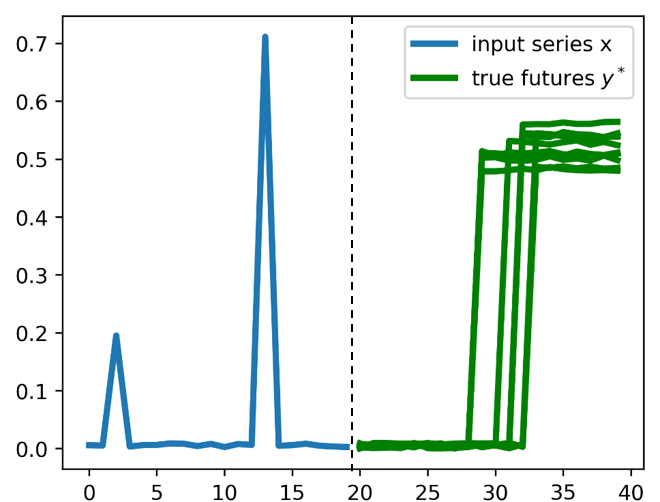}   & 
\hspace{-0.5cm} 
\includegraphics[height=4.6cm]{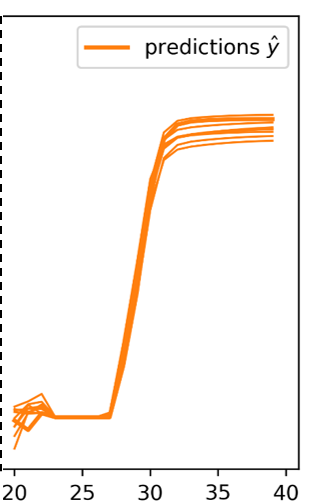}   \\
~ &   \footnotesize{True predictive distribution}    & \hspace{-0.5cm}  \footnotesize{deep stoch. model \cite{yuan2019diverse}}   \\
 \textbf{(a) Deterministic forecasting} & ~ &  \hspace{-5cm} \textbf{(b) Probabilistic forecasting}  \\
\end{tabular}{}

    \caption[MSE limitations in deterministic and probabilistic forecasting.]{\textbf{MSE limitations in deterministic and probabilistic forecasting.} (a) For deterministic forecasting, the three predictions (1,2,3) have the same MSE with respect to the target (in black). However, one would like to favour prediction 2 (correct shape, slight delay) and 3 (correct timing, inaccurate amplitude) over prediction 1 (which is not very informative).
    (b) For probabilistic forecasting, state-of-the-art methods trained with variants of the MSE (\eg \cite{yuan2019diverse,rasul2020multi}) loose the ability to produce sharp forecasts (in orange) compared to the ground truth future trajectories (in green).}
    \label{fig-intro}
\end{figure}

\lettrine[lines=3]{T}ime series forecasting consists in analyzing historical signal correlations to anticipate future behaviour. As discussed in Chapter \ref{chap:related_work}, traditional approaches include linear autoregressive methods \cite{box2015time} or state space models \cite{durbin2012time}, which are simple yet mathematically grounded and benefit from interpretability. They often exploit prior knowledge based on stationarity, \eg by leveraging trend or seasonality to constrain forecasting. 

These grounding assumptions are often violated in many real-world time series that are non-stationary and can present sharp variations such as sudden drops or changes of regime. Long-term multi-step forecasting in this context is particularly challenging and arises in a wide range of important application fields, \eg analyzing traffic flows \cite{li2017diffusion,snyder2019streets}, medical records \cite{chauhan2015anomaly}, predicting sharp variations in financial markets \cite{ding2015deep} or in renewable energy production \cite{vallance2017towards,ghaderi2017deep,leguen-fisheye}, \textit{etc}.

We are interested in forecasting multi-step future trajectories with potentially sharp variations in the deterministic and probabilistic cases. Deep neural networks are an appealing solution for this problem \cite{yu17learning,qin2017dual,lai2018modeling,salinas2017deepar,oreshkin2019n,zhou2020informer}, due to their automatic feature extraction and complex nonlinear time dependencies modelling. However, the verification criteria typically used in applications are not used at training time because they are mostly not differentiable. We may cite for instance the ramp score \cite{vallance2017towards} for assessing the detection of sharp ramping events, or the Time Distortion Index (TDI) \cite{frias2017assessing} for assessing the time delay of a particular predicted event.  

Instead, the huge majority of methods optimize at training time the Mean Squared Error (MSE) or its variants (MAE, quantile loss, \textit{etc}) as a proxy loss function. However, the MSE has important drawbacks in our non-stationary context, as also noted by several other works \cite{vallance2017towards,verbois2020beyond,yang2020verification}. This is illustrated in Figure \ref{fig-intro}. Figure \ref{fig-intro} (a) shows three deterministic predictions, which have the same MSE loss compared to the target step function (in black). Thus, the MSE does not support predictions (2) and (3) over prediction (1), although they clearly are more adequate for regulation purposes because they do anticipate the drop to come, although with a slight delay (2) or with a slightly inaccurate amplitude (3). For probabilistic forecasting (Figure \ref{fig-intro} (b)), current state-of-the art probabilistic methods trained with variants of the MSE tend to produce blurry predictions that do not match the sharp steps of the true futures (in green).

We intend to bridge this train/test criterion gap by incorporating shape and temporal features at training time. In this Chapter, we introduce shape and temporal criteria for training deep forecasting models. We characterize the shape of times series with the Dynamic Time Warping (DTW) \cite{sakoe1990dynamic} algorithm and the temporal shift with the Temporal Distortion Index (TDI) \cite{frias2017assessing}. We provide an unified view of these criteria by formulating them both as dissimilarities (loss functions) and similarities (positive semi-definite kernels). Importantly, we insist on their differentiability, which makes them amenable to gradient-based optimization, and on their efficient computation.

\section{Shape (dis)similarity}

\subsection{Background: Dynamic Time Warping}

To assess the shape similarity between two time series, the popular Dynamic Time Warping (DTW) method \cite{sakoe1990dynamic} seeks a minimal cost alignment for handling time distortions. 
Given two $d$-dimensional time series $\y \in \mathbb{R}^{d \times n}$ and $\z  \in \mathbb{R}^{d \times m}$ of lengths $n$ and $m$, DTW looks for an optimal warping path represented by a binary matrix $\mathbf{A}  \subset \left \{  0,1 \right \}  ^{n \times m}$ where $\mathbf{A}_{ij}=1$ if $\y_i$ is associated to $\z_j$ and 0 otherwise. The set of admissible warping paths $\mathcal{A}_{n,m}$ is composed of paths connecting the endpoints $(1,1)$ to $(n,m)$ with the following authorized moves $\rightarrow, \downarrow, \searrow$. The cost of warping path $\mathbf{A}$ is the sum of the costs along the alignment ; this cost can be written as the scalar product  $\left\langle \mathbf{A}, \mathbf{\Delta}(\y,\z) \right\rangle$, where  $\Delta(\y,\z)$  is a $n \times m$ pairwise dissimilarity matrix whose general term is typically chosen an the Euclidean distance $\mathbf{\Delta}(\y,\z)_{ij} = \Vert \y_i-\z_j \Vert^2_{2}$. DTW computes the minimal cost warping path:
\begin{equation}
\text{DTW}^{\mathbf{\Delta}}(\y, \z) :=\underset{\mathbf{A} \in \mathcal{A}_{n,m}}{\min} \left \langle \mathbf{A},\mathbf{\Delta}(\y, \z) \right \rangle.
\label{eq:dtw}
\end{equation}

\begin{figure}
    \centering
    \includegraphics[width=10cm]{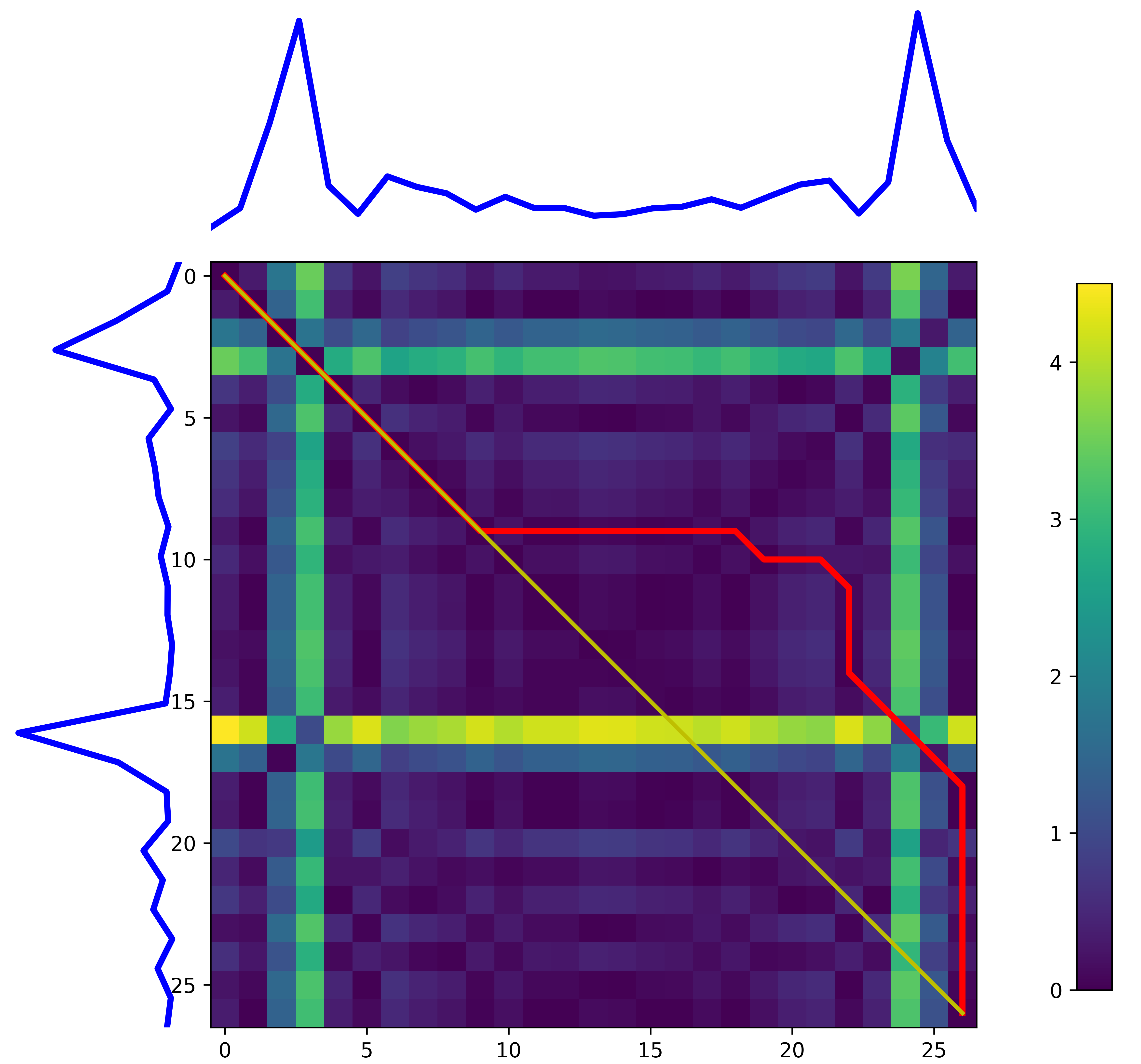}
    \caption[Principle of Dynamic Time Warping (DTW)]{\textbf{Dynamic Time Warping (DTW)} seeks a path of minimal alignment cost (in red) in the pairwise cost matrix between the two time series.}
    \label{fig:dtw}
\end{figure}

Although the cardinality of $\mathcal{A}_{n,m}$ increases exponentially in $\min(n,m)$ \footnote{ $|\mathcal{A}_{n,m}|$ is equal to the Delannoy number $Delannoy(n,m)$ which grows exponentially in $\min(n,m)$}, DTW and the optimal path $\mathbf{A^*}$ can be computed efficiently in $\mathcal{O}(nm)$ by dynamic programming. However, a major limitation of DTW is its non-diffentiability, which prevents its integration in neural network pipelines trained with gradient-based optimization.

\subsubsection{Smooth DTW shape dissimilarity}
\label{sec:soft-dtw}
 
 For handling the non-differentiability of DTW, Cuturi and Blondel \cite{cuturi2017soft} introduced the soft-DTW by replacing the hard-minimum operator by a smooth minimum with the log-sum-exp trick $\min_{\gamma}(a_1,...,a_n) = - \gamma \log(\sum_i^n \exp(-a_i / \gamma) )$:
  \begin{equation}
 \text{DTW}^{\mathbf{\Delta}}_{\gamma}(\y, \z)  :=
 - \gamma  \log \left ( \sum_{\mathbf{A} \in \mathcal{A}_{n,m}} e ^ { - \left \langle \mathbf{A},\mathbf{\Delta}(\y, \z) \right \rangle / \gamma} \right ),
\label{eq:dtwgamma}
 \end{equation}
  where $\gamma > 0$ is a smoothing parameter (when $\gamma \rightarrow 0$, this converges to the true DTW).

$\text{DTW}^{\mathbf{\Delta}}_{\gamma}$ as defined in Eq \ref{eq:dtwgamma} is differentiable with respect to $\mathbf{\Delta}$ (and with respect to both series $\y$ and $\z$ by chain's rule, provided a differentiable cost function $\mathbf{\Delta}$).

We can interpret this relaxed DTW version by considering, instead of the unique optimal path $\mathbf{A}^*$, a Gibbs distribution over possible paths:
\begin{equation}
     p_{\gamma}(\mathbf{A} ; \mathbf{\Delta}) = \frac{1}{Z} \:  e^{- \left \langle \mathbf{A},\mathbf{\Delta}(\y, \z) \right \rangle / \gamma }.
     \label{eq:gibbs}
 \end{equation}
 The soft-DTW is then the negative log-partition of this distribution: $\text{DTW}^{\mathbf{\Delta}}_{\gamma}(\y, \z)  :=  - \gamma \log Z $.

Since $\text{DTW}^{\mathbf{\Delta}}_{\gamma}(\y,\z)$ can take negative values and is not minimized for $\y=\z$, Mensch and Blondel \cite{mensch2018differentiable} normalized the soft-DTW to make it a true divergence. We found experimentally that this does not improve performances and is heavier computationally (see Appendix \ref{app:dilate-div}).

\subsubsection{Shape similarity kernel}
\label{sec:shape-kernel}

Based on the soft-DTW shape dissimilarity defined in Eq \ref{eq:dtwgamma}, we define a shape similarity kernel as follows:
\begin{equation}
    \mathcal{K}_{shape}(\y,\z) = e^{-  \: \text{DTW}^{\mathbf{\Delta}}_{\gamma}(\y,\z) / \gamma}.
    \label{eq:kshape}
\end{equation}
We experiment with the following choices of kernels $\Delta_{ij} = \Delta(\y,\z)_{ij}$:
\begin{itemize}
    \item  Half-Gaussian: $\mathbf{\Delta}_{ij}=  \Vert \y_i-\z_j \Vert^2_2 +  \log (2 - e^{- \Vert \y_i-\z_j \Vert^2_2 })$
    \item L1: $\mathbf{\Delta}_{ij}= |\y_i-\z_j|$ ~~ (for $d=1$)
    \item Euclidean:  $\mathbf{\Delta}_{ij}=  \Vert \y_i-\z_j \Vert^2_2$.
\end{itemize}
$\mathcal{K}_{shape}$ was proven to be positive semi-definite (PSD) for the half-Gaussian\footnote{We denote this kernel "half-Gaussian" since the corresponding $k$ kernel defined in the proof (Appendix \ref{app:proof-ktime}) equals $k(\y_i,\z_j) = e^{- \Delta(\y_i,\z_j)} = \left(\frac{1}{2} e^{-\Vert \y_i-\z_j \Vert^2})\right) \times \left(1 -  \frac{1}{2} e^{-\Vert \y_i-\z_j \Vert^2}\right)^{-1}$} and the L1 kernels \cite{cuturi2007kernel,blondel2020differentiable} and is conjectured to be PSD for the Euclidean kernel \cite{blondel2020differentiable}. Experimentally we observed that these three cost matrices lead to similar behaviour.

\section{Temporal (dis)similarity}

Quantifying the temporal similarity between two time series consists in analyzing the time delays between matched patterns detected in both series. As discussed in introduction, it of great importance for many applications to anticipate sharp variations. 

\subsection{Smooth temporal distortion index}

A common temporal similarity is the Temporal Distortion Index (TDI) \cite{frias2017assessing, vallance2017towards}. The TDI computes the approximate area included between the optimal path $\mathbf{A^*}$ and the first diagonal, characterizing the presence of temporal distortion. A generalized version of the TDI, that we proposed in \cite{leguen19}, can be written:
\begin{equation}
 \text{TDI}^{\mathbf{\Delta, \Omega_{dissim}}}(\y, \z) :=   \langle  \mathbf{A}^*, \mathbf{\Omega_{dissim}}  \rangle \>  ,
 \label{eq:tdi}
\end{equation}
where $ \mathbf{A}^* =  \underset{\mathbf{A} \in \mathcal{A}_{n,m}}{\arg \min} \left \langle \mathbf{A},\mathbf{\Delta}(\y, \z) \right \rangle$ is the DTW optimal path and  $\mathbf{\Omega_{dissim}} \in \mathbb{R}^{n \times m}$ is a matrix penalizing the association between $\y_i$ and $\z_{j}$ for $i \neq j$. We typically choose a quadratic penalization  $\mathbf{\Omega_{dissim}}(i,j) \propto (i-j)^2$, but other variants can encode prior knowledge and penalize more heavily late than early predictions, and \textit{vice-versa}.

The TDI dissimilarity defined in Eq \ref{eq:tdi} is however non-differentiable, since the optimal path $\mathbf{A}^*$ is not differentiable with respect to $\mathbf{\Delta}$. We handle this problem
by defining a relaxed optimal path $\mathbf{A}^*_{\gamma}$ as the gradient of $\text{DTW}_{\gamma}^{\mathbf{\Delta}}$: 
\begin{align}
    \mathbf{A}^*_{\gamma}  := \nabla_{\mathbf{\Delta}} \text{DTW}^{\mathbf{\Delta}}_{\gamma}(\y, \z) =  \frac{1}{Z}  \sum_{\mathbf{A} \in \mathcal{A}_{n,m}} \mathbf{A} \:  e^{- \left \langle \mathbf{A},\mathbf{\Delta}(\y, \z) \right \rangle / \gamma }.
 \label{eq:grad_dtw}
\end{align}
The expression in Eq \ref{eq:grad_dtw} results from a direct computation from Eq. \ref{eq:dtwgamma}. Notice that this soft optimal path corresponds to the expected path $\mathbf{A}^*_{\gamma} = \mathbb{E}_{p_{\gamma}(\cdot ; \mathbf{\Delta})} [\mathbf{A}]$ under the Gibbs distribution in Eq \ref{eq:gibbs}. Note also that $\mathbf{A}^*_{\gamma}$ becomes a soft assignment, \ie $\mathbf{A}^*_{\gamma}(i,j)$ represents the probability for a path to contain the cell $(i,j)$. An illustration of soft optimal paths with the influence of $\gamma$ is given in Figure \ref{fig:dilate_analysis}.

We can now define a differentiable version of the TDI:
\begin{equation}
    \text{TDI}^{\mathbf{\Delta,\Omega_{dissim}}}_{\gamma}(\y,\z) := \left \langle  \mathbf{A}_{\gamma}^* , \mathbf{\Omega_{dissim}}  \right \rangle  = \\
     \dfrac{1}{Z}  \sum_{\mathbf{A} \in \mathcal{A}_{n,m}}  \left \langle  \mathbf{A}, \mathbf{\Omega_{dissim}} \right \rangle  e^{-\frac{ \left \langle \mathbf{A},\mathbf{\Delta}(\y, \z) \right \rangle}{\gamma} }, 
     \label{eq:temporal}
\end{equation}
which corresponds to the expected value of the TDI under the Gibbs distribution.

\subsection{Temporal similarity kernel}

Based on the temporal dissimilarity in Eq \ref{eq:temporal} and the shape similarity kernel in Eq. \ref{eq:kshape}, we can define a time similarity as follows:
\begin{equation}
    \mathcal{K}_{time}(\y,\z) := e^{- \text{DTW}^{\mathbf{\Delta}}_{\gamma}(\y,\z) / \gamma}
  \times \text{TDI}^{\mathbf{\Delta, {\Omega_{sim}}}}_{\gamma} (\y,\z),
  \label{eq:Ktime}
\end{equation}
where in this case, we use a similarity matrix $\mathbf{\Omega_{sim}}$ favoring pairs of time series with low temporal distortion, \ie with an optimal path near the main diagonal. We typically choose a pointwise inverse of $\mathbf{\Omega_{dissim}}$: $\mathbf{\Omega_{sim}}(i,j) = \frac{1}{(i-j)^2+1}$. We prove that $ \mathcal{K}_{time}$ defines a valid PSD temporal kernel (proof in Appendix \ref{app:proof-ktime}). \\

The following table provides an overview of the shape and temporal criteria introduced in this work:\\
\begin{center}
 \begin{tabular}{c|c|c}
  criterion    & differentiable loss  &  PSD similarity kernel  \\ 
  \hline
    shape & $\text{DTW}^{\mathbf{\Delta}}_{\gamma}(\y, \z)$   &  $ e^{-  \: \text{DTW}^{\mathbf{\Delta}}_{\gamma}(\y,\z) / \gamma}$  \\ 
    \hline
    time &   $\text{TDI}^{\mathbf{\Delta,\Omega_{dissim}}}_{\gamma}(\y,\z)$  & 
   
       $e^{-  \text{DTW}^{\mathbf{\Delta}}_{\gamma}(\y,\z) / \gamma  } 
             \times  \text{TDI}^{\mathbf{\Delta, {\Omega_{sim}}}}_{\gamma} (\y,\z)$
\end{tabular}   
\end{center}

\subsection{Efficient forward and backward computation} 
\label{app:efficient-computation}

The direct computation of the shape loss $\text{DTW}^{\mathbf{\Delta}}_{\gamma}$ (Eq \ref{eq:dtwgamma}) and the temporal loss $\text{TDI}^{\mathbf{\Delta,\Omega_{dissim}}}_{\gamma}$ (Eq \ref{eq:temporal}) is intractable, due to the exponential growth of cardinal of $\mathcal{A}_{n,m}$. We provide a careful implementation of the forward and backward passes in order to make learning efficient.\\

\paragraph*{Shape loss:} Regarding $\text{DTW}^{\mathbf{\Delta}}_{\gamma}$, we rely on~\cite{cuturi2017soft} to efficiently compute the forward pass with a variant of the Bellmann dynamic programming approach~\cite{bellman1952theory}. For the backward pass, we implement the recursion proposed in~\cite{cuturi2017soft} in a custom Pytorch loss. This implementation is much more efficient than relying on vanilla auto-differentiation, since it reuses intermediate results from the forward pass.

\paragraph*{Temporal loss:} For $\text{TDI}^{\mathbf{\Delta},\mathbf{\Omega_{dissim}}}_{\gamma}$, note that the bottleneck for the forward pass in Eq \ref{eq:temporal} is to compute $\mathbf{A}^*_{\gamma} = \nabla_{\Delta} \text{DTW}^{\mathbf{\Delta}}_{\gamma}(\y,\z)$, which we implement as explained for the  $\text{DTW}^{\mathbf{\Delta}}_{\gamma}$ backward pass. Regarding $\text{TDI}^{\mathbf{\Delta},\mathbf{\Omega_{dissim}}}_{\gamma}$  backward pass, we need to compute the Hessian $\nabla^2 \text{DTW}^{\mathbf{\Delta}}_{\gamma}(\y,\z)$. We use the method proposed in~\cite{mensch2018differentiable}, based on a dynamic programming implementation that we embed in a custom Pytorch loss. Again, our back-prop implementation allows a significant speed-up compared to standard auto-differentiation. 
The resulting time complexity of both shape and temporal losses for forward and backward is $\mathcal{O}(nm)$. 

\begin{figure}
    \centering
    \includegraphics[width=8cm]{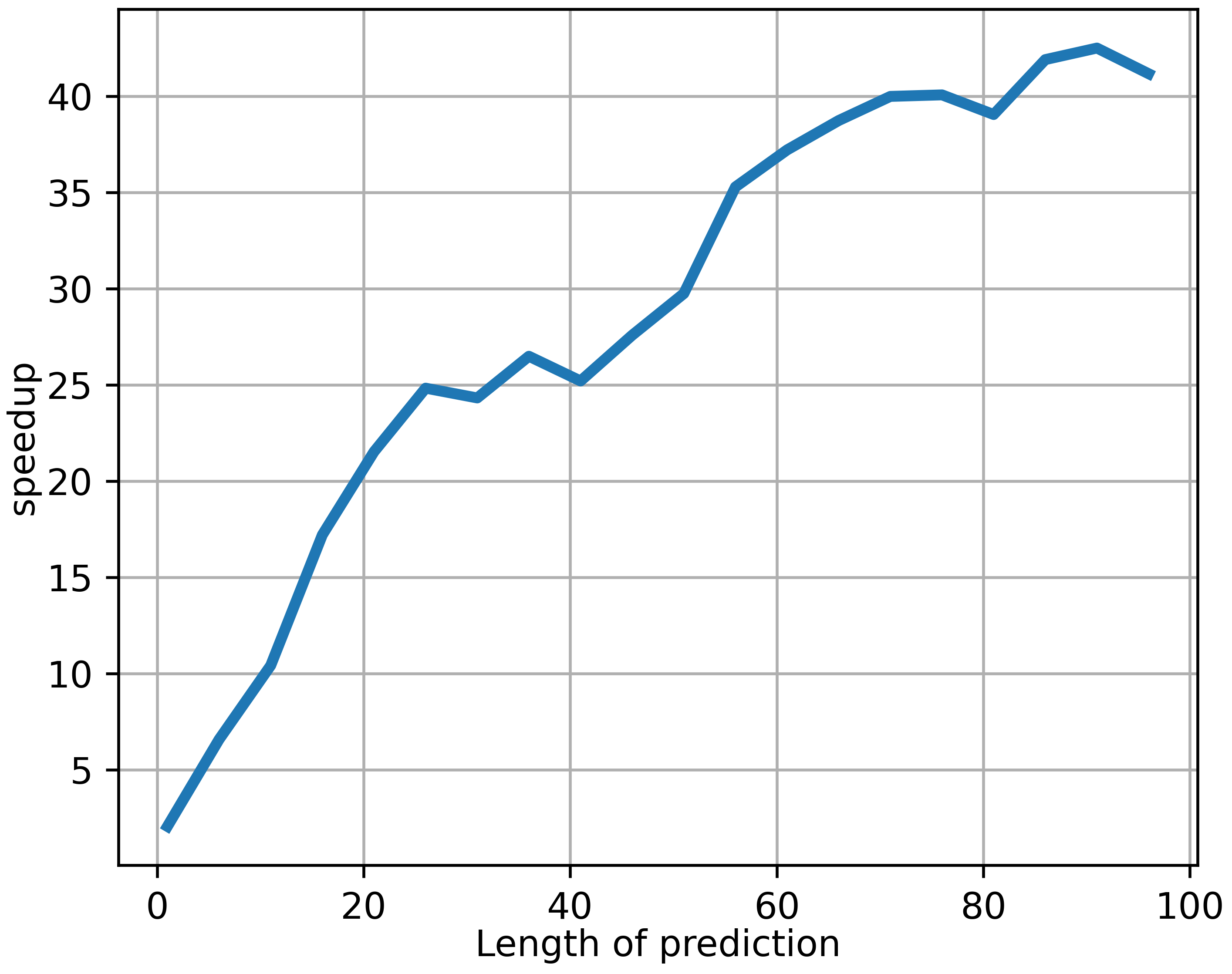}
    \caption{Speedup of the custom forward and backward implementation of the  DILATE loss introduced in Chapter \ref{chap:dilate}.}
    \label{fig:speedup}
\end{figure}

\paragraph*{Custom backward implementation speedup:} We compare in Figure \ref{fig:speedup} the computational time between the standard PyTorch auto-differentiation mechanism and our custom backward pass implementation for calculating $\text{DTW}^{\mathbf{\Delta}}_{\gamma}+\text{TDI}^{\mathbf{\Delta},\mathbf{\Omega_{dissim}}}_{\gamma}$ (we will call this quantity the DILATE loss in the next Chapter). We plot the speedup of our implementation with respect to the prediction length $H$ (averaged over 10 random target/prediction tuples). We notice the increasing speedup with respect to $H$: speedup of $\times$ 20 for 20 steps ahead and up to $\times$ 35 for 100 steps ahead predictions.

\section{Conclusion}

To tackle the multi-step and stationary time series forecasting problem, we question the widely-used MSE training loss that lead to non-sharp predictions. We instead propose to leverage shape and temporal features at training time. In this Chapter, we have introduced differentiable similarities and dissimilarities for characterizing shape accuracy and temporal localization error. Shape is characterized with the Dynamic Time Warping (DTW) \cite{sakoe1990dynamic} algorithm and the temporal error with the Temporal Distortion Index (TDI) \cite{frias2017assessing}. We have provided an unified view of these criteria by formulating them in terms of dissimilarities (loss functions) and similarities (positive semi-definite kernels). We have insisted on their differentiability and efficient computation.  

In subsequent Chapters, we provide two implementations for time series forecasting: the DILATE loss function for deterministic forecasting that ensures both sharp predictions with accurate temporal localization (Chapter \ref{chap:dilate}), and the STRIPE model for probabilistic forecasting with shape and temporal diversity (Chapter \ref{chap:stripe}).

\clearpage{\pagestyle{empty}\cleardoublepage}

\mbox{}
\thispagestyle{empty}
\chapter{Distortion loss with shape and time}
\label{chap:dilate}
\minitoc

\chapabstract{
\begin{center}
   \textsc{Chapter abstract}
\end{center}
\textit{
In this Chapter, we propose a new differentiable loss function, called DILATE, for training deep multi-step time series forecasting models, in a deterministic context. The DILATE loss builds on the shape and temporal dissimilarities introduced in the previous Chapter. DILATE combines two terms for precise shape and temporal localization of non-stationary signals with sudden changes. The DILATE loss is differentiable, enabling to train any deep forecasting model with gradient-based optimization. We also introduce a variant of DILATE, which provides
a smooth generalization of temporally-constrained Dynamic Time Warping (DTW). Extensive experiments on synthetic and real-world datasets show that DILATE is equivalent to the standard MSE loss when evaluated on MSE, and much better when evaluated on several shape and timing metrics. Besides, DILATE improves the performances of state-of-the-art forecasting algorithms trained with the MSE. The work described in this Chapter is based on the following publications:
\begin{itemize}
    \item Vincent Le Guen and Nicolas Thome. "Shape and Time Distortion Loss for Training Deep Time Series Forecasting Models". In Advances in Neural Information Processing Systems (NeurIPS 2020).
    \item Vincent Le Guen and Nicolas Thome. "Deep Time Series Forecasting with Shape and Temporal Criteria". IEEE Transactions on Pattern Analysis and Machine Intelligence, 2022.
\end{itemize}
}
}

\section{Introduction}

\begin{figure}[H]
\begin{center}
    \begin{tabular}{ccc}
  \hspace{-1.0cm}  \includegraphics[width=5.8cm]{{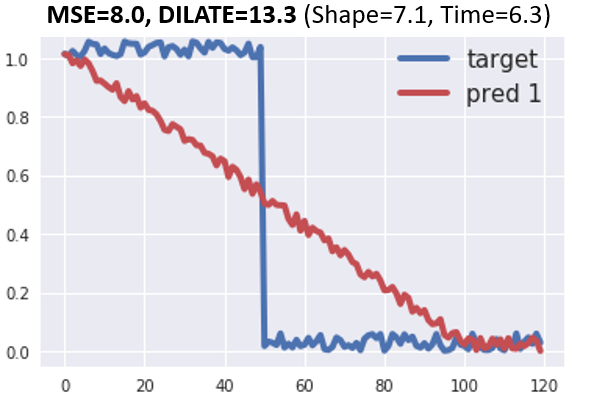}}  & 
    \hspace{-0.4cm} \includegraphics[width=5.8cm]{{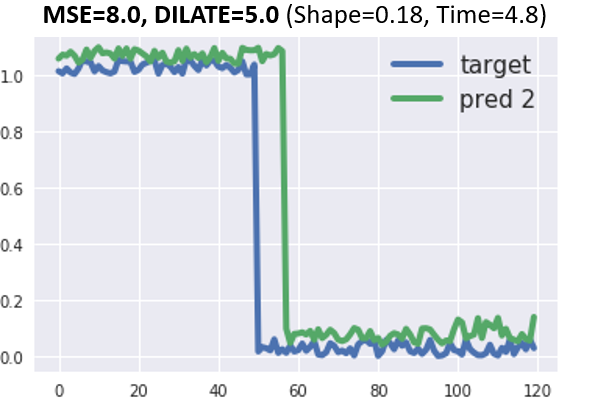}} & 
    \hspace{-0.4cm} \includegraphics[width=5.8cm]{{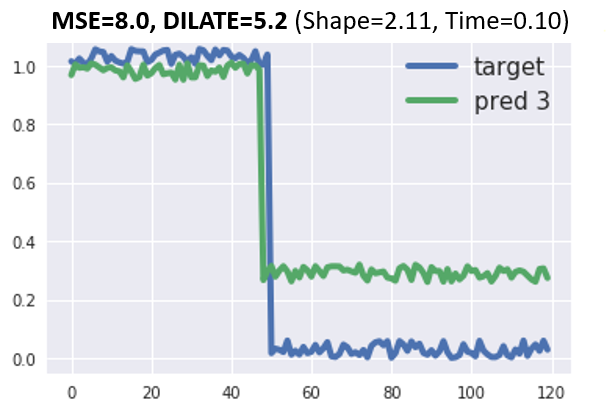}} \\
   \hspace{-1.0cm}   (a) Non informative prediction  & \hspace{-0.4cm}(b) Correct shape, time delay & \hspace{-0.4cm} (c) Correct time, inaccurate shape
    \end{tabular}
\end{center}
    \caption[Limitations of the MSE in deterministic forecasting.]{\textbf{Limitation of the Euclidean (MSE) loss}: when predicting a sudden change (target blue step function), the 3 predictions (a), (b) and (c) have similar MSE but very different forecasting skills. In contrast, the DILATE loss proposed in this work, which disentangles shape and temporal decay terms, supports predictions (b) and (c) over prediction (a) that does not capture the sharp change of regime.}
   \label{fig:intro_dilate}
\end{figure}

\lettrine[lines=3]{A}s discussed in the previous Chapter, the Mean Squared Error (MSE) is inadequate in the context of non-stationary time series with sudden variations, as illustrated in Figure \ref{fig:intro_dilate}. Here, the target ground truth prediction is a step function (in blue), and we present three predictions, shown in Figure \ref{fig:intro_dilate} (a), (b), and (c), which have a similar MSE loss compared to the target, but very different forecasting skills. Prediction (a) is not adequate for regulation purposes since it doesn't capture the sharp drop to come. Predictions (b) and (c) much better reflect the change of regime since the sharp drop is indeed anticipated, 
although with a slight delay (b) or with a slight 
inaccurate amplitude (c).

This Chapter introduces DILATE (DIstortion Loss including shApe and TimE), a new objective function for training deep neural networks in the context of multi-step and non-stationary time series forecasting. DILATE explicitly disentangles into two terms the penalization related to the shape and the temporal localization errors of change detection. The behaviour of DILATE is shown in  Figure \ref{fig:intro_dilate}: whereas the values of our proposed shape and temporal losses are large in Figure \ref{fig:intro_dilate} (a), the shape (resp. temporal) term is small in Figure \ref{fig:intro_dilate} (b) (resp. Figure \ref{fig:intro_dilate} (c)). DILATE combines shape and temporal terms, and is consequently able to output a much smaller DILATE loss for predictions (b) and (c) than for (a), as expected.

We first present the DILATE loss in section  \ref{sec:training_with_dilate}. We also introduce a variant of DILATE, which provides a smooth generalization of temporally-constrained Dynamic Time Warping (DTW) metrics~\cite{sakoe1990dynamic,jeong2011weighted}. Experiments carried out on several synthetic and real non-stationary datasets reveal that 
models trained with DILATE significantly outperform models trained with the MSE loss function when evaluated with shape and temporal distortion metrics, while DILATE maintains very good performance when evaluated with MSE. Finally, we show that DILATE can be used with various network architectures and can outperform on shape and time metrics state-of-the-art models specifically designed for multi-step and non-stationary forecasting.

\section{Training Deep Neural Networks with DILATE}
\label{sec:training_with_dilate}

 Given an input sequence $\x_{1:T}=(\x_1,\dots,\x_T) \in \mathbb{R}^{p \times T}$, the deterministic multi-step time series forecasting problem consists in predicting a $H$-steps future trajectory  $ \hat{\y} = (\hat{\y}_{T+1},\dots, \hat{\y}_{T+H} )  \in  \mathbb{R}^{d \times H}$. As an alternative to the MSE, we introduce here the DIstortion Loss with shApe and TimE (DILATE) for training any deterministic deep multi-step forecasting model. Crucially, the DILATE loss needs to be differentiable in order to train models with gradient-based optimization.
 
\begin{figure*}
    \centering
    \includegraphics[width=17cm]{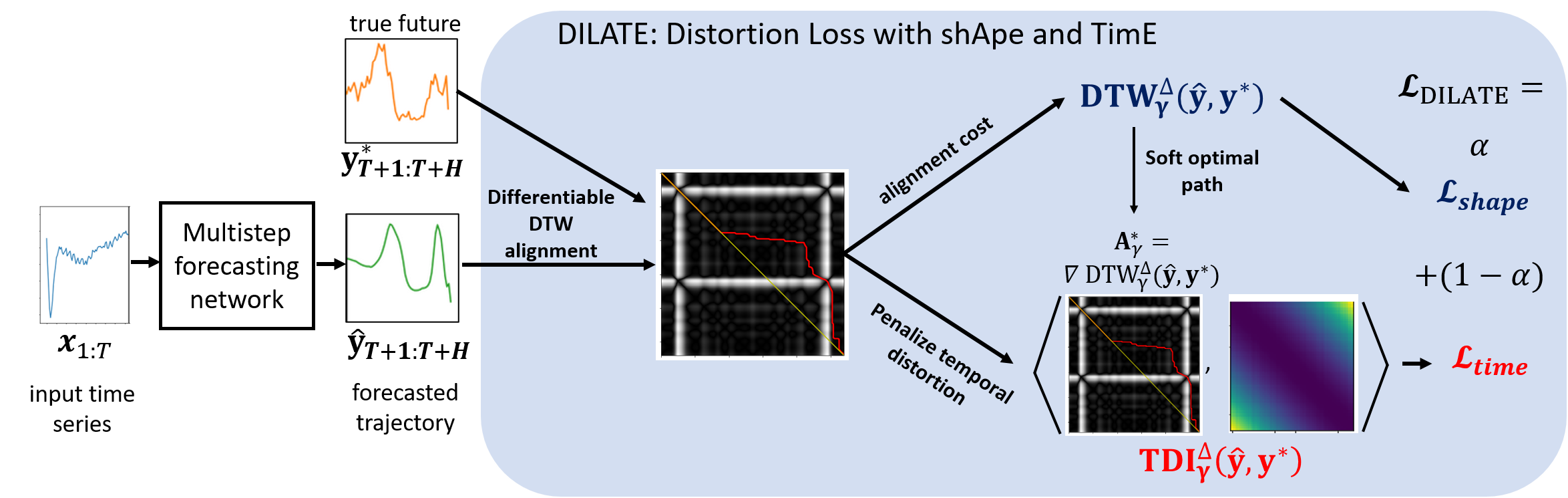}
    \caption[Overview of the DILATE loss.]{\textbf{Overview of the DILATE loss:} $\mathcal{L}_{\text{DILATE}}$ for training deterministic deep time series forecasting models is composed of two terms: $\mathcal{L}_{shape}$ based on the soft DTW and $\mathcal{L}_{time}$ that penalizes the temporal distortions visible on the soft optimal path. The overall loss  $\mathcal{L}_{\text{DILATE}}$ is differentiable, and we provide an efficient implementation of its forward and backward passes.}
    \label{fig:dilate}
\end{figure*}

The DILATE objective function, which compares the prediction $\hat{\y} =  (\hat{\y}_{T+1},\dots, \hat{\y}_{T+H} )$ with the actual ground truth future trajectory $\y^* = (\y^*_{T+1},\dots,\y^*_{T+H})$, is composed of two terms balanced by the hyperparameter $\alpha \in [0,1]$:
\begin{align}
\label{eq:dilate}
\mathcal{L}_{\text{DILATE}}(\hat{\y}, \y^*) &= \alpha~\mathcal{L}_{shape}(\hat{\y}, \y^*) + (1-\alpha)~ \mathcal{L}_{time}(\hat{\y}, \y^*)\\
 &= \alpha ~\text{DTW}^{\mathbf{\Delta}}_{\gamma}(\hat{\y}, \y^*) + (1-\alpha)~ \text{TDI}^{\mathbf{\Delta},\mathbf{\Omega_{dissim}}}_{\gamma}(\hat{\y}, \y^*).
\end{align}

The computational graph of the DILATE loss is illustrated in Figure \ref{fig:dilate}. We use for the shape term $\mathcal{L}_{shape}$ the smooth shape dissimilarity $\text{DTW}^{\mathbf{\Delta}}_{\gamma}$ defined in Eq \ref{eq:dtwgamma} and for the temporal term  $\mathcal{L}_{time}$ the time dissimilarity $\text{TDI}^{\mathbf{\Delta},\mathbf{\Omega_{dissim}}}_{\gamma}$ defined in Eq \ref{eq:temporal}.

\paragraph{Tangled DILATE variant} A variant of our approach to combine shape and temporal penalization would be to incorporate a temporal term inside our smooth $\mathcal{L}_{shape}$ function in Eq \ref{eq:dtwgamma}, leading to a \textit{tangled} version $\mathcal{L}_{\text{DILATE}^t}$: 
\begin{equation}
\mathcal{L}_{\text{DILATE}^t}(\hat{\y}_i, \y^*_{i}) :=  - \gamma  \log \left ( \sum_{\mathbf{A} \in \mathcal{A}_{\tau,\tau}} \exp\left ( - \textstyle \frac{ \left \langle \mathbf{A} , \alpha \mathbf{\Delta}(\hat{\y}_i, \y^*_{i}) + (1-\alpha) \mathbf{\Omega} \right \rangle}{\gamma} \right ) \right ).
\label{eq:smoothwdtw}
\end{equation}

We can notice that Eq \ref{eq:smoothwdtw} reduces to minimizing $\left \langle \mathbf{A} , \alpha \mathbf{\Delta}(\hat{\y}_i, \y^*_{i}) + (1-\alpha) \mathbf{\Omega}  \right \rangle$ when $\gamma \to 0^+$. In this case, $\mathcal{L}_{\text{DILATE}^t}$ can recover DTW variants studied in the literature to bias the computation based on penalizing sequence misalignment, by designing specific $\mathbf{\Omega}$ matrices:
\begin{center}
 \begin{adjustbox}{max width=\linewidth}  
  \begin{tabular}{c|c}
  
  \begin{tabular}{c}
     Sakoe-Chiba DTW \\
      band constraint \cite{sakoe1990dynamic}    
  \end{tabular}

  & $\Omega(i,j) =$ 
  $\begin{cases}
         + \infty  \text{~if~} |i-j|>T       \\
       0 \text{~~ otherwise}    
    \end{cases}$
  \\ \hline
  Weighted DTW \cite{jeong2011weighted}   & $\Omega(i,j) = f(|i-j|)$ \text{~ for $f$ increasing function}
\end{tabular}  
\end{adjustbox}
\end{center}

$\mathcal{L}_{\text{DILATE}^t}$ in Eq \ref{eq:smoothwdtw} enables to train deep neural networks with a smooth loss combining shape and temporal criteria. However, $\mathcal{L}_{\text{DILATE}^t}$ presents limited capacities for disentangling the shape and temporal errors, since the optimal path is computed from both shape and temporal terms, \ie there is no guarantee to recover the true optimal alignment path because of the temporal penalization inside the cost matrix. In contrast, our $\mathcal{L}_{\text{DILATE}}$ loss in Eq \ref{eq:dilate} separates the loss into two shape and temporal components, the temporal penalization being applied to the optimal unconstrained DTW path.

\paragraph{Discussion on most related works}

We review here the most related works that attempt to train deep forecasting models with alternatives to the MSE. For exploiting the shape of future trajectories, recent works have explored smooth approximations of Dynamic Time Warping (DTW) \cite{cuturi2017soft, mensch2018differentiable,abid2018learning,vayer2020time,blondel2020differentiable}. Cuturi and Blondel have proposed the soft-DTW  \cite{cuturi2017soft}, which is differentiable loss function that can be computed by dynamic programming with a quadratic complexity. They have shown convincing experiments on time series classification, clustering under the DTW geometry and early experiments on time series forecasting. The soft-DTW  was further normalized to ensure a non-negative divergence \cite{blondel2020differentiable}. However, since DTW is by design invariant to elastic distortions, it completely ignores the temporal localization of the changes. A differentiable timing error loss function based on DTW on the event (binary) space was proposed in \cite{rivest2019new} ; however it is only applicable for predicting binary time series. Some works explored the use of adversarial losses for time series \cite{yoon2019time,wu2020adversarial}, which can be seen as an implicit way of enforcing semantic criteria learned from data. However, it gives a weaker and non-interpretable control on shape and time criteria and brings additional adversarial training  difficulties.

\section{Experiments}

In this section, we evaluate the relevance of DILATE, both quantitatively and qualitatively, compared to generic as well as recent state-of-the-art models trained with the MSE. We also provide an in-depth analysis of the DILATE loss properties.

\subsection{Datasets}
\label{sec:datasets}

We carry out experiments on 5 synthetic and real-world datasets from various domains to illustrate the broad applicability of our methods. For each dataset, the task is to predict the $H$-steps ahead future trajectory given a $T$-steps context window:

\begin{itemize}
    \item \texttt{Synthetic-det} ($T=20, H=20$): deterministic dataset consisting in predicting sudden changes (step functions) based on an input signal composed of two peaks. This controlled setup was designed to measure precisely the shape and time errors of predictions. We generate 500 times series for train, 500 for validation and 500 for test, with 40 time steps each: the first 20 are the inputs, the last 20 are the targets to forecast. In each series, the input range is composed of 2 peaks of random temporal position $i_1$ and $i_2$ and random amplitude $j_1$ and $j_2$ between 0 and 1, and the target range is composed of a step of amplitude $j_2-j_1$  and  stochastic position $i_2 + (i_2-i_1)+ randint(-3,3)$. All time series are corrupted by an additive Gaussian white noise of variance 0.01.
    
    \item \texttt{ECG5000}  ($T=84, H=56$): this dataset comes from the UCR Time Series Classification Archive \cite{chen2015ucr}, and is composed of 5000 electrocardiograms (ECG) (500 for training, 4500 for testing) of length 140. We take the first 84 time steps (60 \%) as input and predict the last 56 steps (40 \%) of each time series (same setup as in \cite{cuturi2017soft}).
    
    \item \texttt{Traffic} ($T=168, H=24$): this dataset is composed of road occupancy rates (between 0 and 1) from the California Department of Transportation (48 months from 2015-2016) measured every 1h. We work on the first univariate series of length 17544 (with the same 60/20/20 train/valid/test split as in \cite{lai2018modeling}), and we train models to predict the 24 future points given the past 168 points (past week)
    
    \item  \texttt{Electricity} ($T=168, H=24$): this dataset consists in hourly electricity consumption measurements (kWh) from 370 customers.
    
    \item \texttt{ETTh1} \cite{zhou2020informer} ($T=96, H=96$): dataset of hourly Electricity Transformer Temperature measurements, which is an important indicator for electricity grids. This dataset enables to assess the generalization of our approach on much longer term predictions.
\end{itemize}

\subsection{Implementation details}

\paragraph*{Metrics}
To evaluate the benefits of our proposed DILATE training loss, we compare it against the widely used Euclidean (MSE) loss, and the soft-DTW introduced in~\cite{cuturi2017soft,mensch2018differentiable}. We use the following multi-step prediction metrics: MSE, DTW (shape), TDI (temporal). To consolidate the evaluation, we also consider two additional (non differentiable) metrics for assessing shape and time.  For shape, we compute the ramp score \cite{vallance2017towards}. For time, we compute the Hausdorff distance between a set of detected change points in the target signal $\mathcal{T}^*$ and in the predicted signal 
$\hat{\mathcal{T}}$:
\begin{equation}
\text{Hausdorff}(\mathcal{T}^*,\hat{\mathcal{T}}) :=  \max (  \underset{\hat{t} \in \mathcal{ \hat{T} }}{\max}  \underset{t^* \in \mathcal{ T^* }}{\min} |\hat{t}-t^* |  ,  \underset{t^* \in \mathcal{ T^* }}{\max}   \underset{\hat{t} \in \mathcal{ \hat{T} }}{\min} |\hat{t}-t^* | ),
\end{equation}{}
which corresponds to the largest possible distance between a change point and its prediction. Additional details about these external metrics are given in Appendix \ref{app:dilate_metrics}.

\paragraph*{Neural networks architectures:} For the generic neural network architectures, we use a fully connected network (1 layer of 128 neurons), which does not make any assumption on data structure, and a more specialized Seq2Seq model \cite{sutskever2014sequence} with Gated Recurrent Units (GRU) \cite{cho2014learning} with 1 layer of 128 units. 
Each model is trained with PyTorch for a max number of 1000 epochs with Early Stopping with the ADAM optimizer. The smoothing parameter $\gamma$ of DTW and TDI is set to $10^{-2}$. 

\paragraph*{DILATE hyperparameters:} the hyperparameter $\alpha$ balancing $\mathcal{L}_{shape}$ and $\mathcal{L}_{time}$ is determined on a validation set to get comparable DTW shape performance than the $\text{DTW}_{\gamma}^{\mathbf{\Delta}}$ trained model: $\alpha=0.5$  for Synthetic and ECG5000, and 0.8 for Traffic, Electricity and ETTh1. The DTW smoothing parameter $\gamma$ is fixed to $10^{-2}$, as further discussed in section \ref{sec:dilate-analysis}.

Our code implementing DILATE is available on line from: \url{https://github.com/vincent-leguen/DILATE}.

\begin{table*}
      \caption[DILATE forecasting results on generic MLP and RNN architectures.]{\textbf{DILATE forecasting results on generic MLP and RNN architectures}, averaged over 10 runs (mean $\pm$ standard deviation). Metrics are scaled for readability. For each experiment, best method(s) (Student t-test) in bold.}    
      \centering
          \begin{adjustbox}{max width=\linewidth}   
    \begin{tabular}{llccc|ccc}
    \toprule
     \multicolumn{2}{c}{}  &  \multicolumn{3}{c|}{\textbf{Fully connected network (MLP)}}       & \multicolumn{3}{c}{\textbf{Recurrent neural network (Seq2Seq)}}    \\
     \hline
       Dataset      &  \diagbox{Eval}{Train}   &  MSE &   $\text{DTW}_{\gamma}^{\mathbf{\Delta}}$~\cite{cuturi2017soft} &  DILATE (ours)  &  MSE  &   $\text{DTW}_{\gamma}^{\mathbf{\Delta}}$~\cite{cuturi2017soft} &  DILATE (ours) \\ 
       \hline
    ~         & MSE (x1000)  & ~  \textbf{16.5 $\pm$ 1.4}     & ~   48.2 $\pm$  4.0      & ~   \textbf{16.7$\pm$  1.8}        & ~         \textbf{11.0 $\pm$  1.7}        & ~  23.1 $\pm$  4.5              & ~  \textbf{12.1 $\pm$  1.3}          \\
    \texttt{Synthetic} & DTW (x10)     & ~   38.6 $\pm$ 1.28          & ~  \textbf{27.3 $\pm$ 1.37}          & ~  32.1 $\pm$ 5.33       & ~   \textbf{24.6 $\pm$ 1.20}                & ~  \textbf{22.7 $\pm$ 3.55}             & ~  \textbf{23.1 $\pm$ 2.44}           \\
    ~         & TDI (x10)     & ~  15.3 $\pm$ 1.39           & ~  26.9 $\pm$ 4.16           & ~   \textbf{13.8 $\pm$ 0.71}       & ~    17.2 $\pm$ 1.22              & ~      20.0 $\pm$ 3.72          & ~   \textbf{14.8 $\pm$ 1.29}          \\ 
~    & Ramp (x10)  & 5.21 $\pm$ 0.10  &  \textbf{2.04 $\pm$ 0.23}  & 3.41 $\pm$ 0.29  &  5.80 $\pm$ 0.10 & \textbf{4.27 $\pm$ 0.8} & 4.99 $\pm$ 0.46   \\
~ & Hausdorff (x1) & 4.04 $\pm$ 0.28 & 4.71 $\pm$ 0.50  & \textbf{3.71 $\pm$ 0.12}  & 2.87 $\pm$ 0.13 & 3.45 $\pm$ 0.32 & \textbf{2.70 $\pm$ 0.17}  \\
\midrule
     ~         & MSE (x100) & ~  \textbf{31.5 $\pm$ 1.39}   & ~    70.9 $\pm$ 37.2  & ~    37.2 $\pm$ 3.59    & ~    \textbf{21.2 $\pm$ 2.24}    & ~   75.1 $\pm$ 6.30              & ~      30.3 $\pm$ 4.10             \\
    \texttt{ECG}       & DTW (x10)     & ~     19.5 $\pm$ 0.16          & ~    18.4 $\pm$ 0.75         & ~   \textbf{17.7 $\pm$ 0.43}       & ~     17.8 $\pm$ 1.62              & ~      17.1 $\pm$ 0.65           & ~  \textbf{16.1 $\pm$ 0.16}            \\
    ~         & TDI (x10)     & ~     \textbf{7.58 $\pm$ 0.19}          & ~    17.9 $\pm$ 0.7      & ~   \textbf{7.21 $\pm$ 0.89}       & ~    8.27 $\pm$ 1.03              & ~   27.2 $\pm$ 11.1              & ~    \textbf{6.59 $\pm$ 0.79}             \\
    ~ & Ramp (x1) & \textbf{4.9 $\pm$ 0.1}  & 5.1 $\pm$ 0.3  &  \textbf{5.0 $\pm$ 0.1} &    \textbf{4.84 $\pm$ 0.24}              & ~      \textbf{4.79 $\pm$ 0.37}           & ~  \textbf{4.80 $\pm$ 0.25}    \\
    ~ & Hausdorff (x1) & \textbf{4.1 $\pm$ 0.1} & 6.3 $\pm$ 0.6  & 4.7 $\pm$ 0.3  &    \textbf{4.32 $\pm$ 0.51}     & ~   6.16 $\pm$ 0.85               & ~ \textbf{4.23 $\pm$ 0.41}        \\
 \midrule
    ~         & MSE (x1000) & ~    \textbf{6.58 $\pm$ 0.11}           & ~   25.2 $\pm$ 2.3          & ~   19.3 $\pm$ 0.80       & ~        \textbf{8.90 $\pm$ 1.1}           & ~      22.2 $\pm$ 2.6           & ~     \textbf{10.0 $\pm$ 2.6}         \\
    \texttt{Traffic}   & DTW (x100)     & ~   25.2 $\pm$ 0.17            & ~    \textbf{23.4 $\pm$ 5.40}         & ~   \textbf{23.1 $\pm$ 0.41}   & ~    24.6 $\pm$ 1.85              & ~      \textbf{22.6 $\pm$ 1.34}           & ~    \textbf{23.0 $\pm$ 1.62}          \\
    ~         & TDI (x100)     & ~   24.8 $\pm$ 1.1            & ~    27.4 $\pm$ 5.01         & ~  \textbf{16.7 $\pm$ 0.51}        & ~     \textbf{15.4 $\pm$ 2.25}              & ~     22.3 $\pm$ 3.66            & ~    \textbf{14.4$\pm$  1.58}          \\
    ~ & Ramp (x10) & 6.18 $\pm$ 0.1  & \textbf{5.59 $\pm$ 0.1}  &  \textbf{5.6 $\pm$ 0.1}   &  6.29 $\pm$ 0.32             & ~     \textbf{5.78 $\pm$ 0.41}           & ~    \textbf{5.93 $\pm$ 0.24}    \\
    ~ & Hausdorff (x1) & \textbf{1.99 $\pm$ 0.2}  & \textbf{1.91 $\pm$ 0.3} &  \textbf{1.94 $\pm$ 0.2}  &   \textbf{2.16 $\pm$ 0.38}           & ~      \textbf{2.29 $\pm$ 0.33}          & ~     \textbf{2.13 $\pm$ 0.51}   \\ 
\bottomrule
    \end{tabular}
    \end{adjustbox}
    \label{results1}  
\end{table*}

\subsection{DILATE performances on generic architectures}

To demonstrate the broad applicability of our approach, we first perform multi-step forecasting with two generic neural network architectures: a fully connected network (1 layer , which does not make
any assumption on data structure, and a more specialized Seq2Seq model with 1 layer of 128 Gated Recurrent Units (GRU). We perform a Student t-test with significance level 0.05 to highlight the best(s) method(s) in each experiment (averaged over 10 runs). Overall results are presented in Table \ref{results1}. 

\paragraph*{Comparison to MSE training loss:} DILATE outperforms MSE when evaluated on shape (DTW) in all experiments,  with significant differences on 5/6 experiments. When evaluated on time (TDI), DILATE also performs better in all experiments (significant differences on 3/6 tests). Finally, DILATE is equivalent to MSE when evaluated on MSE on 3/6 experiments.

\paragraph*{Comparison to $\text{DTW}_{\gamma}^{\mathbf{\Delta}}$ training loss:} When evaluated on shape (DTW), DILATE performs similarly to  $\text{DTW}_{\gamma}^{\mathbf{\Delta}}$ (2 significant improvements, 1 significant drop and 3 equivalent performances). For time (TDI) and MSE evaluations, DILATE is significantly better than  $\text{DTW}_{\gamma}^{\mathbf{\Delta}}$ in all experiments, as expected.

We can notice that the ramp score (resp. the Haussdorff distance) provides the same trends than the shape metric DTW (resp. the time metric TDI). It reinforces our conclusions and shows that DILATE indeed improves shape and temporal accuracy beyond the metrics being optimized.

We display a few qualitative examples for Synthetic, ECG5000 and Traffic datasets in Figure \ref{fig:dilate_visu} (other examples are provided in Appendix \ref{app:dilate_visus}). We see that MSE training leads to predictions that are non-sharp, making them inadequate in presence of drops or sharp spikes. $\text{DTW}_{\gamma}^{\mathbf{\Delta}}$ leads to very sharp predictions in shape, but with a possibly large temporal misalignment. In contrast, our DILATE loss predicts series that have both a correct shape and precise temporal localization.

\begin{figure}[t]
    \centering
    \includegraphics[width=16cm]{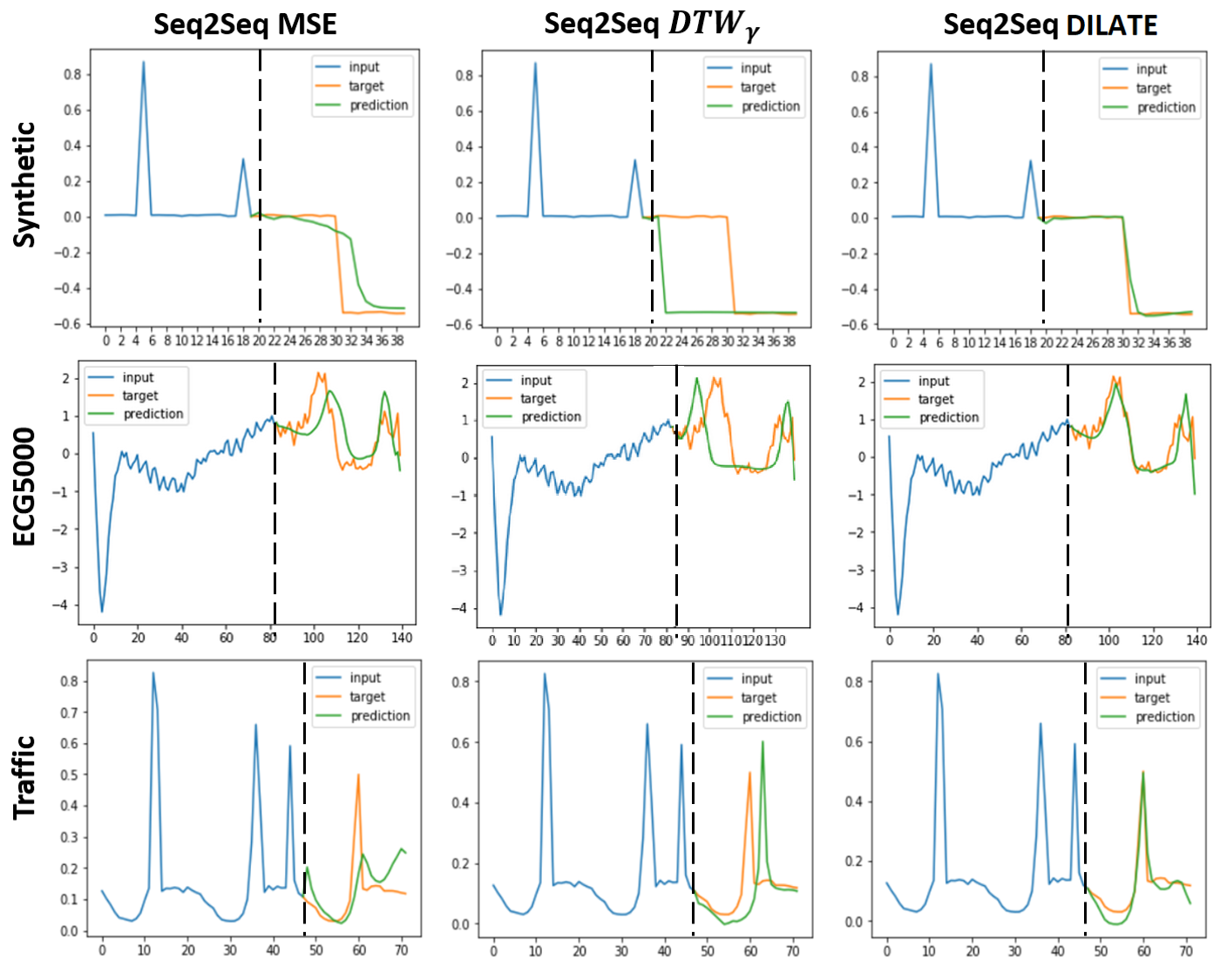}
    \caption[Qualitative prediction results with the DILATE loss.]{\textbf{Qualitative prediction results with the DILATE loss.} For each dataset, the MSE training loss leads to non-sharp predictions, whereas the soft-DTW loss can predict sharp variations but has no control over their temporal localization. In contrast, the DILATE loss produces sharp predictions with accurate temporal localization.}
    \label{fig:dilate_visu}
\end{figure}

\subsection{DILATE performances with state-of-the-art models}

\begin{figure}
\begin{tabular}{cc}
\textbf{N-Beats \cite{oreshkin2019n} MSE} & \textbf{N-Beats \cite{oreshkin2019n} DILATE} \\
\includegraphics[width=8.5cm]{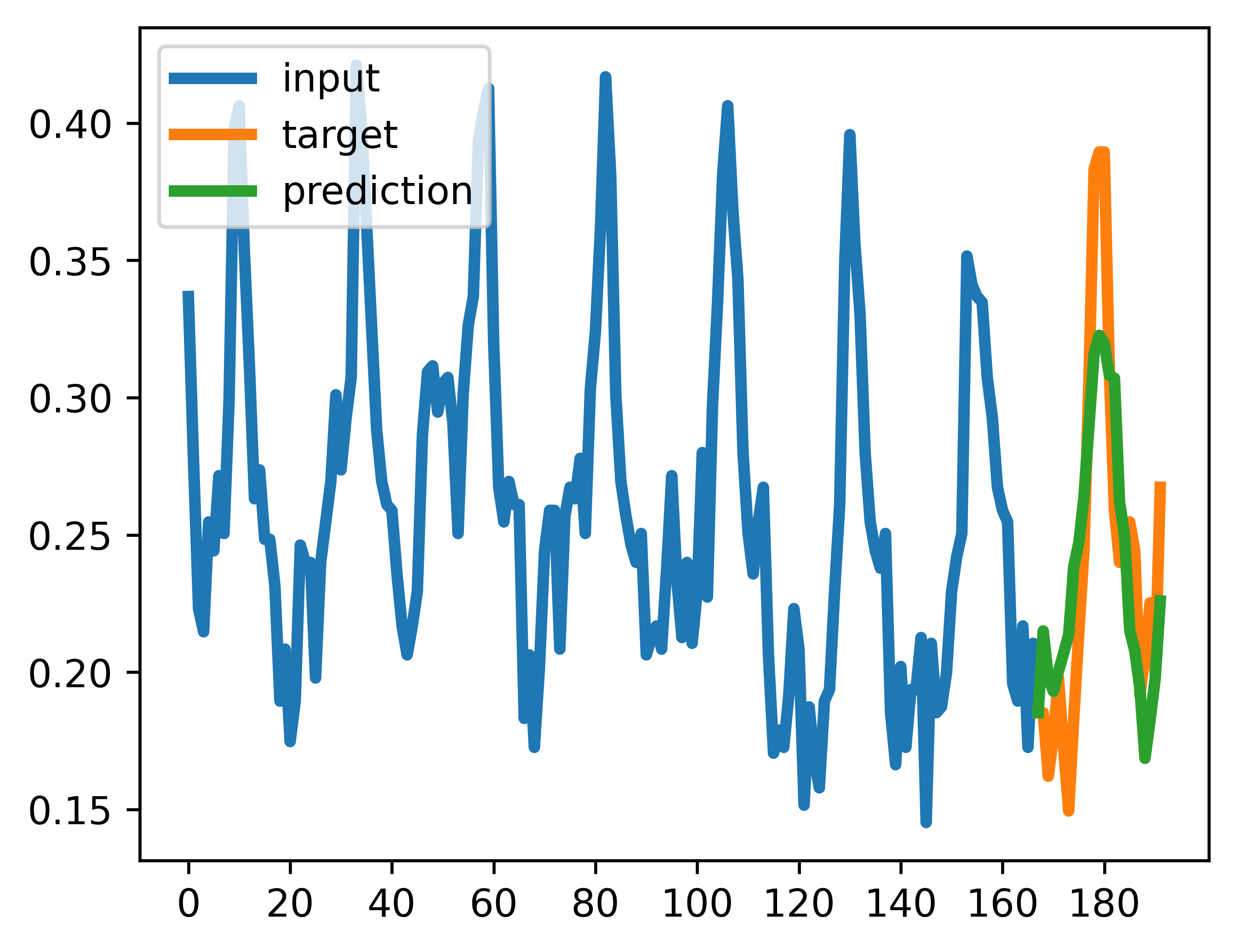}     & \hspace{-0.5cm} \includegraphics[width=8.5cm]{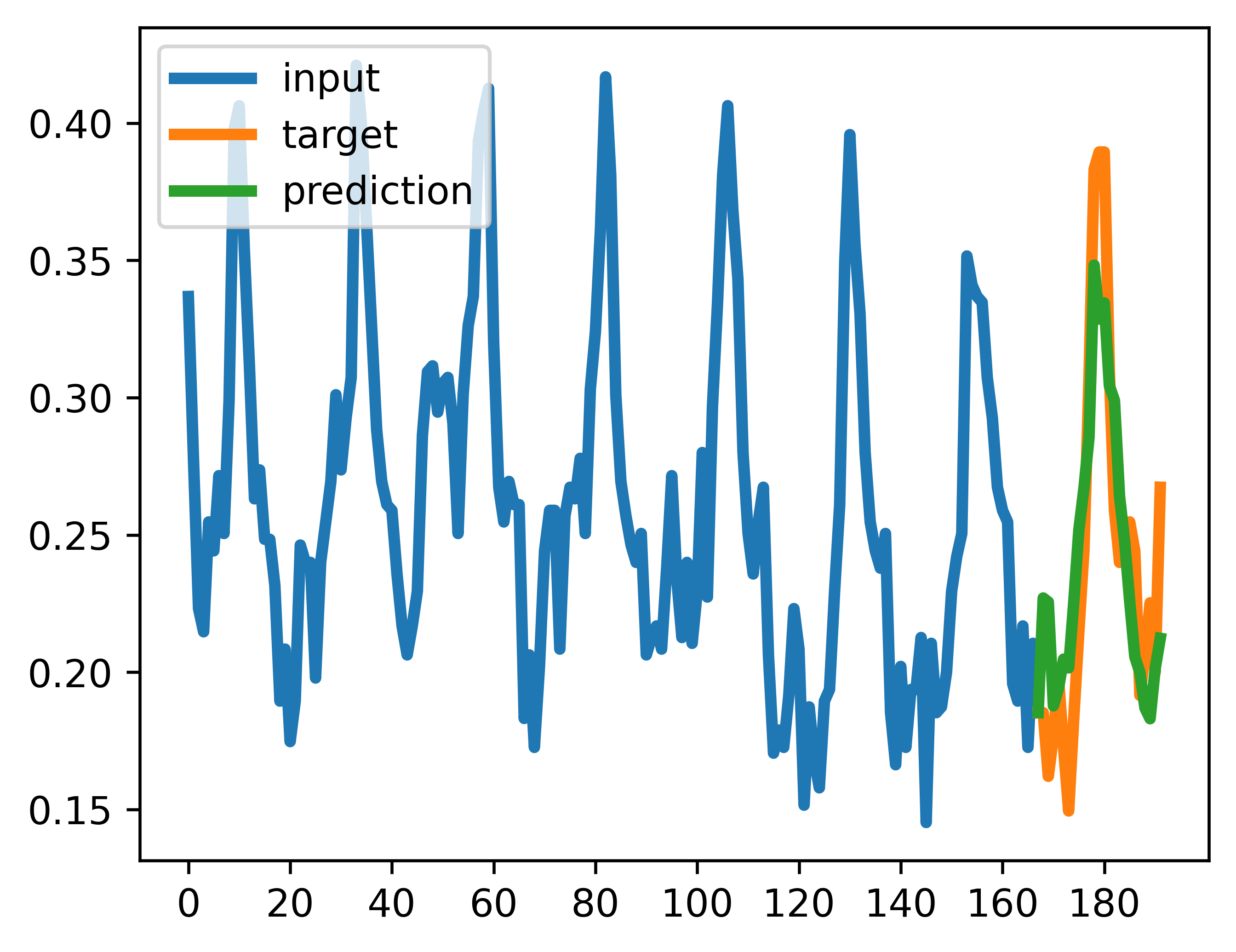} \\
\end{tabular}\
\hspace{-0.7cm}
\caption[DILATE forecasting results on state-of-the-art architectures.]{Qualitative forecasting results comparing the N-Beats model \cite{oreshkin2019n} trained with MSE and the DILATE loss on the \texttt{Electricity} dataset.}
\label{fig:dilate_elec}
\end{figure}

\begin{figure}
\begin{tabular}{cc}
\textbf{N-Beats \cite{oreshkin2019n} MSE} & \textbf{N-Beats \cite{oreshkin2019n} DILATE} \\
\hspace{-0.5cm}
\includegraphics[width=8.6cm]{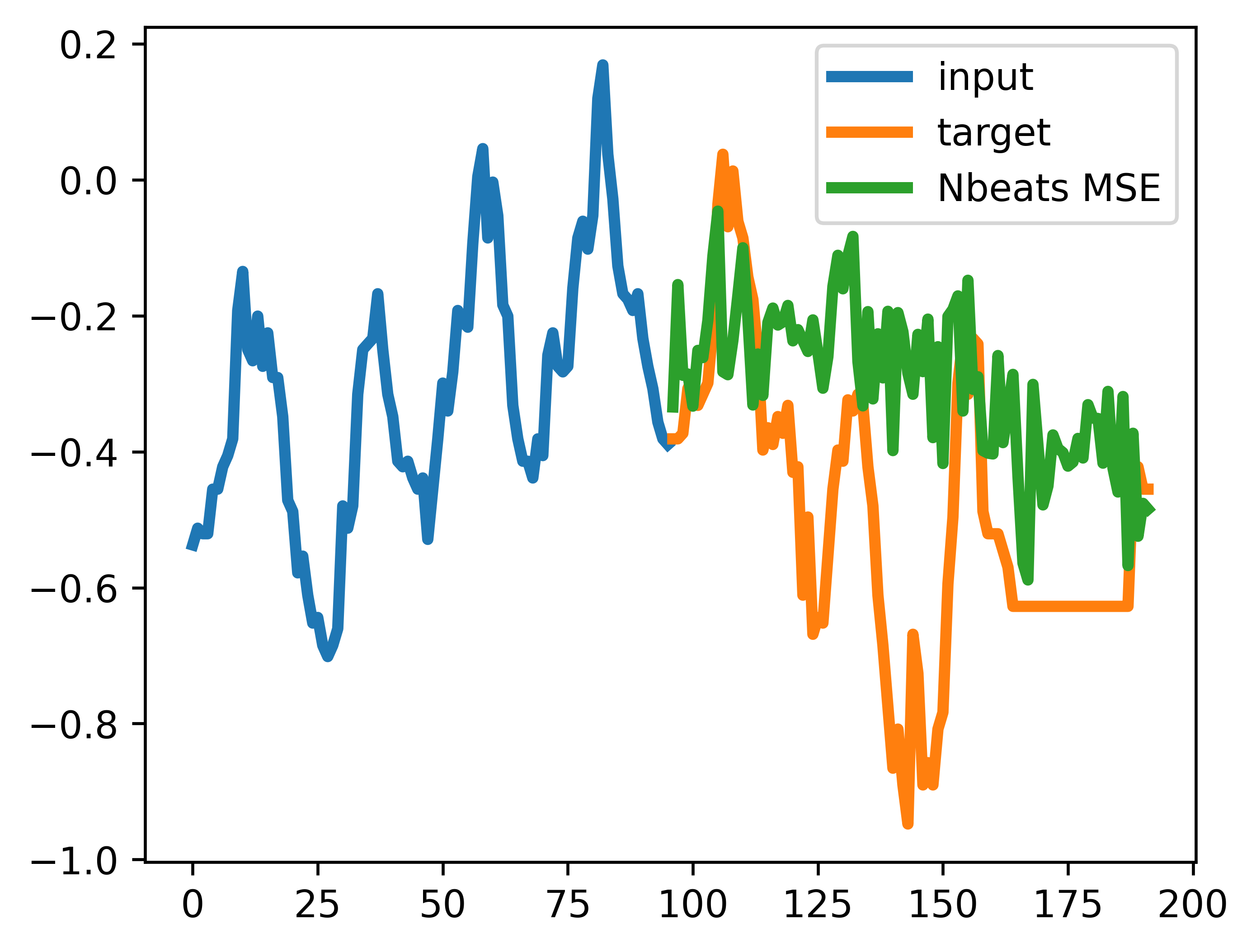}     & \hspace{-0.5cm}
\includegraphics[width=8.6cm]{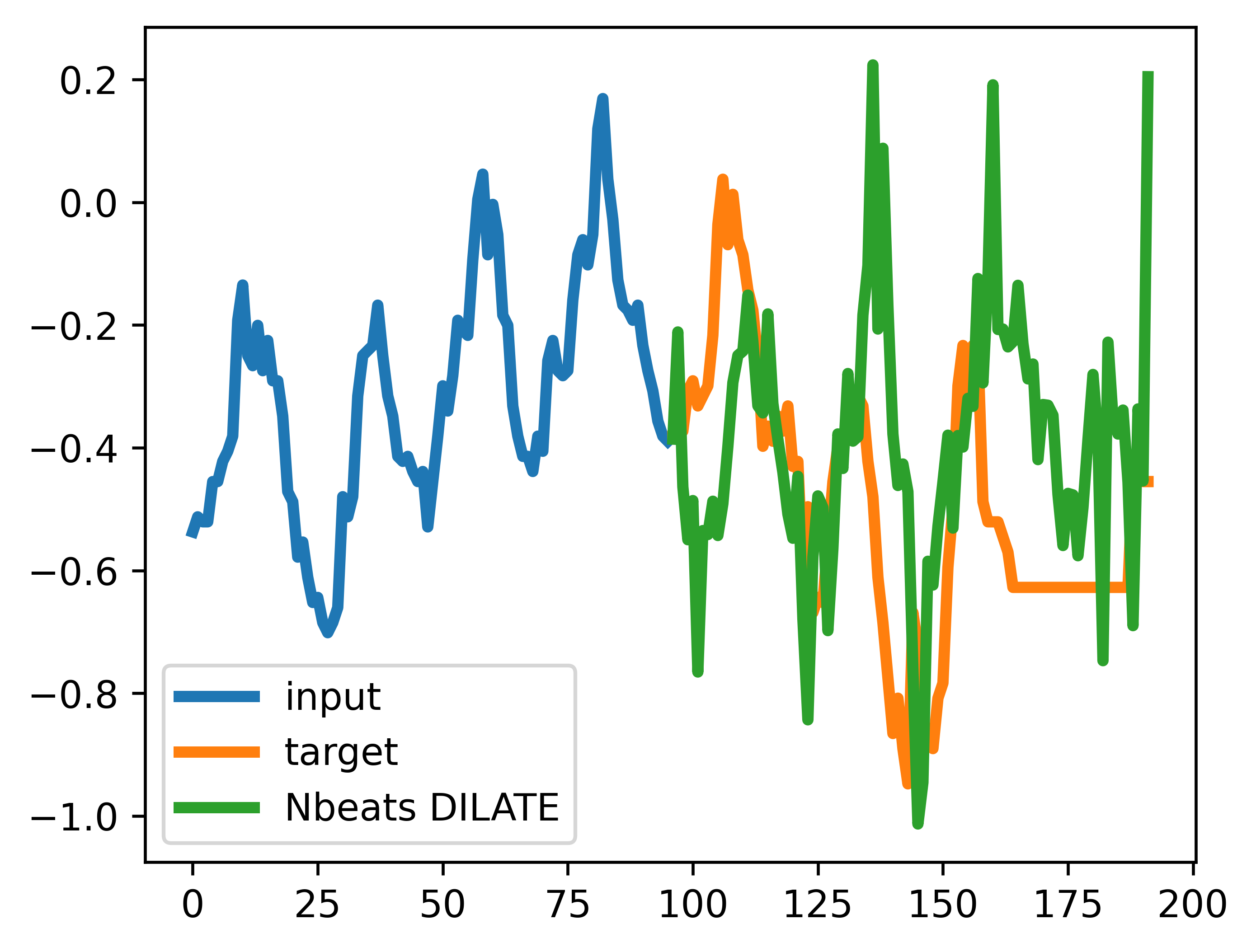}  
\end{tabular}
\caption[DILATE forecasting results on state-of-the-art architectures.]{Qualitative forecasting results comparing the N-Beats model \cite{oreshkin2019n} trained with MSE and the DILATE loss on the \texttt{ETTH1} dataset.}
\label{fig:dilate_etth1}
\end{figure}

Beyond generic forecasting architectures, we show that DILATE can also improve the performances of state-of-the-art deep architectures. We experiment here with two recent and competitive models:  N-Beats \cite{oreshkin2019n} and Informer \cite{zhou2020informer}.  Results in Table \ref{tab:dilate_sota} are consistent with those in Table \ref{results1}: models trained with DILATE improve over MSE in shape (in DTW and ramp score for 6/6 experiments) and time (in TDI for 5/6 and Hausdorff for 4/6 experiments) and are equivalent to MSE when evaluated in MSE (equivalent or better for 3/6 experiments). We provide qualitative predictions of N-Beats on \texttt{Electricity} in Figure \ref{fig:dilate_elec} and \texttt{ETTh1} in Figure \ref{fig:dilate_etth1}. It again confirms that training with DILATE leads to much sharper predictions with a better temporal localization than training with the MSE.

\begin{table*}
    \caption[DILATE forecasting results on state-of-the-art architectures.]{\textbf{DILATE forecasting results on state-of-the-art architectures N-Beats \cite{oreshkin2019n} and Informer \cite{zhou2020informer}}. Evaluation metrics are scaled for readability. Results are averaged over 10 runs, best(s) method(s) in bold (Student t-test).}
    \centering
        \begin{adjustbox}{max width=\linewidth}   
    \begin{tabular}{cccccccc}
    \toprule
      Dataset & Model & MSE  & DTW & Ramp & TDI & Hausdorff & DILATE    \\
      \midrule
\texttt{Synthetic}   & N-Beats \cite{oreshkin2019n} MSE & \textbf{13.6 $\pm$ 0.5} & 24.9 $\pm$ 0.6 & 5.9 $\pm$ 0.1 & \textbf{13.8 $\pm$ 1.1} & \textbf{2.8 $\pm$ 0.1} & \textbf{19.3 $\pm$ 0.5}  \\
 & N-Beats \cite{oreshkin2019n} DILATE & \textbf{13.3 $\pm$ 0.7} & \textbf{23.4 $\pm$ 0.8}  & \textbf{4.8 $\pm$ 0.4}  &  \textbf{14.4 $\pm$ 1.3}  & \textbf{2.7 $\pm$ 0.5}  & \textbf{18.9 $\pm$ 0.8}  \\
  \cdashline{2-8}
   & Informer \cite{zhou2020informer} MSE & \textbf{10.4 $\pm$ 0.3} & 20.1 $\pm$ 1.1  & 4.3 $\pm$ 0.3  & 13.1 $\pm$ 0.9  & \textbf{2.5 $\pm$ 0.1} & 16.6 $\pm$ 0.8   \\
 & Informer \cite{zhou2020informer} DILATE & 11.8 $\pm$ 0.7  &  \textbf{18.5 $\pm$ 1.2}  & \textbf{2.4 $\pm$ 0.3}  & \textbf{11.6 $\pm$ 0.9}  & \textbf{2.4 $\pm$ 0.9} & \textbf{15.1 $\pm$ 0.7} \\
  \midrule
 \texttt{Electricity}  & N-Beats \cite{oreshkin2019n} MSE & \textbf{24.8 $\pm$ 0.4} & \textbf{15.6 $\pm$ 0.2} & \textbf{13.3  $\pm$ 0.3} & 4.6 $\pm$ 0.1 & \textbf{2.6 $\pm$ 0.3}  & \textbf{13.4 $\pm$ 0.2} \\
 & N-Beats \cite{oreshkin2019n} DILATE & 25.8 $\pm$ 0.9  & \textbf{15.5 $\pm$ 0.2} & \textbf{13.3 $\pm$ 0.3} & \textbf{4.4 $\pm$ 0.2} & 3.1 $\pm$ 0.5  & \textbf{13.2 $\pm$ 0.2}  \\
  \cdashline{2-8}
   & Informer \cite{zhou2020informer} MSE & \textbf{38.1 $\pm$ 2.1} & 18.9 $\pm$ 0.6 & 13.2 $\pm$ 0.2 & 6.5 $\pm$ 0.3 & 2.1 $\pm$ 0.2 & 16.4 $\pm$ 0.5  \\
 & Informer \cite{zhou2020informer} DILATE & \textbf{37.8 $\pm$ 0.8} & \textbf{18.5 $\pm$ 0.3} & \textbf{12.9 $\pm$ 0.2} & \textbf{5.7 $\pm$ 0.2} & \textbf{1.9 $\pm$ 0.1}  & \textbf{15.9 $\pm$ 0.3} \\ 
   \midrule 
  \texttt{ETTH1} & N-Beats \cite{oreshkin2019n} MSE & 32.5 $\pm$ 1.4 & 3.9 $\pm$ 0.2 & 13.3 $\pm$ 2.0 & 21.6 $\pm$ 4.3 & \textbf{5.7 $\pm$ 0.7} & 7.4 $\pm$ 1.0  \\
   & N-Beats \cite{oreshkin2019n} DILATE &  \textbf{26.0 $\pm$ 2.8} & \textbf{2.9 $\pm$ 0.1} & \textbf{4.6 $\pm$ 0.6} & \textbf{11.4 $\pm$ 1.7} & \textbf{6.4 $\pm$ 1.0} & \textbf{4.6 $\pm$ 0.4}  \\
     \cdashline{2-8}
    & Informer \cite{zhou2020informer} MSE & \textbf{28.2 $\pm$ 2.6} & 4.3 $\pm$  0.3 & 5.8 $\pm$ 0.1 & 21.6 $\pm$ 3.3 & \textbf{6.6 $\pm$ 1.9} & 7.8 $\pm$ 0.9    \\
 & Informer \cite{zhou2020informer} DILATE & 32.5 $\pm$ 3.8 & \textbf{3.2 $\pm$ 0.3} & \textbf{4.5 $\pm$ 0.3} & \textbf{19.1 $\pm$ 1.9} & \textbf{6.4 $\pm$ 1.0} & \textbf{6.4 $\pm$ 0.6}    \\
 \bottomrule
    \end{tabular}
    \end{adjustbox}
    \label{tab:dilate_sota}
\end{table*}

\subsection{DILATE loss analysis \label{sec:dilate-analysis}}

\begin{figure}[H]
\centering
\begin{tabular}{cc}
    \includegraphics[height=6.1cm]{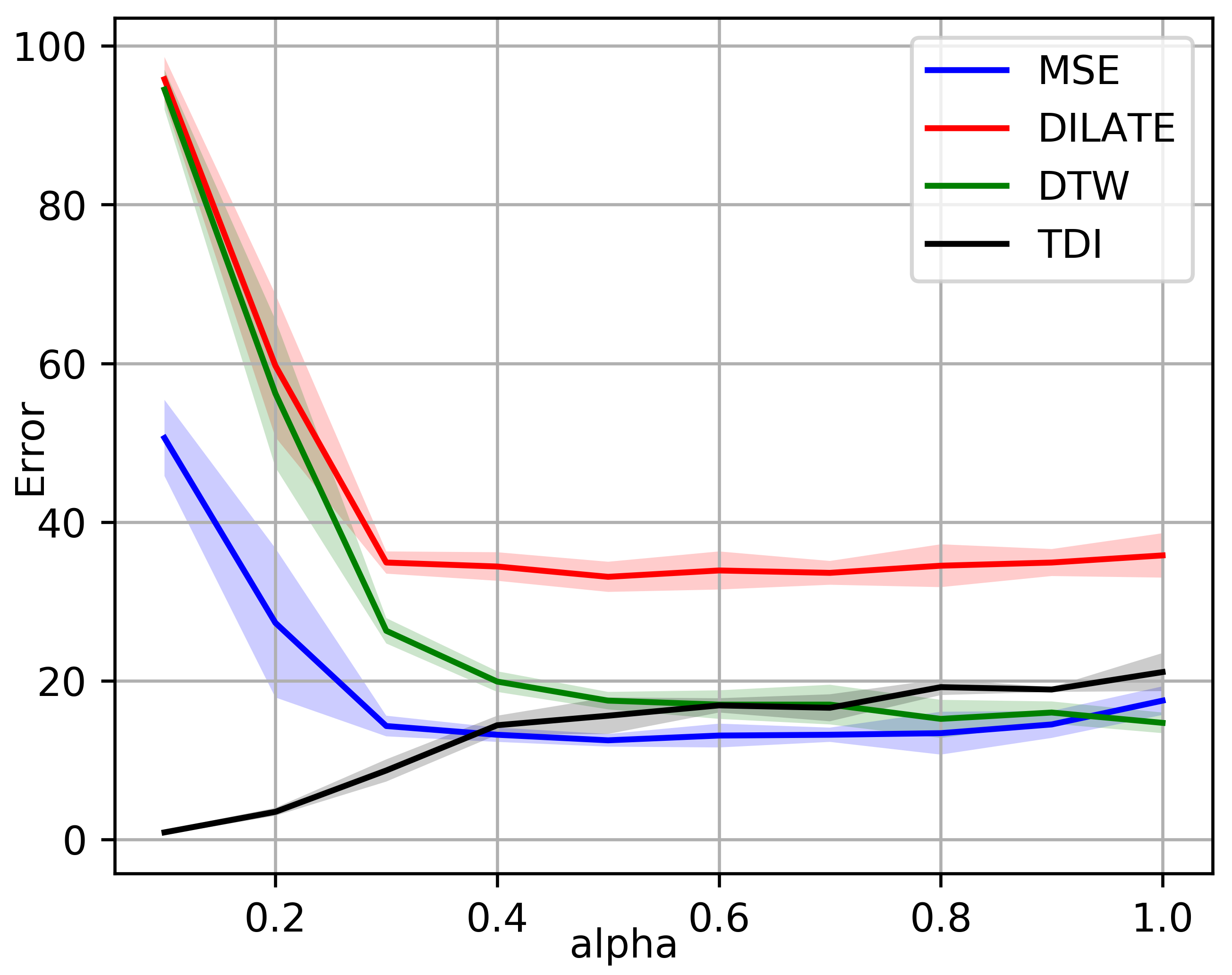} & 
    \includegraphics[height=6.1cm]{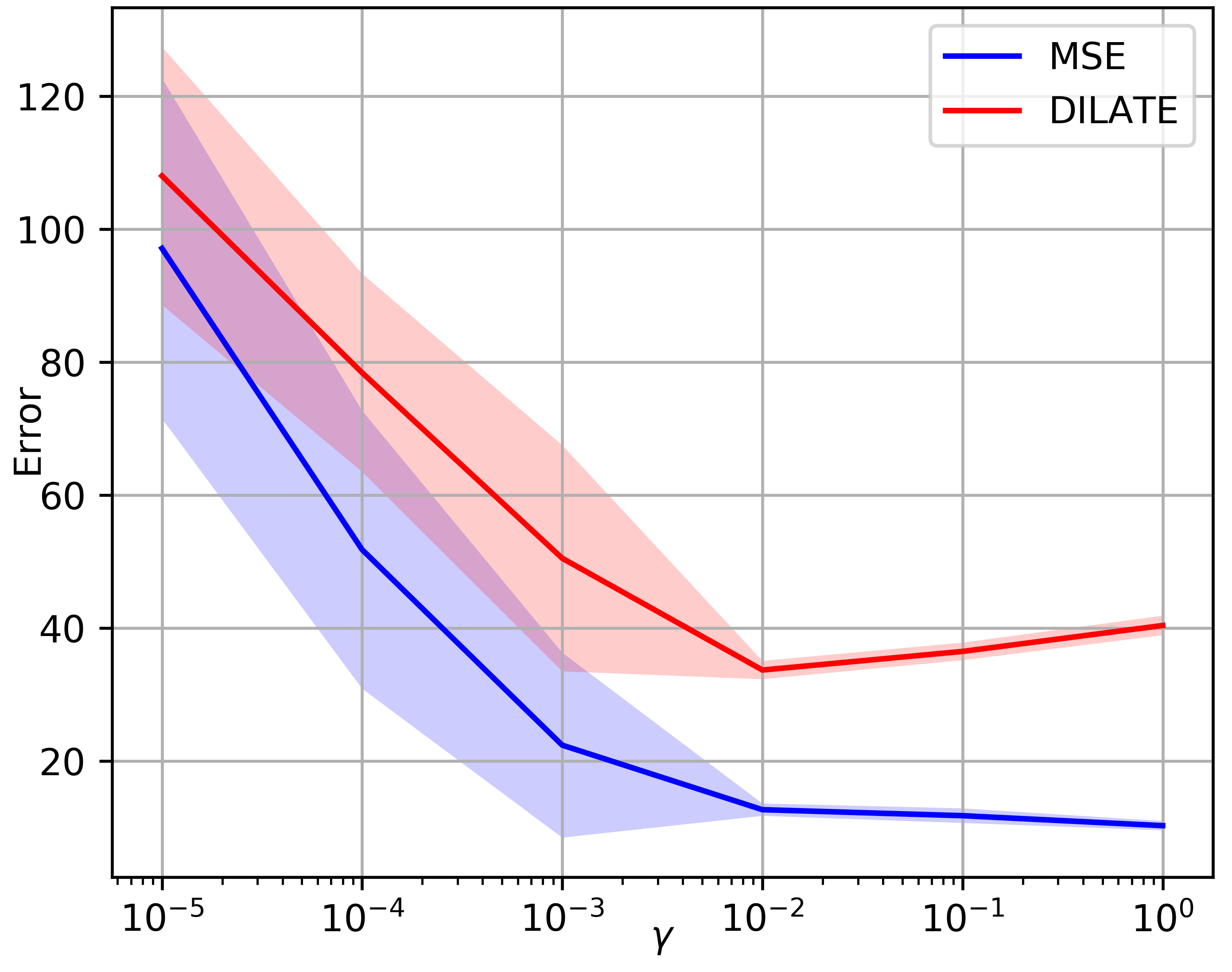}  \\
    (a) Influence of $\alpha$ & (b) Influence of $\gamma$
\end{tabular}
\begin{tabular}{c}
 \includegraphics[width=17cm]{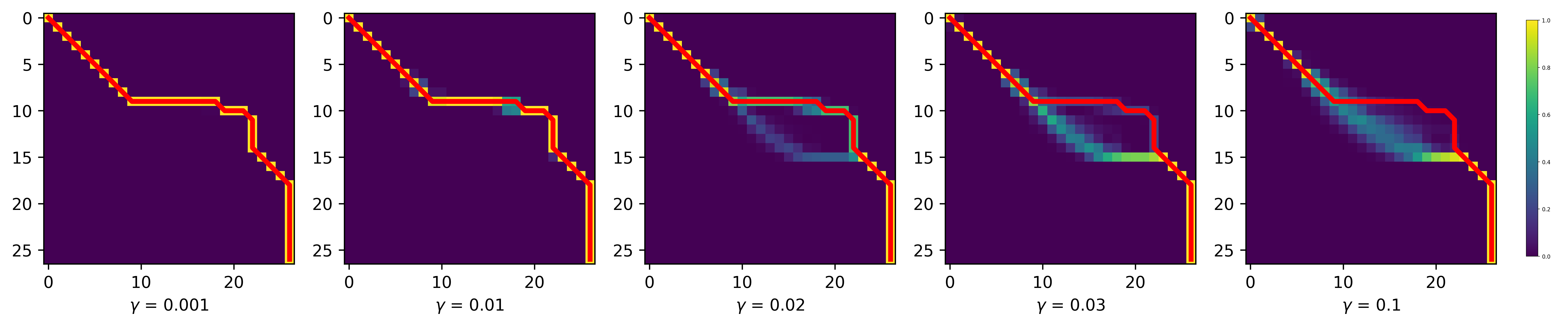} \\
   (c) Influence of $\gamma$ on the soft-DTW optimal path (true path in red)
\end{tabular}
    \caption[DILATE loss analysis.]{\textbf{DILATE loss analysis.} The shaded areas represent $\pm $ std computed over 10 runs.}
    \label{fig:dilate_analysis}
\end{figure}

\paragraph*{Influence of $\alpha$} We analyze in Figure \ref{fig:dilate_analysis} (a) the influence of the tradeoff parameter $\alpha$ when training a Seq2Seq model on the \texttt{Synthetic-det} dataset. When $\alpha=1$, $\mathcal{L}_{\text{DILATE}}$ reduces to $\text{DTW}_{\gamma}^{\mathbf{\Delta}}$, with an accurate shape but a large temporal error. When $\alpha \longrightarrow 0$, we only minimize  $\mathcal{L}_{time}$ without any shape constraint. Both MSE and shape errors explode in this case, illustrating the fact that $\mathcal{L}_{time}$ is only meaningful in conjunction with $\mathcal{L}_{shape}$. Both the MSE and DILATE error curves present a U-shape ; in this case, $\alpha=0.5$ seems an acceptable tradeoff for the \texttt{Synthetic-det} dataset. 

\paragraph*{Influence of $\gamma$} We analyse the influence of the $\text{DTW}_{\gamma}^{\mathbf{\Delta}}$ smoothing parameter $\gamma$ in Figure \ref{fig:dilate_analysis}. We show in Figure \ref{fig:dilate_analysis} (c) the assignment probabilities of the $\text{DTW}_{\gamma}^{\mathbf{\Delta}}$ path between the two test time series from Figure \ref{fig:dtw}, the true DTW path being depicted in red. When $\gamma$ increases, the  $\text{DTW}_{\gamma}^{\mathbf{\Delta}}$ path is more uncertain and becomes multimodal. When $\gamma \rightarrow 0$, the soft DTW converges toward the true DTW. However, we see in Figure \ref{fig:dilate_analysis} (b) that for small $\gamma$ values, optimizing $\text{DTW}_{\gamma}^{\mathbf{\Delta}}$  becomes more difficult, resulting in higher test error and higher variance (on  \texttt{Synthetic-det}). We fixed $\gamma=10^{-2}$ in all our experiments, which yields a good tradeoff between an accurate soft optimal path and a low test error.

\section{Conclusion}

In this Chapter, we have introduced DILATE, a new differentiable loss function for training deep multi-step time series forecasting models. DILATE combines two terms for precise shape and temporal localization of non-stationary signals with sudden changes. We showed that DILATE is comparable to the standard MSE loss when evaluated on MSE, and far better when evaluated on several shape and timing metrics. DILATE compares favourably on shape and timing to state-of-the-art forecasting algorithms trained with the MSE. 

\clearpage{\pagestyle{empty}\cleardoublepage}

\chapter{Probabilistic forecasting with shape and temporal diversity}
\label{chap:stripe}
\chapabstract{

\minitoc

\begin{center}
   \textsc{Chapter abstract}
\end{center}
\textit{
In this Chapter, we address the non-stationary time series forecasting problem in the probabilistic setting. To describe the predictive distribution, our goal is to provide a limited set of diverse and accurate scenarios in terms of shape and temporal localization. We introduce the STRIPE forecasting model for representing structured diversity based on shape and time features, ensuring both probable predictions while being sharp and accurate. STRIPE is a forecasting model which outputs multiple predictions by sampling latent variables. STRIPE is equipped with a diversification mechanism relying on determinantal point processes (DPP). Structured diversity is enforced with two shape and temporal semi-definite kernels. We use the two shape and time kernel of Chapter \ref{chap:criteria}, that we prove to be valid PSD kernels, for enforcing structured diversity. Experiments carried out on synthetic datasets show that STRIPE significantly outperforms baseline methods for representing diversity, while maintaining accuracy of the forecasting model. Finally, experiments on real datasets illustrate that STRIPE is able to outperform state-of-the-art probabilistic forecasting approaches in the best sample prediction.
The work described in this Chapter is based on the following publications:
\begin{itemize}
    \item \cite{leguen20stripe}:  Vincent Le Guen and Nicolas Thome. "Probabilistic Time Series Forecasting with Structured Shape and Temporal Diversity". In Advances in Neural Information Processing Systems (NeurIPS 2020).
    \item \cite{leguen2021deep}:  Vincent Le Guen and Nicolas Thome. "Deep Time Series Forecasting with Shape and Temporal Criteria". IEEE Transactions on Pattern Analysis and Machine Intelligence, 2022. 
\end{itemize}
}

Notation: we describe the STRIPE++ model from the publication \cite{leguen2021deep} that we rename STRIPE in this Chapter. It is an improvement over the STRIPE model published in \cite{leguen20stripe}.  
}

\section{Introduction}

\begin{figure}[H]
\begin{tabular}{cccc}
\hspace{-1cm}
 \includegraphics[height=4.8cm]{images/dilatestripe_fig1a.png}   & \hspace{-0.3cm} 
\includegraphics[height=4.8cm]{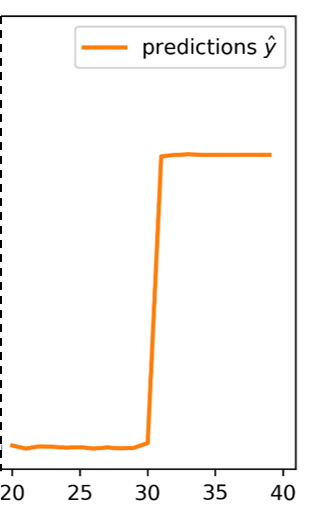}   &  
\hspace{-0.3cm} 
\includegraphics[height=4.8cm]{images/dilatestripe_fig1c.png}   &  
\hspace{-0.3cm} 
\includegraphics[height=4.8cm]{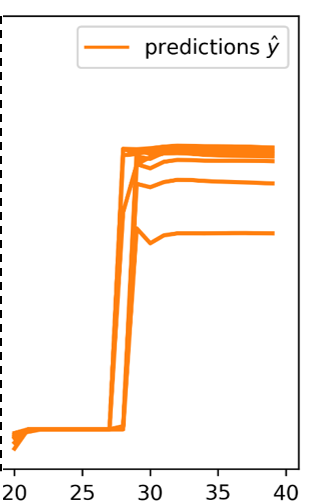}   \\
\hspace{-1.5cm}  (a) True predictive distribution  & \hspace{-0.3cm} (b) DILATE ~\cite{leguen19} & \hspace{-0.3cm} (c) deep stoch model \cite{yuan2019diverse} & \hspace{-0.3cm} (d) STRIPE (ours)  
\end{tabular}{}
    \caption[Probabilistic forecasting motivation.]{\textbf{Probabilistic time series forecasting}:  recent advances include the DILATE loss \cite{leguen19} for enabling sharp predictions (b), but are inadequate for producing diverse forecasts. On the other hand, probabilistic forecasting approaches based on generative models \cite{yuan2019diverse,rasul2020multi} loose the ability to generate sharp forecasts (c). The proposed STRIPE model (d) produces both sharp and diverse future forecasts, matching the ground truth distribution (a).}
    \label{fig:stripe_motivation}
\end{figure}

\lettrine[lines=3]{I}n many applications, producing deterministic forecasts, \ie a single future trajectory, is not sufficient for decision makers, who need information about the forecast's uncertainty. Probabilistic forecasting consists in modelling the conditional predictive distribution of future trajectories given past values. In this work, our goal is to describe this predictive distribution with a small set (\eg $N=10$) of plausible and diverse predictions. This is a different goal than estimating the variance of the predictions or the quantiles of the distribution. Focusing on the non-stationary context with possible sharp variations, the targeted set of predictions should reflect the shape and temporal diversity of the true future trajectories. Our motivation is illustrated in the example of the blue input in Figure \ref{fig:stripe_motivation} (a): we aim at performing predictions covering the full distribution of future trajectories, whose samples are shown in green. 

State-of-the-art methods for time series forecasting currently rely on deep neural networks, which exhibit strong abilities in modelling complex nonlinear dependencies between variables and time. Recently, increasing attempts have been made for improving architectures for accurate predictions \cite{lai2018modeling,sen2019think,li2019enhancing,oreshkin2019n,leguen20phydnet} or for making predictions sharper, \eg  by explicitly modelling dynamics~\cite{chen2018neural,dupont2019augmented,rubanova2019latent,franceschi2020stochastic}, or by designing specific loss functions addressing the drawbacks of blurred prediction with MSE training~\cite{cuturi2017soft,rivest2019new,leguen19,vayer2020time} (\eg with DILATE). Although Figure \ref{fig:stripe_motivation} (b) shows that DILATE produces sharp and realistic forecasts, its deterministic nature leads to to a single trajectory prediction without uncertainty quantification.

Probabilistic methods targeting for producing a diverse set of predictions include generative models \cite{yuan2019diverse,koochali2020if,rasul2020multi} that produce multiple trajectories by sampling from a latent space. These approaches are commonly trained using MSE or variants, and consequently often loose the ability to represent sharp predictions, as shown in Figure~\ref{fig:stripe_motivation} (c) for \cite{yuan2019diverse}. These generative models also lack an explicit structure to control the type of diversity in the latent space.

In this Chapter, we introduce the STRIPE model for including Shape and Time diverRsIty in Probabilistic forEcasting. As shown in Figure \ref{fig:stripe_motivation} (d), this enables to produce sharp and diverse forecasts, which fit well the ground truth distribution of trajectories in Figure \ref{fig:stripe_motivation} (a). STRIPE is a predictive model equipped with a diversification mechanism based on determinantal point processes (DPP). The diversity of predictions is structured with the two shape and temporal semi-definite kernels defined in Chapter \ref{chap:criteria}, and we design explicit schemes to control the quality vs.~ diversity tradeoff. 

We conduct experiments on synthetic datasets to evaluate the ability of STRIPE to match the ground truth trajectory distribution. We show that STRIPE significantly outperforms baseline methods for representing diversity, while maintaining the accuracy of the forecasting model. Experiments on real datasets further show that STRIPE is able to outperform state-of-the-art probabilistic forecasting approaches when evaluating the best sample (\ie diversity), while being equivalent based on its mean prediction (\ie quality).

\section{Related work}

In the Section, we pursue the review from Chapter \ref{chap:related_work} on spatio-temporal forecasting and insist on the most related works for probabilistic forecasting and for imposing structured diversity.

\paragraph{Probabilistic forecasting}
 For describing the conditional distribution of future values given an input sequence, a first class of deterministic methods add variance estimation with Monte Carlo dropout \cite{zhu2017deep,laptev2017time} or predict the quantiles of this distribution \cite{wen2017multi,gasthaus2019probabilistic,wen2019deep} by minimizing the pinball loss \cite{koenker2001quantile,romano2019conformalized} or the continuous ranked probability score (CRPS) \cite{gneiting2007probabilistic}. Other probabilistic methods try to approximate the predictive distribution, \textit{explicitly} with a parametric distribution (\eg Gaussian for DeepAR \cite{salinas2017deepar} and variants \cite{rangapuram2018deep,salinas2019high}), or \textit{implicitly} with a generative model with latent variables (\eg with conditional variational autoencoders (cVAEs) \cite{yuan2019diverse}, conditional generative adversarial networks (cGANs) \cite{koochali2020if}, normalizing flows \cite{rasul2020multi}). However, these methods lack the ability to produce sharp forecasts by minimizing variants of the MSE (pinball loss, gaussian maximum likelihood), at the exception of cGANs - but which suffer from mode collapse that limits predictive diversity. Moreover, these generative models are generally represented by unstructured distributions in the latent space (\eg Gaussian), which do not allow to have an explicit control on the targeted diversity.

\paragraph{Structured diversity for prediction}

For diversifying forecasts, several repulsive schemes were studied such as the variety loss \cite{gupta2018social,thiede2019analyzing} that consists in optimizing the best sample, or entropy regularization \cite{dieng2019prescribed,wang2019nonlinear} that encourages a uniform distribution. Besides, generative models, such as variational autoencoders (VAE) \cite{kingma2013auto}, are widely used for producing multiple predictions through sampling from a latent space. However latent states are typically sampled at test time from a standard Gaussian prior distribution, resulting in an unstructured diversity. To improve this unstructured mechanism, prior works \cite{yuan2019diverse,yuan2020dlow} introduced proposal neural networks for generating the latent variables that are trained with a diversity objective.

As discussed in Chapter \ref{chap:related_work}, determinantal point processes (DPPs) are an appealing mathematical solution for characterizing the diversity of a set of items. Efficient algorithms  maximizing the diversity of a set of items with a given sampling budget.
GDPP \cite{elfeki2018gdpp} proposed by Elfeki \etal is based on matching generated and true sample diversity by aligning the corresponding DPP kernels, and thus limits their use in datasets where the full distribution of possible outcomes is accessible. In contrast, our probabilistic forecasting approach is applicable in realistic scenarios where only a single future trajectory is available for each training sample. 
 Yuan and Kitani \cite{yuan2019diverse} train their proposal neural networks with a DPP diversity loss. Although we share with  \cite{yuan2019diverse} the goal to use DPP as diversification mechanism for future trajectories, the main limitation in~\cite{yuan2019diverse} is to use the MSE loss for training the predictor and the MSE kernel for  diversification, leading to blurred prediction, as illustrated in Figure~\ref{fig:stripe_motivation} (c). In contrast, we design specific shape and time DPP kernels and we show the necessity to decouple the criteria used for quality and diversity.

\begin{figure*}
    \centering
    \includegraphics[width=17cm]{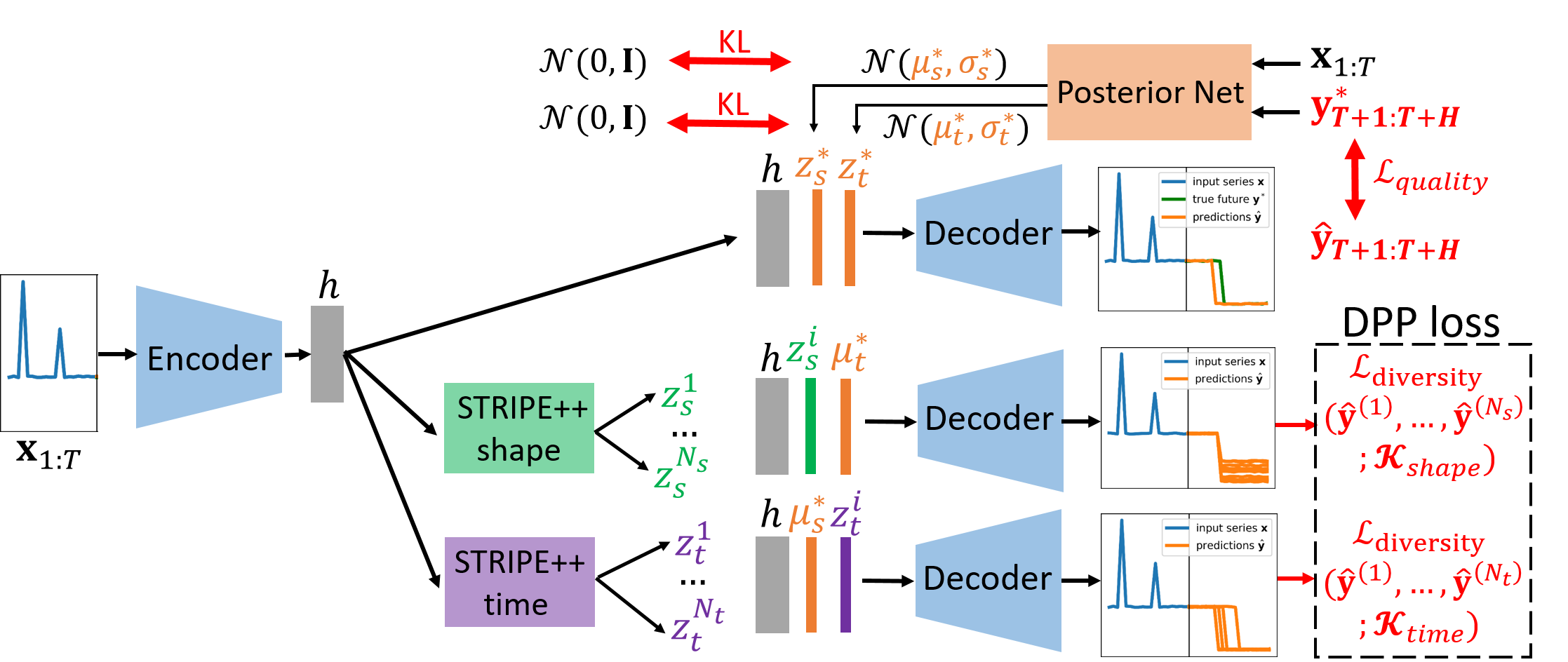}
    \caption[Overview of the STRIPE model]{\textbf{Overview of the STRIPE model:} STRIPE builds on a forecasting architecture trained with a quality loss $\mathcal{L}_{quality}$ enforcing sharp predictions.
    The latent state is disentangled into a deterministic part $h$ from the encoder and two stochastic codes $z_s$ and $z_t$ that account for the shape and time variations. First step (Figure upper part), we train the predictor with a quality loss, the stochastic codes are sampled from a posterior network. Second step (bottom), we diversify the predictions with two STRIPE shape and time proposal networks trained with a DPP diversity loss (keeping the encoder and decoder frozen).}
    \label{fig:stripe}
\end{figure*}

\section{Probabilistic forecasting with structured diversity}
\label{sec:stripe}

We consider the multi-step and non-stationary time series forecasting problem in the probabilistic case. Given an input sequence $\x_{1:T}=(\x_1,\dots,\x_T) \in \mathbb{R}^{p \times T}$, we aim at describing the conditional predictive distribution of future trajectories with a set  of $N$ future trajectories  $ \{ \hat{\y}^{(i)} \}_{i=1..N}  \in  \mathbb{R}^{d \times H}$ (corresponding to diverse scenarii sampled from the true future distribution $\hat{\y}^{(i)} \sim p(\cdot |\x_{1:T})$). 

We introduce the STRIPE framework (Shape and Time diverRsIty in Probabilistic
forEcasting), that extends STRIPE \cite{leguen20stripe}. Depicted in Figure \ref{fig:stripe}, STRIPE builds upon a general multi-step forecasting pipeline: the input time series $\x_{1:T}$ is fed into an encoder that summarizes the input into a latent vector $h$. This context vector $h$ is then transformed by a decoder into a future trajectory. 

The key idea of STRIPE is to augment the deterministic latent state $h$ with stochastic diversifying variables $z_s$ (resp. $z_t$) meant to capture the shape (resp. temporal) variations of the future time series. We distinguish two phases for training the overall model: (i) we train the predictor with a quality loss and (ii) we train the diversifying STRIPE mechanism with a DPP diversity loss (with the weights of the predictor frozen). For both of these steps, we detail now how the diversifying variables are sampled.

\subsection{Training the predictor with a quality loss}
For training the predictor (upper part in Figure \ref{fig:stripe}) with possibly multiple admissible futures as supervision, we take inspiration from the probabilistic U-Net \cite{kolh-probunet} and introduce a posterior network from which to sample the diversifying variables $z_s^*$ and $z_t^*$ (which represent the shape and temporal variant attached to a particular future $\y^*$). The posterior net outputs the parameters $\mu_s^*$ and $\sigma_s^*$ of a  Gaussian distribution  $\mathcal{N}(\mu_s^*,\sigma_s^*)$ for parameterizing  the shape posterior distribution  $q(z_s | \x,\y^*)$ (and similarly for the temporal posterior distribution).

To train this generative model (encoder, decoder and posterior networks), we resort to variational inference \cite{kingma2013auto} and maximize the evidence lower bound (ELBO) of the log-likelihood, or equivalently, minimize the following prediction loss over all training examples:
\begin{equation}
    \mathcal{L}_{prediction}(\hat{\y},\y^*) = \mathcal{L}_{quality}(\hat{\y},\y^*) \; + \\  \text{KL} \left( q(z_s | \x,\y^*)\;||\;p(z_s) \right) + \text{KL}\left( q(z_t | \x,\y^*)\;||\;p(z_t) \right).
\end{equation}
In our non-stationary context, we choose the DILATE loss for $\mathcal{L}_{quality}$, in order to guarantee sharp predictions with accurate temporal localization. The Kullback-Leibler (KL) losses enforce that the shape posterior distribution $q(z_s | \x,\y^*)$ matches a prior distribution $p(z_s)$ (we use a Gaussian prior $\mathcal{N}(0,\mathbf{I})$, which is a common choice in variational inference).

\subsection{Training the STRIPE diversification mechanism}

For including structured shape and temporal diversity (lower part in Figure \ref{fig:stripe}), we introduce two proposal neural networks STRIPE$_{\text{shape}}$ and STRIPE$_{\text{time}}$ that aim to produce a set of $N_s$ shape latent codes $\left\{z_s^i\right\}_{{i=1..N_s}} \in \mathbb{R}^k$ (resp. $N_t$ time codes  $\left\{z_t^i\right\}_{{i=1..N_t}} \in \mathbb{R}^k$) dedicated to generate diverse trajectories in terms of shape (resp. time). 

When training STRIPE$_{\text{shape}}$ (the description for STRIPE$_{\text{time}}$ is similar), we concatenate $h$ with the posterior time latent code $\mu_t^*$ and the $N_s$ shape latent codes $z_s^i$ provided by STRIPE$_{\text{shape}}$, which leads to $N_s$ future trajectories $\hat{\y}^{i} = \text{Decoder}\left( (h, z_s^i , \mu_t^*) \right)$, $i=1..N_s$\footnote{If there exists multiple futures as supervision, we repeat this operation for each posterior latent code $\mu_t^{*,j}$ (it corresponds to consider each tuple $(\x_{1:T},\y^{*,j})$ as a separate training example).}. The shape diversity of this set of $N_s$ trajectories is then enforced by a shape diversity loss that we describe below.\\

\paragraph*{DPP diversity loss:} We resort to determinantal point processes (DPP) for their appealing properties for maximizing the diversity of a set of items $\mathcal{Y} = \left\{  \y_1,...,\y_N \right\}$ given a fixed sampling budget $N$ and for structuring diversity via the choice of the DPP kernel. Following \cite{yuan2019diverse}, we minimize the negative expected cardinality of a random subset $Y$ from the DPP:
\begin{align}
    \mathcal{L}_{diversity}(\mathcal{Y} ; \mathbf{K}) &= -\mathbb{E}_{Y \sim \text{DPP}(\mathcal{K})} |Y| \\ &= - Tr(\mathbf{I}-(\mathbf{K}+\mathbf{I})^{-1}).
    \label{eq:ldiversity}
\end{align}{}
Intuitively, a larger expected cardinality means a more diverse sampled set according to kernel $\mathcal{K}$. This loss is differentiable and can be computed in closed form.\\

\paragraph*{Quality regularizer in the DPP:} When training the shape and time proposal networks with the diversity loss, we do not have control over the quality of predictions, which can deteriorate to improve diversity. To address this, we introduce a quality regularization term in the DPP kernels. Crucially, we decouple the criteria used for quality (DILATE) and diversity (shape or time). $\mathcal{K}_{shape}$ maximizes the shape (DTW) diversity, while maintaining a globally low DILATE loss (thus playing on the temporal localization to ensure a good tradeoff). This contrasts with \cite{yuan2019diverse} which uses the same MSE criterion for both quality and diversity (see Figure \ref{fig:stripe_analysis} (b) for a detailed analysis). In practice,  we introduce a quality vector $\mathbf{q}= (q_1,\dots,q_{N_s})$ between the prediction $\hat{\y}^i$ and the ground truth $\y^*$ \footnote{If there are multiple futures as supervision, we again consider each tuple (input sequence, possible future) as a separate training example.}. We choose $q_i = \mu  (1 - \text{DILATE}(\hat{\y}^i, \y^*))$, where $\mu > 0$ is a hyperparameter to tune the influence of the quality regularization. The modified shape kernel becomes (and similarly for the time kernel):
\begin{equation}
\Tilde{\textbf{K}}_{shape} = \text{Diag}(\textbf{q}) ~ \textbf{K}_{shape} ~ \text{Diag}(\textbf{q}).
\label{eq:kshape-tilde}
\end{equation}
This decomposition enables to sample sets of items of both high quality and diversity:
\begin{equation}
    \mathcal{P}_{\mathbf{\Tilde{K}}}(\mathbf{Y}=Y) \propto \left( \prod_{i \in Y} q_i^2  \right) \det(\mathbf{K}_Y).
\end{equation}{}

 We then train STRIPE$_{\text{shape}}$ by applying the shape kernel $\Tilde{\textbf{K}}_{shape}$  (Eq \ref{eq:kshape-tilde}) to the set of $N_s$ shape future trajectories $\mathcal{L}_{diversity}(\hat{\y}^{1},\dots,\hat{\y}^{N_s} ; \Tilde{\mathbf{K}}_{shape})$ and STRIPE$_{\text{time}}$ by applying the  time kernel $\Tilde{\textbf{K}}_{time}$ to the set of $N_t$ time future trajectories $\mathcal{L}_{diversity}(\hat{\y}^{1},\dots,\hat{\y}^{N_t} ; \Tilde{\mathbf{K}}_{time})$.

 \subsection{Diverse trajectory generation at test time}
 
 At test time, the posterior network is discarded and we only rely on the trained encoder, STRIPE$_{\text{shape}}$, STRIPE$_{\text{time}}$ proposal networks and decoder to generate future predictions. More precisely, we combine the shape and temporal proposals $\left\{ z_s^i \right\}_{i=1..N_s}$ and $\left\{ z_t^j \right\}_{j=1..N_t}$  to obtain $N_s \times N_t$ predictions $\hat{\y}^{i,j} =  \text{Decoder}((h,z_s^i,z_t^j))$.

\section{Experiments \label{sec:stripe_expe}}

We firstly assess the ability of STRIPE to capture the full predictive distribution of future trajectories. To do so, we need for evaluation the ground truth set of admissible futures for a given input; we construct here the \texttt{Synthetic-prob} dataset designed for this purpose. Secondly, on a more realistic setting where we only know one future for each input, we evaluate STRIPE on the \texttt{Traffic} and \texttt{Electricity} datasets with the best (resp. the mean) sample metrics as a proxy for diversity (resp. quality). We describe the implementation details and neural network architectures (encoder, decoder, posterior net and STRIPE proposal network) in Appendix \ref{app:stripe}.

\subsection{Full predictive distribution evaluation on \texttt{Synthetic-prob}}
\label{sec:stripe-synth}

\paragraph*{Dataset:} In this Chapter, we build the \texttt{Synthetic-prob} ($T=20, H=20$) dataset with multiple admissible futures for each input series. This is a variant of \texttt{Synthetic-det} used in Chapter \ref{chap:dilate} where for each input series, we generate 10 different future series of length 20 by adding noise on the step amplitude and localization. A sample from this dataset can be observed in Figure \ref{fig:stripe_motivation} (a). The dataset is composed of $100 \times 10=1000$ time series for each train/valid/test split.

\paragraph*{Metrics:} To assess the discrepancy between the predicted and true distributions of futures trajectories, we define the two following measures $\text{H}_{quality}(\ell)$ and $\text{H}_{diversity}(\ell)$ ($\ell = $ DTW, TDI or DILATE in our experiments):
\begin{align}
        \text{H}_{quality}(\ell) &:= \mathbb{E}_{\x \in \mathcal{D}_{test}} \mathbb{E}_{\hat{\y}}  \left[  \underset{\y \in F(\x)}{\inf} \: \ell(\hat{\y},\y) \right]   \\
         \text{H}_{diversity}(\ell) &:= \mathbb{E}_{\x \in \mathcal{D}_{test}} \mathbb{E}_{\y \in F(\x)}  \left[  \underset{\hat{\y}}{\inf} \: \ell(\hat{\y},\y) \right] .
\end{align}
$\text{H}_{quality}$ penalizes forecasts $\hat{\y}$ that are far away from a ground truth future of $\x$ denoted $\y \in F(\x)$ (similarly to the \textit{precision} concept in pattern recognition) whereas $\text{H}_{diversity}$ penalizes when a true future is not covered by a forecast (similarly to \textit{recall}). As a tradeoff balancing quality and diversity, we compute the F1 score defined in Eq \ref{eq:F1score}:
\begin{equation}
             F1 \text{ score} = \frac{2 ~ \text{H}_{quality}(\ell)  \cdot  \text{H}_{diversity}(\ell) }{ \text{H}_{quality}(\ell) +  \text{H}_{diversity}(\ell)} \label{eq:F1score}.
\end{equation}
In addition, we also use the continuous ranked probability score (CRPS)  which is a standard \textit{proper scoring rule} \cite{gneiting2007probabilistic} for assessing probabilistic forecasts \cite{gasthaus2019probabilistic}. Intuitively, the CRPS is the pinball loss integrated over all quantile levels. A key property is that the CRPS attains its minimum when the predicted future distribution equals the true future distribution, making this metric particularly adapted to our context.

\paragraph*{Forecasting results:} We compare in Table \ref{tab:stripe} our method to 4 recent competing diversification mechanisms (variety loss \cite{thiede2019analyzing}, entropy regularisation \cite{dieng2019prescribed}, diverse DPP \cite{yuan2019diverse} and GDPP \cite{elfeki2018gdpp}) based on a conditional variational autoencoder (cVAE) backbone trained with DILATE. We observe that STRIPE obtains the global best performances by improving diversity by a large amount ($\text{H}_{diversity}(\text{DILATE)}$=17.9) compared to the backbone cVAE DILATE ($\text{H}_{diversity}(\text{DILATE)}$=33.9) and to other diversification schemes (the best competitor GDPP \cite{elfeki2018gdpp} attains $\text{H}_{diversity}(\text{DILATE)}$=23.9).
This highlights the relevance of the structured shape and time diversity. We can also notice that, in contrast to competing diversification schemes that improve diversity at the cost of a loss in quality, STRIPE maintains high quality predictions. STRIPE is only beaten in $\text{H}_{quality}(\text{DILATE)}$ by GDPP \cite{elfeki2018gdpp}, but this method is significantly worse than STRIPE in diversity, and GDPP requires full future distribution supervision, which it not applicable in real datasets (see section \ref{sec:stripe_real_datasets}). All in all, the F1 scores summarize the quality vs.~ diversity tradeoffs, and STRIPE gets the best F1 DILATE score. Moreover, STRIPE outperforms all other methods with the CRPS metric, indicating that the predicted future trajectory distribution is closer to the ground truth one.

\begin{table*}[t]
        \caption[STRIPE forecasting results on the \texttt{Synthetic-prob} dataset.]{\textbf{STRIPE forecasting results on the \texttt{Synthetic-prob} dataset with multiple futures}, averaged over 5 runs (mean $\pm$ std). Best equivalent methods (Student t-test) shown in bold. Metrics are scaled (MSE $\times$ 1000, DILATE $\times 100$, CRPS $\times$ 1000).}   
    \begin{adjustbox}{max width=\linewidth}        
    \begin{tabular}{ccccccccccc}
       \toprule 
   \multicolumn{1}{c}{}  & \multicolumn{3}{c}{$\text{H}_{quality}(\cdot) \; (\downarrow)$} & \multicolumn{3}{c}{$\text{H}_{diversity}(\cdot) \; (\downarrow)$} & \multicolumn{3}{c}{$ F1 \text{ score} \; (\downarrow)$}   &  \multicolumn{1}{c}{CRPS ($\downarrow$)} \\ 
       \cmidrule(lr){2-4}   \cmidrule(lr){5-7}   \cmidrule(lr){8-10} 
 Methods   & DTW & TDI   & DILATE  & DTW & TDI  & DILATE & DTW & TDI & DILATE \\ 
        \midrule
  cVAE DILATE   & \textbf{11.7 $\pm$ 1.5}  &  9.4 $\pm$ 2.2   & \textbf{14.2 $\pm$ 1.5} & 18.8 $\pm$ 1.3 & 48.6 $\pm$ 2.2   &   33.9 $\pm$ 3.9 &  14.4  &  15.7  & 20.0  & 62.2 $\pm$ 4.2  \\
   variety loss \cite{thiede2019analyzing} DILATE  & 15.6 $\pm$ 3.4  & 10.2 $\pm$ 1.1  & 16.8 $\pm$ 0.9 & 22.7 $\pm$ 4.1  & 37.7 $\pm$ 4.9   & 30.8 $\pm$ 1.0 & 18.5  & 16.1 & 21.7  & 62.6 $\pm$ 3.0  \\
    Entropy reg. \cite{dieng2019prescribed} DILATE  & 13.8 $\pm$ 3.1  & 8.8 $\pm$ 2.2  & \textbf{15.0 $\pm$ 1.6} & 20.4 $\pm$ 2.8  & 42.0 $\pm$ 7.8  & 32.6 $\pm$ 2.3 & 16.5  & 14.5 & 20.5 & 62.4 $\pm$ 3.9   \\
  Diverse DPP \cite{yuan2019diverse} DILATE  &  \textbf{12.9 $\pm$ 1.2}  & 9.8 $\pm$ 2.1  & 15.1 $\pm$ 1.5 & 18.6 $\pm$ 1.6 &  42.8 $\pm$ 10.1 & 31.3 $\pm$ 5.7 & 15.2 & 15.9 & 20.4 & 60.7 $\pm$ 1.6 \\
 GDPP  \cite{elfeki2018gdpp} DILATE   &  14.8 $\pm$ 2.9  & 11.7 $\pm$ 8.4  &    \textbf{14.4 $\pm$ 2.1} & 20.8 $\pm$ 2.4  & 25.2 $\pm$ 7.2  &   23.9 $\pm$ 4.5  & 17.3  & 15.9 & 17.9 & 63.4 $\pm$ 6.4  \\
STRIPE  & 13.5 $\pm$ 0.5 & 9.2 $\pm$ 0.5 & \textbf{15.0 $\pm$ 0.3} & \textbf{12.9 $\pm$ 0.3} & 16.3 $\pm$ 1.2 & \textbf{17.9 $\pm$ 0.6} & \textbf{13.2} & 11.7 & \textbf{16.3} & \textbf{48.6 $\pm$ 0.6}   \\
\bottomrule
    \end{tabular}
    \end{adjustbox}
    \label{tab:stripe}
\end{table*}{}

\subsection{State-of-the-art comparison on real-world datasets}
\label{sec:stripe_real_datasets}

\begin{table*}
\caption[STRIPE probabilistic forecasting results on \textsc{Traffic} and \textsc{Electricity}.]{\textbf{Probabilistic forecasting results on the \texttt{Traffic} and \texttt{Electricity} datasets}, averaged over 5 runs (mean $\pm$ std). Metrics are scaled for readability. Best equivalent method(s) (Student t-test) shown in bold.} 
   \label{tab:stripe_sota}    
 \centering              
\setlength{\tabcolsep}{6.8pt}
    \begin{adjustbox}{max width=\linewidth}
    \begin{tabular}{ccccc|cccc}
    \toprule
    \multicolumn{1}{c}{} & \multicolumn{4}{c|}{\texttt{Traffic}} & \multicolumn{4}{c}{\texttt{Electricity}}   \\
    \multicolumn{1}{c}{} & \multicolumn{2}{c}{MSE} &  \multicolumn{2}{c|}{DILATE}   & \multicolumn{2}{c}{MSE} &  \multicolumn{2}{c}{DILATE}  \\ 
         \cmidrule(lr){2-3}   \cmidrule(lr){4-5}      \cmidrule(lr){6-7}   \cmidrule(lr){8-9} 
 Method   & mean   & best & mean & best   & mean & best & mean  & best   \\ 
 \midrule
 Nbeats \cite{oreshkin2019n} MSE  &  - & 7.8 $\pm$ 0.3 & - & 22.1 $\pm$ 0.8  & - & 24.8 $\pm$ 0.4 & - & 20.2 $\pm$ 0.3    \\
 Nbeats \cite{oreshkin2019n} DILATE  & - & 17.1 $\pm$ 0.8 & - & 17.8 $\pm$ 0.3  & - & 25.8 $\pm$ 0.9 & - & 19.9 $\pm$ 0.5 \\
  \midrule
   Deep AR \cite{salinas2017deepar} & 15.1 $\pm$ 1.7 & \textbf{6.6 $\pm$ 0.7} & 30.3 $\pm$ 1.9 & 16.9 $\pm$ 0.6  &   67.6 $\pm$ 5.1  & 25.6 $\pm$ 0.4 & 59.8 $\pm$ 5.2  & 17.2 $\pm$ 0.3    \\
   cVAE DILATE & \textbf{10.0 $\pm$ 1.7}  & 8.8 $\pm$ 1.6  & \textbf{19.1 $\pm$ 1.2} & 17.0 $\pm$ 1.1   &  \textbf{28.9 $\pm$ 0.8}    &  27.8 $\pm$ 0.8 & 24.6 $\pm$ 1.4   & 22.4 $\pm$ 1.3   \\
   Variety loss \cite{thiede2019analyzing} & \textbf{9.8 $\pm$ 0.8} & 7.9 $\pm$ 0.8  & \textbf{18.9 $\pm$ 1.4}  & 15.9 $\pm$ 1.2  & 29.4 $\pm$ 1.0  & 27.7 $\pm$ 1.0  & 24.7 $\pm$ 1.1  & 21.6 $\pm$ 1.0    \\
   Entropy regul. \cite{dieng2019prescribed} & 11.4 $\pm$ 1.3   & 10.3 $\pm$ 1.4  & \textbf{19.1 $\pm$ 1.4}  & 16.8 $\pm$ 1.3  & 34.4 $\pm$ 4.1  &  32.9 $\pm$ 3.8  & 29.8 $\pm$ 3.6  & 25.6 $\pm$ 3.1  \\
   Diverse DPP \cite{yuan2019diverse}  & 11.2 $\pm$ 1.8  & 6.9 $\pm$ 1.0   & 20.5 $\pm$ 1.0  & 14.7 $\pm$ 1.0   &  31.5 $\pm$ 0.8  & 25.8 $\pm$ 1.3  & 26.6 $\pm$ 1.0  & 19.4 $\pm$ 1.0    \\ 
  STRIPE & \textbf{10.0 $\pm$ 0.2} & \textbf{6.7 $\pm$ 0.3} & \textbf{19.0 $\pm$ 0.2} & \textbf{14.1 $\pm$ 0.3}  & \textbf{29.5 $\pm$ 0.3} & \textbf{23.6 $\pm$ 0.4} & \textbf{24.1 $\pm$ 0.2}  & 17.3 $\pm$ 0.4  \\
  \bottomrule
    \end{tabular}
    \end{adjustbox}
\end{table*}{}

We evaluate here the performances of STRIPE on the two challenging real-world datasets \texttt{Traffic} and \texttt{Electricity} commonly used as benchmarks in the time series forecasting literature \cite{yu2016temporal,salinas2017deepar,lai2018modeling,rangapuram2018deep,leguen19,sen2019think} and described in Chapter \ref{chap:dilate}. Contrary to the \texttt{Synthetic-prob} dataset, we only dispose of one future trajectory sample $\y^{*}_{T+1:T+\tau}$ for each input series $\x_{1:T}$. In this case, the metric $\text{H}_{quality}$ (resp. $\text{H}_{diversity}$) defined in section \ref{sec:stripe-synth} reduces to the mean sample (resp. best sample), which are common for evaluating stochastic forecasting models \cite{babaeizadeh2017stochastic,franceschi2020stochastic}. 

Results in Table \ref{tab:stripe_sota} reveal that STRIPE outperforms all other baselines in the best sample (evaluated in MSE or DILATE). Our method even outperforms in the best sample the state-of-the-art N-Beats algorithm \cite{oreshkin2019n} (either trained with MSE or DILATE), which is dedicated to producing high quality deterministic forecasts. In terms of quality (evaluation with the mean sample), STRIPE gets the best (or equivalently best) results in all cases. This contrasts to competing diversification methods, \eg Diverse DPP \cite{yuan2019diverse}, that deteriorate quality to improve diversity. Finally we notice that STRIPE is consistently better in diversity and quality than the state-of-the art probabilistic deep AR method \cite{salinas2017deepar}.

\begin{figure*}
\centering
\begin{tabular}{cc}
   \includegraphics[width=8cm]{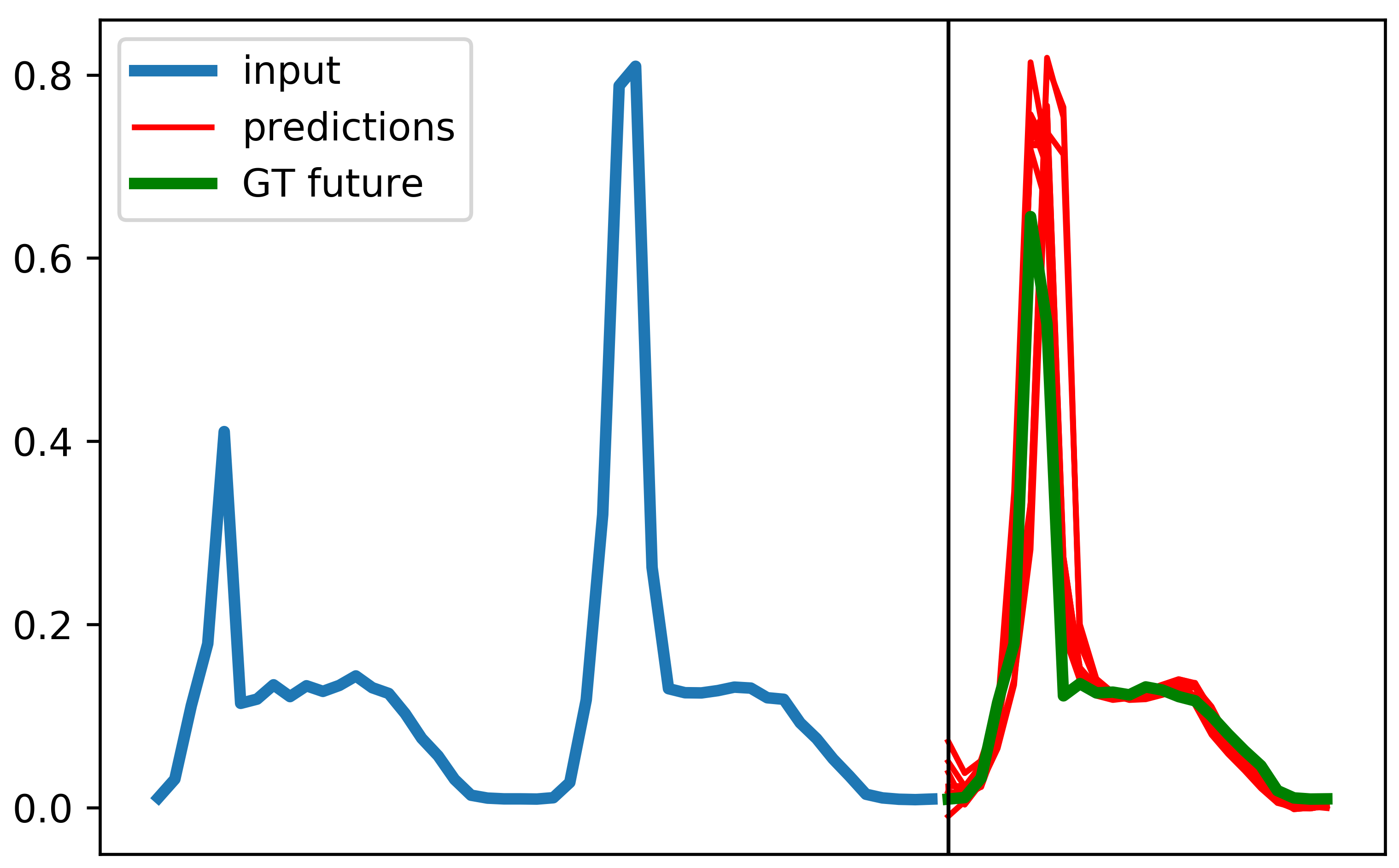}   &    \includegraphics[width=8cm]{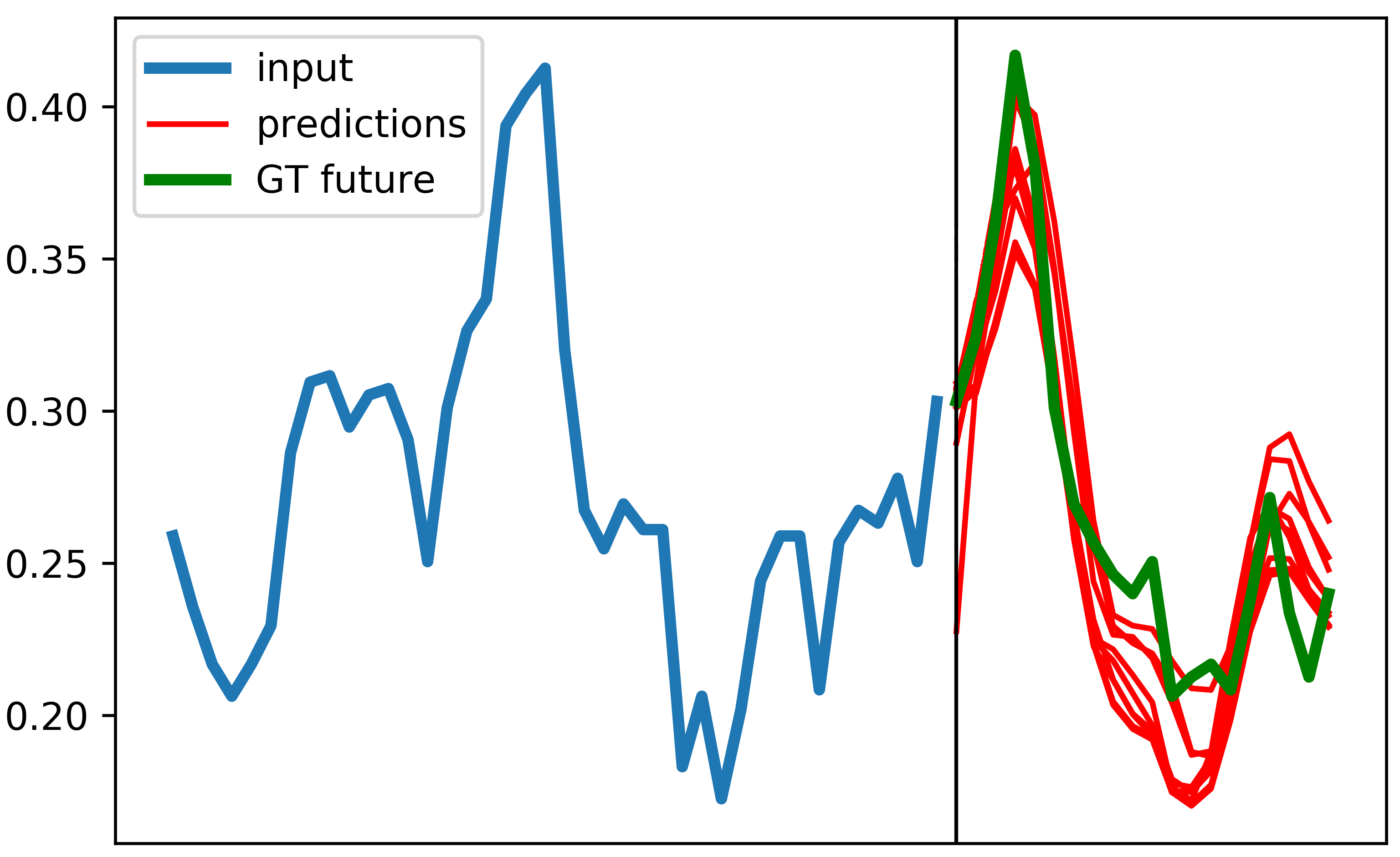} \\
   (a) \texttt{Traffic} & (b) \texttt{Electricity}
\end{tabular}
    \caption[STRIPE qualitative predictions on Traffic and Electricity.]{STRIPE qualitative predictions on datasets \texttt{Traffic} (a) and \texttt{Electricity} (b).}
    \label{fig:stripe_visus}
\end{figure*}

We display a few qualitative forecasting examples of STRIPE on Figure \ref{fig:stripe_visus}. We observe that STRIPE predictions are both sharp and accurate: both the shape diversity (amplitude of the peaks) and temporal diversity match the ground truth future.

\subsubsection{STRIPE analysis: quality-diversity cooperation}

\begin{figure}[H]
\centering
    \begin{tabular}{c|c}
    \includegraphics[width=7.5cm]{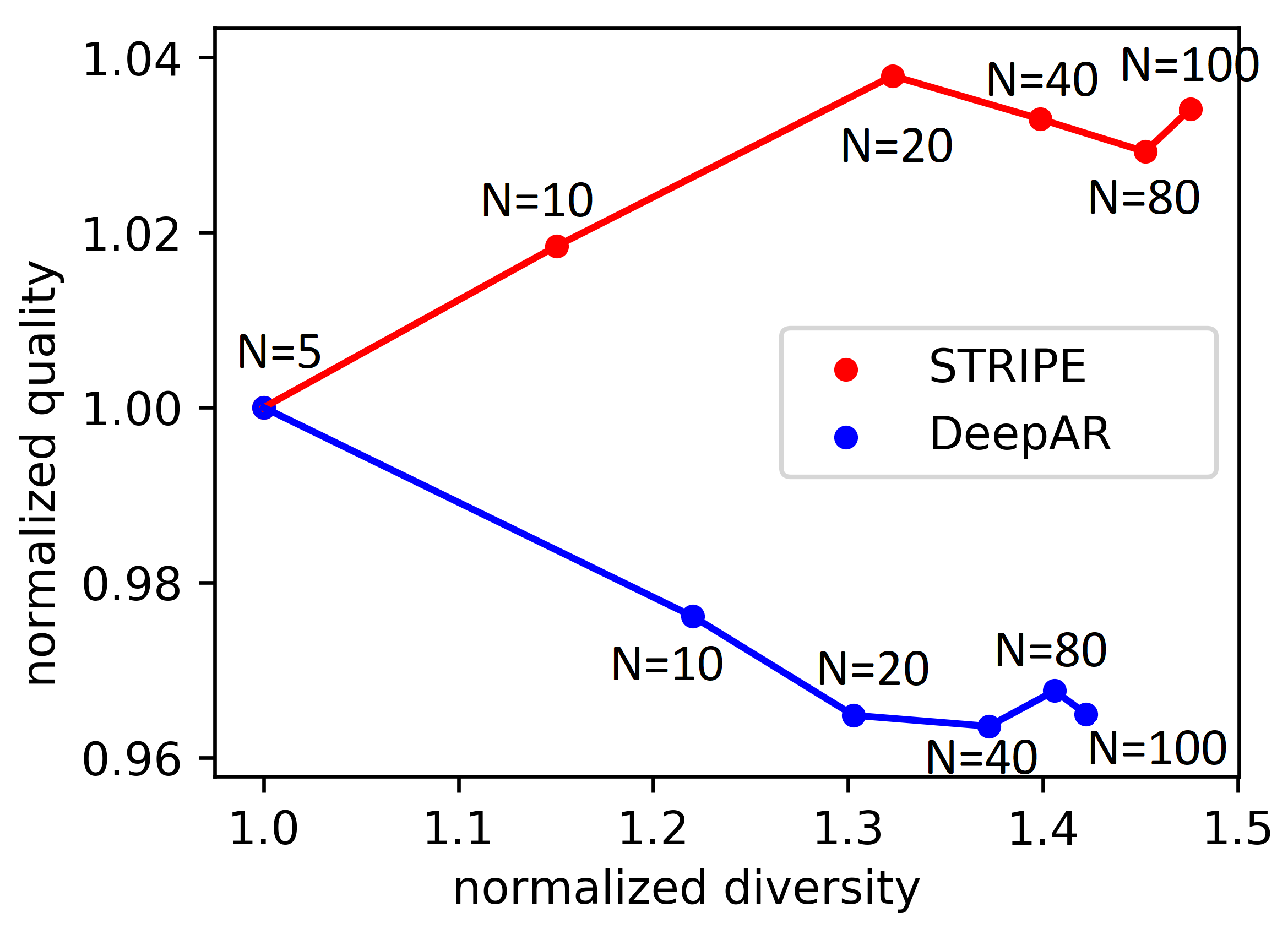}  &  \includegraphics[width=6cm]{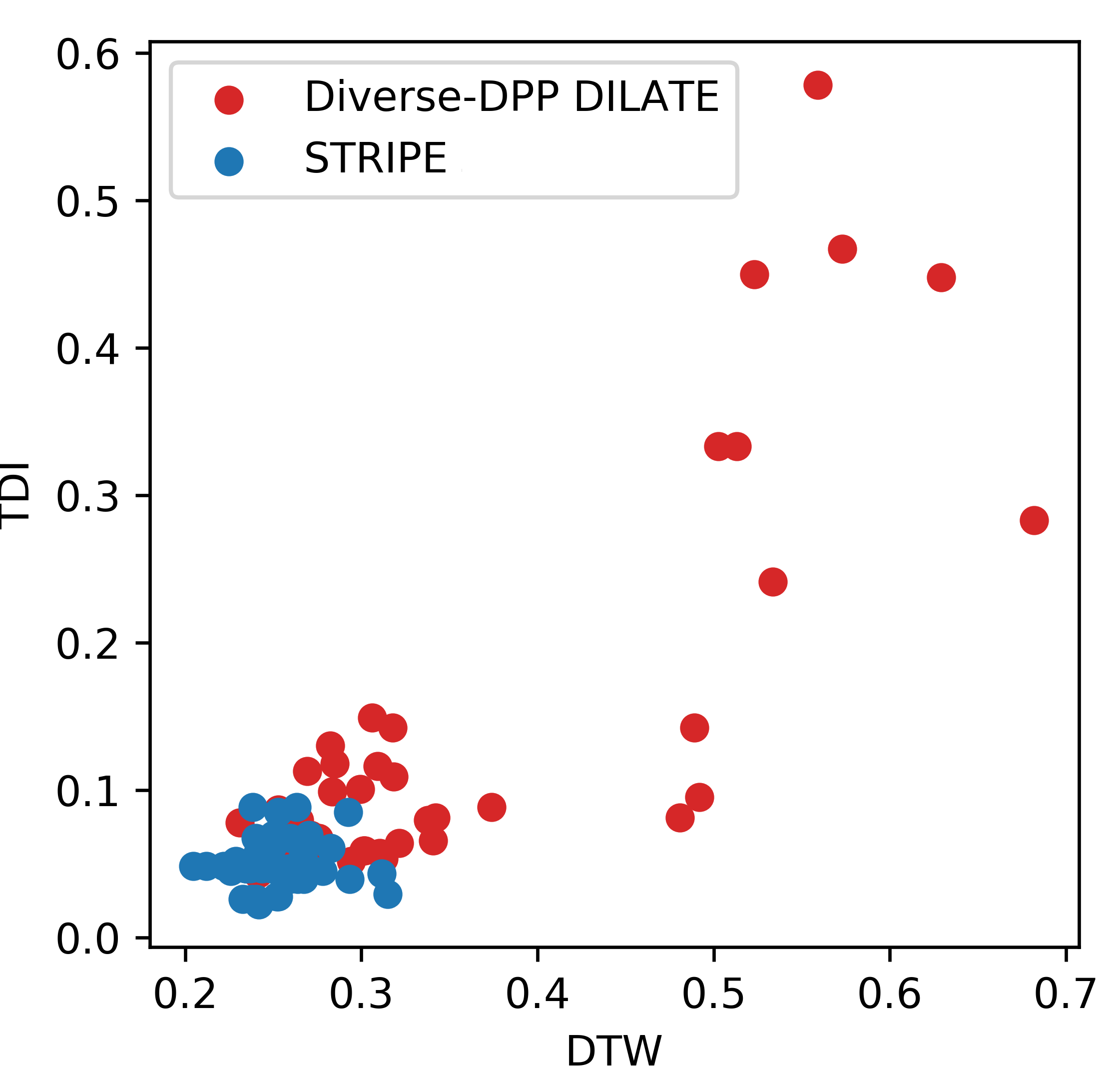}  \\
     (a)    & (b)
    \end{tabular}

\caption[STRIPE analysis.]{\textbf{STRIPE analysis:} (a) Influence of the number $N$ of trajectories on quality (higher is better) and diversity for the \texttt{Synthetic-prob} dataset. (b) Scatterplot of 50 predictions in the plane (DTW,TDI), comparing STRIPE v.s. Diverse DPP DILATE \cite{yuan2019diverse}.}    
    \label{fig:stripe_analysis}
\end{figure}

We analyze here the quality-diversity tradeoff with respect to the  number $N$ of sampled future trajectories. In Figure \ref{fig:stripe_analysis} (a) we represent the evolution of performances when $N$ increases from 5 to 100 on the synthetic-prob dataset. As expected, the normalized DILATE diversity  $\text{H}_{diversity}(5)/\text{H}_{diversity}(N)$ (higher is better) increases with $N$ for both STRIPE and deepAR models \cite{salinas2017deepar}. However we remark that STRIPE does not deteriorate normalized quality (which even increases slightly), in contrast to deepAR which does not have control over the targeted diversity. This again confirms the relevance of our approach that effectively combines an adequate quality loss function and a structured diversity mechanism. 

We also highlight the importance to separate the criteria for enforcing quality and diversity. In Figure \ref{fig:stripe_analysis}, we represent 50 predictions from the models Diverse DPP DILATE \cite{yuan2019diverse} and STRIPE in the plane (DTW,TDI). Diverse DPP DILATE \cite{yuan2019diverse} uses a DPP diversity loss based on the DILATE kernel, which is the same than for quality. We clearly see that the two objectives conflict: this model increases the DILATE diversity (by increasing the variance in the shape (DTW) or the time TDI) components) but a lot of these predictions have a high DILATE loss (worse quality). In contrast, STRIPE predictions are diverse in DTW and TDI, and maintain an overall low DILATE loss. STRIPE succeeds in recovering a set of good tradeoffs between shape and time leading a low DILATE loss. 

\section{Conclusion}

In this Chapter, we have presented STRIPE, a probabilistic time series forecasting method that introduces structured shape and temporal diversity based on determinantal point processes. Diversity is controlled via two proposed differentiable positive semi-definite kernels for shape and time and exploits a forecasting model with a disentangled latent space. Experiments on synthetic and real-world datasets confirm that STRIPE leads to more diverse forecasts without sacrificing on quality. Ablation studies also reveal the crucial importance to decouple the criteria used for quality and diversity.

\clearpage{\pagestyle{empty}\cleardoublepage}

\partabstract{
\vspace{1cm}
\begin{center}
    \textsc{Abstract}\\
\end{center}
\vspace{1cm}
In this part, we are interested in designing Machine Learning (ML) / Model-Based (MB) augmented models by leveraging incomplete physical knowledge formalized through ODE/PDE. Since physical laws are often not directly applicable at the pixel level nor be sufficient for predicting the whole content of future images in generic videos, we propose to learn a latent space where we suppose that physical dynamics apply. We introduce the PhyDNet model (Chapter \ref{chap:phydnet}), which is a two-branch recurrent neural network. One branch is responsible for modelling the physical dynamics while the other branch captures the complementary information required for accurate prediction.  We show that PhyDNet reaches state-of-the-art performances on several video prediction benchmarks. Going further, we concentrate on the ML/MB decomposition problem discussed in Chapter \ref{chap:intro}, which is ill-posed and admits an infinity of solutions. We introduce a principled learning framework, called APHYNITY (Chapter \ref{chap:aphynity}). Inspired by the least-action principle, APHYNITY minimizes the norm of the data-driven complement under the constraint of perfect prediction of the augmented model. We provide a theoretical analysis of the decomposition and show that we can ensure existence and uniqueness decomposition guarantees, under mild conditions. We show on several challenging physical dynamics that APHYNITY ensures better forecasting and parameter identification performances than MB or ML models alone, and that competing MB/ML hybrid methods.
}
\part{Physics-informed forecasting with incomplete knowledge}
\label{part:part2}

\chapter{Disentangling physical from residual dynamics for video prediction}
\label{chap:phydnet}
\chapabstract{

\minitoc

\begin{center}
   \textsc{Chapter abstract}
\end{center}
\textit{
In this Chapter, we address the video prediction problem with deep learning. To constrain the challenging generation of high-dimensional images at the pixel level, we propose to incorporate physical knowledge described by partial differential equations (PDEs). However, since physics is too restrictive for describing the full visual content of generic videos, we introduce in this Chapter PhyDNet, a two-branch deep architecture, which disentangles PDE dynamics from unknown complementary information. The physical branch is composed of a new recurrent physical cell (PhyCell), inspired from data assimilation techniques, that performs PDE-constrained prediction in latent space. Extensive experiments conducted on four various datasets show the very good performances reached by PhyDNet. Ablation studies also highlight the important gain brought out by both disentanglement and PDE-constrained prediction. Finally, we show that PhyDNet presents interesting features for dealing with  missing data and long-term forecasting.\\
The work described in this Chapter is based on the following publication:
\begin{itemize}
    \item \cite{leguen20phydnet}: Vincent Le Guen and Nicolas Thome. "Disentangling Physical Dynamics from Unknown Factors for Unsupervised Video Prediction". In Proceedings of the IEEE/CVF Conference on Computer Vision and Pattern Recognition (CVPR 2020).
\end{itemize}
}
}

\section{Introduction}
\label{sec:intro}

\lettrine[lines=3]{V}ideo forecasting consists in predicting the future content of a video conditioned on previous frames. This is of crucial importance in various contexts, such as weather forecasting \cite{xingjian2015convolutional}, autonomous driving \cite{kwon2019predicting}, reinforcement learning \cite{oh2015action}, robotics \cite{finn2016unsupervised}, or action recognition \cite{liu2017video}. 
In this work, we focus on unsupervised video prediction, where the absence of semantic labels to drive predictions exacerbates the challenges of the task. 
In this context, a key problem is to design video prediction methods able to represent the complex dynamics underlying raw data.

State-of-the-art methods for training such complex dynamical models currently rely on deep learning, with specific architectural choices based on 2D/3D convolutional~\cite{mathieu2015deep,vondrick2016generating} or recurrent neural networks~\cite{wang2017predrnn,wang2018predrnn++,wang2019memory}. To improve predictions, recent methods use adversarial training \cite{mathieu2015deep,vondrick2016generating,kwon2019predicting}, stochastic models \cite{castrejon2019improved,minderer2019unsupervised,franceschi2020stochastic}, constraint predictions by using geometric knowledge \cite{finn2016unsupervised,jia2016dynamic,xue2016visual} or by disentangling factors of variation \cite{villegas2017decomposing,tulyakov2018mocogan,denton2017unsupervised,hsieh2018learning}.

Another appealing way to model the video dynamics is to exploit prior physical knowledge, \eg formalized by partial differential equations (PDEs) \cite{de2017deep,seo2019differentiable}. Recently, interesting connections between residual networks and PDEs have been drawn \cite{weinan2017proposal,lu2018beyond,chen2018neural}, enabling to design physically-constrained machine learning frameworks~\cite{raissi2018deep,de2017deep,seo2019differentiable,rudy2017data}. 
These approaches are very successful for modelling physical systems, when the underlying dynamics is well described by the physical equations in the input space~\cite{raissi2018deep,rudy2017data,long2018pde}. However, such assumption is rarely fulfilled in the pixel space for predicting generalist videos.
  
\begin{figure}
    \centering
    \includegraphics[width=11cm]{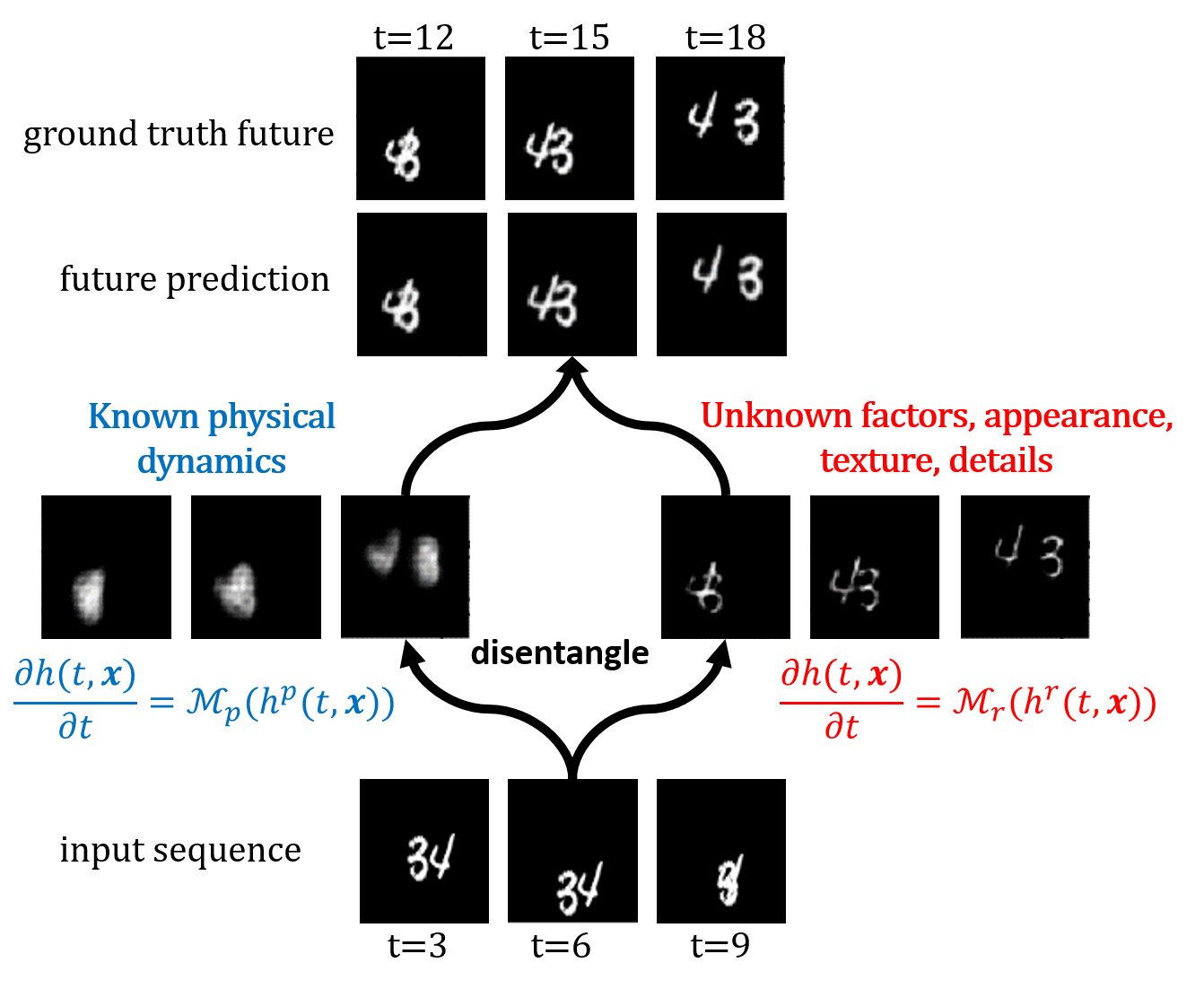}
        \caption[Overview of the PhyDNet model.]{PhyDNet is a deep model mapping an input video into a latent space $\bm{\mathcal{H}}$, from which future frame prediction can be accurately performed. PhyDNet learns $\bm{\mathcal{H}}$ in an unsupervised manner, such that physical dynamics and unknown factors necessary for prediction, \eg appearance, details, texture, are disentangled. \vspace{-0.1cm}} 
    \label{fig:fig1}
\end{figure}

 In this work,  we introduce PhyDNet, a deep model dedicated to perform accurate future frame predictions from generalist videos. In such a context, physical laws do not apply in the input pixel space; the goal of PhyDNet is to learn a semantic latent space $\bm{\mathcal{H}}$ in which they do, and are disentangled from other factors of variation required to perform future prediction. Prediction results of PhyDNet when trained on Moving MNIST~\cite{srivastava2015unsupervised} are shown in Figure \ref{fig:fig1}. The left branch represents the physical dynamics in $\bm{\mathcal{H}}$; when decoded in the image space, we can see that the corresponding features encode approximate segmentation masks predicting digit positions on subsequent frames. 
  On the other hand, the right branch extracts residual information required for prediction, here the precise appearance of the two digits. Combining both representations eventually makes accurate prediction successful.
  
Our contributions to the unsupervised video prediction problem with PhyDNet can be summarized as follows: 
\begin{itemize}
\item We introduce a global sequence to sequence two-branch deep model (section~\ref{sec:3.1}) dedicated to jointly learn the latent space $\bm{\mathcal{H}}$ and to disentangle physical dynamics from residual information, the latter being modeled by a data-driven (ConvLSTM~\cite{xingjian2015convolutional}) method. \vspace{-0.05cm}

\item Physical dynamics is modelled by a new recurrent physical cell, PhyCell (section~\ref{section:phycell}), discretizing a broad class of PDEs in $\bm{\mathcal{H}}$. PhyCell is based on a prediction-correction paradigm inspired from the data assimilation community \cite{asch2016data},~enabling robust training with missing data and for long-term forecasting. \vspace{-0.05cm}

\item Experiments (section~\ref{section4}) reveal that PhyDNet outperforms state-of-the-art methods on four generalist datasets: this is, as far as we know, the first physically-constrained model able to show such capabilities. We highlight the importance of both disentanglement and physical prediction for optimal performances.
\end{itemize}

\section{Related work}
\label{sec:sota}

We review here related multi-step video prediction approaches dedicated to long-term forecasting. We also focus on unsupervised training, \ie only using input video data and without manual supervision based on semantic labels.

Deep neural networks have recently achieved state-of-the-art performances for data-driven video prediction. Seminal works include the application of sequence to sequence LSTM  or Convolutional variants~\cite{srivastava2015unsupervised,xingjian2015convolutional}, adopted in many studies \cite{finn2016unsupervised,lu2017flexible,xu2018structure}. Further works explore different architectural designs based on Recurrent Neural Networks (RNNs) \cite{wang2017predrnn,wang2018predrnn++,oliu2018folded,wang2019memory,wang2018eidetic} and 2D/3D ConvNets \cite{mathieu2015deep,vondrick2016generating,reda2018sdc,byeon2018contextvp}. Dedicated loss functions \cite{cuturi2017soft,leguen19} and Generative Adversarial Networks (GANs) have been investigated for sharper predictions \cite{mathieu2015deep,vondrick2016generating,kwon2019predicting}. However, the problem of conditioning GANs with prior information, such as physical models, remains an open question. 

To constrain the challenging generation of high dimensional images at the pixel level, several methods rather use domain-specific knowledge such as predicting geometric transformations between frames \cite{finn2016unsupervised,jia2016dynamic,xue2016visual}, estimating the optical flow \cite{patraucean2015spatio,luo2017unsupervised,liu2017video,liang2017dual,li2018flow} or exploiting the semantics of the scene \cite{bei2021learning}. This is very effective for short-term prediction, but degrades quickly when the video content evolves, where more complex models and memory about dynamics are required. 

 Another line of work consists in disentangling independent factors of variations in order to apply the prediction model on lower-dimensional representations. A few approaches explicitly model interactions between objects inferred from an observed scene \cite{eslami2016attend,kosiorek2018sequential,ye2019compositional}. Relational reasoning, often implemented with graphs \cite{battaglia2016interaction,kipf2018neural,sanchez2018graph,palm2018recurrent,van2018relational}, can account for basic physical laws, \eg drift, gravity, spring \cite{watters2017visual,wu2017learning,mrowca2018flexible}. However, these methods are object-centric, only evaluate on controlled settings and are not suited for general real-world video forecasting.
Other disentangling approaches factorize the video into independent components \cite{villegas2017decomposing,tulyakov2018mocogan,denton2017unsupervised,hsieh2018learning,gao2019disentangling}. Several disentanglement criteria are used, such as content/motion \cite{villegas2017decomposing,lee2021video} or deterministic/stochastic \cite{denton2017unsupervised}. In specific contexts, the prediction space can be structured using additional information, \eg with human pose \cite{villegas2017learning,walker2017pose} or key points \cite{minderer2019unsupervised}, which imposes a severe overhead on the annotation budget. In this work, we share with these works the motivation to use disentangled representations, but we disentangle incomplete physical dynamics from residual information required for prediction.

\paragraph{Deep Kalman filters}
To handle unobserved phenomena, state space models, in particular the Kalman filter \cite{kalman1960new}, have been recently integrated with deep learning, by modelling dynamics in learned latent space \cite{Krishnan2015DeepKF,watter2015embed,haarnoja2016backprop,fraccaro2017disentangled,becker2019recurrent}. The Kalman variational autoencoder \cite{fraccaro2017disentangled} separates state estimation in videos from dynamics with a linear gaussian state space model. The Recurrent Kalman Network \cite{becker2019recurrent} uses a factorized high dimensional latent space in which the linear Kalman updates are simplified and don't require computationally-heavy covariance matrix inversions. These methods inspired by the data assimilation community \cite{asch2016data,bocquet2019data} have advantages in missing data or long-term forecasting contexts due to their mechanisms decoupling latent dynamics and input assimilation. However, they assume simple latent dynamics (linear) and don't include any physical prior.

\begin{figure}[t]
    \centering
    \begin{tabular}{cc}
    \hspace{-2cm}
        \includegraphics[width=6cm]{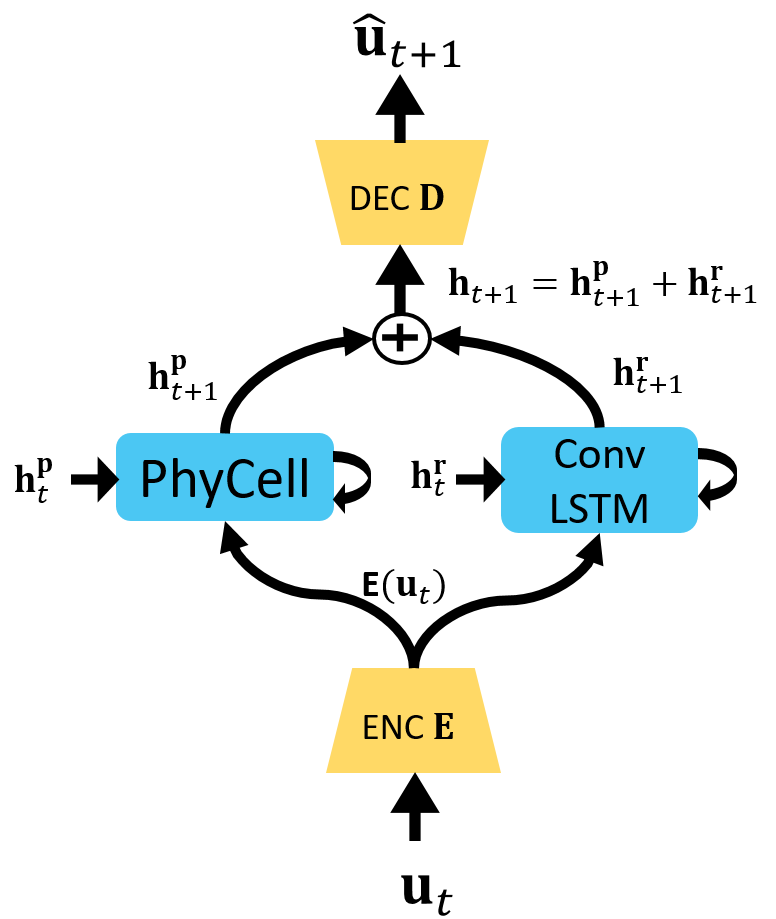} & \includegraphics[width=11cm]{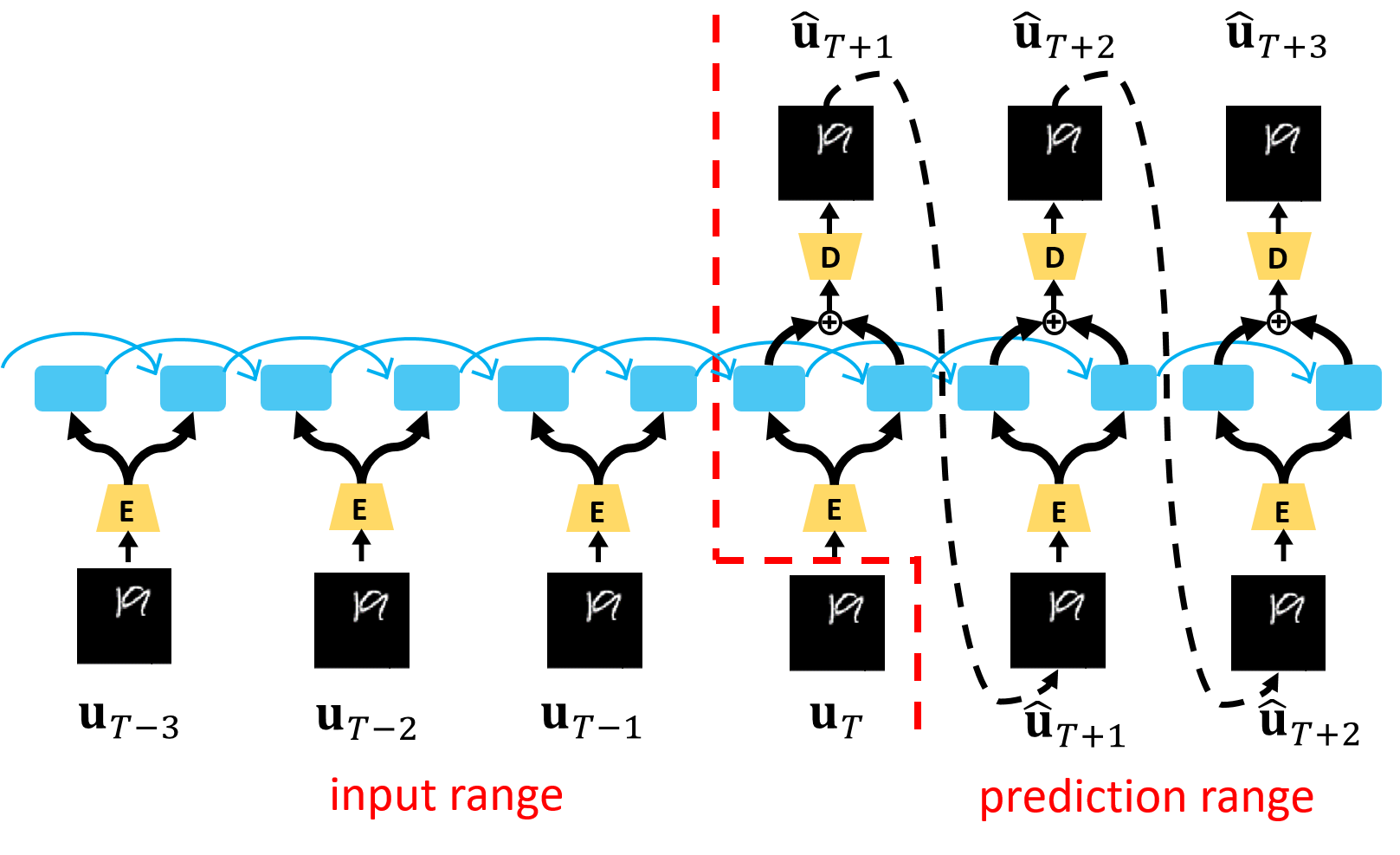}  \\
        \hspace{-0.5cm}\textbf{(a) PhyDNet disentangling cell} & \textbf{(b) Global Seq2Seq architecture} \vspace{0.2cm}
    \end{tabular}{}
    \caption[Proposed PhyDNet deep model for video forecasting.]{\textbf{Proposed PhyDNet deep model for video forecasting.} (a) The core of PhyDNet is a recurrent block projecting input images $\mathbf{u_t}$ into a latent space $\bm{\mathcal{H}}$, where two recurrent neural networks disentangle physical dynamics (PhyCell, section \ref{section:phycell}) from residual information (ConvLSTM). Learned physical $\mathbf{h}^{\mathbf{p}}_{t+1}$ and residual $\mathbf{h}^{\mathbf{r}}_{t+1}$ representations are summed before decoding to predict the future image $\hat{\mathbf{u}}_{t+1}$. (b) Unfolded in time, PhyDNet forms a sequence to sequence (seq2seq) architecture suited for multi-step video prediction. Dotted arrows mean that  predictions are reinjected as next input only for the ConvLSTM branch, and not for PhyCell, as explained in section \ref{sec:training}.}
    \label{fig:fig2}
\end{figure}

\section{PhyDNet model for video forecasting}
\label{section3}

We introduce PhyDNet, a model dedicated to video prediction, which leverages physical knowledge on dynamics, and disentangles it from other unknown factors of variations necessary for accurate forecasting. To achieve this goal, we introduce a disentangling architecture (section~\ref{sec:3.1}), and a new physically-constrained recurrent cell (section~\ref{section:phycell}).

\paragraph*{Problem statement:} As discussed in introduction, physical laws do not apply at the pixel level for general video prediction tasks. However, we  assume that there exists a conceptual latent space $\bm{\mathcal{H}}$ in which physical dynamics and residual factors are linearly disentangled. Formally, let us denote as  $\mathbf{u}= \mathbf{u}(t,\mathbf{x})$ the frame of a video sequence at time $t$, for spatial coordinates $\mathbf{x}=(x,y)$. $\mathbf{h}(t,\mathbf{x}) \in \bm{\bm{\mathcal{H}}}$ is the latent representation of the video up to time $t$, which decomposes as $\mathbf{h}=\mathbf{h^p}+\mathbf{h^r}$, where $\mathbf{h^p}$ (resp. $\mathbf{h^r}$) represents the physical (resp. residual) component of the disentanglement. The video evolution in the latent space $\bm{\bm{\mathcal{H}}}$ is thus governed by the following partial differential equation (PDE):
\begin{equation}
\!\!\!\dfrac{\partial \mathbf{h}(t,\mathbf{x})}{\partial t} \! = \!\frac{\partial \mathbf{h^p}}{\partial t} \!+\! \frac{\partial \mathbf{\mathbf{h^r}}}{\partial t} \!:=\! \bm{\mathcal{M}}_{p}(\mathbf{h^p},\mathbf{u}) + \bm{\mathcal{M}}_{r}(\mathbf{\mathbf{h^r}},\mathbf{u}). \!\!\!
\label{eq:eq1}
\end{equation}
$\bm{\mathcal{M}}_p(\mathbf{h^p},\mathbf{u})$ and $\bm{\mathcal{M}}_r(\mathbf{h^r},\mathbf{u})$ represent physical and residual dynamics in the latent space $\bm{\bm{\mathcal{H}}}$.

\subsection{PhyDNet disentangling architecture}
\label{sec:3.1}

The main goal of PhyDNet is to learn the mapping from input sequences to a latent space which approximates the disentangling properties formalized in Eq \ref{eq:eq1}. 

To reach this objective, we introduce a recurrent bloc which is shown in Figure \ref{fig:fig2} (a). A video frame $\mathbf{u}_t$ at time $t$ is mapped by a deep convolutional encoder $\mathbf{E}$ into a latent space representing the targeted space $\bm{\mathcal{H}}$. $\mathbf{E}(\mathbf{u}_t)$ is then used as input for two parallel recurrent neural networks, incorporating this spatial representation into a dynamical model. 

The left branch in Figure \ref{fig:fig2} (a) models the latent representation $\mathbf{h^p}$ fulfilling the physical part of the PDE in Eq (\ref{eq:eq1}), \ie $\frac{\partial \mathbf{h^p}(t,\mathbf{x})}{\partial t} = \bm{\mathcal{M}}_{p}(\mathbf{h^p},\mathbf{u})$. This PDE is modeled by our recurrent physical cell described in section \ref{section:phycell}, PhyCell, which leads to the computation of $\mathbf{h}^{\mathbf{p}}_{t+1}$ from $\mathbf{E}(\mathbf{u}_t)$ and $\mathbf{h}_t^{\mathbf{p}}$. From the machine learning perspective, PhyCell leverages physical constraints to limit the number of model parameters, regularizes training and improves generalization.

The right branch in Figure \ref{fig:fig2} (a) models the latent representation $\mathbf{h^r}$ fulfilling the residual part of the PDE in Eq \ref{eq:eq1}, \ie $\frac{\partial \mathbf{h^r}(t,\mathbf{x})}{\partial t} = \bm{\mathcal{M}}_{r}(\mathbf{h^r},\mathbf{u})$. Inspired by wavelet decomposition \cite{mallat1999wavelet} and recent semi-supervised works \cite{robert2018hybridnet}, this part of the PDE corresponds to unknown phenomena, which do not correspond to any prior model, and is therefore entirely learned from data. We use a generic recurrent neural network for this task, \eg ConvLSTM \cite{xingjian2015convolutional} for videos, which computes $\mathbf{h}_{t+1}^{\mathbf{r}}$ from $\mathbf{E}(\mathbf{u}_t)$ and $\mathbf{h}_{t}^{\mathbf{r}}$. 

$\mathbf{h}_{t+1}=\mathbf{h}_{t+1}^{\mathbf{p}} +\mathbf{h}_{t+1}^{\mathbf{r}}$ is the combined representation processed by a deep decoder $\mathbf{D}$ to forecast the image $\mathbf{\hat{u}}_{t+1}$.

Figure~\ref{fig:fig2} (b) shows the "unfolded" PhyDNet. An input video $\mathbf{u}_{1:T} = (\mathbf{u}_1,...,\mathbf{u}_T) \in \mathbb{R}^{T\times n \times m \times c}$ with spatial size $n \times m$ an d $c$ channels is projected into $\bm{\mathcal{H}}$ by the
encoder $\mathbf{E}$ and processed by the recurrent block unfolded in time. This forms a  Sequence To Sequence architecture~\cite{sutskever2014sequence} suited for multi-step prediction, outputting H future frame predictions $\mathbf{\hat{u}}_{T+1:T+H}$. Encoder, decoder and recurrent block parameters are all trained end-to-end, meaning that PhyDNet learns itself without supervision the latent space $\bm{\mathcal{H}}$ in which physics and residual factors are disentangled.

\subsection{PhyCell: a deep recurrent physical model}
\label{section:phycell}

PhyCell is a new physical cell, whose dynamics is governed by the PDE response function $\bm{\mathcal{M}}_p(\mathbf{h^p},\mathbf{u})$\footnote{In the sequel, we drop the index $\mathbf{p}$ in $\mathbf{h^p}$ for the sake of simplicity}:
 \begin{equation}
     \bm{\mathcal{M}}_p(\mathbf{h},\mathbf{u}) := \Phi(\mathbf{h})+ \mathcal{C}(\mathbf{h},\mathbf{u}) ,
\label{eq:Mp}
\end{equation}
where $\Phi(\mathbf{h})$ is a physical predictor modelling only the latent dynamics and $C(\mathbf{h},\mathbf{u})$ is a correction term  modelling the interactions between latent state and input data. 

\paragraph*{Physical predictor:} $\Phi(\mathbf{h})$ in Eq~(\ref{eq:Mp}) is modeled as follows: 
 \begin{equation}
    \Phi(\mathbf{h}(t,\mathbf{x})) = \sum_{i,j: i+j \leq q}  c_{i,j} \dfrac{\partial^{i+j} \mathbf{h}}{\partial x^i \partial y^j}(t,\mathbf{x}).
    \label{eq:phi}
\end{equation}
$\Phi(\mathbf{h}(t,\mathbf{x}))$ in Eq \ref{eq:phi} combines the spatial derivatives with coefficients $c_{i,j}$ up to a certain differential order $q$. This generic class of linear PDEs subsumes a wide range of classical physical models, \eg the heat equation, the wave equations, the advection-diffusion equations.

\paragraph*{Correction:} $\mathcal{C}(\mathbf{h},\mathbf{u})$ in Eq \ref{eq:Mp} takes the following form: 
\begin{equation}
    \!\!\!\!\!\!\mathcal{C}(\mathbf{h},\mathbf{u}) \!:=\! \mathbf{K}(t,\mathbf{x})\odot \left[\mathbf{E} (\mathbf{u}(t,\mathbf{x})) \!-\! (\mathbf{h}(t,\mathbf{x}) \!+\! \Phi(\mathbf{h}(t,\mathbf{x} ))\right].
    \label{eq:corrcont}
\end{equation}
Eq \ref{eq:corrcont} computes the difference between the latent state after physical motion $\mathbf{h}(t,\mathbf{x}) + \Phi(\mathbf{h}(t,\mathbf{x}))$ and the embedded new observed input $\mathbf{E}(\mathbf{u}(t,\mathbf{x}))$. $\mathbf{K}(t,\mathbf{x})$ is a gating  factor, where $\odot$ is the Hadamard product.

\subsubsection{Discrete PhyCell} 
\label{sec:discretephicell}
\begin{figure}
    \centering
    \includegraphics[width=12cm]{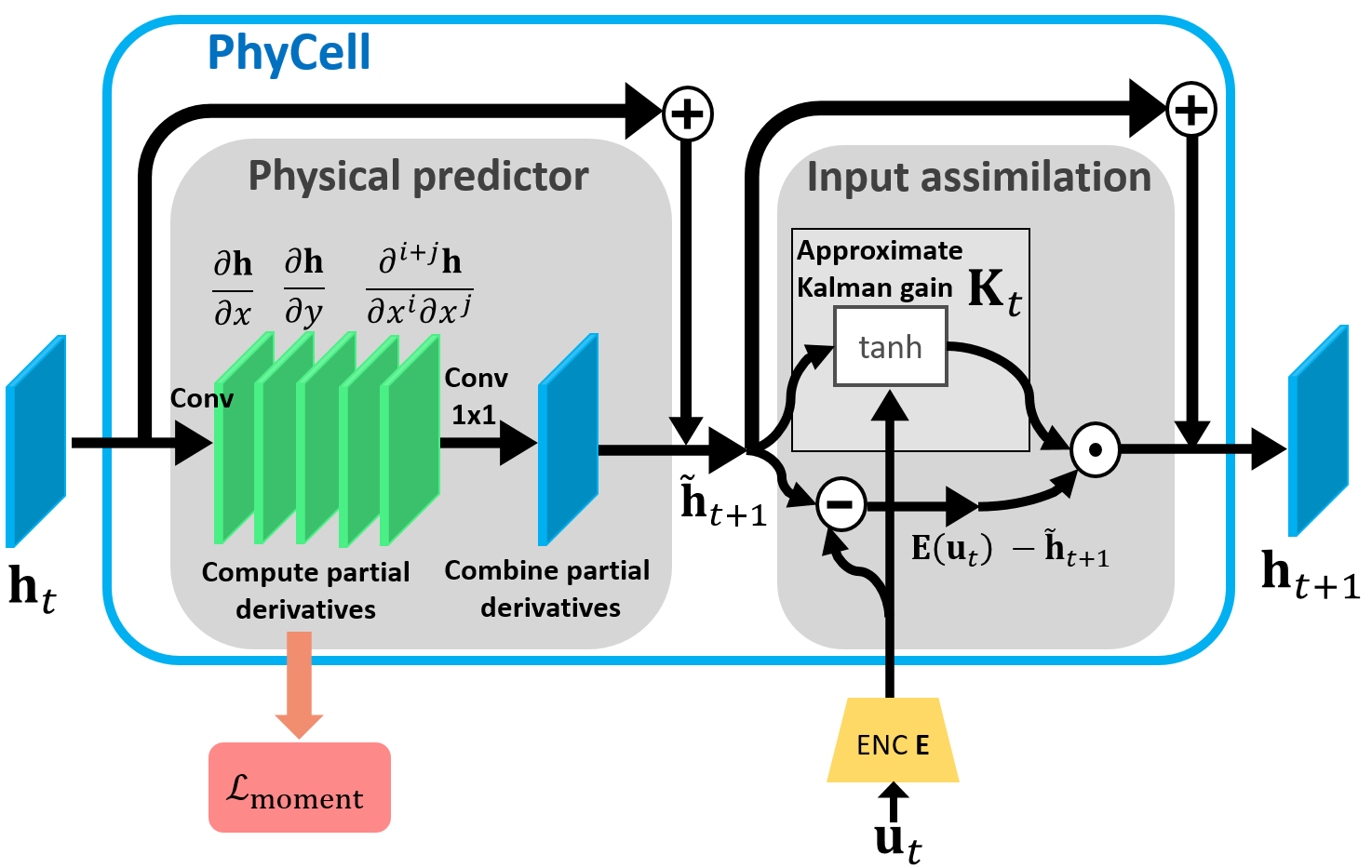}
    \caption[Description of the PhyCell predictor.]{PhyCell recurrent cell implements a two-steps scheme: physical prediction with convolutions for approximating and combining spatial derivatives (Eq \ref{eq:prediction} and Eq \ref{eq:phi}), and input assimilation as a correction of latent physical dynamics driven by observed data (Eq \ref{eq:correction}). During training, the filter moment loss in red (Eq \ref{eq:lmoment}) enforces the convolutional filters to approximate the desired differential operators.}
    \label{fig:phicell}
\end{figure}

We discretize the  continuous time PDE in Eq \ref{eq:Mp} with the standard forward Euler numerical scheme \cite{lu2018beyond}, leading to the discrete time PhyCell (derivation in Appendix \ref{app:phycell-deriv}):
\begin{equation}
    \mathbf{h}_{t+1} = (1-\mathbf{K}_t) \odot \left(\mathbf{h}_t + \Phi(\mathbf{h}_t) \right) + \mathbf{K}_t \odot \mathbf{E}(\mathbf{u}_t).
    \label{eq:physical_cell}
\end{equation}
Depicted in Figure \ref{fig:phicell}, PhyCell is an atomic recurrent cell for building physically-constrained RNNs. In our experiments, we use one layer of PhyCell but one can also easily stack several PhyCell layers to build more complex models, as done for stacked RNNs \cite{wang2017predrnn,wang2018predrnn++,wang2019memory}. To gain insight into PhyCell in Eq~(\ref{eq:physical_cell}), we write the equivalent two-steps form:

\begin{empheq}[left=\empheqlbrace]{alignat=2}
&   \tilde{\mathbf{h}}_{t+1} \!= \mathbf{h}_{t} +  \Phi(\mathbf{h}_{t})   &  \!\!\!\quad \text{\small{\textbf{Prediction}\!}} \label{eq:prediction}\\
&   \mathbf{h}_{t+1} \!= \tilde{\mathbf{h}}_{t+1}  + \mathbf{K}_t \odot \left( \mathbf{E}(\mathbf{u}_t) - \tilde{\mathbf{h}}_{t+1} \right). & \!\!\! \quad \text{\small{\textbf{Correction}\!}} \label{eq:correction}
\end{empheq}

The prediction step in Eq \ref{eq:prediction} is a physically-constrained motion in the latent space, computing the intermediate representation $\tilde{\mathbf{h}}_{t+1}$. Eq \ref{eq:correction} is a correction step  incorporating input data. This prediction-correction formulation is reminiscent of the way to combine numerical models with observed data in the data assimilation community \cite{asch2016data,bocquet2019data}, \eg with the Kalman filter \cite{kalman1960new}. We show in section \ref{sec:training} that this decoupling between prediction and correction can be leveraged to robustly train our model in  long-term forecasting and missing data contexts. $\mathbf{K}_t$ can be interpreted as the Kalman gain controlling the trade-off between both steps.

\subsubsection{PhyCell implementation}
We now specify how the physical predictor $\Phi$ in Eq \ref{eq:prediction} and the correction Kalman gain $\mathbf{K}_t$ in Eq \ref{eq:correction} are implemented.

\paragraph*{Physical predictor:} We implement $\Phi$ using  a convolutional neural network (left gray box in Figure \ref{fig:phicell}), based on the connection between convolutions and differentiations \cite{dong2017image,long2018pde}.

This offers the possibility to learn a class of filters approximating each partial derivative in Eq \ref{eq:phi}, which are constrained by a kernel moment loss, as detailed in section \ref{sec:training}. As noted by~\cite{long2018pde}, the flexibility added by this constrained learning strategy gives better results for solving PDEs than handcrafted derivative filters.
Finally, we use $1 \times 1$ convolutions to linearly combine these derivatives with $c_{i,j}$ coefficients in Eq \ref{eq:phi}. 

\paragraph*{Kalman gain:}
We approximate $\mathbf{K}_t$ in Eq \ref{eq:correction} by a gate with learned  convolution kernels $\mathbf{W}_h$, $\mathbf{W}_u$ and bias $\mathbf{b}_k$:
\begin{equation}
\mathbf{K}_t =  \tanh \left( \mathbf{W}_{h} * \tilde{\mathbf{h}}_{t+1} + \mathbf{W}_{u} * \mathbf{E}(\mathbf{u}_t) + \mathbf{b}_k \right).
\label{eq:kalman_gain}
\end{equation}
Note that if $\mathbf{K}_t = \mathbf{0}$, the input is not accounted for and the dynamics follows the physical predictor; if $\mathbf{K}_t = 1 $, the latent dynamics is resetted and only driven  by the input. This is similar to gating mechanisms in LSTMs or GRUs. 

\paragraph*{Discussion:} With specific $\Phi$ predictor,
$\mathbf{K}_t$ gain and  encoder $\mathbf{E}$, PhyCell recovers recent models from the literature:
\begin{table}[H]
    \centering
        \begin{adjustbox}{max width=\columnwidth}
    \begin{tabular}{c|ccc}
    model & $\Phi$ & $\mathbf{K}_t$ & $\mathbf{E}$ \\ \hline
    PDE-Net \cite{long2019pde}    &  Eq \ref{eq:prediction} & $\mathbf{0}$ & $\mathbf{Id}$     \\ \hline
    Advection-diffusion & advection-diffusion & $\mathbf{0}$ &  $\mathbf{Id}$ \\ 
    flow~\cite{de2017deep}  & predictor & &  \\ \hline
    Recurrent Kalman Filter \cite{becker2019recurrent}  &  locally linear, no   & approximate   & deep encoder   \\
            ~   & physical constraint     & Kalman gain  &  \\ \hline
            PhyDNet (ours) &  Eq \ref{eq:prediction}  & Eq \ref{eq:kalman_gain}  & deep encoder
    \end{tabular}
    \vspace{-0.2cm}
    \label{tab:my_label}
    \end{adjustbox}
\end{table}

PDE-Net~\cite{long2018pde} directly works on raw pixel data (identity encoder $\mathbf{E}$) and assumes Markovian dynamics (no correction, $\mathbf{K}_t\!\!\!=\!\!\!\mathbf{0}$): the model solves the autonomous PDE $\frac{\partial \mathbf{u}}{\partial t}=\Phi(\mathbf{u})$ given in Eq \ref{eq:prediction} but in pixel space. This prevents from modelling time-varying PDEs such as those tackled in this work, \eg  varying advection terms. 
The flow model in \cite{de2017deep} uses the closed-form solution of the advection-diffusion equation as predictor ; it is however limited only to this PDE, whereas PhyDNet models a much broader class of PDEs. The Recurent Kalman Filter (RKF) \cite{becker2019recurrent} also proposes a prediction-correction scheme in a deep latent space, but their approach does not include any prior physical information, and the prediction step is locally linear, whereas we use deep models. An approximated form of the covariance matrix is used for estimating $\mathbf{K}_t$ in \cite{becker2019recurrent}, which we find experimentally inferior to our gating mechanism in Eq \ref{eq:kalman_gain}.

\subsection{Training}
\label{sec:training}

Given a training set of $N$ videos $\bm{\mathcal{D}} =  \left \{ \mathbf{u}^{(i)} \right \} _{i=\{1:N \}}$ and PhyDNet parameters $\mathbf{w}= (\mathbf{w_p},\mathbf{w_r},\mathbf{w_s})$, where $\mathbf{w_p}$ (resp.  $\mathbf{w_r}$) are parameters of the PhyCell (resp. residual) branch,  and $\mathbf{w_s}$ are encoder and decoder shared parameters, we minimize the following objective function:
\begin{equation}
    \mathcal{L}(\bm{\mathcal{D}},\mathbf{w}) = \mathcal{L}_{\text{image}}(\bm{\mathcal{D}},\mathbf{w}) + \lambda  \mathcal{L}_{\text{moment}}(\mathbf{w_p}).
\end{equation}
We use the $L^2$ loss for the image reconstruction loss $\mathcal{L}_{\text{image}}$, as commonly done in the literature \cite{wang2017predrnn,wang2018predrnn++,oliu2018folded,wang2018eidetic,wang2019memory}. 

$\mathcal{L}_{\text{moment}}(\mathbf{w_p})$ imposes physical constraints on the $k^2$ learned filters  $  \left\{ \mathbf{w}^k_{p,i,j}\right\}_{i,j \leq k}$, such that each $\mathbf{w}^k_{p,i,j}$ of size $k \times k$ approximates $\frac{\partial^{i+j}}{\partial x^i y^j}$. This is achieved by using a loss based on the moment matrix $\mathbf{M}(\mathbf{w}^k_{p,i,j})$~\cite{long2019pde}, representing the order of the filter differentiation~\cite{dong2017image}.  $\mathbf{M}(\mathbf{w}^k_{p,i,j})$ is compared to a target moment matrix $\mathbf{\Delta}^k_{i,j}$ (see $\mathbf{M}$ and $\mathbf{\Delta}$ computations in Appendix \ref{app:moment-matrix}), leading to: 
 \begin{equation}
   \mathcal{L}_{\text{moment}} = \sum\limits_{i \leq k} \sum\limits_{j \leq k} ||\mathbf{M}(\mathbf{w}^k_{p,i,j}) - \mathbf{\Delta}^k_{i,j} ||_F .
   \label{eq:lmoment}
 \end{equation}

\paragraph*{Prediction mode:} An appealing feature of PhyCell is that we can use and train the model in a "prediction-only" mode by setting $\mathbf{K}_t = \mathbf{0}$ in Eq \ref{eq:correction}, \ie by only relying on the physical predictor $\Phi$ in Eq \ref{eq:prediction}. It is worth mentioning that the "prediction-only" mode is not applicable to standard Seq2Seq RNNs: although the decomposition in Eq \ref{eq:Mp} still holds, \ie $\bm{\mathcal{M}}_r(\mathbf{h},\mathbf{u}) = \Phi(\mathbf{h})+ \mathcal{C}(\mathbf{h},\mathbf{u})$, the resulting predictor is naive and useless for multi-step prediction  $\mathbf{\tilde{h}}_{t+1}=0$, see Appendix \ref{sec:pdernn}).

Therefore, standard RNNs are not equipped to deal with unreliable input data $\mathbf{u}_t$. We show in section~\ref{sec:expe_prediction} that the gain of PhyDNet over those models increases in two important contexts with unreliable inputs: multi-step prediction and dealing with missing data.

\section{Experiments}
\label{section4}

\subsection{Experimental setup}

We evaluate PhyDNet on four datasets from various origins. 

\paragraph{Moving MNIST} is a standard benchmark in video prediction \cite{srivastava2015unsupervised} consisting in two random MNIST digits bouncing on the walls of a $64 \times 64$ grid. We predict 10 future frames given 10 input frames. Training sequences are generated on the fly and the test set of 10000 sequences is provided by \cite{srivastava2015unsupervised}. 

\paragraph{Traffic BJ} consists in traffic flow data collected by taxicabs in Beijing \cite{zhang2017deep}. Each $32 \times 32$ image is a 2-channels heat map with leaving/entering traffic.
~Video prediction on such real-world complex data require modelling transport phenomena and traffic diffusion.
~Following the setting of \cite{zhang2017deep,wang2019memory,wang2018eidetic}, we predict 4 future frames given 4 input frames. 

\paragraph{SST} consists in daily Sea Surface Temperature (SST) data from the sophisticated simulation engine NEMO (Nucleus for European Modelling of the Ocean), as in \cite{de2017deep}. SST evolution is governed by the physical laws of fluid dynamics. We predict 4 frames of size $64 \times 64$ given 4 input frames. 

\paragraph{Human 3.6} contains 3.6 million images of human actions \cite{ionescu2013human3}, with complex 3D articulated motions. Following the setting of \cite{wang2019memory}, we use only the "walking" scenario with subjects S1, S5, S6, S7, S8 for training, and S9, S11 for testing. We predict 4 future images of size $128 \times 128  \times 3$ given 4 input images. \vspace{0.1cm}\\


\paragraph*{Network architectures and training:}
PhyDNet shares a common backbone architecture for all datasets where the physical branch contains 49 PhyCells filters (with kernel of size $7 \times 7$) and the residual branch is composed of a 3-layers ConvLSTM with 128 filters in each layer. We set up the trade-off parameter between $\mathcal{L}_{\text{image}}$ and $\mathcal{L}_{\text{moment}}$ to $\lambda=1$. Detailed architectures and $\lambda$ impact are given in Appendix \ref{app:phydnet-impl}. Our code is available at \url{https://github.com/vincent-leguen/PhyDNet}. 

\paragraph*{Evaluation metrics:} We follow evaluation metrics commonly used in state-of-the-art video prediction methods: the Mean Squared Error (MSE), Mean Absolute Error (MAE) and the Structural Similarity (SSIM) \cite{wang2004image} that computes the perceived image quality with respect to a reference. Metrics are averaged for each frame of the output sequence. Lower MSE, MAE and higher SSIM indicate better performances.

\subsection{State of the art comparison}

We evaluate PhyDNet against strong recent baselines, including very competitive data-driven RNN architectures: ConvLSTM  \cite{xingjian2015convolutional}, PredRNN \cite{wang2017predrnn}, Causal LSTM \cite{wang2018predrnn++}, Memory in Memory (MIM) \cite{wang2019memory}. We also compare to methods dedicated to specific datasets: DDPAE \cite{hsieh2018learning}, a disentangling method specialized and state-of-the-art on Moving MNIST ; and the physically-constrained advection-diffusion flow model \cite{de2017deep} that is state-of-the-art for the SST dataset.

\begin{table}[b!]
    \caption[Quantitative forecasting results of the PhyDNet model.]{Quantitative forecasting results of PhyDNet compared to baselines using various datasets. Numbers are copied from original or citing papers. * corresponds to results obtained by running online code from the authors. The first five baseline are general deep models applicable to all datasets, whereas DDPAE \cite{hsieh2018learning} (resp. advection-diffusion flow \cite{de2017deep}) are specific state-of-the-art models for Moving MNIST (resp. SST). Metrics are scaled to be in a similar range across datasets to ease comparison.}
    \begin{adjustbox}{max width=\linewidth}
    \begin{tabular}{l|lll|lll|lll|lll}
  \toprule
     \multicolumn{1}{c}{} &  \multicolumn{3}{|c}{\textbf{Moving MNIST}} &  \multicolumn{3}{|c}{\textbf{Traffic BJ}} &  \multicolumn{3}{|c}{\textbf{Sea Surface Temperature}}  &  \multicolumn{3}{|c}{\textbf{Human 3.6}}  \\ 
     \midrule
Method  & MSE & MAE & SSIM & MSE $\times 100$  & MAE & SSIM & MSE $\times 10$ & MAE & SSIM  & MSE  / 10 & MAE $/ 100$ & SSIM \\
  \midrule

    ConvLSTM \cite{xingjian2015convolutional} & 103.3 & 182.9 & 0.707  & $48.5^*$ & $17.7^*$ & $0.978^*$ & $45.6^*$ & $63.1^*$ & $0.949^*$ & $50.4^*$  & $18.9^*$ & $0.776^*$  \\ 
    PredRNN \cite{wang2017predrnn}  & 56.8 & 126.1 & 0.867 & 46.4 & $17.1^*$ & $0.971^*$ & 41.9 & 62.1  & 0.955 & 48.4 & 18.9 & 0.781 \\ 
    Causal LSTM \cite{wang2018predrnn++}  & 46.5  & 106.8  & 0.898 & 44.8 & $16.9^*$ & $0.977^*$  & $39.1^*$  & $62.3^*$ & $0.929^*$ & 45.8  & 17.2   & 0.851 \\     
    MIM \cite{wang2019memory}  & 44.2  & 101.1 & 0.910 & 42.9 & $16.6^*$ & $0.971^*$ & $42.1^*$ & $60.8^*$  & $0.955^*$  & 42.9 & 17.8 & 0.790 \\
    E3D-LSTM \cite{wang2018eidetic}  & 41.3  & 86.4  & 0.920  & $43.2^*$ & $16.9^*$  & $0.979^*$   & $34.7^*$ & $59.1^*$  & $0.969^*$  & 46.4  & 16.6  & 0.869 \\ \hline
    Advection-diffusion \cite{de2017deep}  & -  &  - &    -& - & - &- &  $34.1^*$ & $54.1^*$ & $0.966^*$ & - & - &-  \\ 
    DDPAE \cite{hsieh2018learning}  & 38.9 & $90.7^*$ & $0.922^*$  & - &- &- &- &- &- &- &- &- \\ 
    \midrule
    \textbf{PhyDNet}  & \textbf{24.4} & \textbf{70.3} & \textbf{0.947} & \textbf{41.9} & \textbf{16.2} & \textbf{0.982}    & \textbf{31.9} & \textbf{53.3} & \textbf{0.972} & \textbf{36.9} & \textbf{16.2} & \textbf{0.901} \\ 
 \bottomrule
    \end{tabular}
    \end{adjustbox}
    \label{tab:res1}   
\end{table}

Overall results presented in Table \ref{tab:res1} reveal that PhyDNet outperforms significantly all baselines on all four datasets. The performance gain is large with respect to state-of-the-art general RNN models, with a gain of 17 MSE points for Moving MNIST, 6 MSE points for Human 3.6, 3 MSE points for SST and 1 MSE point for Traffic BJ. In addition, PhyDNet also outperforms specialized models: it gains 14 MSE points compared to the disentangling DDPAE model \cite{hsieh2018learning} specialized for Moving MNIST, and 2 MSE points compared to the advection-diffusion model \cite{de2017deep} dedicated to SST data. PhyDNet also presents large and consistent gains in SSIM, indicating that image quality is greatly improved by the physical regularization. Note that for Human 3.6, a few approaches use specific strategies dedicated to human motion with additional supervision, \eg human pose in \cite{villegas2017learning}. We perform similarly to \cite{villegas2017learning} using only unsupervised training, as shown in Appendix \ref{app:compa-villegas}. This is, to the best of our knowledge, the first time that physically-constrained deep models reach state-of-the-art performances on generalist video prediction datasets.

\begin{figure*}[ht]
    \centering
    \includegraphics[width=\linewidth]{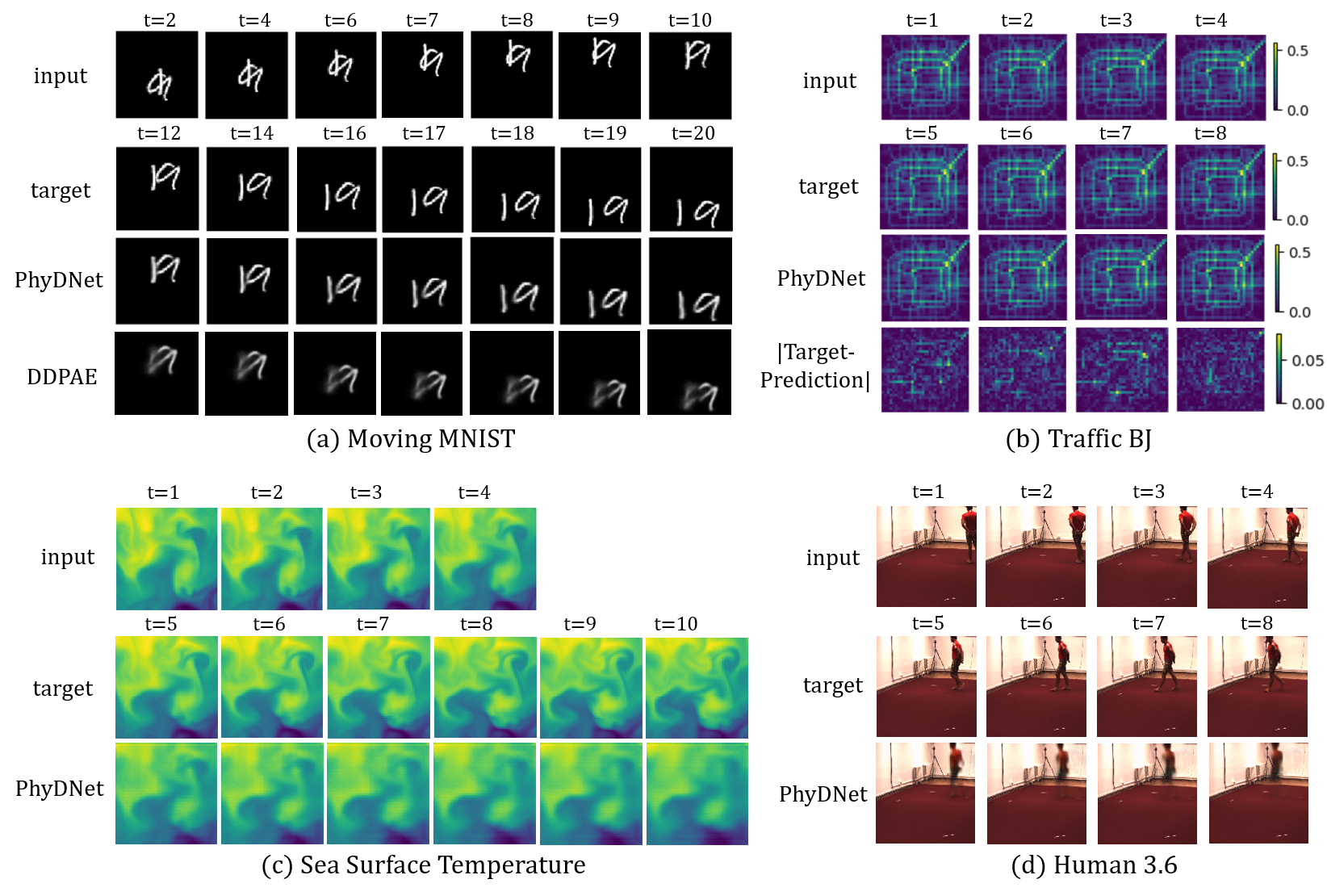}
    \caption[Qualitative prediction results of PhyDNet.]{Qualitative results of the predicted frames by PhyDNet for all datasets. First line is the input sequence, second line the target and third line PhyDNet prediction. For Moving MNIST, we add a fourth line with the comparison to DDPAE \cite{hsieh2018learning} and for Traffic BJ the difference $|\text{Prediction-Target}|$ for better visualization.}
    \label{fig:visus}
\end{figure*}

In Figure \ref{fig:visus}, we provide qualitative prediction results for all datasets, showing that PhyDNet properly forecasts future images for the considered horizons: digits are sharply and accurately predicted for Moving MNIST in (a), the absolute traffic flow error is low and approximately spatially independent in (b), the evolving physical SST phenomena are well anticipated in (c) and the future positions of the person is accurately predicted in (d). We add in Figure \ref{fig:visus}(a) a qualitative comparison to DDPAE \cite{hsieh2018learning}, which fails to predict the future frames properly. Since the two digits overlap in the input sequence, DPPAE is unable to disentangle them. In contrast, PhyDNet successfully learns the physical dynamics of the two digits in a disentangled latent space, leading a correct prediction. In Appendix \ref{app:phydnet-visu}, we detail this comparison to DPPAE, and provide additional visualizations for all datasets. 

\subsection{Ablation Study}

\begin{table}
    \caption[Ablation study of the PhyDNet model.]{An ablation study shows the consistent performance gain on all datasets of our physically-constrained PhyCell vs the general purpose ConvLSTM, and the additional gain brought up by the disentangling architecture PhyDNet. * corresponds to results obtained by running online code from the authors.}
    \begin{adjustbox}{max width=\linewidth}
    \begin{tabular}{l|lll|lll|lll|lll}
  \toprule
     \multicolumn{1}{c}{} &  \multicolumn{3}{|c|}{\textbf{Moving MNIST}} &  \multicolumn{3}{|c|}{\textbf{Traffic BJ}} &  \multicolumn{3}{|c|}{\textbf{Sea Surface Temperature}} &  \multicolumn{3}{|c}{\textbf{Human 3.6}}  \\ 
  \midrule
    Method  & MSE & MAE & SSIM & MSE $\times$ 100 & MAE & SSIM & MSE $\times$ 10 & MAE & SSIM & MSE $/$ 10 & MAE $/$ 100 & SSIM \\ 
    \midrule
    ConvLSTM  & 103.3  & 182.9  & 0.707  & $48.5^*$ & $17.7^*$  & $0.978^*$  &  $45.6^*$ & $63.1^*$ & $0.949^*$  & $50.4^*$ & $18.9^*$  & $0.776^*$ \\
    PhyCell    & 50.8 & 129.3  & 0.870  & 48.9  & 17.9 & 0.978  & 38.2  & 60.2 & 0.969  & 42.5  & 18.3 & 0.891 \\ 
    PhyDNet   & \textbf{24.4}  & \textbf{70.3}  & \textbf{0.947}  &  \textbf{41.9} & \textbf{16.2} & \textbf{0.982}  & \textbf{31.9}  & \textbf{53.3}  & \textbf{0.972}  & \textbf{36.9} & \textbf{16.2} & \textbf{0.901}  \\ 
 \bottomrule
    \end{tabular}
    \end{adjustbox}

  \label{tab:ablation}  
\end{table}

We perform here an ablation study to analyse the respective contributions of physical modelling and disentanglement. Results are presented in Table \ref{tab:ablation} for all datasets. We see that a 1-layer PhyCell model (only the left branch of PhyDNet in Figure \ref{fig:fig2}(b)) outperforms a 3-layers ConvLSTM (50 MSE points gained for Moving MNIST, 8 MSE points for Human 3.6, 7 MSE points for SST and equivalent results for Traffic BJ), while PhyCell  has much fewer parameters (270,000 \textit{vs.} 3 million parameters). This confirms that PhyCell is a very effective recurrent cell that successfully incorporates physical prior in deep models.  When we further add our disentangling strategy with the two-branch architecture (PhyDNet), we have another performance gap on all datasets (25 MSE points for Moving MNIST, 7 points for Traffic and SST, and 5 points for Human 3.6), which proves that physical modelling is not sufficient by itself to perform general video prediction and that learning unknown factors is necessary.

To complement the discussion of Table \ref{tab:ablation}, we give here in Table \ref{tab:nb-parameters} the approximate number of models parameters of trained models:

\begin{table}[H]
    \centering
        \caption{Number of parameters of models trained on Moving MNIST.}
    \begin{tabular}{c|c}
    \toprule
    method     &  number of parameters \\ 
    \midrule
        ConvLSTM & $3 . 10^6$ \\
        PhyCell & $370 . 10^3$ \\
        PhyDNet & $3 . 10^6$ \\
        \bottomrule
    \end{tabular}
    \label{tab:nb-parameters}
\end{table}
We see that a 1-layer PhyCell with 49 filters has far fewer parameters than a 3-layers ConvLSTM (with 128 filters in each layer) and obtains far better results (gain of 50 MSE points). Then PhyDNet with approximately the same number of parameters as ConvLSTM (3 million) again improves the performances by 25 MSE points, reaching a state-of-the-art MSE score of 24.4.

We qualitatively analyze in Figure~\ref{fig:ablation} partial predictions of PhyDNet for the physical branch  $\hat{\mathbf{u}}^{\mathbf{p}}_{t+1} = \mathbf{D}(\mathbf{h}^{\mathbf{p}}_{t+1})$ and residual branch  $\hat{\mathbf{u}}^{\mathbf{r}}_{t+1} = \mathbf{D}(\mathbf{h}^{\mathbf{r}}_{t+1})$. As noted in Figure \ref{fig:fig1} for Moving MNIST, $\mathbf{h^p}$ captures coarse localisations of objects, while $\mathbf{h^r}$ captures fine-grained details that are not useful for the physical model. Additional visualizations for the other datasets are provided in Appendix \ref{app:phydnet-visu}.

\begin{figure}
    \centering
    \includegraphics[width=11cm]{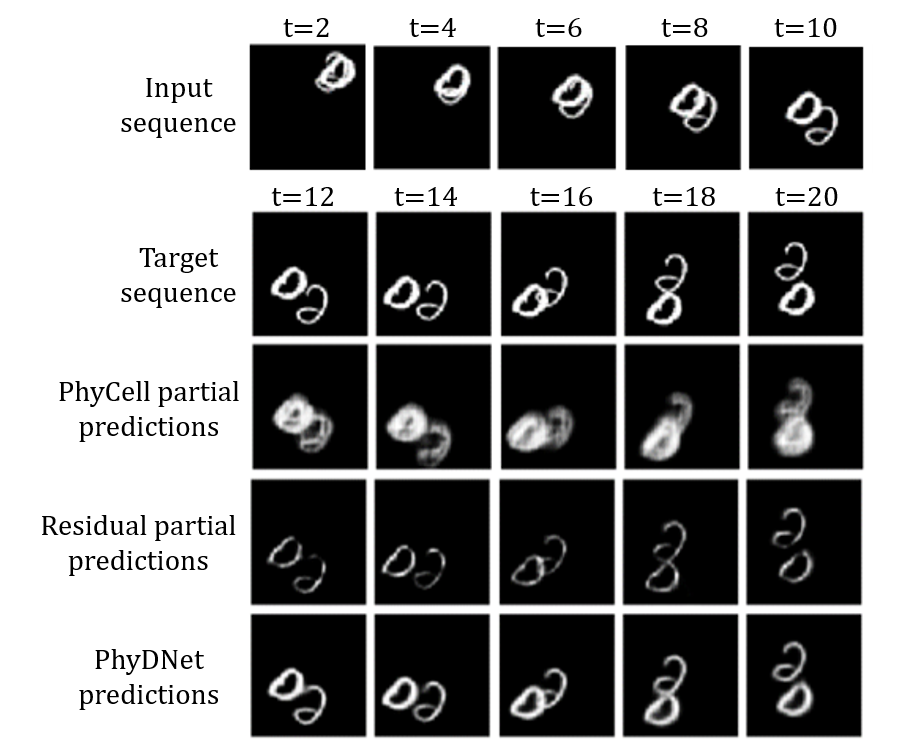}
    \caption[Qualitative ablation results on Moving MNIST.]{Qualitative ablation results on Moving MNIST: partial predictions show that PhyCell captures coarse localisation of digits, whereas the ConvLSTM branch models the fine shape details of digits. Every two frames are displayed.}
    \label{fig:ablation}
\end{figure}

\paragraph{Influence of physical regularization}

We conduct in Table \ref{tab:ablation2} a finer ablation on Moving MNIST to study the impact of the physical regularization $\mathcal{L}_{\text{moment}}$ on the performance of PhyCell and PhyDNet. When we disable $\mathcal{L}_{\text{moment}}$ for training PhyCell, performances improve by 7 points in MSE. This underlines that physical laws alone are too restrictive for learning dynamics in a general context, and that complementary factors should be accounted for.  
On the other side, when we disable $\mathcal{L}_{\text{moment}}$ for training our disentangled architecture PhyDNet, performances decrease by 5 MSE points ($29$ \textit{vs} $24.4$) compared to the physically-constrained version. This proves that physical constraints are relevant, but should be incorporated carefully in order to make both branches cooperate. This enables to leverage physical prior, while keeping remaining information necessary for pixel-level prediction. Same conclusions can be drawn for the other datasets, see Appendix \ref{app:phydnet-influence}.

\begin{table}[H]
\centering
    \caption{Influence of physical regularization for Moving MNIST.}
    \begin{adjustbox}{max width=\columnwidth}
    \begin{tabular}{l|lll}
\toprule
    Method  & MSE & MAE & SSIM  \\ 
    \midrule
    PhyCell & 50.8 & 129.3  & 0.870    \\ 
    PhyCell without $\mathcal{L}_{\text{moment}}$  & 43.4  & 112.8  & 0.895   \\ 
    PhyDNet &  \textbf{24.4}  & \textbf{70.3}  & \textbf{0.947}   \\     
    PhyDNet without $\mathcal{L}_{\text{moment}}$ & 29.0  & 81.2 & 0.934   \\ 
\bottomrule
    \end{tabular}
    \end{adjustbox}
    \label{tab:ablation2} 
\end{table}

\subsection{PhyCell analysis}
\label{sec:expe_prediction}

\paragraph*{Physical filter analysis}

\begin{figure}[H]
    \centering
    \begin{tabular}{cccc}
   \hspace{-0.5cm} \includegraphics[height=3.3cm]{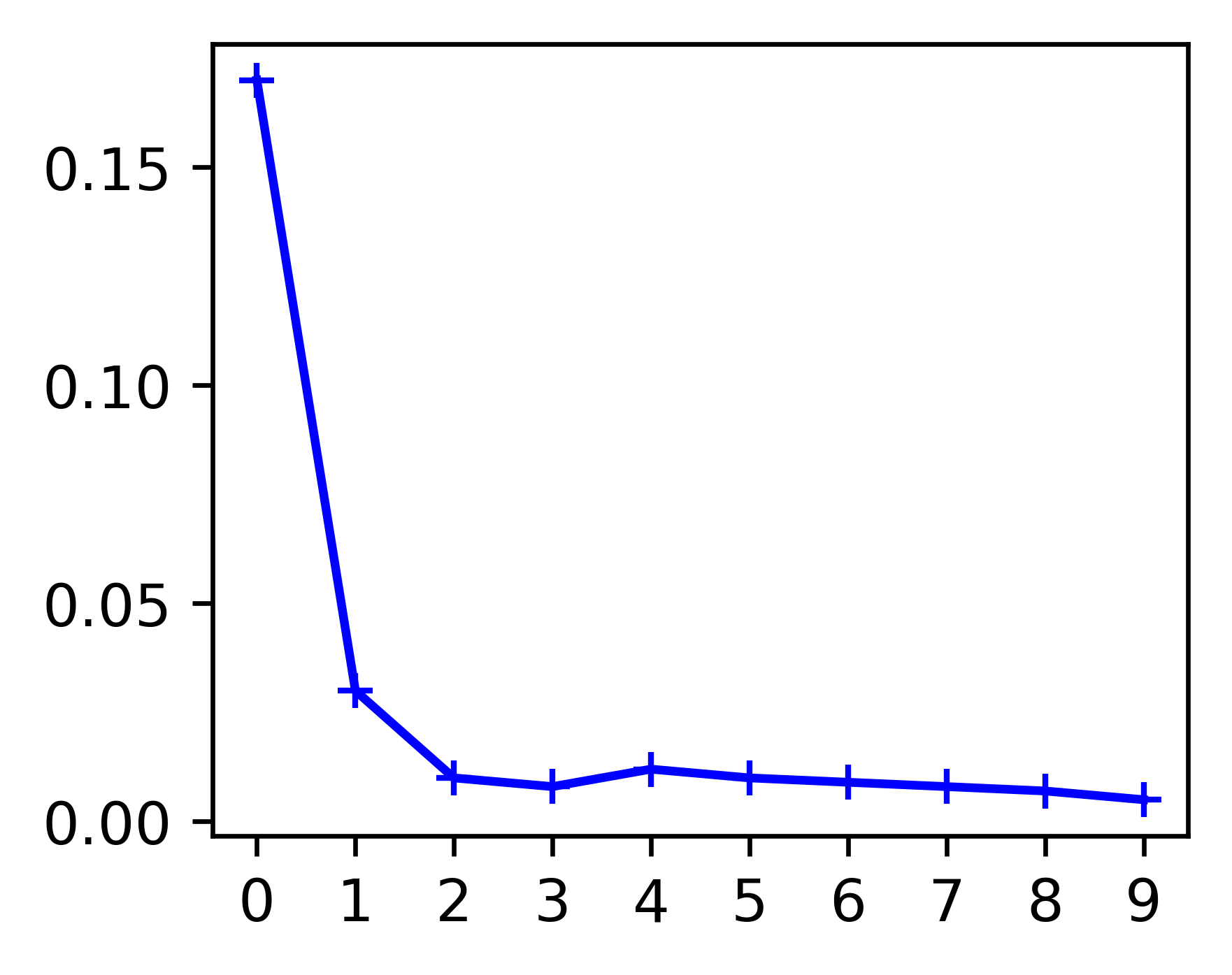}      & \hspace{-0.5cm}
    \includegraphics[height=3.3cm]{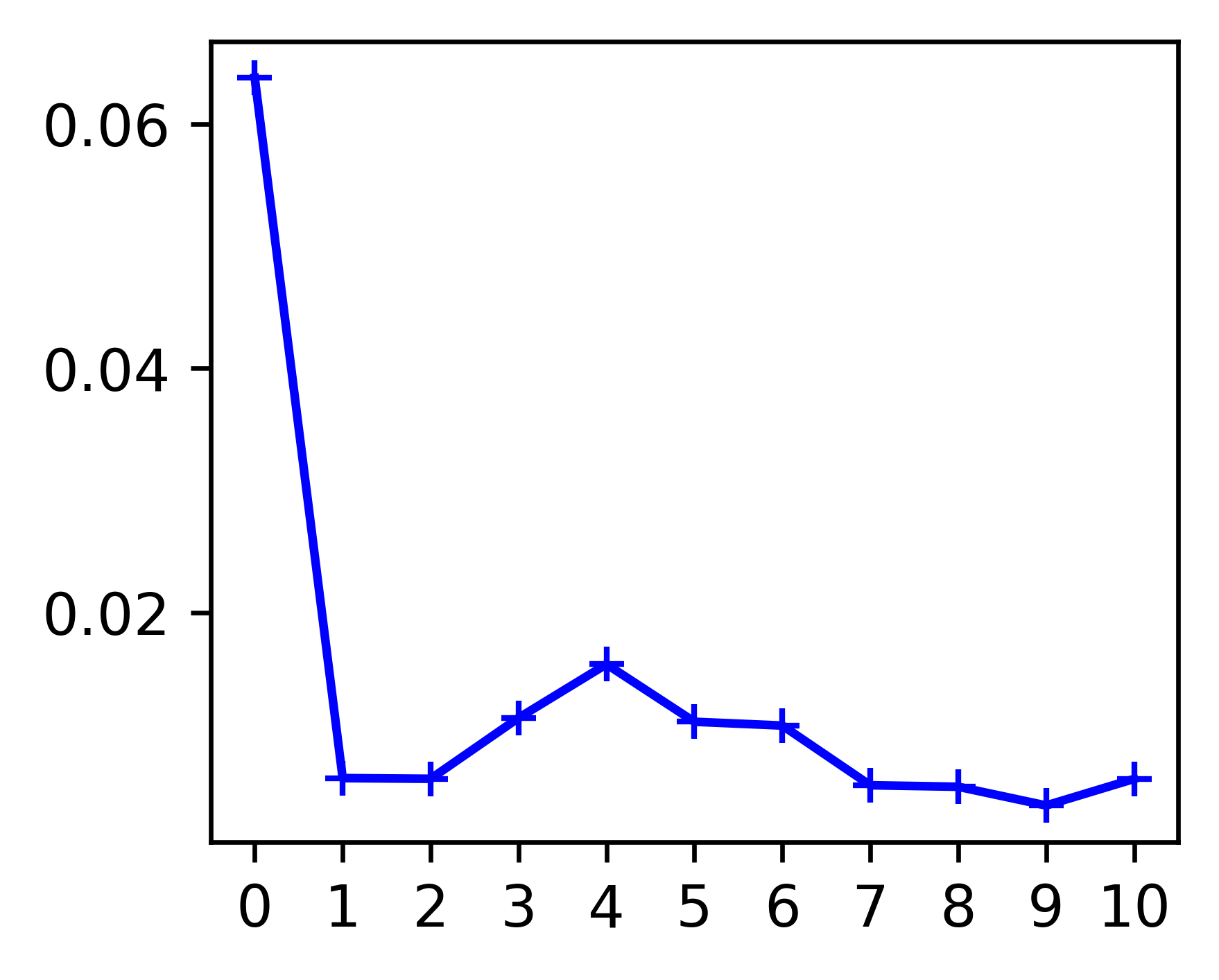}  & 
     \includegraphics[height=3.3cm]{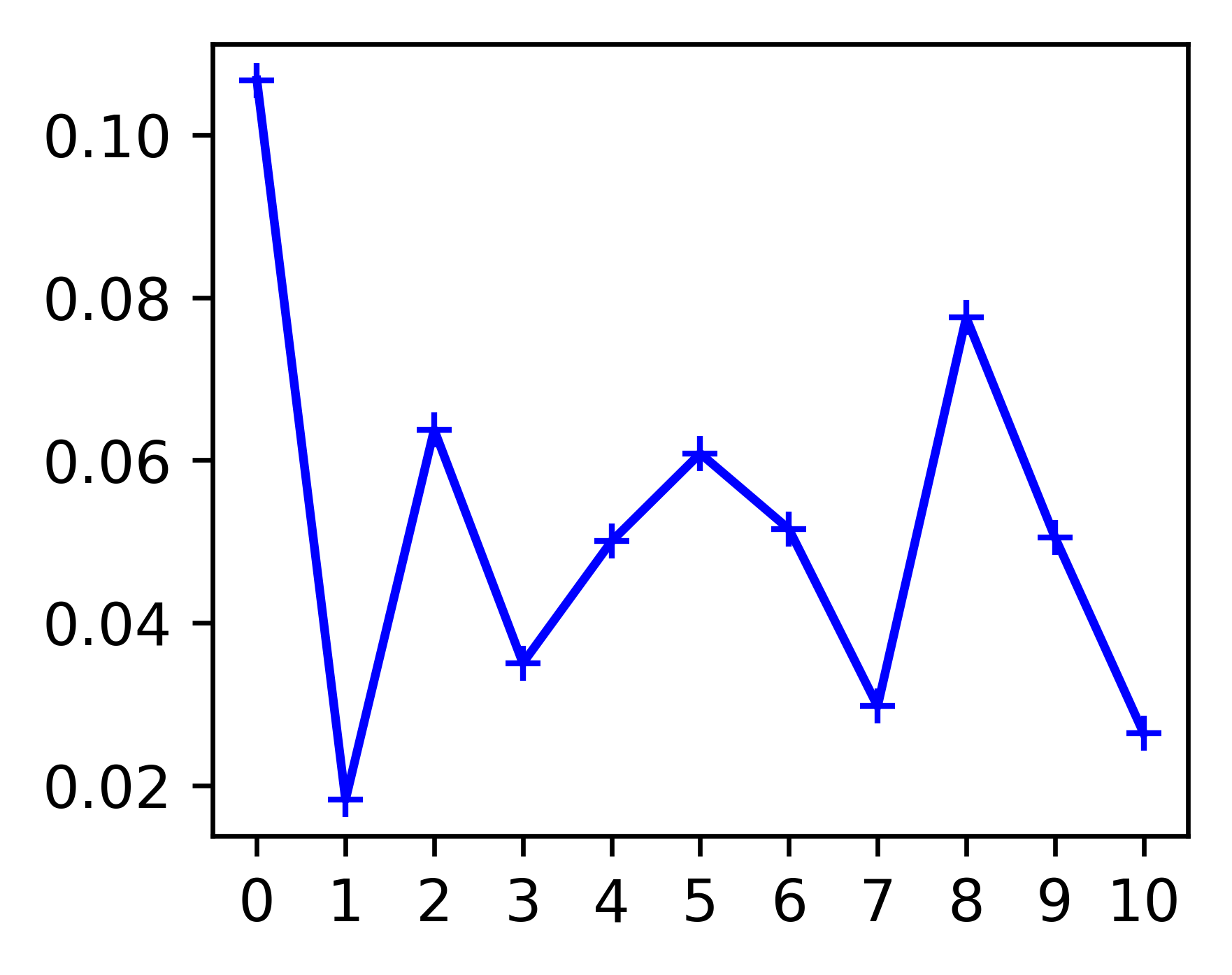} &
      \hspace{-0.5cm} \includegraphics[height=3.3cm]{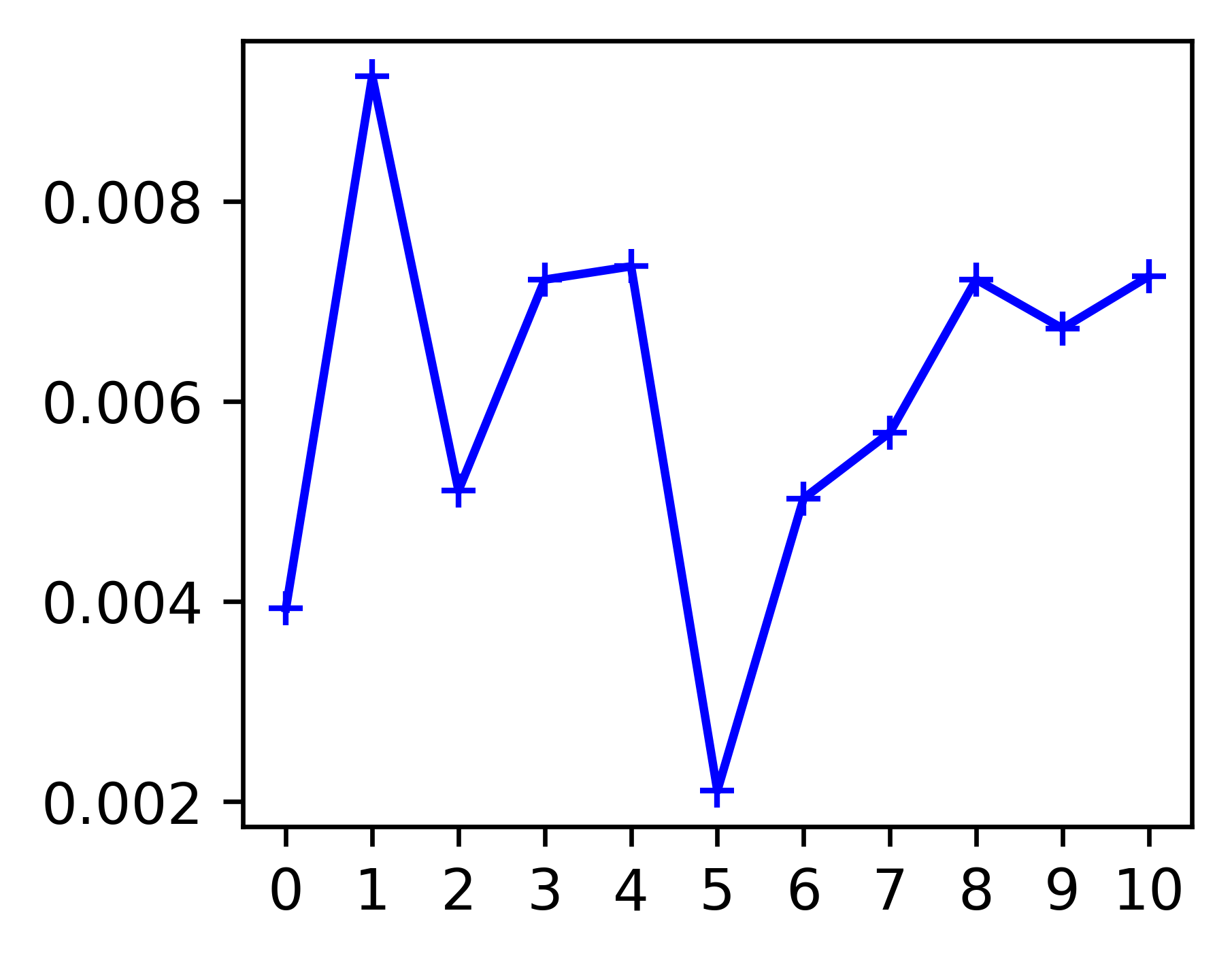} \\
     Moving MNIST & Traffic BJ &  SST  & Human 3.6
    \end{tabular}{}
    \caption{Mean amplitude of the combining coefficients $c_{i,j}$ with respect to the order of the differential operators approximated.}
    \label{fig:cij}
\end{figure}

With the same general backbone architecture, PhyDNet can express different PDE dynamics associated to the underlying phenomena by learning specific $c_{i,j}$ coefficients combining the partial derivatives in Eq (\ref{eq:phi}). In Figure \ref{fig:cij}, we display the mean amplitude of the learned coefficients $c_{i,j}$ with respect to the order of differentiation. For Moving MNIST, the $0^{th}$ and $1^{st}$ orders are largely dominant, meaning a purely advective behaviour coherent with the piecewise-constant translation dynamics of the dataset. For Traffic BJ and SST, there is also a global decrease in amplitude with respect to order, we nonetheless notice a few higher order terms appearing to be useful for prediction. 
For Human 3.6, where the nature of the prior motion is less obvious, these coefficients are more spread across order derivatives.

\paragraph*{Dealing with unreliable inputs} 
\label{sec:lt-forecasting}

\begin{figure}[H]
    \centering
    \begin{tabular}{cc}
    \hspace{-0.8cm} \includegraphics[width=6cm]{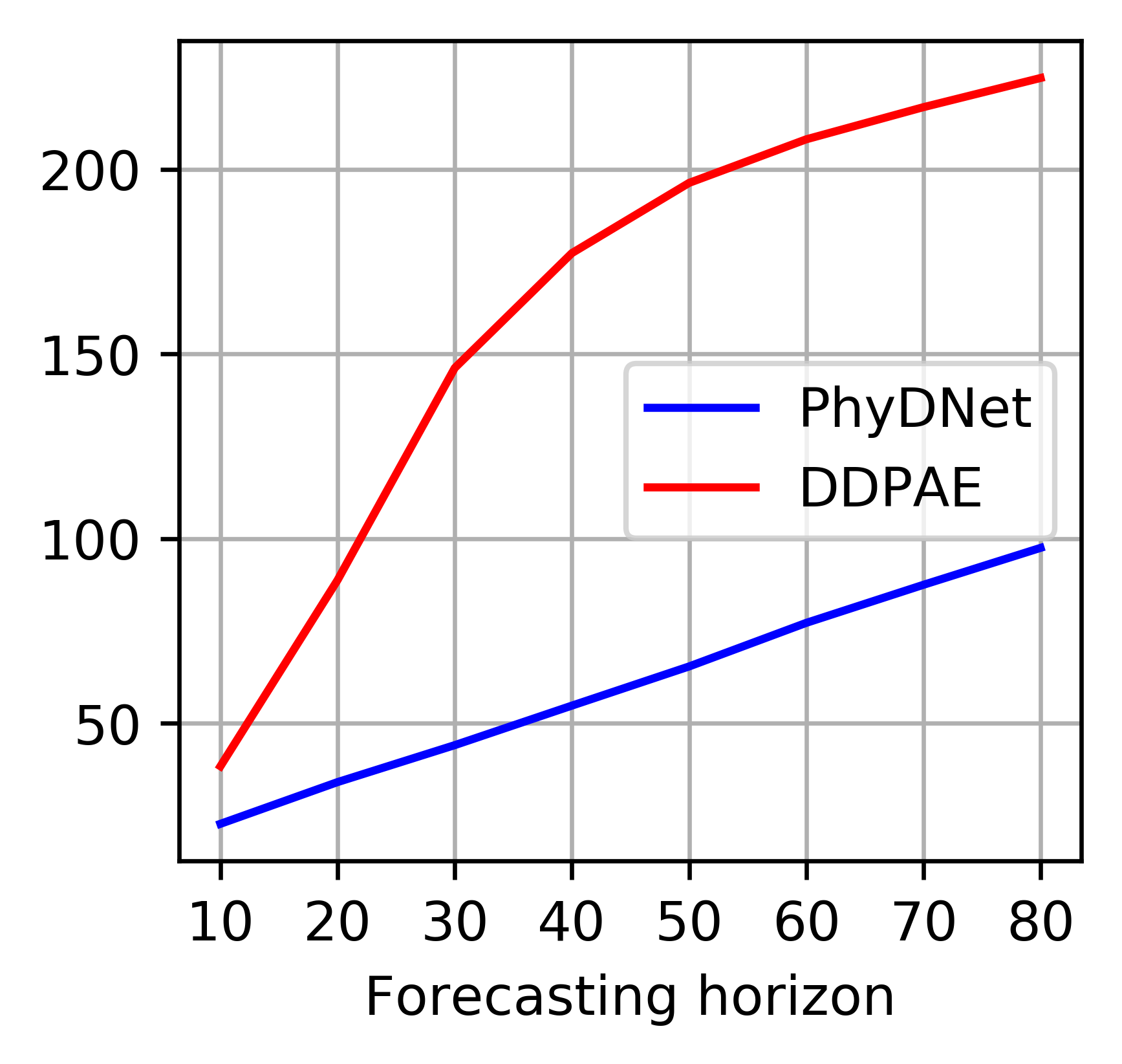}     &  \hspace{-0.5cm} \includegraphics[width=6cm]{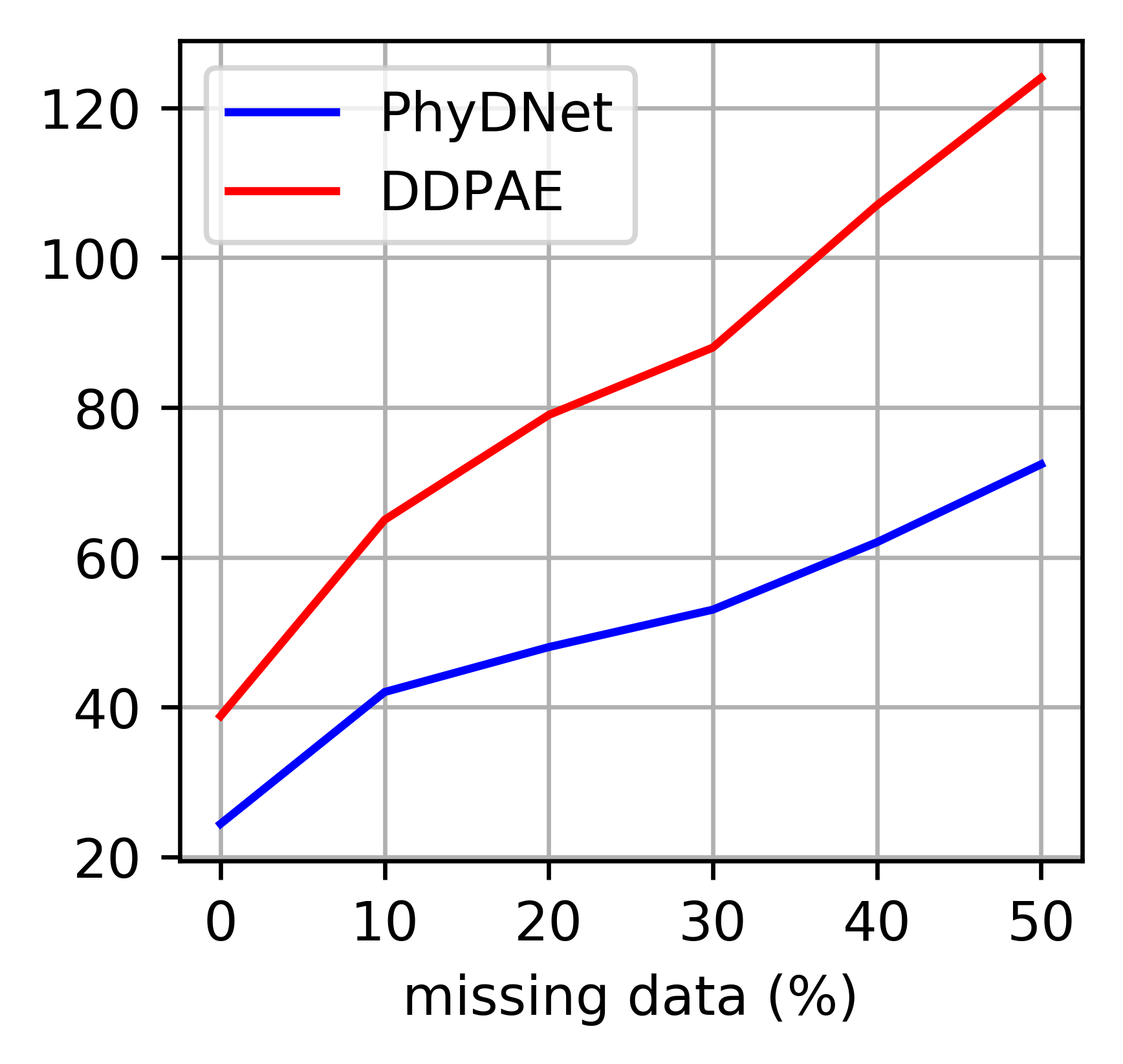} \\
     \hspace{-0.8cm} \includegraphics[width=6cm]{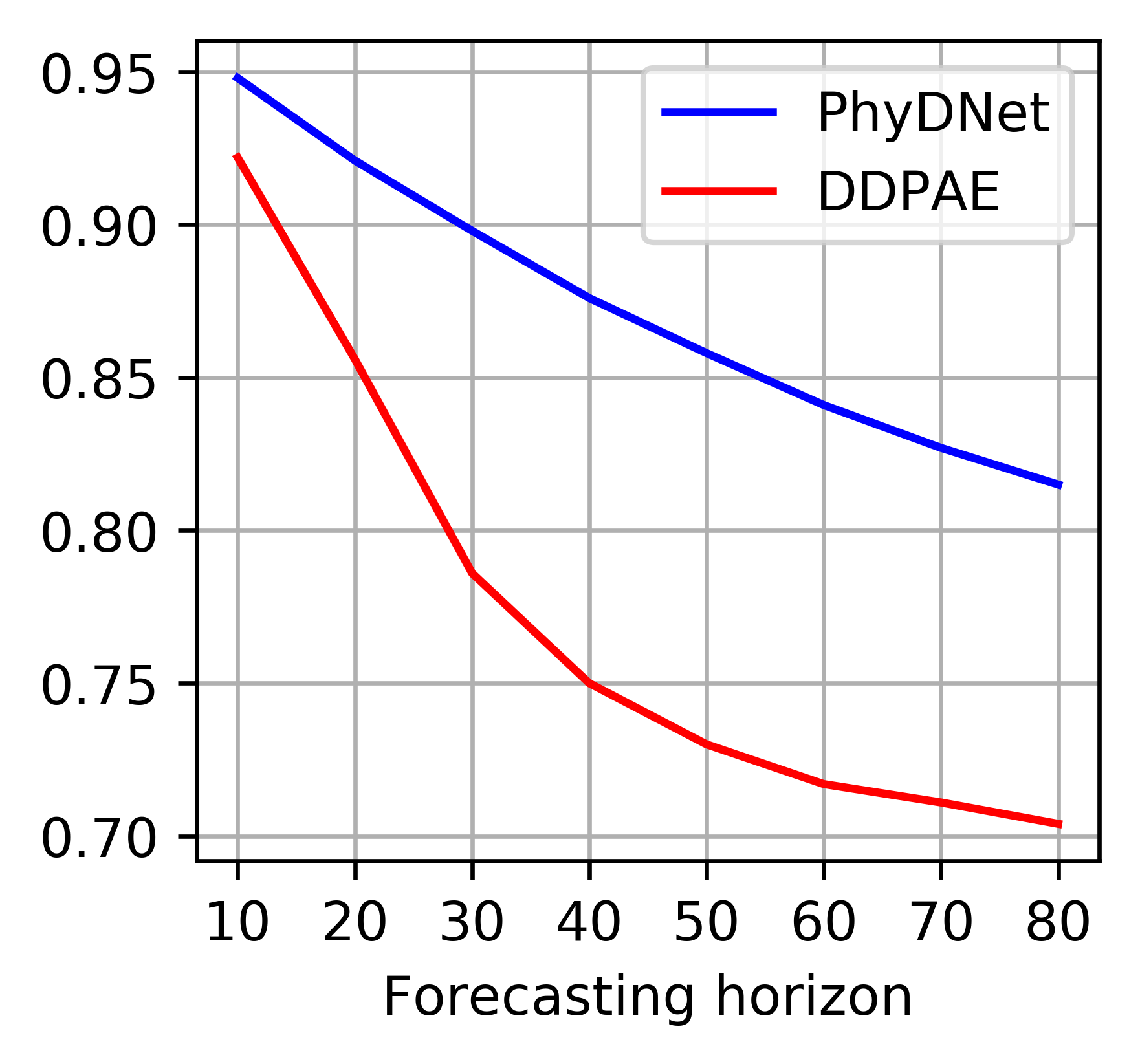}     &  \hspace{-0.5cm} \includegraphics[width=6cm]{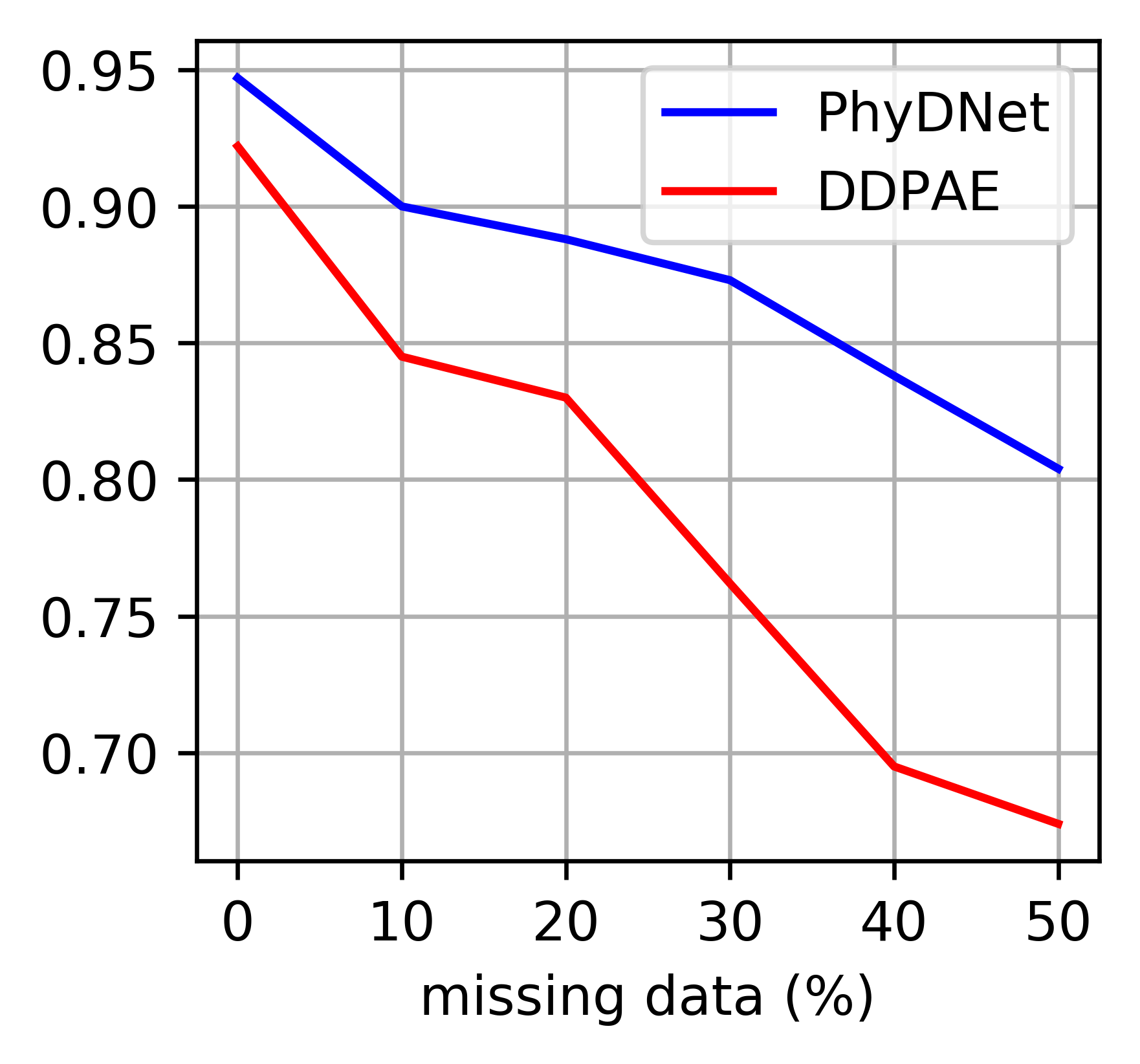} \\   
        (a) Long-term forecasting & (b) Missing data 
    \end{tabular} \\

    \caption{MSE comparison between PhyDNet and DDPAE \cite{hsieh2018learning} when dealing with unreliable inputs, for long-term forecasting (a) and in presence of missing data (b).}
    \label{fig:long-term}
\end{figure}

We explore here the robustness of PhyDNet when dealing with unreliable inputs, that can arise in two contexts: long-term forecasting and missing data. As explained in section~\ref{sec:training}, PhyDNet can be used in a prediction mode in this context, limiting the use of unreliable inputs, whereas general RNNs cannot. To validate the relevance of the prediction mode, we compare PhyDNet to DDPAE \cite{hsieh2018learning}, based on a standard RNN (LSTM) as predictor module. 
Figure \ref{fig:long-term} presents the results evaluated in MSE and SSIM obtained by PhyDNet and DDPAE on Moving MNIST.

For long-term forecasting, we evaluate the performances of both methods far beyond the prediction range seen during training (up to 80 frames), as shown in Figure \ref{fig:long-term}(a). We can see that the performance drop (MSE increase rate) is approximately linear for PhyNet, whereas it is much more pronounced for DDPAE. For example, PhyDNet for 80-steps prediction reaches similar performances in MSE than DDPAE for 20-steps prediction. This confirms that PhyDNet can limit error accumulation during forecasting by using a powerful dynamical model.

Finally, we evaluate the robustness of PhyDNet on DDPAE on missing data, by varying the ratio of missing data (from 10 to 50\%) in input sequences during training and testing. 
A missing input image is replaced with a default value (0) image. In this case, PhyCell 
relies only on its latent dynamics by setting $\mathbf{K}_t=0$, whereas DDPAE takes the null image as input. Figure \ref{fig:long-term}(b) shows that the performance gap between PhyDNet and DDPAE increases with the percentage of missing data. 

\section{Conclusion}

We have proposed PhyDNet, a new model for disentangling prior dynamical knowledge from other factors of variation required for video prediction. PhyDNet enables to apply PDE-constrained prediction beyond fully observed physical phenomena in pixel space, and to outperform state-of-the-art performances on four generalist datasets. Our introduced recurrent physical cell for modelling PDE dynamics generalizes recent models and offers the appealing property to decouple prediction from correction.

\clearpage{\pagestyle{empty}\cleardoublepage}

\mbox{}
\thispagestyle{empty}
\chapter{Augmenting incomplete physical models for complex dynamics forecasting}
\label{chap:aphynity}
\chapabstract{

\minitoc

\begin{center}
   \textsc{Chapter abstract}
\end{center}
\textit{
Forecasting complex dynamical phenomena in settings where only partial knowledge of their dynamics is available is a prevalent problem across various scientific fields. \eg in climate. While purely data-driven approaches are arguably insufficient in this context, standard physical modelling based approaches tend to be over-simplistic, inducing non-negligible errors. In this Chapter, we introduce the APHYNITY framework, a principled approach for augmenting \textit{incomplete} physical dynamics described by differential equations with deep data-driven models. It consists in decomposing the dynamics into two components: a physical component accounting for the dynamics for which we have some prior knowledge, and a data-driven component accounting for errors of the physical model. The learning problem is carefully formulated such that the physical model explains as much of the data as possible, while the data-driven component only describes information that cannot be captured by the physical model, no more, no less. This not only provides the existence and uniqueness for this decomposition, but also ensures interpretability and benefits generalization. Experiments made on three important use cases, each representative of a different family of physical phenomena, show that APHYNITY can efficiently leverage approximate physical models to accurately forecast the evolution of the system and correctly identify relevant physical parameters.\\
The work described in this Chapter is based on the following publication:
\begin{itemize}
    \item \cite{leguen2021augmenting,Yin_2021_jstat}: Yuan Yin$^*$, Vincent Le Guen$^*$, Jeremie Dona$^*$, Ibrahim Ayed$^*$, Emmanuel de Bézenac$^*$, Nicolas Thome and Patrick Gallinari. "Augmenting Physical Models with Deep Networks for Complex Dynamics Forecasting", In International Conference on Learning Representations (ICLR 2021, oral presentation), JSTAT 2021.
\end{itemize}
}
}

\section{Introduction\label{sec:intro}}

\lettrine[lines=3]{M}odelling and forecasting complex dynamical systems is a major challenge in domains such as environment and climate~\cite{rolnick2019tackling}, health science~\cite{choi2016retain}, and in many industrial applications~\cite{toubeau2018deep}. As explained in Chapter \ref{chap:intro}, Model-Based (MB) approaches typically rely on partial or ordinary differential equations (PDE/ODE) and stem from a deep understanding of the underlying physical phenomena. Machine learning (ML) and deep learning methods are more prior agnostic yet have become state-of-the-art for several spatio-temporal prediction. However, pure ML methods are still limited for modelling complex physical dynamics, and cannot properly generalize to new conditions unlike MB approaches.

Combining the MB and ML paradigms is an emerging trend to develop the interplay between the two paradigms. For example, \cite{brunton2016discovering,long2018pde} learn the explicit form of PDEs directly from data, \cite{raissi2017physics,sirignano2018dgm} use NNs as implicit methods for solving PDEs, \cite{seo2020} learn spatial differences with a graph network, \cite{ummenhofer2020} introduce continuous convolutions for fluid simulations, \cite{de2017deep} learn the velocity field of an advection-diffusion system,
\cite{greydanus2019hamiltonian,chen2019symplectic} enforce conservation laws in the network architecture or in the loss function. 

The large majority of aforementioned ML/MB hybrid approaches assume that the physical model adequately describes the observed dynamics. This assumption is, however, commonly violated in practice.  This may be due to various factors, \eg idealized assumptions and difficulty to explain processes from first principles \cite{gentine}, computational constraints prescribing a fine grain modelling of the system \cite{epnet}, unknown external factors, forces and sources which are present \cite{Large2004}.

\begin{figure}
\centering
\vspace{-0.6cm}
\begin{tabular}{ccc} 
 \hspace{-0.5cm} \includegraphics[height=4.5cm]{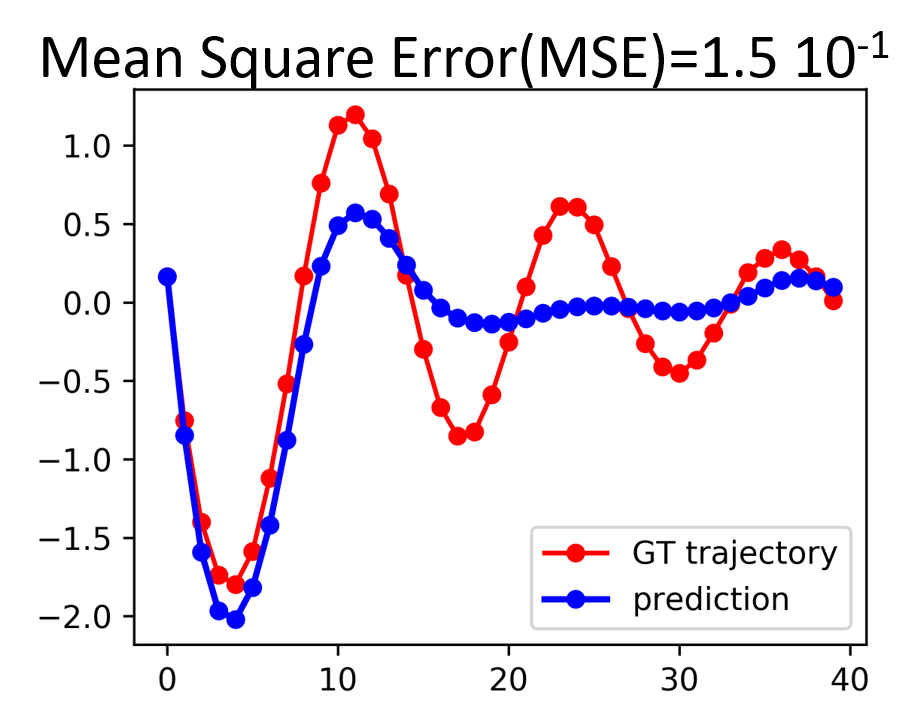} &  \hspace{-0.5cm} \includegraphics[height=4.5cm]{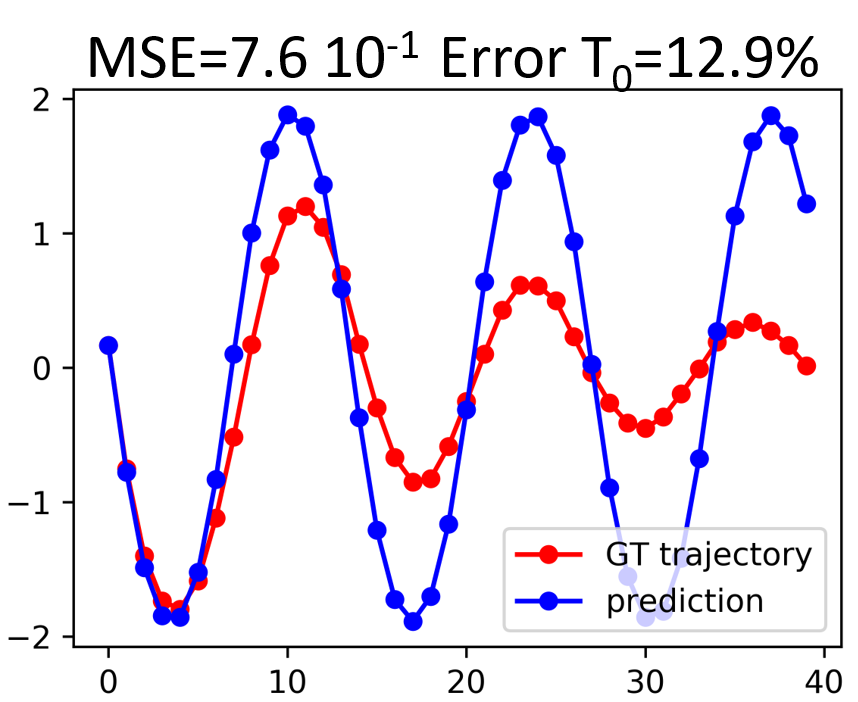} &
 \hspace{-0.5cm}
 \includegraphics[height=4.5cm]{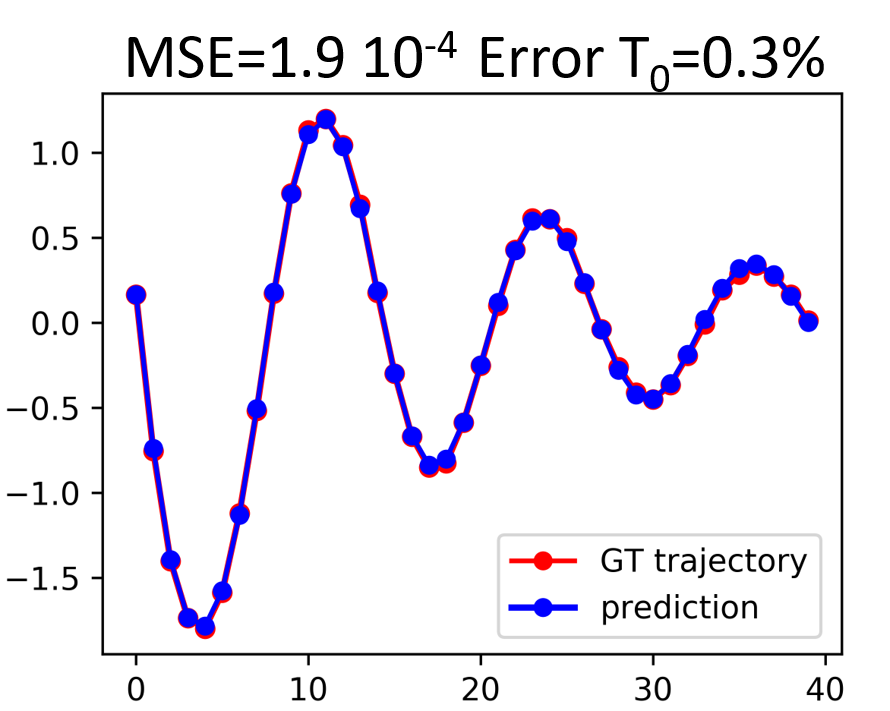}\\
(a) Data-driven Neural ODE  & (b) Simple physical model & (c) Our APHYNITY framework
\end{tabular}
    \caption[APHYNITY motivation.]{Predicted dynamics for the damped pendulum vs. ground truth (GT) trajectories $\nicefrac{\diff^2 \theta}{\diff t^2} + \omega_0^2 \sin \theta + \alpha \nicefrac{\diff \theta}{\diff t} = 0$. We show that in (a) the data-driven approach~\cite{chen2018neural} fails to properly learn the dynamics due to the lack of training data, while in (b) an ideal pendulum cannot take friction into account. The proposed APHYNITY shown in (c) augments the over-simplified physical model in (b) with a data-driven component. APHYNITY improves both forecasting (MSE) and parameter identification (Error $T_0$) compared to (b).}
    \label{fig:comparison_data_phys_coop}
\end{figure}

In this Chapter, we aim at leveraging prior dynamical ODE/PDE knowledge in situations where this physical model is \textit{incomplete}, \ie unable to represent the whole complexity of observed data. To handle this case, we introduce a principled learning framework to Augment incomplete PHYsical models for ideNtIfying and forecasTing complex dYnamics~(APHYNITY). The rationale of APHYNITY, illustrated in Figure~\ref{fig:comparison_data_phys_coop} on the pendulum problem, is to \textit{augment} the physical model when---and only when---it falls short.

Designing a general method for combining ML and MB approaches is still a widely open problem, and a clear problem formulation for the latter is lacking \cite{Reichstein2019}. Our contributions towards these goals are the following:
\begin{itemize}
\item We introduce a simple yet principled framework for combining both approaches. We decompose the data into a physical and a data-driven term such that the data-driven component only models information that cannot be captured by the physical model. We provide existence and uniqueness guarantees~(Section~\ref{subsec:decomp}) for the decomposition given mild conditions, and show that this formulation ensures interpretability and benefits generalization.

\item We propose a trajectory-based training formulation~(Section~\ref{subsec:learning}) along with an adaptive optimization scheme~(Section~\ref{subsec:optim}) enabling end-to-end learning for both physical and deep learning components. This allows APHYNITY to \textit{automatically} adjust the complexity of the neural network to different approximation levels of the physical model, paving the way to flexible learned hybrid models.
    
\item We demonstrate the generality of the approach on three use cases (reaction-diffusion, wave equations and the pendulum) representative of different PDE families (parabolic, hyperbolic), having a wide spectrum of application domains, \eg acoustics, electromagnetism, chemistry, biology, physics~(Section~\ref{sec:expes}). We show that APHYNITY is able to achieve performances close to complete physical models by augmenting incomplete ones, both in terms of forecasting accuracy and physical parameter identification. Moreover, APHYNITY can also be successfully extended to the non-stationary dynamics context (Section \ref{sec:aph-nonstat}).
\end{itemize}

\section{Related work}
\label{sec:related-work}

\paragraph{Correction in data assimilation} As discussed in Chapter \ref{chap:related_work}, data assimilation techniques such as the Kalman filter \cite{kalman1960new,becker2019recurrent} assume that the prediction errors correspond to noise. These errors are modelled probabilistically as random variables, and an optimal correction step is derived after each prediction step. In this sequential two-step scheme, also arising commonly in robotics and optimal control \cite{chen2004disturbance,li2014disturbance}, there is no cooperation between prediction and correction. The originality of APHYNITY is to leverage model-based prior knowledge by augmenting it with neurally parameterized dynamics; the residual does not corresponds to noise but to an unknown or unmodelled part of the dynamical model. APHYNITY also ensures an optimal cooperation between the prior model and the augmentation.

\paragraph{Augmented physical models} Combining physical models with machine learning (\textit{gray-box or \textit{hybrid}} modelling) was first explored from the 1990's: \cite{psichogios1992hybrid,thompson1994modeling,rico1994continuous} use neural networks to predict the unknown parameters of physical models. The challenge of proper MB/ML cooperation was already raised as a limitation of gray-box approaches but not addressed. Moreover these methods were evaluated on specific applications with a residual targeted to the form of the equation.
In the last few years, there has been a renewed interest in deep hybrid models bridging data assimilation techniques and machine learning to identify complex PDE parameters using cautiously constrained forward model \cite{long2018pde,de2017deep}.

Recently, some approaches have specifically targetted the ML/MB cooperation in the case of incomplete physical models. HybridNet~\cite{long2018hybridnet} and PhICNet~\cite{saha2020phicnet} both use data-driven networks to learn additive perturbations or source terms to a given PDE. The former considers the favorable context where the perturbations can be accessed, and the latter the special case of additive noise on the input. \cite{wang2019integrating,neural20} propose several empirical fusion strategies with deep neural networks but lack theoretical groundings. Crucially, all the aforementioned approaches do not address the issues of uniqueness of the decomposition or of proper cooperation for correct parameter identification. Besides, we found experimentally that this vanilla cooperation is inferior to the APHYNITY learning scheme in terms of forecasting and parameter identification performances~(see experiments in Section~\ref{sec:results}).

\section{The APHYNITY Model}
\label{sec:model}

In the following, we study dynamics driven by an equation of the form:
\begin{equation}
\label{eq:ode}
\frac{\diff X_t}{\diff t} = F(X_t)    
\end{equation}
defined over a finite time interval $[0,T]$, where the state $X$ is either vector-valued, \ie we have $X_t\in\R^d$ for every $t$ (pendulum equations in Section \ref{sec:expes}), or $X_t$ is a $d$-dimensional vector field over a spatial domain $\Omega\subset\R^k$, with $k\in\{2,3\}$, \ie $X_t(x)\in\R^d$ for every $(t,x)\in[0,T]\times\Omega$ (reaction-diffusion and wave equations in Section \ref{sec:expes}). 
We suppose that we have access to a set of observed trajectories $\D = \{X_\cdot:[0,T]\rightarrow\A \ |\ \forall t\in[0,T], \nicefrac{\diff X_t}{\diff t} = F(X_t)\}$, where $\A$ is the set of $X$ values (either $\R^d$ or vector field). In our case, the unknown $F$ has $\A$ as domain and we only assume that $F\in\F$, with $(\F, \|\cdot\|)$ a normed vector space.

\begin{figure}
    \centering
    \includegraphics[width=\linewidth]{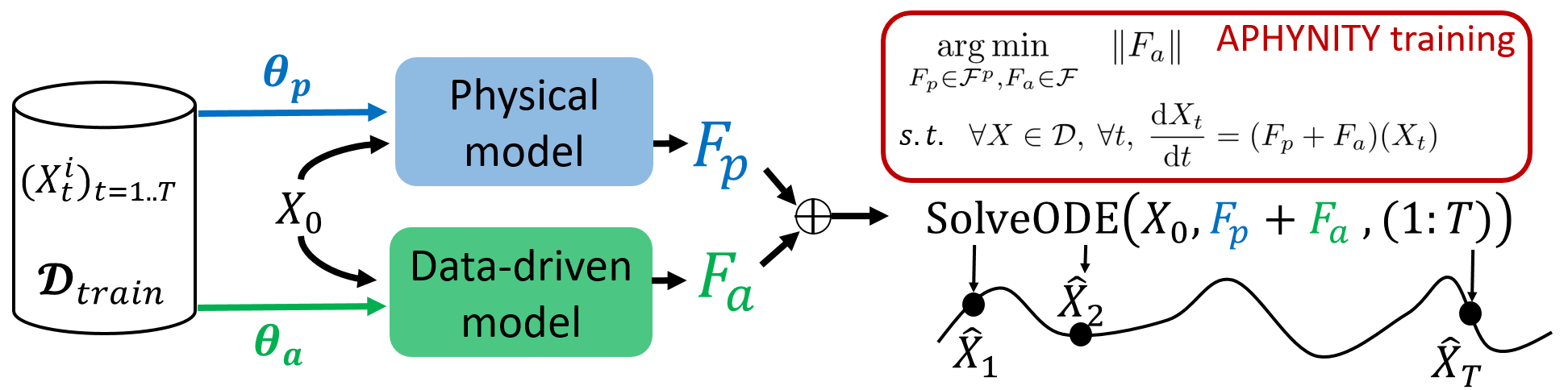}
    \caption[Principle of the APHYNITY framework.]{The APHYNITY model for learning complex dynamical systems augments an approximate physical model $F_p$ by a deep data-driven model $F_a$. We propose a decomposition fulfilling uniqueness guarantees (Section~\ref{subsec:decomp}). We introduce a trajectory-based formulation for learning the joint ODE $\frac{\diff X_t}{\diff t}=(F_p+F_a)(X_t)$, which leads to different and experimentally better identification results than the physical model $F_p$  (Section~\ref{subsec:learning}). APHYNITY is learned end-to-end with an adaptive optimization algorithm (Section \ref{subsec:optim}) ensuring a meaningful cooperation between physics and augmentation.}
    \label{fig:fig_aphynity}
\end{figure}

The overall APHYNITY approach is illustrated in Figure \ref{fig:fig_aphynity}.

\subsection{Decomposing dynamics into physical and augmented terms\label{subsec:decomp}}

As introduced in \ref{sec:intro}, we consider the common situation where incomplete information is available on the dynamics, under the form of a family of ODEs or PDEs characterized by their temporal evolution $F_p\in\F_p\subset\F$. The APHYNITY framework leverages the knowledge of $\F_p$ while mitigating the approximations induced by this simplified model through the combination of physical and data-driven components. $\F$ being a vector space, we can write:
\[
F = F_p + F_a,
\]
where $F_p\in\F_p$ encodes the incomplete physical knowledge and $F_a\in\F$ is the  data-driven augmentation term complementing $F_p$. The incomplete physical prior is supposed to belong to a known family, but the physical parameters~(\eg propagation speed for the wave equation) are unknown and need to be estimated from data. Both $F_p$ and $F_a$ parameters are estimated by fitting the trajectories from $\D$.

The decomposition $F = F_p + F_a$ is in general not unique. For example, all the dynamics could be captured by the $F_a$ component. This decomposition is thus ill-defined, which hampers the interpretability and the extrapolation abilities of the model. In other words, one wants the estimated parameters of $F_p$ to be as close as possible to the true parameter values of the physical model and $F_a$ to play only a complementary role w.r.t $F_p$, so \textit{as to model only the information that cannot be captured by the physical prior}. For example, when $F\in\F_p$, the data can be fully described by the physical model, and in this case it is sensible to desire $F_a$ to be nullified; this is of central importance in a setting where one wishes to identify physical quantities, and for the model to generalize and extrapolate to new conditions. In a more general setting where the physical model is incomplete, the action of $F_a$ on the dynamics, as measured through its norm, should be as small as possible.

This general idea is embedded in the following optimization problem:
\begin{equation}
\label{eq:aphynity-opt}
\underset{F_p\in\F_p, F_a\in\F}{\min} ~~~\left\Vert  F_a  \right\Vert ~~~ 
\mathrm{subject~to} ~~~~ \forall X\in\D, \forall t, \frac{\diff X_t}{\diff t} =(F_p+F_a)(X_t).
\end{equation}

The originality of APHYNITY is to leverage model-based prior knowledge by augmenting it with neurally parameterized dynamics. It does so while ensuring optimal cooperation between the prior model and the augmentation.

A first key question is whether the minimum in Eq \ref{eq:aphynity-opt} is indeed well-defined, in other words whether there exists indeed a decomposition with a minimal norm $F_a$. The answer actually depends on the geometry of $\F_p$, and is formulated in the following proposition proven in Appendix~\ref{app:proof}:
\begin{prop}[Existence of a minimizing pair]\label{prop:exist_unique}
If $\F_p$ is a proximinal set\footnote{\label{fn:proximal-chebyshev}A proximinal set is one from which every point of the space has at least one nearest point. A Chebyshev set is one from which every point of the space has a unique nearest point. More details in Appendix~\ref{app:chebyshev}.}, there exists a decomposition minimizing Eq \ref{eq:aphynity-opt}.
\end{prop}
Proximinality is a mild condition which, as shown through the proof of the proposition, cannot be weakened. It is a property verified by any boundedly compact set. In particular, it is true for closed subsets of finite dimensional spaces. However, if only existence is guaranteed, while forecasts would be expected to be accurate, non-uniqueness of the decomposition would hamper the interpretability of $F_p$ and this would mean that the identified physical parameters are not uniquely determined. 

It is then natural to ask under which conditions solving problem Eq \ref{eq:aphynity-opt} leads to a unique decomposition into a physical and a data-driven component. The following result provides guarantees on the existence and uniqueness of the decomposition under mild conditions. The proof is given in Appendix~\ref{app:proof}:
\begin{prop}[Uniqueness of the minimizing pair]\label{prop:unique}
If $\F_p$ is a Chebyshev set\textcolor{red}{\footnotemark[1]}, Eq \ref{eq:aphynity-opt} admits a unique minimizer. The $F_p$ in this minimizer pair is the metric projection of the unknown $F$ onto $\F_p$.
\end{prop}
The Chebyshev assumption condition is strictly stronger than proximinality but is still quite mild and necessary. Indeed, in practice, many sets of interest are Chebyshev, including all closed convex spaces in strict normed spaces and, if $\F = L^2$, $\F_p$ can be any closed convex set, including all finite dimensional subspaces. In particular, all examples considered in the experiments are Chebyshev sets.

Propositions \ref{prop:exist_unique} and \ref{prop:unique} provide, under mild conditions, the theoretical guarantees for the  APHYNITY formulation to infer the correct MB/ML decomposition, thus enabling both recovering the proper physical parameters and accurate forecasting.

\subsection{Solving APHYNITY with deep neural networks\label{subsec:learning}}
In the following, both terms of the decomposition are parametrized and are denoted as $F_p^{\theta_p}$ and $F_a^{\theta_a}$. Solving APHYNITY then consists in estimating the parameters $\theta_p$ and $\theta_a$. $\theta_p$ are the physical parameters and are typically low-dimensional, \eg 2 or 3 in our experiments for the considered physical models. For $F_a$, we need sufficiently expressive models able to optimize over all $\F$: we thus use deep neural networks, which have shown promising performances for the approximation of differential equations~\cite{raissi2017physics,ayed2019learning}. 

When learning the parameters of $F_p^{\theta_p}$ and $F_a^{\theta_a}$, we have access to a finite dataset of trajectories discretized with a given temporal resolution $\Delta t$: $\D_{\text{train}} = \{(X^{(i)}_{k\Delta t})_{0\leq k\leq \left \lfloor{\nicefrac{T}{\Delta t}}\right \rfloor} \}_{1\leq i\leq N}$. Solving Eq  \ref{eq:aphynity-opt} requires estimating the state derivative $\nicefrac{\diff X_t}{\diff t}$ appearing in the constraint term. One solution is to approximate this derivative using \eg finite differences as in \cite{brunton2016discovering,greydanus2019hamiltonian,cranmer2020lagrangian}. This numerical scheme requires high space and time resolutions in the observation space in order to get reliable gradient estimates. Furthermore it is often unstable, leading to explosive numerical errors as discussed in Appendix~\ref{app:der_superv}. We propose instead to solve Eq \ref{eq:aphynity-opt} using an integral trajectory-based approach: we compute $\widetilde{X}
^i_{k\Delta t, X_0}$ from an initial state $X^{(i)}_0$ using the current $F_p^{\theta_p} + F_a^{\theta_a}$ dynamics, then enforce the constraint $\widetilde{X}^i_{k\Delta t, X_0}=X^i_{k\Delta t}$. This leads to our final objective function on $(\theta_p, \theta_a)$: 
\begin{equation}
\label{eq:opt_final}
\underset{\theta_p, \theta_a}{\min} ~~~\left\Vert  F_a^{\theta_a}  \right\Vert ~~~
\mathrm{subject~to} ~~~~ \forall i, \forall k, \widetilde{X}^{(i)}_{k\Delta t} = X^{(i)}_{k\Delta t},
\end{equation}
where  $\widetilde{X}^{(i)}_{k\Delta t}$ is the approximate solution of the integral $ \int_{X_0^{(i)}}^{X_0^{(i)}+k \Delta t} (F_p^{\theta_p} + F_a^{\theta_a})(X_s) \diff X_s$ obtained by a differentiable ODE solver. 

In our setting, where we consider situations for which $F^{\theta_p}_p$ only partially describes the physical phenomenon, this coupled ML/MB formulation leads to different parameter estimates than using the MB formulation alone, as analyzed more thoroughly in Appendix~\ref{app:alt_methods}.

Interestingly, our experiments show that using this formulation also leads to a better identification of the physical parameters $\theta_p$ than when fitting the simplified physical model $F^{\theta_p}_p$ alone~(Section~\ref{sec:expes}). With only an incomplete knowledge on the physics, $\theta_p$ estimator will be biased by the additional dynamics which needs to be fitted in the data. 
Appendix~\ref{app:ablation} also confirms that the integral formulation gives better forecasting results and a more stable behavior than supervising over finite difference approximations of the derivatives.

\subsection{Adaptively constrained optimization\label{subsec:optim}}

The formulation in Eq \ref{eq:opt_final} involves constraints which are difficult to enforce exactly in practice.
We considered a variant of the method of multipliers~\cite{constrained_optim} which uses a sequence of Lagrangian relaxations $\mathcal{L}_{\lambda_j}(\theta_p, \theta_a)$:
 \begin{equation}
 \label{eq:opt_final_relaxed}
\mathcal{L}_{\lambda_j}(\theta_p, \theta_a) = \|F_a^{\theta_a}\| + \lambda_j \cdot \mathcal{L}_{traj}(\theta_p, \theta_a),
 \end{equation}

where $\mathcal{L}_{traj}(\theta_p, \theta_a) = \sum_{i=1}^N\sum_{h=1}^{T/\Delta t} \|X^{(i)}_{h\Delta t} -  \widetilde{X}^{(i)}_{h\Delta t}   \|$.

\begin{algorithm}[H]
\SetAlgoLined
Initialization: $\lambda_0\geq0, \tau_1 >0, \tau_2>0$\;
\For{epoch = $1:N_{epochs}$} {
\For{iter in $1:N_{iter}$}{
\For{batch in $1:B$}{
$\theta_{j+1} = \theta_j - \tau_1 \nabla \left[ \lambda_j\mathcal{L}_{traj}(\theta_j) + \left\Vert F_a \right\Vert  \right]  \;$
}
 }
 $\lambda_{j+1} =  \lambda_j +$ $\tau_2\mathcal{L}_{traj}(\theta_{j+1}) \;$
 }
  \caption{\label{alg:optim} APHYNITY}
\end{algorithm}

 This method needs an increasing sequence $(\lambda_j)_j$ such that the successive minima of $\mathcal{L}_{\lambda_j}$ converge to a solution~(at least a local one) of the constrained problem in Eq \ref{eq:opt_final}. We select $(\lambda_j)_j$ by using an iterative strategy: starting from a value $\lambda_0$, we iterate, minimizing $\mathcal{L}_{\lambda_j}$ by gradient descent\footnote{Convergence to a local minimum isn't necessary, a few steps are often sufficient for a successful optimization.}, then update $\lambda_j$ with:
$\lambda_{j+1} = \lambda_j + \tau_2\mathcal{L}_{traj}(\theta_{j+1})$, where $\tau_2$ is a chosen hyper-parameter and $\theta = (\theta_p, \theta_a)$. This procedure is summarized in Algorithm~\ref{alg:optim}. This adaptive iterative procedure allows us to obtain stable and robust results, in a reproducible fashion, as shown in the experiments.

\section{Experimental validation\label{sec:expes}}

We validate our approach on 3 classes of challenging physical dynamics: the damped pendulum, reaction-diffusion, and wave propagation, representative of various application domains such as chemistry, biology or ecology (for reaction-diffusion) \cite{cantrell2004spatial,chung2007bifurcation,volpert2009reaction}  and earth physic, acoustic, electromagnetism or even neuro-biology (for waves equations) \cite{Slater1937, NUNEZ1974}.

~The last two dynamics are described by PDEs and thus in practice should be learned from very high-dimensional vectors, discretized from the original compact domain. This makes the learning much more difficult than from the one-dimensional pendulum case. For each problem, we investigate the cooperation between physical models of increasing complexity encoding incomplete knowledge of the dynamics~(denoted \textit{Incomplete physics} in the following) and data-driven models. We show the relevance of APHYNITY~(denoted \textit{APHYNITY models}) both in terms of forecasting accuracy and physical parameter identification.

\subsection{Experimental setting\label{subsec:exp_sett}}

We describe the three families of equations studied in the experiments. In all experiments, $\F=\mathcal{L}^2(\A)$ where $\A$ is the set of all admissible states for each problem, and the $\mathcal{L}^2$ norm is computed on $\D_{train}$ by: $\|F\|^2 \approx \sum_{i,k}\|F(X^{(i)}_{k\Delta t})\|^2$. All considered sets of physical functionals $\F_p$ are closed and convex in $\F$ and thus are Chebyshev. In order to enable the evaluation on both prediction and parameter identification, all our experiments are conducted on simulated datasets with known model parameters. Each dataset has been simulated using an appropriate high-precision integration scheme for the corresponding equation. All solver-based models take the first state $X_0$ as input and predict the remaining time-steps by integrating $F$ through the same differentiable generic and common ODE solver~(4th order Runge-Kutta)\footnote{This integration scheme is then different from the one used for data generation, the rationale for this choice being that when training a model one does not know how exactly the data has been generated.}. Implementation details and architectures are given in Appendix~\ref{app:implementation}.

\paragraph{Damped pendulum:}
The evolution of a damped pendulum is governed by the ODE $\frac{\diff^2\theta}{\diff t^2} + \omega_0^2 \sin \theta  +\lambda \frac{\diff\theta}{\diff t} = 0$, where $\theta(t)$ is the angle, $\omega_0 = \frac{2 \pi}{T_0}$ is the proper pulsation~($T_0$ being the period) and $\lambda$ is the damping coefficient. With the state $X = (\theta, \frac{\diff\theta}{\diff t})$, the ODE can be written as in Eq \ref{eq:ode} with
$ F : X \mapsto ( \frac{\diff\theta}{\diff t} ,  - \omega_0^2 \sin \theta  - \lambda \frac{\diff\theta}{\diff t})$.

\noindent We consider the following physical models of increasing complexity:
\begin{itemize}[leftmargin=*,align=left]
\itemsep0em
    \item \textit{Hamiltonian models~}\cite{greydanus2019hamiltonian,toth2019hamiltonian}, an energy conservative approximation of the system, with $\F_p = \{F^{\mathcal{H}}_p: \linebreak (u,v) \mapsto (\partial_y \mathcal{H}(u,v), -\partial_x \mathcal{H}(u,v))\ |\ \mathcal{H}\in H^1(\R^2)\}$ where $H^1(\R^2)$ is the first order Sobolev space. 
    \item \textit{Param ODE ($\omega_0$)}, the pendulum without friction, with $\F_p = \{F^{\omega^2_0}_p:(u,v) \mapsto (v,- \omega_0^2 \sin u)\  |\ \linebreak \omega^2_0\geq\omega_{\min}^2\}$.  
    \item \textit{Param ODE ($\omega_0, \lambda$)}, the full pendulum equation (but with unknown parameters), with $\F_p=\{F^{\omega^2_0,\lambda}_p:(u,v)\mapsto(v,-\omega_0^2 \sin u - \lambda v)\ |\ \omega^2_0\geq\omega_{\min}^2, \lambda\geq\lambda_{\min}>0\}$.
\end{itemize}

\paragraph{Reaction-diffusion equations:}
We  consider a 2D FitzHugh-Nagumo type model~\cite{klaasen1984fitzhugh}. The system is driven by the PDE 
\(
        \frac{\partial u}{\partial t} = a\Delta u + R_u(u,v; k),
    \frac{\partial v}{\partial t} = b\Delta v + R_v(u,v)\)
where $a$ and $b$ are respectively the diffusion coefficients of $u$ and $v$, $\Delta$ is the Laplace operator. The local reaction terms are $R_u(u,v; k) = u - u^3 - k - v, R_v(u,v) = u - v$. The state is $X=(u,v)$ and is defined over a compact rectangular domain $\Omega$ with periodic boundary conditions.

\noindent The considered physical models are: 
\begin{itemize}[leftmargin=*,align=left]
\itemsep0em
    \item \textit{Param PDE ($a,b$)} with unknown ($a,b$) diffusion terms and without reaction terms:
    $\F_p = \{F_p^{a,b}:(u,v) \mapsto (a\Delta u, b\Delta v)\ | \ a\geq a_{\min} >0, b\geq b_{\min}>0\}$; 
    \item \textit{Param PDE ($a,b,k$)} the full PDE with unknown parameters: 
    $\F_p = \{F_p^{a,b,k} : (u,v)\mapsto \linebreak (a\Delta u + R_u(u,v;k),   b\Delta v + R_v(u,v)\ |\ a\geq a_{\min} > 0, b\geq b_{\min} > 0, k\geq k_{\min} > 0\}$.
\end{itemize}

\paragraph{Damped wave equations:}
We investigate the following 2-dimensional damped-wave PDE:
\(
    \frac{\partial^2 w}{\partial t^2} - c^2\Delta w + k \frac{\partial w}{\partial t}=0 
\)
where $k$ is the \textit{damping coefficient}. The state is $X=(w, \frac{\partial w}{\partial t}
)$ and, as for reaction-diffusion, we consider a compact spatial domain $\Omega$ with Neumann homogeneous boundary conditions. Note that this damping differs from the pendulum case, as its effect is global.

\noindent The considered physical models are:
\begin{itemize}[leftmargin=*,align=left]
\itemsep0em
    \item \textit{Param PDE ($c$)}, without damping term and $\F_p = \{F^c_p: (u, v) \mapsto \linebreak (v, c^2 \Delta u) \ |\ c\geq c_{\min} > 0\}$;
    \item \textit{Param PDE ($c,k$)} with $\F_p = \{F^{c,k}_p:(u, v) \mapsto (v, c^2 \Delta u - kv)\ | \linebreak  \ c\geq c_{\min} > 0, k\geq k_{\min} > 0\}$.
\end{itemize}

\paragraph{Baselines}
As purely data-driven baselines, we use Neural ODE~\cite{chen2018neural} for the three problems and PredRNN++~(\cite{wang2018predrnn++}, for reaction-diffusion only) which are competitive models for datasets generated by differential equations and  for spatio-temporal data. As ML/MB methods, in the ablations studies (see Appendix \ref{app:ablation}), we compare for all problems, to the vanilla ML/MB cooperation scheme found in \cite{wang2019integrating,neural20}. We also show results for \textit{True PDE/ODE}, which corresponds to the equation for data simulation~(which do not lead to zero error due to the difference between simulation and training integration schemes). For the pendulum, we compare to Hamiltonian neural networks \cite{greydanus2019hamiltonian,toth2019hamiltonian} and to the the deep Galerkin method (DGM) \cite{sirignano2018dgm}. See additional details in Appendix~\ref{app:implementation}.

\subsection{Results}
\label{sec:results}

\begin{table}[t!]
    \centering
    \vspace{-0.6cm}
    \footnotesize
\setlength{\tabcolsep}{8.5pt}
    \caption[Forecasting and identification results with APHYNITY.]{Forecasting and identification results on the (a) damped pendulum, (b) reaction-diffusion and (c) wave equation datasets. We set for (a) $T_0=6$, $\alpha=0.2$, for (b) $a=1\times 10^{-3}, b=5\times 10^{-3}, k=5\times 10^{-3}$, and for (c) $c=330$, $k=50$ as true parameters. $\log$ MSEs are computed respectively over 40, 25 and 25 predicted time-steps. \%Err param. averages the results when several physical parameters are present.  For each level of incorporated physical knowledge, equivalent best results according to a Student t-test are shown in bold. n/a corresponds to non-applicable cases. 
    \label{tab:pendulum}}
    \begin{tabular}{cclccc}
    \toprule
Dataset & & Method & $\log$ MSE & \%Err param. & $\|F_a\|^2$ \\ 
    \midrule
    
\multirowcell{9}{\parbox{1cm}{\centering \tiny (a) \\Damped pendulum}} & \centering\tiny Data-driven & Neural ODE \cite{chen2018neural}
& -2.84$\pm$0.70 & n/a & n/a   \\ \cmidrule{2-6}
&\multirowcell{5}{\parbox[c]{1cm}{\centering \tiny Incomplete physics}}  & Hamiltonian \cite{toth2019hamiltonian} 
& -0.35$\pm$0.10  & n/a & n/a \\
&& APHYNITY Hamiltonian   & \textbf{-3.97$\pm$1.20}    & n/a & 623 \\ \cdashline{3-6}\noalign{\vskip 0.2ex}
&& Param ODE ($\omega_0$) & -0.14$\pm$0.10  & 13.2  & n/a   \\
&& Deep Galerkin Method ($\omega_0$) \cite{sirignano2018dgm}  & -3.10$\pm$0.40 & 22.1 & n/a \\
&& APHYNITY Param ODE ($\omega_0$) &  \textbf{-7.86$\pm$0.60}  & \textbf{4.0}  &132  \\ \cmidrule{2-6}
&\multirowcell{5}{\parbox{0.8cm}{\centering\tiny Complete physics}} & Param ODE ($\omega_0, \alpha$)  & \textbf{-8.28$\pm$0.40}  & \textbf{0.45}   & n/a    \\
&& Deep Galerkin Method ($\omega_0,\alpha$) \cite{sirignano2018dgm}  & -3.14$\pm$0.40 & 7.1 & n/a \\
&& APHYNITY Param ODE ($\omega_0, \alpha$)  & \textbf{-8.31$\pm$0.30} & \textbf{0.39} & 8.5  \\ \cdashline{3-6}\noalign{\vskip 0.2ex}
&& True ODE  & \textbf{-8.58$\pm$0.20}  & n/a & n/a \\ 
&& APHYNITY True ODE   & \textbf{-8.44$\pm$0.20}  & n/a & 2.3  \\  \midrule

\multirowcell{8}{\parbox{0.7cm}{\centering\tiny (b) Reaction-diffusion}} & \multirowcell{2}{\parbox{0.9cm}{\centering\tiny Data-driven}} &  
Neural ODE \cite{chen2018neural} 
& -3.76$\pm$0.02 & n/a & n/a \\
&& PredRNN++ \cite{wang2018predrnn++}
& -4.60$\pm$0.01  & n/a & n/a\\\cmidrule{2-6}
& \multirowcell{2}{\parbox{0.9cm}{\centering\tiny Incomplete physics}} &Param PDE ($a,b$) & -1.26$\pm$0.02 & 67.6 & n/a\\ 
&& APHYNITY Param PDE ($a,b$) & \textbf{-5.10$\pm$0.21} & \textbf{2.3} &  67 \\\cmidrule{2-6}
&\multirowcell{4}{\parbox{0.9cm}{\centering\tiny Complete physics}}&  Param PDE ($a,b,k$) & \textbf{-9.34$\pm$0.20} & 0.17 & n/a\\
&&  APHYNITY Param PDE ($a,b,k$) & \textbf{-9.35$\pm$0.02} & \textbf{0.096} & 1.5e-6\\ 
  \cdashline{3-6}\noalign{\vskip 0.2ex}
&&  True PDE & -8.81$\pm$0.05 & n/a  & n/a\\ 
&&  APHYNITY True PDE & \textbf{-9.17$\pm$0.02} & n/a & 1.4e-7\\ \midrule

\multirowcell{8}{\parbox{0.8cm}{\centering\tiny (c)\\ Wave equation}} &\centering\tiny Data-driven & Neural ODE \cite{chen2018neural}
& -2.51$\pm$0.29 & n/a  & n/a  \\ \cmidrule{2-6}
&\multirowcell{2}{\parbox{0.9cm}{\centering\tiny Incomplete physics}} & Param PDE ($c$) & 0.51$\pm$0.07 & 10.4 & n/a  \\
&& APHYNITY Param PDE ($c$) & \textbf{-4.64$\pm$0.25} & \textbf{0.31} & 71. \\ \cmidrule{2-6}
&\multirowcell{4}{\parbox{0.8cm}{\centering\tiny Complete physics}} & Param PDE $(c,k)$ & -4.68$\pm$0.55 & 1.38 & n/a \\
&& APHYNITY Param PDE $(c, k)$ & \textbf{-6.09$\pm$0.28} &  \textbf{0.70} & 4.54 \\ \cdashline{3-6}\noalign{\vskip 0.2ex}
&& True PDE  & -4.66$\pm$0.30 & n/a & n/a \\ 
&& APHYNITY True PDE & \textbf{-5.24$\pm$0.45} & n/a & 0.14 \\
\bottomrule
\end{tabular}
    \label{tab:results-for-all}
    \vspace{-0.2cm}
\end{table}

We analyze and discuss below the results obtained for the three kind of dynamics. We successively examine different evaluation or quality criteria. The conclusions are consistent for the three problems, which allows us to highlight clear trends for all of them. 

\paragraph{Forecasting accuracy:}
The data-driven models do not perform well compared to \textit{True PDE/ODE} (all values are test errors expressed as $\log$ MSE): -4.6 for PredRNN++ vs. -9.17 for reaction-diffusion, -2.51 vs. -5.24 for wave equation, and -2.84 vs. -8.44 for the pendulum in Table~\ref{tab:results-for-all}. The Deep Galerkin method for the pendulum in complete physics \textit{DGM ($\omega_0,\alpha$)}, being constrained by the equation, outperforms Neural ODE but is far inferior to APHYNITY models. In the incomplete physics  case, \textit{DGM ($\omega_0$)} fails to compensate for the missing information. The \textit{incomplete physical models}, \textit{Param PDE ($a,b$)} for the reaction-diffusion, \textit{Param PDE ($c$)} for the wave equation, and \textit{Param ODE ($\omega_0$)} and \textit{Hamiltonian models} for the damped pendulum, have even poorer performances than purely data-driven ones, as can be expected since they ignore important dynamical components, \eg friction in the pendulum case. Using APHYNITY with these imperfect physical models greatly improves forecasting accuracy in all cases, significantly outperforming purely data-driven models, and reaching results often close to the accuracy of the true ODE, when APHYNITY and the true ODE models are integrated with the same numerical scheme~(which is different from the one used for data generation, hence the non-null errors even for the true equations), \eg -5.92 vs. -5.24 for wave equation in Table~\ref{tab:results-for-all}. This clearly highlights the capacity of our approach to augment incomplete physical models with a learned data-driven component.

\paragraph{Physical parameter estimation:}
Confirming the phenomenon mentioned in the introduction and detailed in Appendix~\ref{app:alt_methods}, incomplete physical models can lead to bad estimates for the relevant physical parameters: an error respectively up to 67.6\% and 10.4\% for parameters in the reaction-diffusion and wave equations, and an error of more than 13\% for parameters for the pendulum in Table~\ref{tab:results-for-all}. APHYNITY is able to significantly improve physical parameters identification: 2.3\% error for the reaction-diffusion, 0.3\% for the wave equation, and 4\% for the pendulum. This validates the fact that augmenting a simple physical model to compensate its approximations is not only beneficial for prediction, but also helps to limit errors for parameter identification when dynamical models do not fit data well. This is crucial for interpretability and explainability of the estimates.

\paragraph{Ablation study:}
  We conduct ablation studies to validate the importance of the APHYNITY augmentation compared to a naive strategy consisting in learning $F=F_p+F_a$ without taking care on the quality of the decomposition, as done in \cite{wang2019integrating,neural20}. Results shown in Table \ref{tab:results-for-all} of Appendix \ref{app:ablation} show a consistent gain of APHYNITY for the three use cases and for all physical models:  for instance for \textit{Param ODE ($a,b$)} in reaction-diffusion, both forecasting performances ($\log \text{MSE}=$-5.10 vs. -4.56) and identification parameter (Error$=$ 2.33\% vs. 6.39\%) improve. Other ablation results are provided in Appendix \ref{app:ablation} showing the relevance of the the trajectory-based approach described in Section~\ref{subsec:learning}~(vs supervising over finite difference approximations of the derivative $F$). 

\paragraph{Flexibility:}
When applied to complete physical models, APHYNITY does not degrade accuracy, contrary to a vanilla cooperation scheme (see ablations in Appendix \ref{app:ablation}). This is due to the least action principle of our approach: when the physical knowledge is sufficient for properly predicting the observed dynamics, the model learns to ignore the data-driven augmentation. This is shown by the norm of the trained neural net component $F_a$, which is reported in Table~\ref{tab:results-for-all} last column: as expected, $\|F_a\|^2$ diminishes as the complexity of the corresponding physical model increases, and, relative to incomplete models, the norm becomes very small for complete physical models~(for example in the pendulum experiments, we have $\Vert F_a \Vert= 8.5$ for the APHYNITY model to be compared with 132 and 623 for the incomplete models). Thus, we see that the norm of $F_a$ is a good indication of how imperfect the physical models $\F_p$ are. It highlights the flexibility of APHYNITY to successfully adapt to very different levels of prior knowledge. Note also that APHYNITY sometimes slightly improves over the true ODE, as it compensates the error introduced by different numerical integration methods for data simulation and training (see Appendix~\ref{app:implementation}).

\begin{figure}[t!]
    \centering
    \subfloat[Param PDE ($a$, $b$), diffusion-only
    ]{
    \includegraphics[width=0.32\linewidth]{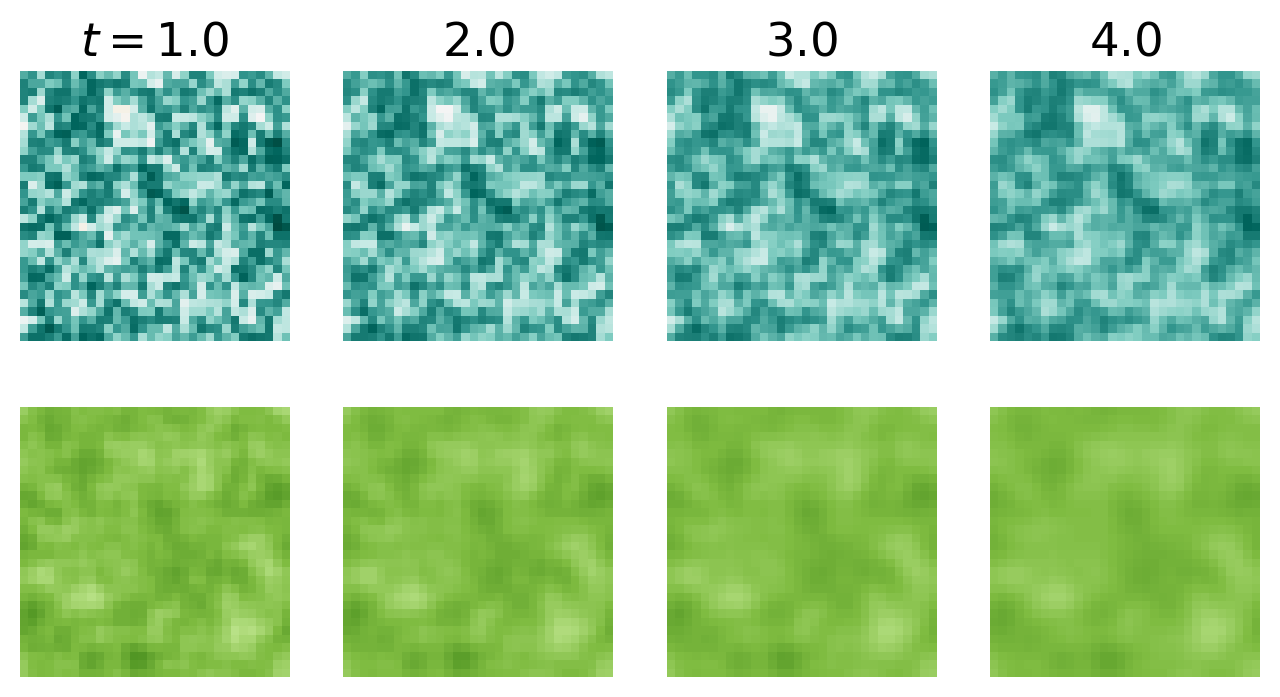}}
      \hfill
    \subfloat[APHYNITY Param PDE ($a$, $b$)
    ]{
    \includegraphics[width=0.32\linewidth]{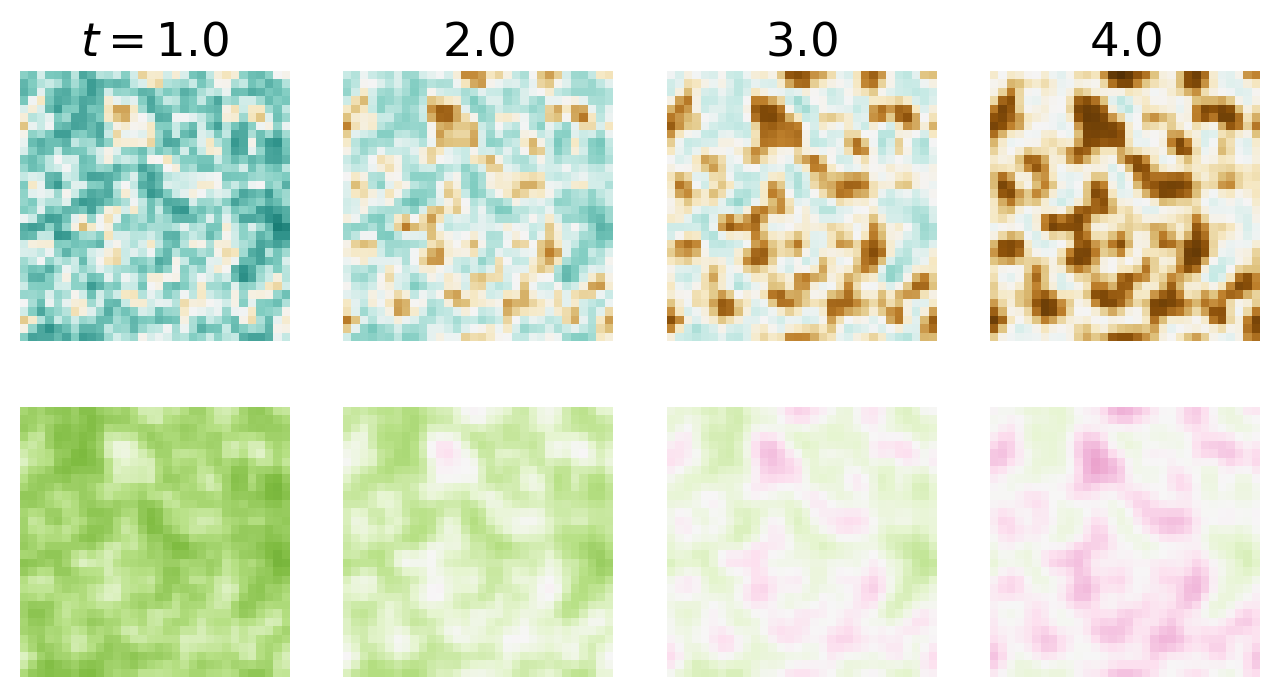}}\hfill
    \subfloat[Ground truth simulation
    ]{
    \includegraphics[width=0.32\linewidth]{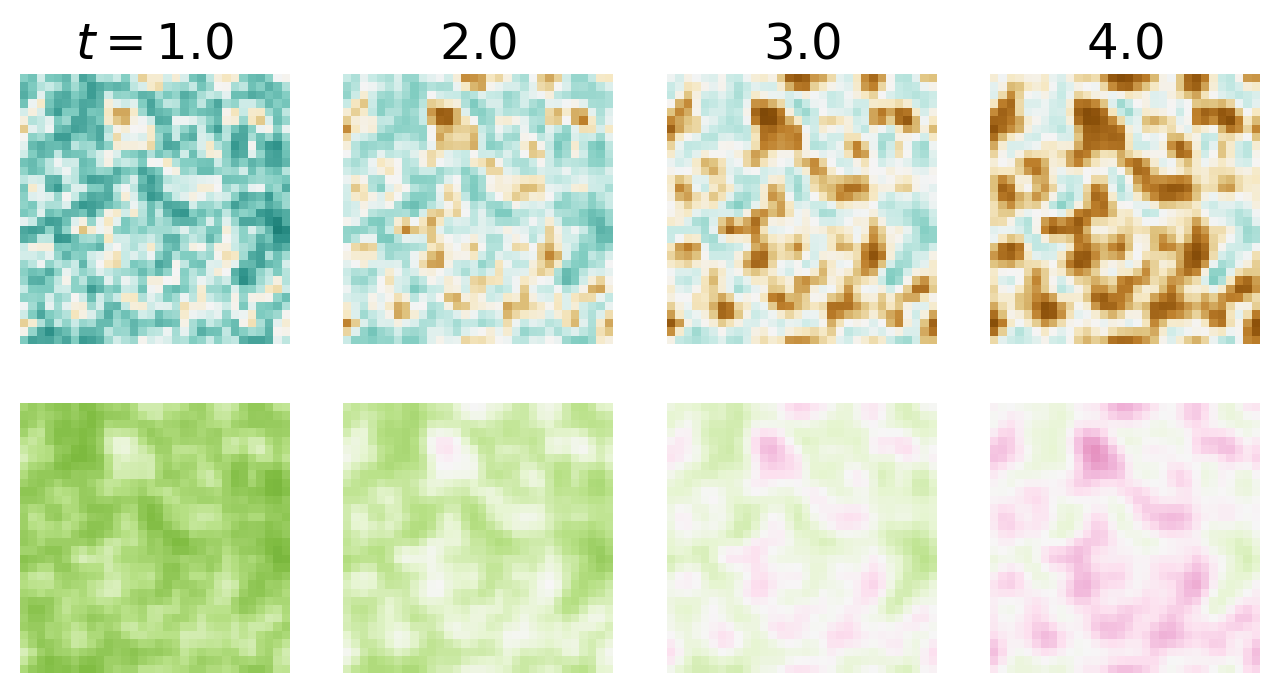}}
    \caption[Qualitative results on the reaction-diffusion equations.]{Comparison of predictions of two components $u$ (top) and $v$ (bottom) of the reaction-diffusion system. Note that $t=4$ is largely beyond the dataset horizon ($t=2.5$).
    \label{fig:reaction-diffusion-demo}
    }
\end{figure}

\begin{figure}[t!]
    \centering
    \subfloat[Neural ODE
    ]{
    \includegraphics[width=0.32\linewidth]{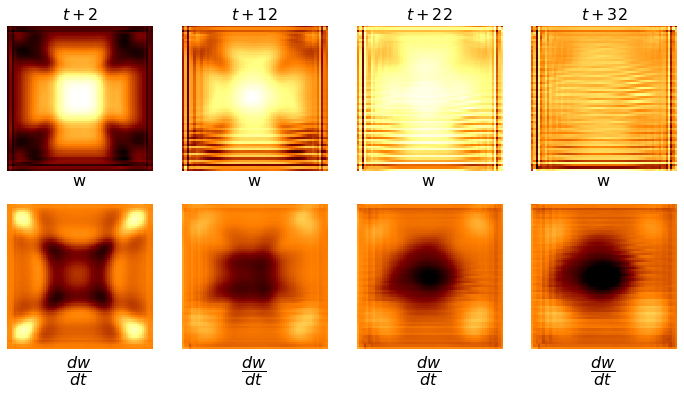}}
    \hfill
    \subfloat[APHYNITY Param PDE ($c$)
    ]{
    \includegraphics[width=0.32\linewidth]{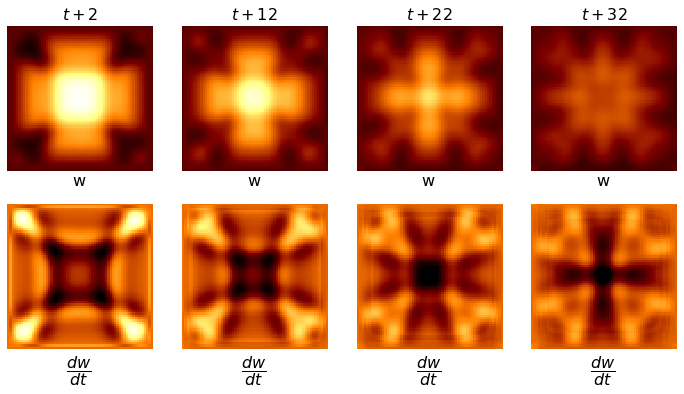}}\hfill
    \subfloat[Ground truth simulation
    ]{
    \includegraphics[width=0.32\linewidth]{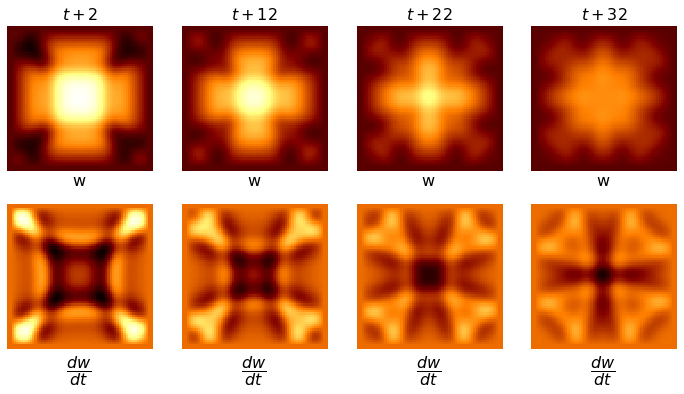}}
    \caption[Qualitative results on the wave equations.]{Comparison between the prediction of APHYNITY when $c$ is estimated and Neural ODE for the damped wave equation. Note that $t+32$, last column for (a, b, c) is already beyond the training time horizon ($t+25$), showing the consistency of APHYNITY method.\label{fig:wave-damped-demo}
    }
\end{figure}

\paragraph{Qualitative visualizations:}
Results in Figure~\ref{fig:reaction-diffusion-demo} for reaction-diffusion show that the incomplete diffusion parametric PDE in Figure~\ref{fig:reaction-diffusion-demo}(a) is unable to properly match ground truth simulations: the behavior of the two components in Figure~\ref{fig:reaction-diffusion-demo}(a) is reduced to simple independent diffusions due to the lack of interaction terms  between $u$ and $v$. By using APHYNITY in Figure~\ref{fig:reaction-diffusion-demo}(b), the correlation between the two components appears together with the formation of Turing patterns, which is very similar to the ground truth. This confirms that $F_a$ can learn the reaction terms and improve prediction quality. In Figure~\ref{fig:wave-damped-demo}, we see for the wave equation that the data-driven Neural ODE model fails at approximating $\nicefrac{\diff w}{\diff t}$ as the forecast horizon increases: it misses crucial details for the second component $\nicefrac{\diff w}{\diff t}$ which makes the forecast diverge from the ground truth. APHYNITY incorporates a Laplacian term as well as the data-driven $F_a$ thus capturing the damping phenomenon and succeeding in maintaining physically sound results for long term forecasts, unlike Neural ODE.

\begin{figure}[h]
\centering
\subfloat[$a=0.33\times 10^{-3}, b=0.94\times 10^{-3} $, diffusion estimated with Param PDE $(a,b)$]{\includegraphics[width=0.5\textwidth]{images/aphynity_reacdiff_physical.png}}\hfill
\subfloat[$a=0.97\times 10^{-3}, b=4.75\times 10^{-3}$, diffusion estimated with APHYNITY Param PDE $(a,b)$]{\includegraphics[width=0.5\textwidth]{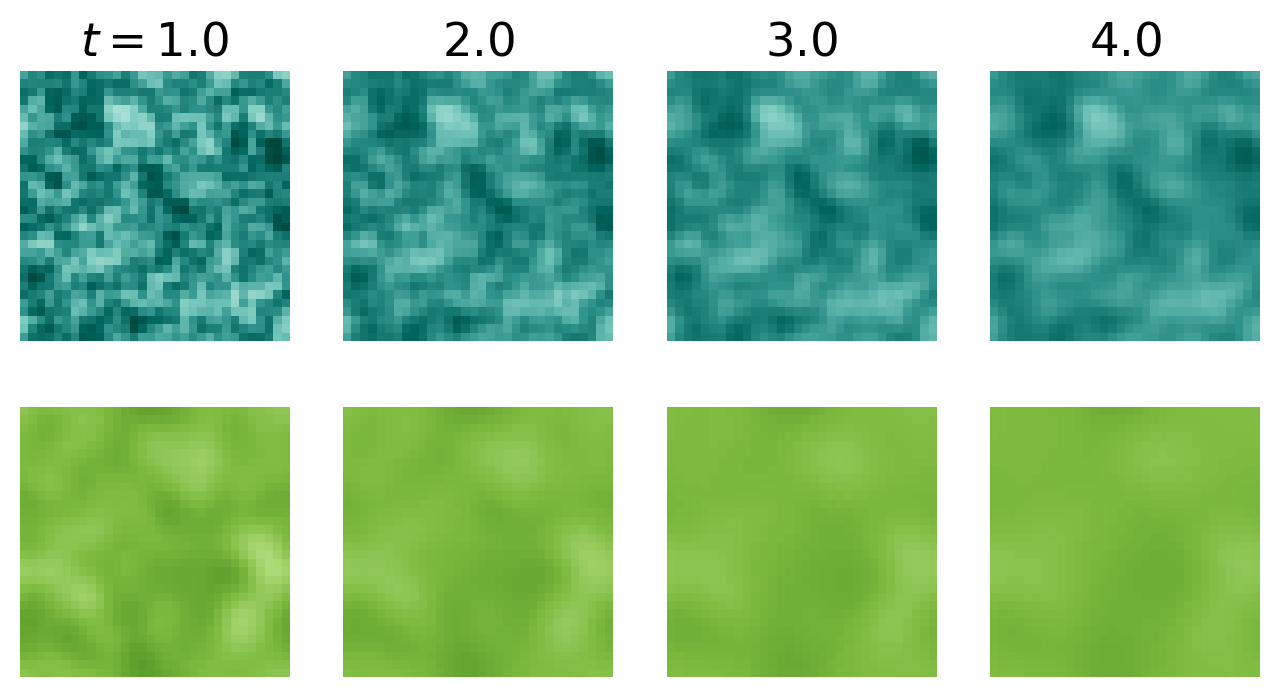}} \hfill
\subfloat[$a=1.0\times 10^{-3}, b=5.0\times 10^{-3}$, true diffusion]{\includegraphics[width=0.5\textwidth]{images/aphynity_reacdiff_aphynity_phys_est.png}}

\caption[Qualitative analysis on the reaction-diffusion equations.]{Diffusion predictions using coefficient learned with (a) incomplete physical model Param PDE $(a,b)$ and (b) APHYNITY-augmented Param PDE$(a,b)$, compared with the (c) true diffusion\label{fig:comp-diffusion}}
\end{figure}

\paragraph{Additional illustrations:}
We give further visual illustrations to demonstrate how the estimation of parameters in incomplete physical models is improved with APHYNITY. For the reaction-diffusion equation, we show that the incomplete parametric PDE underestimates both diffusion coefficients. The difference is visually recognizable between the poorly estimated diffusion (\Figref{fig:comp-diffusion}(a)) and the true one (\Figref{fig:comp-diffusion}(c)) while APHYNITY gives a fairly good estimation of those diffusion parameters as shown in \Figref{fig:comp-diffusion}(b).

\subsection{Extension to non-stationary dynamics}
\label{sec:aph-nonstat}

We evaluate here the applicability of APHYNITY in a more challenging setting where physical parameters of the equations vary in each sequence. For the damped pendulum equations, instead of fixed parameters ($T_0=6, \alpha=0.2$) and varying initial conditions (Section \ref{sec:results}), we vary both the parameters ($T_0, \alpha$) and the initial conditions between trajectories.

We simulate 500/50/50 trajectories for the train/valid/test sets. For each trajectory, the period $T_0$ (resp. the damping coefficient $\alpha)$ are sampled uniformly in the range $[3,10]$ (resp. $[0,0.5]$). 

We train models that take the first 20 steps as input and predict the next 20 steps. To account for the varying ODE parameters between sequences, we use an encoder that estimates the parameters based on the first 20 timesteps. In practice, we use a recurrent encoder composed of 1 layer of 128 GRU units. The output of the encoder is fed as additional input to the data-driven augmentation models and to an MLP with final softplus activations to estimate the physical parameters when necessary ($\omega_0 \in \mathbb{R}_+$ for Param ODE ($\omega_0$), $(\omega_0,\alpha) \in \mathbb{R}_+^2$ for Param ODE ($\omega_0,\alpha$)). 

In this varying ODE context, we also compare to the state-of-the-art univariate time series forecasting method N-Beats \cite{oreshkin2019n}.

Results shown in Table \ref{tab:pendulum-encoder} are consistent with those presented in Section \ref{sec:results}. Pure data-driven models Neural ODE \cite{chen2018neural} and N-Beats \cite{oreshkin2019n} fail to properly extrapolate the pendulum dynamics. Incomplete physical models (Hamiltonian and ParamODE ($\omega_0$)) are even worse since they do not account for friction. Augmenting them with APHYNITY significantly and consistently improves forecasting results and parameter identification.

We provide similar experiments for the reaction-diffusion and wave equations in Appendix \ref{app:additional}.

\begin{table}[H]
    \centering
\setlength{\tabcolsep}{6.8pt}

    \caption[APHYNITY results on the damped pendulum with varying parameters.]{Forecasting and identification results on the damped pendulum dataset with different parameters for each sequence. log MSEs are computed over 20 predicted time-steps. For each level of incorporated physical knowledge, equivalent best results according to a Student t-test are shown in bold. n/a corresponds to non-applicable cases.  \label{tab:pendulum-encoder}}
 \begin{adjustbox}{max width=\linewidth}    
    \begin{tabular}{clcccc}
    \toprule
    & Method & $\log$ MSE & \%Error $T_0$ & \%Error $\alpha$ & $\|F_a\|^2$  \\ 
    \midrule
\multirowcell{2}{\parbox{0.9cm}{\centering\tiny data-driven}} & Neural ODE \cite{chen2018neural}  & -4.35$\pm$0.9 & n/a & n/a & n/a   \\
 &  N-Beats \cite{oreshkin2019n} & -4.57$\pm$0.5  & n/a & n/a & n/a   \\
  \midrule
\multirowcell{4}{\parbox{0.9cm}{\centering\tiny Incomplete physics}}  & Hamiltonian \cite{greydanus2019hamiltonian}  & -1.31$\pm$0.4  & n/a & n/a  & n/a \\
 & APHYNITY Hamiltonian   & \textbf{-4.72$\pm$0.4}    & n/a & n/a & 5.6$\pm$0.6    \\  
  \cdashline{2-6}\noalign{\vskip 0.2ex}
 & Param ODE ($\omega_0$) & -2.66$\pm$0.9  & 21.5$\pm$19    & n/a & n/a   \\
 & APHYNITY Param ODE ($\omega_0$) &  \textbf{-5.94$\pm$0.7}  & \textbf{5.0$\pm$1.8}  & n/a &  0.49$\pm$0.1  \\  
  \midrule
\multirowcell{4}{\parbox{0.8cm}{\centering\tiny Complete physics}} & Param ODE ($\omega_0, \alpha$)  & \textbf{-5.71$\pm$0.4}  & 4.08$\pm$0.8  & 152$\pm$129   & n/a    \\
 & APHYNITY Param ODE ($\omega_0, \alpha$)  & \textbf{-6.22$\pm$0.7} & \textbf{3.26$\pm$0.6}    &  \textbf{62$\pm$27}   & (5.39$\pm$0.1)e-10  \\ 
  \cdashline{2-6}\noalign{\vskip 0.2ex}
&True ODE  & \textbf{-8.58$\pm$0.1}  & n/a &n/a  & n/a \\ 
 & APHYNITY True ODE   & \textbf{-8.58$\pm$0.1}  & n/a & n/a  & (2.15$\pm$1.6)e-4  \\
  \bottomrule
    \end{tabular}
  \end{adjustbox}  
\end{table}

\section{Conclusion}
 \label{discussion}
In this Chapter, we have introduced the APHYNITY framework that can efficiently augment approximate physical models with deep data-driven networks, performing similarly to models with full-known dynamics. We have exhibited the superiority of APHYNITY over data-driven, incomplete physics, and state-of-the-art approaches combining ML and MB methods, both in terms of forecasting and parameter identification on three various classes of physical systems. Besides, APHYNITY is flexible enough to adapt to different approximation levels of prior physical knowledge.

\clearpage{\pagestyle{empty}\cleardoublepage}

\mbox{}
\thispagestyle{empty}
\partabstract{
\vspace{1cm}
\begin{center}
    \textsc{Abstract}\\
\end{center}
\vspace{1cm}
In this final part, we tackle the industrial solar energy forecasting problem with fisheye images that we briefly discussed in Chapter \ref{chap:intro}. We first present in details the use-case, and review the existing traditional methods and the early deep learning approaches (Chapter \ref{chap:overview_fisheye}). We also propose a first data-driven deep learning model for solar irradiance estimation and prediction and discuss its limitations. In Chapter \ref{chap:phydnet_fisheye}, we investigate the model-based machine learning cooperation studied in this thesis for improving the model. We propose a new physically-constrained architecture adapted from our PhyDNet video prediction model (Chapter \ref{chap:phydnet}). We also evaluate the use of our DILATE loss (Chapter \ref{chap:dilate}) for enforcing predictions with accurate shape and temporal localization, and of our APHYNITY framework (Chapter \ref{chap:aphynity}) for optimal ML/MB decomposition.
}
\part{Application to solar irradiance forecasting}
\label{part:part3}

\chapter{Overview of solar irradiance forecasting}
\label{chap:overview_fisheye}

\chapabstract{

\minitoc

\begin{center}
   \textsc{Chapter abstract}
\end{center}
\textit{
In this Chapter, we describe in details the industrial solar irradiance forecasting problem with fisheye images at EDF. We first review the traditional image processing and machine learning techniques, and the early deep learning approaches that have recently shown promising results. We propose a first deep learning model for estimating and predicting the future solar irradiance, which will be a strong deep baseline for the following Chapter.\\
The work described in this Chapter is based on the following publication:
\begin{itemize}
    \item  \cite{leguen-gretsi} Vincent Le Guen and Nicolas Thome. "Prévision de l’irradiance solaire par réseaux de neurones profonds à l’aide de caméras au sol". In: GRETSI 2019.
\end{itemize}
}
}

\section{Introduction}

\begin{figure}[b!]
    \centering
    \includegraphics[width=12cm]{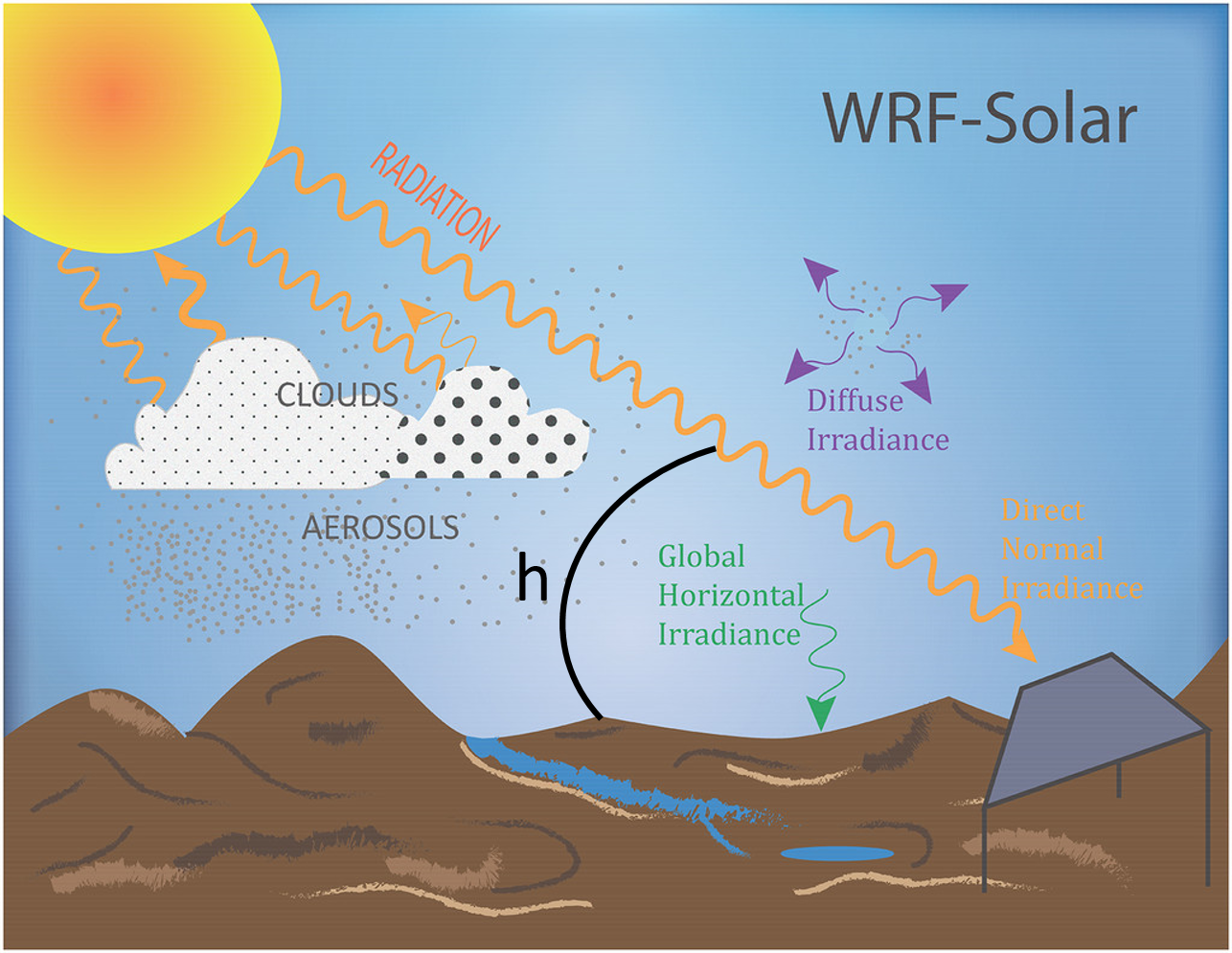}
    \caption[The different components of solar irradiance.]{The different components of solar irradiance. Figure taken from \cite{jimenez2016wrf}.}
    \label{fig:solar-irradiance}
\end{figure}

\lettrine[lines=3]{T}o tackle climate change and limit global warming, major world economies agreed in 2015 at the Paris climate conference (COP21) on a restrictive plan to reduce greenhouse gas emissions. In the energy sector, this reinforced massive investments towards renewable energy generation such as solar or wind. However, a limitation of solar and wind energies is their intermittent and non-controllable nature, in contrast to conventional fossil fuel or nuclear energy. This causes major challenges for their integration at scale in the existing electricity grid, since electricity production and consumption must be balanced at every time. Therefore, accurately forecasting the intermittent energy production at various time horizons (from seconds to a few days) becomes a crucial aspect of the energy transition. Many applications could benefit from improved solar energy forecasts, such as the development of smart grids, hybrid solar/conventional power systems, or energy trading.

\subsection{The solar irradiance components}

In this thesis, we are interested in forecasting the solar irradiance, which corresponds to the incoming power of electromagnetic radiation received from the sun (expressed in $W/m^2$). The Global Horizontal Irradiance (GHI) can be decomposed into the Direct Normal Irradiance (DNI) directly coming from the sun perpendicularly to the photovoltaic (PV) panels, and the Diffuse Horizontal Irradiance (DHI) coming from the diffusion by the clouds and aerosols of the atmosphere or reflection from the ground (see Figure \ref{fig:solar-irradiance}):
\begin{equation}
    \text{GHI} = \text{DHI} + \sin h  \times \text{DNI}
\end{equation}
where $h$ is the solar elevation angle.

The GHI is the main quantity of interest in this thesis, since it is directly linked to the electric power production expressed in Watts, by knowing the technology, orientation of the photovoltaic panels and the ambient temperature. In practice, before applying any statistical method, the solar irradiance is often normalized by a clear-sky model corresponding to the theoretical irradiance received in cloudless conditions. This normalization compensates for the inherent seasonality of the solar irradiance. In this thesis, we use the ESRA (European Solar Radiation Atlas) clear sky model \cite{rigollier2000clear} and we denote KGHI the GHI normalized by its clear-sky values.

\subsection{The different data sources for solar irradiance forecasting}

For solar energy, the main source of variability comes from the occlusion of the sun by clouds. We presented in Figure \ref{fig:types-observations} the main classes of methods for forecasting solar irradiance. Although statistical time series forecasting can be directly applied on the 1D solar irradiance series, this strategy is blind to the motions of clouds and thus cannot properly anticipate the variations. To understand the spatio-temporal dynamics of clouds, current methods rely on weather forecasts or sky image analysis. Numerical weather forecasts solve the equations of physics to forecast the dynamics of the atmosphere ; they have a spatial resolution of around 1km and a temporal resolution of 1 to 2h for the AROME model of Meteo France. For shorter forecasting horizons, satellite images can be exploited to provide irradiance forecasts up to a few hours with a 15 min granularity and a 1km spatial scale.

\begin{figure}[h]
    \centering
    \includegraphics[width=16cm]{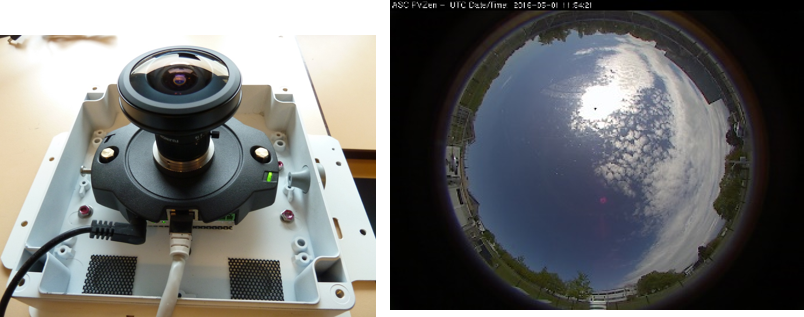}
    \caption{Fisheye camera and fisheye image for short-term solar irradiance forecasting.}
    \label{fig:fisheye-camera}
\end{figure}

For very short-term horizons (< 20min) at the scale of a PV plant, fisheye cameras pointed towards the sky (see Figure \ref{fig:fisheye-camera}) have become popular in recent years \cite{gauchet2012surface,chu2013hybrid,chu2016sun,marquez2013intra,schmidt2016evaluating,kuhn2018validation}. They offer an hemispheric view of the sky that enables to assess the evolution of the cloud cover. 

\begin{figure}
    \centering
    \includegraphics[width=10cm]{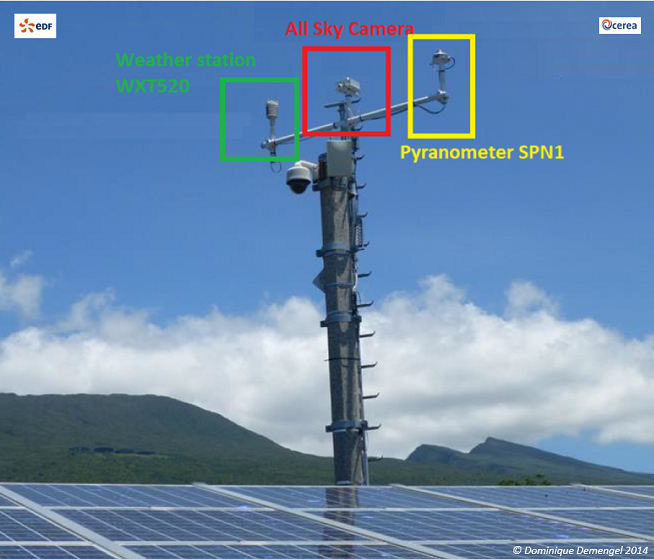}
    \caption{EDF scientific test site at La Reunion Island composed of a fisheye camera, a pyranometer and a weather station mounted above a PV power plant.}
    \label{fig:meteo_campaign}
\end{figure}

\subsection{Meteorological campaign at EDF R\&D with fisheye images}

EDF Research and Development (R\&D) has led a meteorological campaign since 2010 at La Réunion Island with fisheye cameras (Axis PTZ 212) and pyranometers (SPN1) for measuring ground truth solar irradiance (see Figure \ref{fig:fisheye-camera} and Figure \ref{fig:meteo_campaign}). A database of more than 7 million images every 10s and corresponding irradiance measurements was collected. The objective is to forecast solar irradiance with fisheye cameras only, which are much cheaper than pyranometers and provide an additional spatial information compared to irradiance time series.


\section{Related work}

 In the Section, we review the main existing methods for short-term solar irradiance forecasting.

\paragraph{Persistence and statistical models}
For very-short term forecasting, a first natural baseline is the persistence, which assumes that the current irradiance level (normalized by the clear-sky) will persist. Persistence is often a competitive baseline, with optimal performances in clear-sky conditions. However, persistence does not anticipate variations by definition.
Other statistical models \cite{diagne2013review,wolff2016statistical,mellit2008artificial} use local information (\eg past irradiance values, PV production, temperature, weather forecasts) to capture statistical patterns and predict future values with regression or clustering algorithms. However, these methods do not observe the cloud motion and thus fail to anticipate variations due to sun occlusions.

\paragraph{Ground-based images}
For assessing the cloud coverage and anticipating short-term variations due to sun occlusions, researchers have investigated sky imagery with ground-based cameras from the 2010's. Earlier works have used specific scientific instruments, such as the Total Sky Imager in \cite{chow2011intra,marquez2013intra} (spherical mirror with a camera pointing downwards) or suntrackers. Since, low-cost webcam cameras have encountered a great success, leading to a soaring interest from the community \cite{gauchet2012surface,chu2013hybrid,chu2016sun,schmidt2016evaluating}.

Although many hardware and algorithmic variants exist (\eg additional sensors, multiple cameras for stereo estimation), all these methods mainly follow a similar traditional image processing pipeline:
\begin{enumerate}
    \item Camera calibration for determining the distortion parameters of the fisheye objective,
    \item Fisheye image acquisition at fixed intervals (\eg every 10s, 1min), sometimes with several expositions and High Dynamic Range (HDR) processing,
    \item Image segmentation with thresholds based on color ratios or other photometric properties. Thresholds are either handcrafted or adaptative. The segmentation can be used for deriving a binary cloud map, or for deriving image features.
    \item Cloud motion estimation with optical flow,
    \item Cloud motion propagation into the future to generate a predicted irradiance map.
\end{enumerate}
However sophisticated the processing pipeline may be, the challenges of the problem remain: the clouds follow a complex stochastic motion with abrupt variations that is hard to extrapolate. All these methods also rely on some manual engineering that only remains valid in a limited range of conditions.

\paragraph{Deep learning for solar irradiance forecasting}

Since a few years, deep learning has become an appealing alternative to replace the whole conventional pipeline with a model learned end-to-end from raw fisheye images \cite{pothineni2018kloudnet,zhang2018deep,spiess2019learning,sun2019short,nie2020pv,paletta2020temporally,zhen2021ultra}. However, as Paletta \etal \cite{paletta2021benchmarking} has highlighted, standard deep learning methods still struggle to properly understand the cloud motion and do not anticipate sharp variations.

\section{Proposed models for solar irradiance estimation and forecasting}

In this Section, we introduce two deep learning models: one for solar irradiance estimation, the second for forecasting. We define estimation as the prediction of the irradiance $r_T$ associated with the image $I_T$. Forecasting corresponds to predicting the future irradiance $r_{T+H}$ (or the complete future trajectory $r_{T+1},\cdots,r_{T+H}$) given a sequence of past images $(I_1, \cdots, I_T)$.

\subsection{Solar irradiance estimation}

\begin{figure}
    \centering
    \includegraphics[width=10cm]{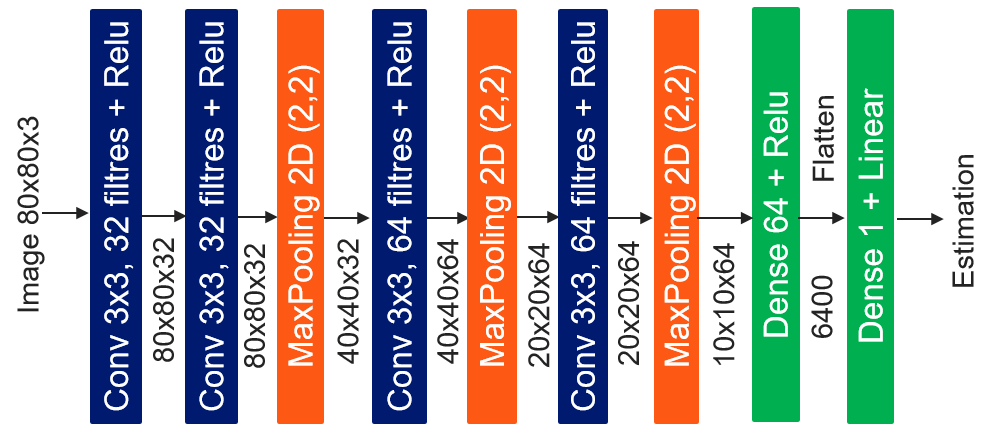}
    \caption{Small convolutional network used for solar irradiance estimation.}
    \label{fig:convnet-small}
\end{figure}

For solar irradiance estimation, we use a convolutional neural network that takes as input a fisheye image (without preprocessing) and outputs the estimated solar irradiance for that image. We first propose a handcrafted convolutional architecture (shown in Figure \ref{fig:convnet-small}) working on RBG images resized at $80\times 80$ pixels. This model has approximately 470,000 parameters.

We also propose a much larger model relying on the DenseNet architecture \cite{huang2017densely} that has reached state-of-art performances on the ImageNet image classification task. The model works with higher resolution images, resized at $224 \times 224$ pixels. For adapting the model to this regression task, we replace the final classification layers by fully-connected layers for outputting one irradiance value. The overall model has approximately 18 Million parameters.

\subsection{Solar irradiance forecasting}

\begin{figure}[H]
    \centering
    \includegraphics[width=14cm]{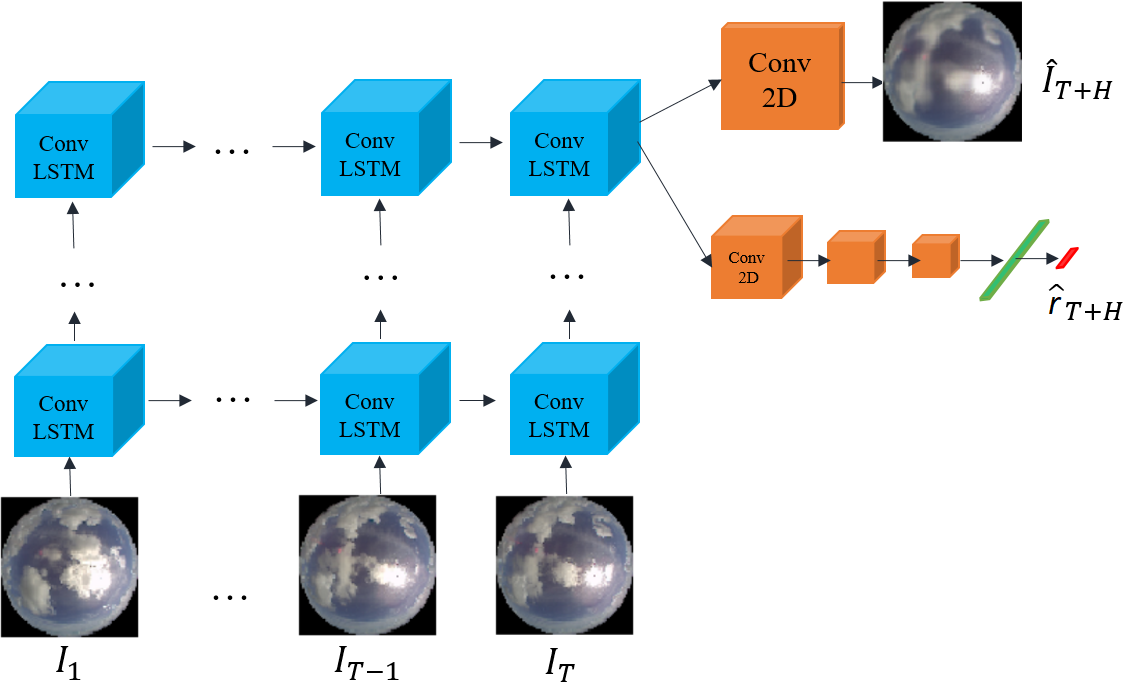}
    \caption{Proposed architecture for solar irradiance forecasting based on the ConvLSTM model \cite{xingjian2015convolutional}.}
    \label{fig:convlstm-fisheye}
\end{figure}

To forecast solar irradiance, we propose a neural network architecture relying on the ConvLSTM model \cite{xingjian2015convolutional} which is a strong baseline for deep video prediction. Depicted in Figure \ref{fig:convlstm-fisheye}, our architecture is composed of a ConvLSTM encoder that reads a sequence of $T$ past fisheye images $(I_{1},\cdots,I_{T-1}, I_T)$ and encodes them into a context vector. The network has two output branches: one for predicting the future solar irradiance $\hat{r}_{T+F}$ at a given horizon $T$ and the other for the future fisheye image $\hat{I}_{T+H}$. 

We empirically verified that this multi-task objective improves performances compared to forecasting irradiance only, due to the richer supervision signal and the cooperation between both tasks.

Our forecasting model is composed of 4 stacked ConvLSTM layers acting on input images resized to $80\times 80$ pixels.

\section{Experimental results}

\subsection{Fisheye image dataset}

We conduct experiments on the fisheye image dataset collected by EDF at La Reunion Island. For the estimation task, we use a training set composed of 4,190,064 images from the years 2012 to 2015, and a test set of 1,265,717 images in the year 2016. Images are processed from a solar elevation of 10°, and all irradiance measurements are normalized by the ESRA clear-sky model \cite{rigollier2000clear}. We use images resized at $80 \times 80$ pixels for the ConvNet model and $224 \times 224$ for the DenseNet model. 

For the forecasting task, the training set is composed of 180,000 sequences of 10 images spaced by 1min (with the associated ground truth solar irradiance measurements) from the years 2014 to 2016, and the test set of 20,000 sequences during the year 2013 on the same site. We use images resized at $80 \times 80$ pixels. We keep 5 images for the input range and predict the 5 following images and solar irradiances.

\subsection{Solar irradiance estimation results}

We present in Table \ref{tab:fisheye-estimation} the estimation results for the KGHI. We have trained two DenseNet models: one that only predicts the KGHI and the other that jointly predicts the KGHI and KDHI. We compare our proposed deep models with the baseline previously developed at EDF R\&D \cite{gauchet2012surface}. This traditional method segments the fisheye images with thresholds on the R-B difference and the lumimance, defines 5 features based on the segmentation ratios and applies a Nadaraya-Watson kernel regression \cite{nadaraya1964estimating} for estimating the irradiance.

We evaluate the performances with the normalized Mean Absolute Error (nMAE) and normalized Root Mean Squared Error (nRSME). Normalization is performed by dividing by the mean KGHI value over the training set.

\begin{table}[H]
\centering
\caption{KGHI estimation results on the test set.}
\label{tab:fisheye-estimation}
\begin{tabular}{ccc}
\toprule
model & nMAE & nRMSE \\ \midrule
Baseline & 14.9 \% & 21.6 \% \\
ConvNet KGHI & 6.59 \% & 10.3 \% \\
DenseNet KGHI & 2.91 \% & 5.27 \% \\
DenseNet KGHI + KDHI & \textbf{2.90 \%} & \textbf{4.83 \%} \\
\bottomrule
\end{tabular}
\end{table}

Results show that the ConvNet model (depicted in Figure \ref{fig:convnet-small}) yields a large performance improvement (from 21.6 \% to 10.3 \% in nRMSE) over the baseline. Going deeper with the DenseNet model further gives a large gap in performances (5.27 \%). It confirms the ability of deep learning to automatically learn a representation space for approximating a complex mapping from a large dataset of annotated images. Finally, we observe that the DenseNet model that jointly estimates the KGHI and KDHI gives the best performances (4.83 \%), indicating that exploiting the correlations between both irradiance components helps in better estimating the KGHI. Intuitively, for two images with similar GHI but different cloud conditions, the differences of diffuse irradiance (DHI) should help to learn more specific cloud features that better generalize for estimating different test images.

\begin{figure}[H]
    \centering
    \includegraphics[width=\linewidth]{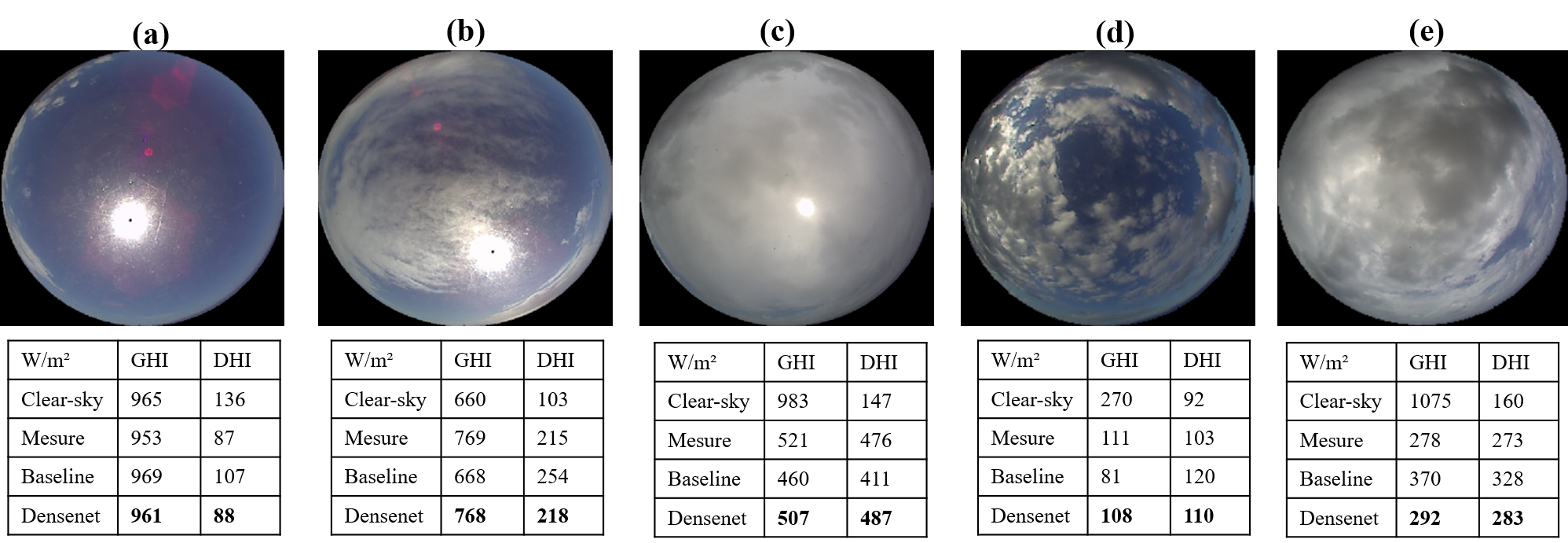}
    \caption{Qualitative fisheye estimation results of the GHI and DHI.}
    \label{fig:fisheye-estimation}
\end{figure}

We display in Figure \ref{fig:fisheye-estimation} a few qualitative estimation examples. We can see for several sky conditions that the DenseNet estimations are very close to the measurements, both in GHI and DHI. Interestingly, the difference with the baseline is much higher when the diffuse irradiance (DHI) is high, \eg for images (c) and (e). It can be explained by the difficulty of the image segmentation with handcrafted thresholds on clouds with different levels of gray; the deep learning approach better learns features for representing the shades of clouds, supervised by the GHI and DHI values.  

\subsection{Solar irradiance forecasting results}

We then evaluate the forecasting performances of our method on the fisheye image dataset. We compare our ConvLSTM architecture with the optical flow baseline previously developed at EDF R\&D \cite{gauchet2012surface} (sketched in Figure \ref{fig:mldl_diff}), and with the (smart) persistence which consists in copying the current value as the forecast for the future timestep (for quantities normalized by the clear sky). 

Global results presented in Table \ref{tab:irradiance-forecasting} show that our proposed deep forecasting model outperforms both the optical flow baseline and the persistence. However, the performance gap with the traditional method is narrower than for estimation, revealing the difficulty of the forecasting task.

\begin{figure}[H]
    \centering
    \includegraphics[width=\linewidth]{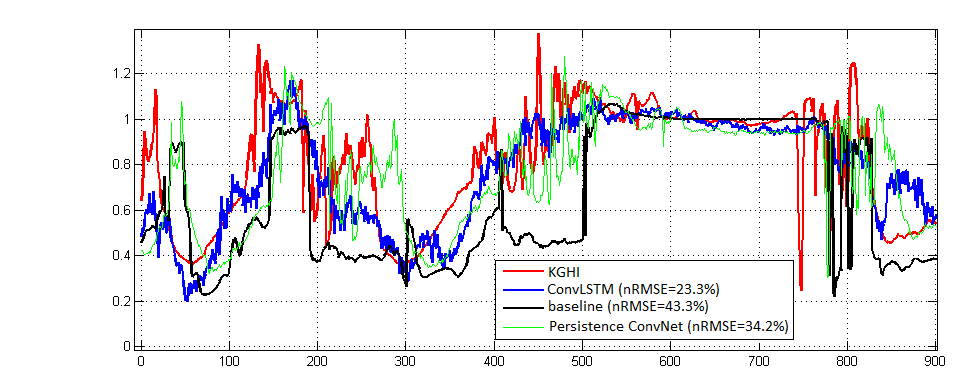}
    \caption{Qualitative KGHI forecasting results at 5min on a particular day.}
    \label{fig:fisheye_conv_forecasting}
\end{figure}

To further analyse the differences, we display in Figure \ref{fig:fisheye_conv_forecasting} the model predictions on a particular day of the test set. We can see that the ConvLSTM predictions are much closer to the KGHI ground truth than the optical flow baseline and than the persistence ConvNet (which corresponds to applying the estimation ConvNet).

Interestingly, the optical flow baseline has a worse RMSE than the persistence. However, the optical flow method shows a better ability to anticipate sharp variations (\eg around the timestep 200), and is therefore better suited for the industrial application. It confirms that the MSE and variants are not adapted to train and evaluate models in this non-stationary context with abrupt changes, which has motivated the contributions of this thesis. In the following Chapter, we will train and evaluate models with our proposed shape and temporal criteria to improve models in this context.

\begin{table}[H]
    \centering
        \caption{Forecasting performances of the KGHI  (normalized Global Horizontal Irradiance) at a 5min horizon.}
    \begin{tabular}{c|c}
     \toprule
     Method &   normalized RMSE   \\ 
         \midrule
    Optical flow baseline  & 32.9 \%  \\
    Persistence  & 28.5  \%  \\
    ConvLSTM (ours)  & \textbf{26.6 \%} \\
     \bottomrule
    \end{tabular}
    \label{tab:irradiance-forecasting}
\end{table}{}

\section{Conclusion}

In this Chapter, we have presented the solar irradiance forecasting problem with fisheye images at EDF, and reviewed the existing methods (traditional and deep). We have proposed first deep models for estimating and forecasting the solar irradiance, that have reached state-of-the-art results compared to traditional methods. However, for the forecasting task, there still exists room for improvement, in particular for modelling the sharp variations and the complex nonlinear cloud dynamics. These limitations will be addressed in the next Chapter.

\clearpage{\pagestyle{empty}\cleardoublepage}

\mbox{}
\thispagestyle{empty}
\chapter{Deep learning for solar irradiance forecasting}
\label{chap:phydnet_fisheye}

\chapabstract{

\minitoc

\begin{center}
   \textsc{Chapter abstract}
\end{center}
\textit{
In this Chapter, we continue on the solar irradiance forecasting problem described in the previous Chapter. Based on the observation that common deep learning methods struggle to properly predict the complex cloud dynamics, we apply here the methodological innovations presented in parts \ref{part:part1} and \ref{part:part2} of this thesis. We first propose a new physically-constrained deep forecasting architecture based on our PhyDNet model for video prediction. We show that it significantly boosts performances compared to the model of the previous Chapter and to other state-of-the-art deep models. Then we apply the proposed DILATE training loss function (Chapter \ref{chap:dilate}) for enforcing predictions with accurate shape and temporal localization. We also apply our APHYNITY framework (Chapter \ref{chap:aphynity}) that guarantees an optimal ML/MB decomposition. We discuss the benefits brought up by each of these mechanisms.\\
The work described in this Chapter is based on the following publication:
\begin{itemize}
    \item \cite{leguen-fisheye}: Vincent Le Guen and Nicolas Thome. "A Deep Physical Model for Solar Irradiance Forecasting With Fisheye Images". In: Proceedings of the IEEE/CVF Conference on Computer Vision and Pattern Recognition Workshops 2020 (OmniCV 2020 workshop)
\end{itemize}
}
}

\section{Introduction}

\lettrine[lines=3]{A}s discussed in the previous Chapter, forecasting solar irradiance with fisheye images remains a very difficult task for pure deep learning methods, because of the complex non-stationary motion of clouds. In this Chapter, we adapt the methodological contributions of this thesis, namely the DILATE loss function (Chapter \ref{chap:dilate}), the PhyDNet video prediction model (Chapter \ref{chap:phydnet}) and the APHYNITY framework (Chapter \ref{chap:aphynity}), for solving this problem.

\section{Proposed forecasting models}

Given a dataset of fisheye images $\mathbf{u}_{1:T} = (\mathbf{u}_1,...,\mathbf{u}_T)$ and associated solar irradiance measurements $r_t$, our goal is to forecast the future irradiance $r_{T+H}$ for a given horizon $H$. 
First, we briefly review the PhyDNet model  (Section \ref{sec:reviewphydnet}) and propose an improvement to the architecture for better disentangling the physical and residual components (Section \ref{sec:phydnet-improvement}). Then, we propose two implementations of the PhyDNet model for solar irradiance forecasting (Section \ref{sec:phydnet-solar}). The PhyDNet-monostep model is a direct adaptation of the architecture introduced in the previous Chapter, where the ConvLSTM is replaced by PhyDNet; we call this model PhyDNet-mono since we directly predict the future irradiance at the desired horizon $r_{T+H}$. We also propose the PhyDNet-multistep model, that forecasts the entire trajectory up to the desired horizon $(r_{T+1}, \cdots, r_{T+H})$. This multistep extension allows to exploit the whole intermediate trajectory for learning, for example by using the DILATE loss that compares multistep time series.

\subsection{Review of the PhyDNet model}
\label{sec:reviewphydnet}

As described in Chapter \ref{chap:phydnet}, PhyDNet \cite{leguen20phydnet} is a deep architecture that leverages partial differential equations (PDEs) for video prediction. Since physics alone is not sufficient for accurate predictions at the pixel level, PhyDNet aims at learning a latent space $\bm{\mathcal{H}}$ that linearly disentangles physical dynamics from residual factors (such as texture, details,...). The latent state $\mathbf{h}$ is decomposed into physical and residual components $\mathbf{h} = \mathbf{h^p} + \mathbf{h^r}$, and follows the dynamics: 
\begin{equation}
\!\!\!\dfrac{\partial \mathbf{h}(t,\mathbf{x})}{\partial t} \! = \!\frac{\partial \mathbf{h^p}}{\partial t} \!+\! \frac{\partial \mathbf{\mathbf{h^r}}}{\partial t} \!:=\! \bm{\mathcal{M}}_{p}(\mathbf{h^p},\mathbf{E(u)}) + \bm{\mathcal{M}}_{r}(\mathbf{\mathbf{h^r}},\mathbf{E(u)}). \!\!\!
\label{eq:eq1}
\end{equation}
The physical model $\bm{\mathcal{M}}_p$ is composed of a PDE in latent space $\Phi_p(\mathbf{h^p})$ and a correction term $\mathcal{C}_p(\mathbf{h^p},\mathbf{E(u)})$ with input data (embedded by encoder $\mathbf{E}$): $\bm{\mathcal{M}}_p(\mathbf{h^p},\mathbf{E(u)}) = \Phi_p(\mathbf{h^p})+ \mathcal{C}_p(\mathbf{h^p},\mathbf{E(u)})$. The physical predictor $\Phi_p$ encodes a general class of linear PDEs up to a differential order $q$:
 \begin{equation}
    \Phi_p(\mathbf{h^p}(t,\mathbf{x})) = \sum_{i,j: i+j \leq q}  c_{i,j} \dfrac{\partial^{i+j} \mathbf{h^p}}{\partial x^i \partial y^j}(t,\mathbf{x}).
    \label{eq:phi}
\end{equation}
Partial derivatives are computed by constrained convolutions as in PDE-Net \cite{long2018pde} and combined by learned coefficients $c_{ij}$. Discretizing the PDE $\frac{\partial \mathbf{h^p}}{\partial t}(t,\mathbf{x})= \bm{\mathcal{M}}_p(\mathbf{h^p},\mathbf{E(u)})$ with the Euler numerical scheme leads to a recurrent neural network cell (PhyCell). PhyCell performs a physical prediction step in latent space (Eq \ref{eq:prediction_phycell}) followed by a correction with embedded input data $\mathbf{E}(\mathbf{u}_t)$ (Eq \ref{eq:correction_phycell}), with a tradeoff controlled by the learned Kalman gain $\mathbf{K}_t$. 
\begin{empheq}[]{alignat=2}
&   \tilde{\mathbf{h}}^\mathbf{p}_{t+1} \!= \mathbf{h}^{\mathbf{p}}_{t} +  \Phi_p(\mathbf{h}^{\mathbf{p}}_{t})   &  \!\!\!\quad \text{\small{\textbf{Prediction}\!}} \label{eq:prediction_phycell}\\
&   \mathbf{h}^{\mathbf{p}}_{t+1} \!= \tilde{\mathbf{h}}^{\mathbf{p}}_{t+1}  + \mathbf{K}_t \odot \left( \mathbf{E}(\mathbf{u}_t) - \tilde{\mathbf{h}}^{\mathbf{p}}_{t+1} \right). & \!\!\! \quad \text{\small{\textbf{Correction}\!}} \label{eq:correction_phycell}
\end{empheq}
The residual model $\bm{\mathcal{M}}_r(\mathbf{h^p},\mathbf{E(u)})$ captures the unknown factors related to unmodelled physics, \eg appearance, texture, and is fully learned from data (implemented by a general ConvLSTM  \cite{xingjian2015convolutional}).

\begin{figure*}
    \centering
    \includegraphics[width=\linewidth]{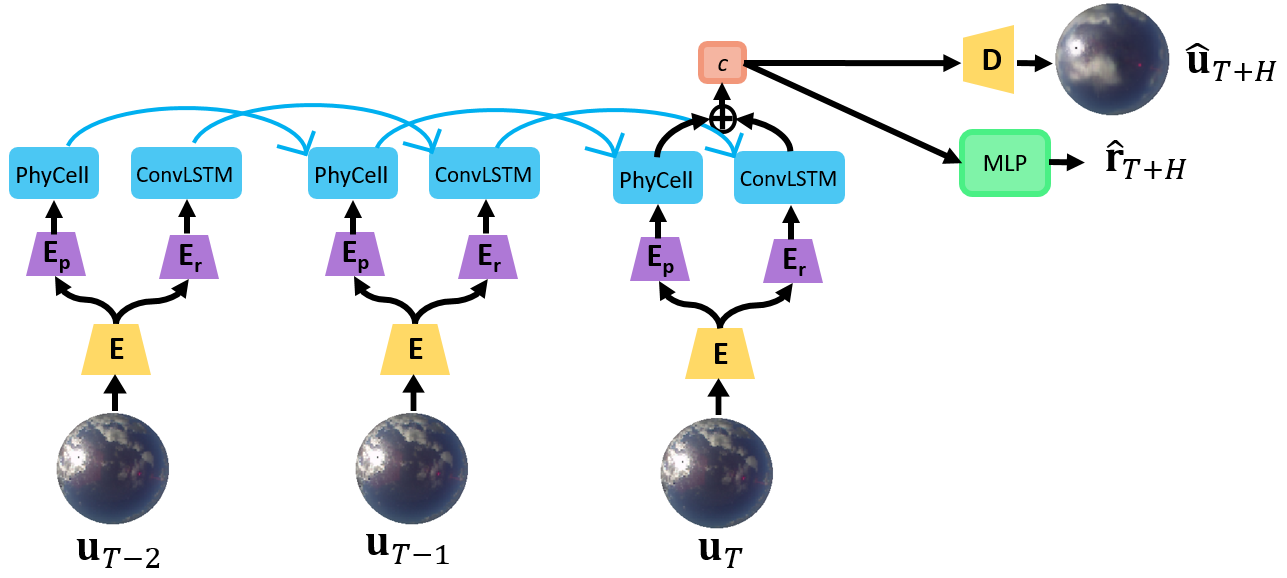}
    \caption[PhyDNet-monostep architecture for solar irradiance forecasting.]{\textbf{PhyDNet-monostep architecture for solar irradiance forecasting.} Input images are embedded by an encoder $\mathbf{E}$ in a common latent space, followed by specific encoders $\mathbf{E_p}$ and $\mathbf{E_r}$ for extracting physical and residual features. PhyDNet recurrent model is unfolded in time and computes a context vector $c=\mathbf{D_p}(\mathbf{h}^\mathbf{p}_T) +\mathbf{D_r}(\mathbf{h}^\mathbf{r}_T$), which is used for predicting the future irradiance $\hat{r}_{T+H}$ and image  $\hat{\mathbf{u}}_{T+H}$.}
    \label{fig:phydnet-mono}
\end{figure*}{}

\subsection{PhyDNet model with separate encoders and decoders}
\label{sec:phydnet-improvement}

One limitation of PhyDNet model is that images $\mathbf{u}_t$ are embedded by the encoder $\mathbf{E}$ in a common latent space for correcting the dynamics of both physical $\mathcal{C}_p(\mathbf{h^p},\mathbf{E(u)})$ and residual models  $\mathcal{C}_r(\mathbf{h^r},\mathbf{E(u)})$. This limits the disentangling ability of PhyDNet since $\mathbf{E}(\mathbf{u}_t)$ contains both physical and residual features. We thus propose to learn separate latent spaces for both branches, via additional specific encoders $(\mathbf{E_p},\mathbf{E_r})$ and decoders $(\mathbf{D_p},\mathbf{D_r})$, leading to the following dynamical model:
\begin{equation}
\!\!\!\dfrac{\partial \mathbf{h}(t,\mathbf{x})}{\partial t} \! =\! \bm{\mathcal{M}}_{p}(\mathbf{h^p},\mathbf{E_p \circ E(u)}) + \bm{\mathcal{M}}_{r}(\mathbf{\mathbf{h^r}},\mathbf{E_r \circ E(u)}). \!\!\!
\label{eq:eq1}
\end{equation}
$\mathbf{E_p}$ aims at learning a specific image embedding for controlling the physical dynamics in latent space with correction features uniquely related to physics (and similarly for $\mathbf{E_r}$). 

\noindent In the following, we denote this model as PhyDNet-dual.

\begin{figure}
    \centering
    \includegraphics[width=16cm]{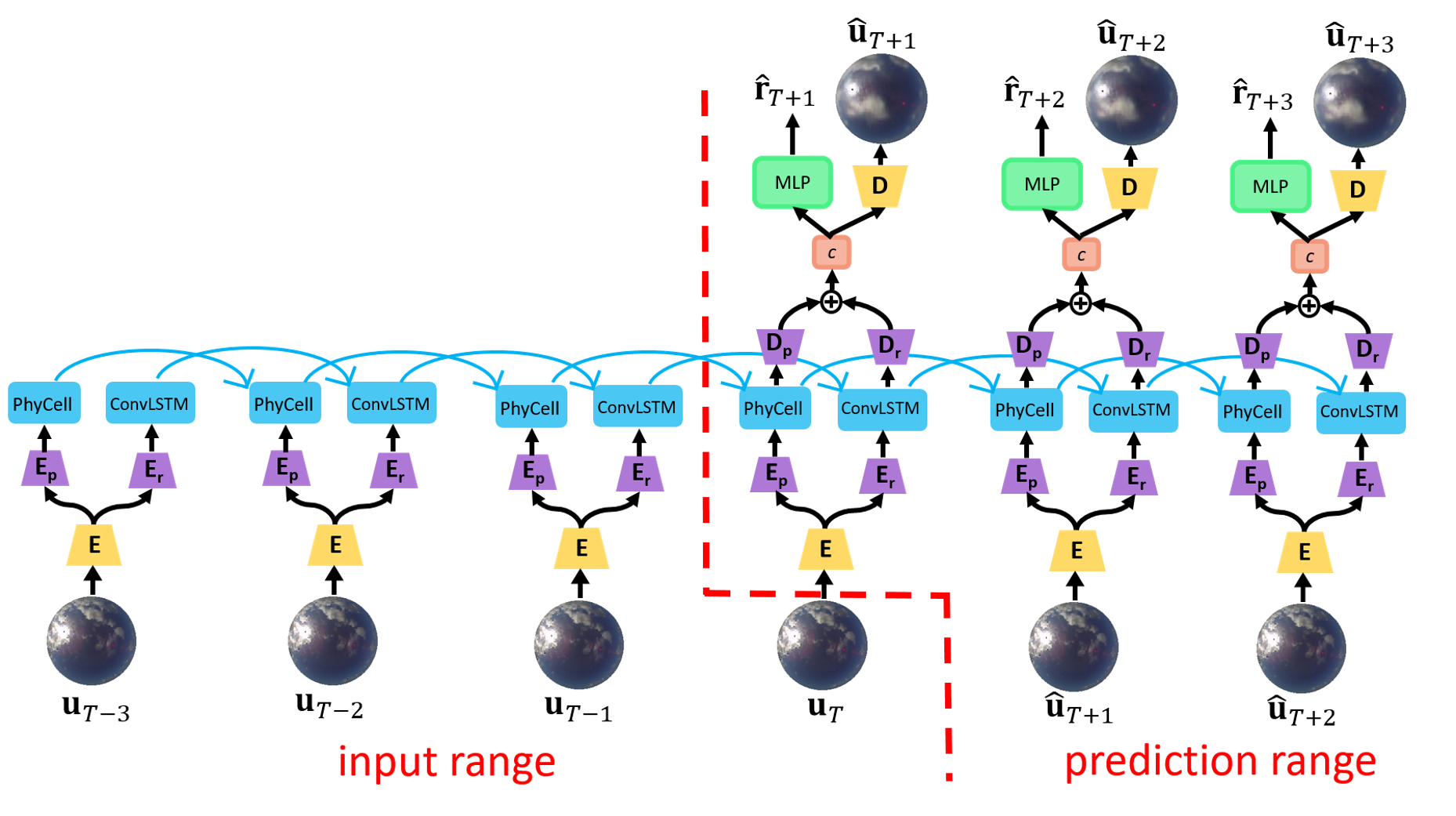}
    \caption[PhyDNet-multistep architecture for solar irradiance forecasting.]{\textbf{PhyDNet-multistep architecture for solar irradiance forecasting.} This is a Sequence To Sequence architecture with the PhyDNet recurrent neural network. Contrary to PhyDNet-monostep, this model predicts the future solar irradiance and image for each time step of the prediction range.}
    \label{fig:phydnet-multi}
\end{figure}

\subsection{PhyDNet for solar irradiance forecasting}
\label{sec:phydnet-solar}

We first propose the PhyDNet-monostep architecture, which is a direct adaptation of the forecasting model described in Chapter \ref{chap:overview_fisheye}. Depicted in Figure \ref{fig:phydnet-mono}, we replace the ConvLSTM encoding the input sequence $\mathbf{u}_{1:T}$ by the PhyDNet-dual encoder, allowing to extract physically-constrained features. The final physical and residual latent states are decoded by their respective specific decoders $\mathbf{D_p}$ and $\mathbf{D_r}$ and then summed to get a context vector $c=\mathbf{D_p}(\mathbf{h}^\mathbf{p}_T) +\mathbf{D_r}(\mathbf{h}^\mathbf{r}_T$). Then a multi-layer perceptron (MLP) uses the input context $c$ to forecast the future irradiance $\hat{r}_{T+H}$, and the global decoder $D$ simultaneously forecasts the future image $\mathbf{D}(c) = \hat{\mathbf{u}}_{T+H}$. 

We also propose the PhyDNet-multistep shown in Figure \ref{fig:phydnet-multi}. Instead of directly forecasting the future values from the last step of the input range, PhyDNet-multistep is composed of a PhyDNet-dual recurrent decoder. It provides future image and irradiance predictions for each time step of the prediction range $(T+1,\cdots, T+H)$. This multi-step strategy allows to supervise the model based on a whole predicted trajectory: we evaluate in the experiments the application of the DILATE training loss function instead of the MSE.

\section{Experimental results}
We conduct experiments on the same fisheye dataset as in the previous Chapter. The training dataset for solar irradiance forecasting is composed of 180,000 sequences of 10 images spaced by 1min (with the associated ground truth solar irradiance measurements) from the years 2014 to 2016 at La Reunion Island, and the evaluation dataset of 20,000 sequences during the year 2013 on the same site. We keep 5 images for the input range and predict the 5 following images and solar irradiances. Images are resized at $80 \times 80$ pixels.

\subsection{Irradiance forecasting with PhyDNet}

We forecast solar irradiance at a 5min horizon, given a 5min past context. We compare quantitatively the proposed PhyDNet models against recent competitive video prediction baselines: ConvLSTM \cite{xingjian2015convolutional} (which corresponds to the model presented in Chapter \ref{chap:overview_fisheye}) and PredRNN \cite{wang2017predrnn}. Each baseline is adapted in the same way for solar irradiance forecasting, in the monostep or multistep settings.

We report in Table \ref{tab:irradiance} the normalized RMSE\footnote{nRMSE = Root Mean Squared Error normalized by the mean value of the quantity on the train set, expressed as a percentage.} for the predicted irradiance (KGHI)  $\hat{r}_{T+\text{5min}}$.

\begin{table}[H]
    \centering
        \caption{Solar irradiance (KGHI) forecasting at a 5min horizon.}
    \begin{tabular}{c|c}
     \toprule
        &  irradiance nRMSE   \\ 
         \midrule
    PhyDNet-monostep irradiance only    & 27.8 \% \\
    ConvLSTM-monostep \cite{xingjian2015convolutional}  & 26.6 \%  \\
    PredRNN-monostep \cite{wang2017predrnn}  & 25.1 \% \\
    PhyDNet-monostep \cite{leguen20phydnet}    & 24.4 \%  \\
    PhyDNet-dual-monostep  & 23.5 \%   \\ 
    PhyDNet-dual-multistep &   \textbf{21.5 \%} \\
     \bottomrule
    \end{tabular}
    \label{tab:irradiance}
\end{table}{}

The first line in Table \ref{tab:irradiance} corresponds to a PhyDNet-monostep that only predicts the future irradiance  $\hat{r}_{T+\text{5min}}$ and not the future image. It gives the worst performances among compared models, indicating that the the joint image-irradiance multitask setting provides a better supervision for training the forecasting model. All the other models in Table  \ref{tab:irradiance} jointly predict future images and irradiances.

We observe that, in the monostep setting, the PhyDNet recurrent neural network gives better results (24.4\%) compared to the ConvLSTM (26.6 \%) and PredRNN (25.1 \%). It shows that integrating physical dynamics greatly helps in modelling the cloud motion. With the separate encoders and decoders, PhyDNet-dual-monostep further improves the performances (23.5 \%). Finally, we see that with the multistep strategy, PhyDNet-dual-multistep provides another large improvement (21.5 \%). The supervision coming for a complete trajectory of future images and irradiances significantly boosts the training process.

\begin{figure}
    \centering
    \includegraphics[width=15cm]{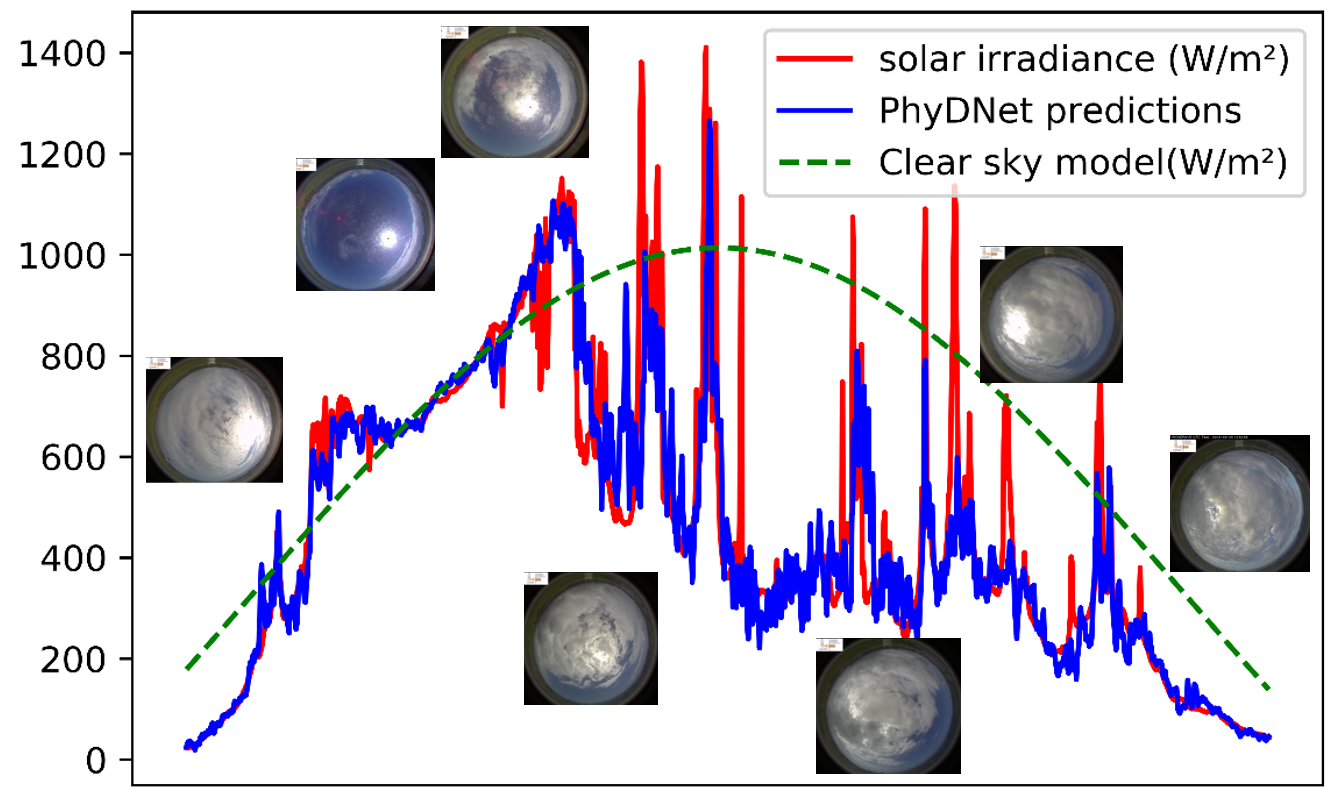}
    \caption[Short-term forecasting with fisheye images.]{5min ahead solar irradiance forecasts from fisheye images. Our proposed deep model leveraging physical prior knowledge accurately predicts the sharp intra-day solar irradiance fluctuations.}
    \label{fig:fisheye-qualitative}
\end{figure}

We provide in Figure \ref{fig:fisheye-qualitative} a qualitative illustration of the  5min GHI predictions of the PhyDNet-dual-multistep predictions on a particular day. We see that our model closely follows the ground truth measurements and is able to successfully anticipates the sharp irradiance fluctuations, despite the fast alternation of clouds and sun. 

\subsection{Applications of DILATE and APHYNITY}

We evaluate here the application of the DILATE loss function (Chapter \ref{chap:dilate}) and APHYNITY framework (Chapter \ref{chap:aphynity}) introduced in this thesis. 

We use the DILATE loss at training time instead of the MSE for the predicted irradiance time series (5 predicted points in the future). We experimentally fixed the hyperparameter $\alpha$ balancing the shape and temporal term to 0.95, which yields the best results. 

For APHYNITY, we minimize the norm of the residual hidden state $\mathbf{h^r}$ for all time steps. Note that contrary to the APHYNITY models presented in Chapter \ref{chap:aphynity}, we do not use here the NeuralODE for extrapolating the trajectory in latent space, but the PhyDNet recurrent neural network. Exploiting a NeuralODE integration is a promising way for future works.

Forecasting results are presented in Table \ref{tab:irradiance-dilate-aphynity}. We compare the application of DILATE, APHYNITY and the combination of both mechanisms. We can see that these 3 variants lead to approximately similar performances: they improve slightly over the PhyDNet-dual-multistep baseline in normalized RMSE and in the DILATE objective (confirmed with the shape and temporal metrics). 

\paragraph{Discussion} The performance improvement due to DILATE and APHYNITY exists, but is rather small compared to the performance gap due to the architecture design of PhyDNet-dual and to the multistep training scheme. We discuss here the possible reasons. Concerning DILATE, we apply the loss in our experiments on predicted trajectories of 5 timesteps. This is rather small compared to our experiments in Chapter \ref{chap:dilate} (the shortest trajectories have 20 timesteps for the \texttt{Synthetic} dataset). For shorter trajectories, dynamic time warping is less relevant, and the sharp variations are more difficult to visualize. Augmenting the forecasting horizon of our method and reducing the time interval between images (up of the 10s sampling frequency) are interesting future directions for better exploiting the DILATE loss.

Regarding APHYNITY, the physical model used in PhyDNet is a class of linear PDEs. This is a very coarse physical prior, more general than in the experiments presented in Chapter \ref{chap:aphynity}. Moreover, due to the non-observed prior, the physical model is applied in a learned latent space which is not explicitly controlled, contrary to the fully-visible setting in Chapter \ref{chap:aphynity}. This may explain why optimizing the ML/MB decomposition leads to less improvement. An appealing future direction would be to exploit more specific physical laws modelling the cloud motion and/or a more precise description of the input space where the physical laws apply.

\begin{table}[H]
    \centering
    \caption{Evaluation of the DILATE loss and the APHYNITY framework on the 5-min solar irradiance forecasting problem.}    
    \begin{tabular}{cccccc}
    \toprule
         & nRMSE & DTW & TDI & DILATE & Ramp score  \\
         \midrule
    PhyDNet-dual-multistep  & 21.5 \% & 34.1 & 63.3 &  97.4  & 78.6 \\
    DILATE  & \textbf{21.2 \%} & \textbf{33.6} & 63.0 & 96.6 & \textbf{77.3}  \\
    APHYNITY & 21.4 \% & 34.2 & 62.2 & 96.4  & \textbf{77.3} \\
    APHYNITY + DILATE & \textbf{21.2 \%} & \textbf{33.6} & \textbf{61.5} & \textbf{95.1} & 77.9 \\
    
    \bottomrule
    \end{tabular}
    \label{tab:irradiance-dilate-aphynity}
\end{table}

\subsection{Video prediction}

We then evaluate PhyDNet-dual-multistep on the video prediction task. Given 5 input images with a 1 min interval, we forecast the 5 future images up to $t_0 + 5\text{min}$. We compare PhyDNet-dual-multistep with ConvLSTM and Memory In Memory (MIM) \cite{wang2019memory}. 
Evaluation metrics are the mean squared error (MSE), mean absolute error (MAE) and the structural similarity index SSIM (higher is better). Results shown in Table \ref{tab:video-prediction} reveal that PhyDNet-dual-multistep outperforms both baselines for all metrics. It confirms that incorporating physical prior information for modelling cloud motion is beneficial compared to fully data-driven algorithms.

\begin{table}[H]
  \caption{Quantitative video prediction results.}
    \centering
    \begin{tabular}{c|c|c|c}
     \toprule
         &  MSE & MAE & SSIM  \\
         \midrule
    ConvLSTM \cite{xingjian2015convolutional} & 83.1  & 681 & 0.845  \\
    MIM \cite{wang2019memory}  & 68.6  & 635   & 0.840  \\
    PhyDNet-dual-multistep & $\mathbf{68.1}$ & $\mathbf{629}$  & $\mathbf{0.862}$   \\ 
     \bottomrule
    \end{tabular}
  
    \label{tab:video-prediction}
\end{table}{}

\begin{figure*}
    \centering
    \includegraphics[width=\linewidth]{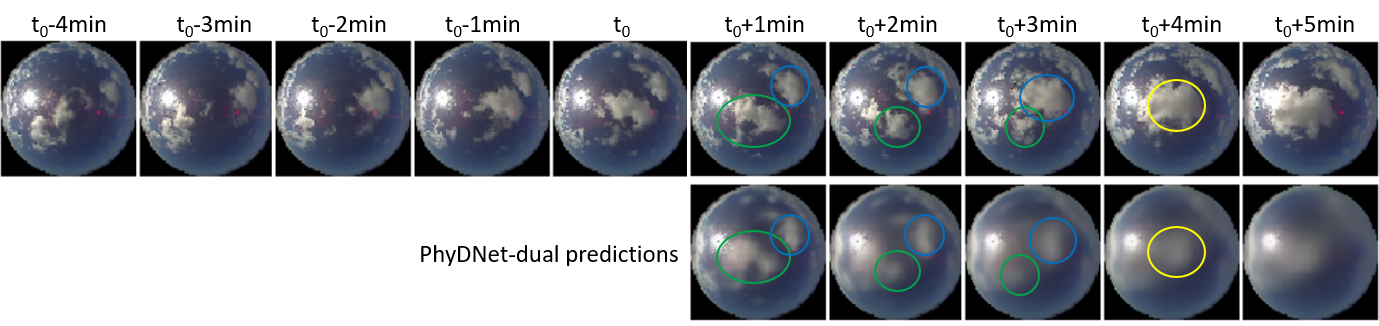}
    \caption[Qualitative fisheye video forecasting results.]{Qualitative fisheye video forecasting results up to 5min horizon. The proposed model successfully predicts the motion of the blue and green clouds that move nearer and finally merge into the yellow cloud.}
    \label{fig:video-prediction}
\end{figure*}{}

We show in Figure \ref{fig:video-prediction} a video prediction example of PhyDNet-dual model. The future of this sequence presents 2 clouds (circled in blue and green) moving closer between $t_0$ and $t_0+3\text{min}$ and finally merging at time $t_0+4\text{min}$. We observe that PhyDNet-dual predicts the same outcome with a good accuracy on cloud location, although clouds become blurry because of uncertainty.

\begin{figure*}
    \centering
    \includegraphics[width=16cm]{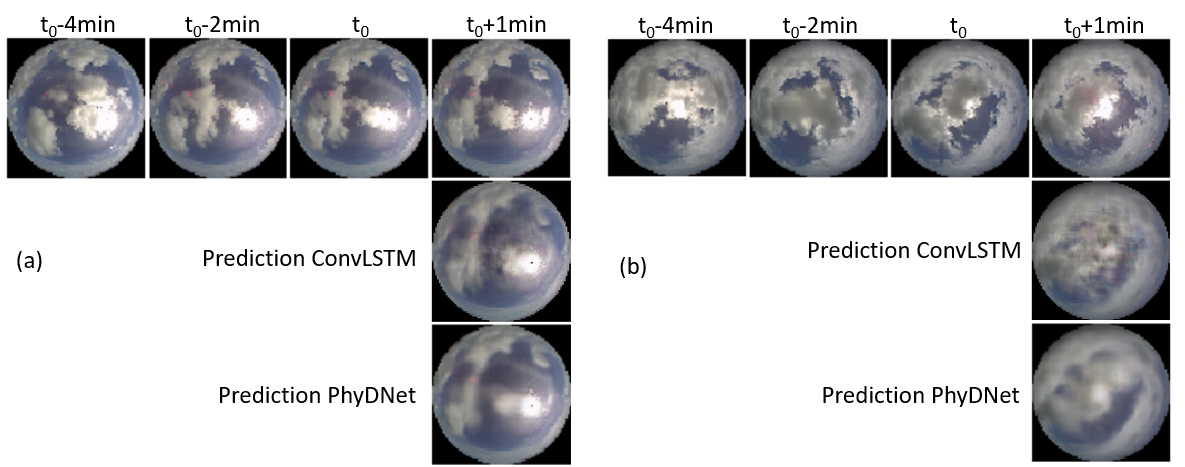}
    \caption{Qualitative forecasting comparison between PhyDNet-dual-multistep and ConvLSTM.}
    \label{fig:ablation_fisheye}
\end{figure*}

In Figure \ref{fig:ablation_fisheye}, we provide a particular comparison to ConvLSTM \cite{xingjian2015convolutional}, which forms the residual branch of PhyDNet.   In sequence (a), we see that the shape of the small cloud getting nearer the sun is much better predicted by PhyDNet-dual. In sequence (b), the sun will reappear 1 min in the future. PhyDNet-dual provides a better anticipation by prediction a bright spot at the sun location and better defined cloud shapes. It confirms that incorporating physical dynamics greatly improves the predictions of natural phenomena, with a small amount of additional parameters with respect to ConvLSTM. 

{}

\section{Conclusion}

In this Chapter, we have explored the methodological contributions of this thesis for solving the solar irradiance forecasting problem at EDF. We have proposed an improvement of our PhyDNet video prediction model that we have adapted for this task. The PhyDNet model greatly improves the performances compared to competitive pure data-driven, confirming the benefits of the MB/ML integration. We have also highlighted the crucial importance of making multistep instead of monostep predictions. Furthermore, we have applied the DILATE loss function and the APHYNITY framework, which further improve the forecasting performances, albeit slightly.

\clearpage{\pagestyle{empty}\cleardoublepage}


\chapter{Conclusion and perspectives}
\label{chap:conclusion}
\minitoc

\section{Summary of contributions}

\lettrine[lines=3]{F}rom a general perspective,  we have explored in this thesis how to incorporate prior knowledge into machine learning for improving spatio-temporal forecasting models. More specifically, we have studied two important scientific challenges.

\subsection{Multistep forecasting of non-stationary dynamics}

In many real-world applications, time series present non-stationary dynamics with possible sharp variations, \eg traffic flows, financial stocks, or solar irradiance time series. Current state-of-the art deep learning methods for multistep deterministic and probabilistic forecasting struggle to properly predict these abrupt events: their predictions often smooth the sharp variations and/or present a temporal missalignment. One of the reasons is that most works focus on neural network architecture design and overlook the choice of the training loss function. The dominantly used loss function is the mean squared error (MSE), that is unable to take into account global information about the multistep dynamics.

In this thesis, we have shown that this is possible to design dedicated multistep loss functions to impose a certain desired behaviour to the output. For time series, we focus on shape and temporal criteria that are commonly used as assessment metrics in applications. In Chapter \ref{chap:criteria}, we have drawn a panorama of shape and temporal criteria based on smooth approximations of Dynamic Time Warping (DTW) and Time Distortion Index (TDI). We have expressed them both as dissimilarities (loss functions) and similarities (positive semi-definite kernels). We have insisted on their differentiability, which is an important requirement for training models with gradient-based optimization, and propose optimized implementations of these losses for efficient back-propagation training.

We have then applied the proposed shape and time differentiable criteria to two spatio-temporal forecasting contexts. In Chapter \ref{chap:dilate}, we have introduced a differentiable loss function (DILATE), that combines a shape term and a temporal term, for training any deep forecasting model to produce multistep deterministic forecasts. We have shown that training with DILATE produces sharper predictions with a better temporal localization than training with the standard MSE, while maintaining the performances with MSE evaluation. 

In Chapter \ref{chap:stripe}, we have proposed the STRIPE model for probabilistic forecasting. In order to produce a limited set of possible scenarios that reflect the shape and temporal variability of ground truth trajectories, the STRIPE model is equipped with a diversification mechanism that structures the output diversity. This is done with a diversity loss relying on determinantal point processes (DPP), with two shape and temporal criteria introduced in Chapter \ref{chap:criteria}. STRIPE leads to more diverse forecasts according to shape and temporal criteria without sacrificing on quality. We have also revealed the crucial importance to decouple the criteria used for quality and diversity. 

\subsection{Exploiting incomplete prior physical knowledge in machine learning models}

The extrapolation task underlying spatio-temporal forecasting is quite different and much more challenging for pure data-driven methods than the perception tasks at the origin of the impressive success of deep learning. For example, forecasting complex natural dynamics such as climate remains out of the scope of pure machine learning. An appealing solution is to incorporate external physical knowledge, which is an old research problem that is still open today. In this thesis, we have particularly focused on exploiting \textit{incomplete} physical knowledge, in contrast to mainstream methods that suppose a full prior knowledge. The incomplete case can stem from the difficulty of the phenomenon that remains elusive to a complete description from physical laws, \eg for modelling all the complex interacting phenomena for predicting the evolution of the atmosphere, or from a non-observable prior context, \ie when the  dynamical model does not apply directly in the input space.

In Chapter \ref{chap:phydnet}, we have tackled the problem of generic video prediction. It is an example of a non-observable prior context: although there often exists some physical dynamical prior, for example on the motion of clouds in fisheye images, physical laws do not directly apply at the pixel level. The dynamical model is meaningful in a space where the clouds have previously been identified and segmented. We have introduced the PhyDNet prediction model that automatically learns a latent space in which we suppose that a class of linear partial differential equations apply. PhyDNet is a two-branch architecture: the first branch captures the physical dynamics. Since this prior knowledge is often insufficient to fully describe the content of videos, PhyDNet is composed of a second branch for modelling the complementary information necessary for accurate prediction (\eg texture, details, \etc). We have highlighted the ability of PhyDNet to properly disentangle the physical dynamics from these unknown factors. 

In Chapter \ref{chap:aphynity}, we have further delved into the question of augmenting incomplete physical models with deep data-driven counterparts. This is an area that has been explored by very few works up to now, and mostly empirically. We have proposed the APHYNITY framework, that consists in decomposing the dynamics in two components: a physical component accounting for the dynamics for which we have some prior knowledge, and a data-driven component accounting for insufficiencies of the physical model. APHYNITY is a principled learning framework minimizing the norm of the data-driven augmentation, that theoretically guarantees a unique decomposition under mild assumptions. APHYNITY is able to seamlessly adapt to different approximation levels of prior physical knowledge, covering the whole range of Machine Learning /Model-Based methods presented in Chapter \ref{chap:intro}. We have  exhibited the superiority of APHYNITY over data-driven, incomplete physics, and state-of-the-art approaches combining ML and MB methods, both in terms of forecasting and parameter identification on three various classes of physical systems. 

\subsection{Solar irradiance forecasting with fisheye images}

Finally, we have proposed solutions to the industrial solar irradiance forecasting problem with fisheye images raised at EDF. In Chapter \ref{chap:overview_fisheye}, we have presented the challenges of the problem and proposed a first deep learning model for estimating and forecasting solar irradiance. We have also discussed the limitations of standard deep learning forecasting approaches in this context, that have motivated the contributions of this thesis.

In Chapter \ref{chap:phydnet_fisheye}, we have applied the methodological contributions exposed in parts \ref{part:part1} and \ref{part:part2} of this thesis. We have improved and adapted our PhyDNet model for physically-constrained fisheye image prediction. The PhyDNet model greatly improves the performances compared to competitive pure data-driven, confirming the benefits of the physical knowledge integration. Furthermore, we have applied the DILATE loss function and the APHYNITY framework, leading to another (relatively small) performance gain.

\section{Perspectives}

We present here a non-exhaustive list of possible future research directions for different time horizons.

\subsection{Directions for improving solar irradiance forecasting}

\paragraph{Application of DILATE and APHYNITY} 

As discussed in Chapter \ref{chap:phydnet_fisheye}, the main performance improvements compared to pure deep learning methods stem from the application of our physically-constrained PhyDNet architecture. The application of the DILATE loss and the APHYNITY framework further improve the performances, but less significantly.

Concerning the DILATE loss function, we have applied in our experiments the loss on future trajectories of 5 timesteps, which is rather small compared to the experiments in Chapter \ref{chap:dilate}  (the shortest trajectories have 20 timesteps for the \texttt{Synthetic} dataset). For short trajectories, the sharp variations are harder to visualize and  the use of dynamic time warping (DTW) is less relevant. To fully exploit the capacity of the DILATE loss, an interesting perspective is to augment the length of future trajectories, by reducing the processing interval between images or by augmenting the forecasting horizon.

Regarding APHYNITY, we use in the PhyDNet model a very general physical prior model: a class of linear PDEs. This is a weaker prior than those used in Chapter \ref{chap:aphynity}. Moreover, due to the non-observability of the prior, the physical model is applied in a learned latent space which is not explicitly controlled, contrary to the fully-visible setting in Chapter \ref{chap:aphynity}. This may explain why the Machine Learning / Model Based decomposition is more challenging to optimize. An interesting future direction would be to exploit more specific physical laws modelling the cloud motion and/or a more precise description of the input space where the physical laws apply. 

\paragraph{Probabilistic forecasting} 

In this thesis, we have forecasted solar irradiance in a deterministic manner with the PhyDNet model. An interesting future work is to extend our contributions on probabilistic forecasting to this problem. An adaptation of the STRIPE model would provide to the decision makers a small set of possible scenarios about the cloud motion (for example if the clouds will occlude the sun or not, and at what temporal horizon).

\paragraph{Handling the rotational distortion of fisheye images} 

 Fisheye images present a rotational symmetry along the vertical axis. Clouds in linear translation are observed as a curved motion in fisheye images. To handle this distortion induced by the fisheye  camera objective, some forecasting methods preprocess fisheye images by projecting them in a plane where a translational cloud motion is linear. In this thesis, we have instead directly processed raw fisheye images with general convolutional layers commonly used in computer vision for encoding translation equivariance. Future works include applying the plane projection or polar transformation \cite{paletta2021spin} as preprocessing, or evaluating more dedicated neural network layers that handle rotation equivariance, such as spherical CNNs \cite{cohen2016group,cohen2018spherical}.

\subsection{Applications of deep augmented physical models}

\subsubsection*{Non-stationary dynamics forecasting}

In this thesis, our contributions towards non-stationary dynamics forecasting concern rethinking the training process by including shape and temporal criteria, and are thus agnostic to the forecasting architectures. An interesting future perspective would be to also incorporate prior knowledge in the model architectures, as studied in part \ref{part:part2} of this thesis. For time series, leveraging trend, seasonality and extrinsic prior knowledge (such as special events) \cite{laptev2017time} could help to better model the non-stationary abrupt changes and measure their impact on diversity and model confidence \cite{gal2016dropout,corbiere2019addressing}. The combination between a traditional forecasting model with interpretable and controlled factors (\eg a ARIMA model) and a data-driven augmentation network would be a possible application case for APHYNITY.

\subsubsection*{Optical flow}

Optical flow estimation is a long-standing problem in computer vision, consisting in estimating the motion field between two frames. It is a core building block for many applications, such as image compression or object tracking. For example, optical flow is used to understand the cloud motion in traditional forecasting methods with fisheye images.

Traditional methods for optical flow, \eg the Lucas-Kanade \cite{lucas1981iterative}
and the Horn-Schunk \cite{horn1981determining}  models, are based on the brightness constancy assumption $I_1(\mathbf{x}) = I_2(\mathbf{x}+w)$ that states that the pixel intensity is preserved after advection by the flow field $w$. Linearising this equation leads to the celebrated optical flow PDE:
\begin{equation}
    \frac{\partial I}{\partial t} (t,\mathbf{x}) = - w(t,\mathbf{x}) \cdot \nabla I (t,\mathbf{x}).
    \label{eq:flot}
\end{equation}

The PDE in Eq \ref{eq:flot} is a simplified physical model, since the brightness constancy assumption is violated in several conditions, \eg in presence of occluded objects, local, global illumination changes or specular reflexions.

Other traditional methods exploit different prior physical models for optical flow in specific contexts, \eg the PDE continuity equation for fluid flows \cite{corpetti2002dense}.

More recently, deep learning approaches have proposed learning optical flow in an end-to-end fashion and have become state-of-the-art \cite{flownet,sun2018pwc,raft,stone2021smurf}. Two classes of methods exist: supervised and unsupervised ones. In the supervised context \cite{flownet,sun2018pwc,raft}, deep learning methods do not exploit the brightness constancy hypothesis anymore, or indirectly (through the computation of a cost volume). Instead, they rely on large synthetic datasets of annotated image pairs, making their generalization to real-world datasets not obvious.

On the other side, unsupervised deep learning approaches \cite{jason2016back,liu2020learning,stone2021smurf} are closer in spirit to traditional approaches. Without ground truth labels for optical flow, they rely on a photometric reconstruction loss. The reason deep unsupervised methods outperform traditional methods is that they fully exploit the correlations from the training dataset, instead of independently optimizing a flow field for each image pair. Typical photometric losses include  the L1 loss that directly assumes intensity constancy, or more robust losses such the Charbonnier loss, the structural similarity (SSIM) \cite{jonschkowski2020matters}
or the census loss \cite{meister2018unflow} that is robust to global illumination changes. Although adequate losses may address some limitations of the brightness constancy assumption, they do not overcome all failure cases. Therefore the photometric constancy assumption also represents a simplified physical model.

In this context, an appealing research perspective is to explicitly exploit the simplified optical flow PDE in Eq \ref{eq:flot} in a deep augmented model. This is a favorable case for the application of our APHYNITY framework. This ML/MB integration could regularize and boost the performances of deep supervised estimation models, in particular for generalizing to new datasets. It could also be applied in a semi-supervised context, where the learned data-driven augmentation could complement the simplified photometric constancy for non-annotated images.

\subsubsection*{Model-Based Reinforcement Learning}

 Reinforcement Learning (RL) \cite{sutton2018reinforcement} is a branch of machine learning that studies how autonomous agents make decisions in the environment in order to maximize their cumulative reward. Combined with deep learning, RL has encountered impressive successes for example by reaching super-human performance at the game of Go \cite{silver2017mastering}.

There are two main modelling approaches in RL: \textit{model-based} and \textit{model-free}. In the model-based approach, the agent uses an internal predictive model of the world to simulate the consequences of its actions, and choose the best action accordingly. In contrast, in the model-free approach, the control policy is learned directly from experienced trajectories, without any dynamical model.

\begin{figure}[H]
    \centering
    \includegraphics[width=16cm]{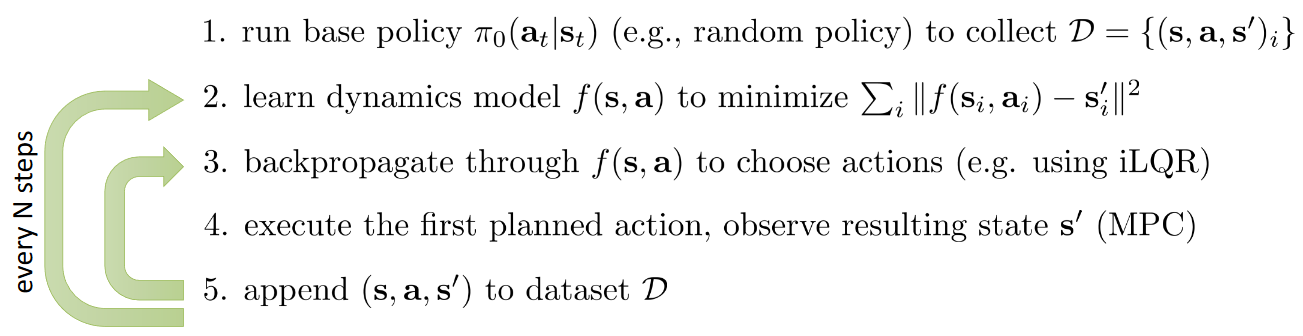}
    \caption[Principle of Model-Based Reinforcement Learning.]{Principle of Model-Based Reinforcement Learning.}
    \label{fig:MBRL}
\end{figure}

The principle of Model-Based Reinforcement Learning (MBRL) is illustrated in Figure \ref{fig:MBRL}. It consists in planning through a dynamical model $f(\mathbf{s}_t,\mathbf{a}_t)$, where $\mathbf{s}_t$ is the current state and $\mathbf{a}_t$ the chosen action. The dynamical model is learned to minimize the future (discounted) cumulative cost:
\begin{equation}
\label{eq:opt}
\underset{a_{t_0},...,a_{\infty}}{\min} ~~~  \sum_{t=t_0}^{\infty} \gamma^{t-t_0} c(\hat{\mathbf{s}}_t, \mathbf{a}_t)  ~~~ 
\mathrm{subject~to} ~~~~  \forall t \geq t_0, \frac{\diff \mathbf{s}_t}{\diff t} =f(\mathbf{s}_t,\mathbf{a}_t).
\end{equation}
where $c$ is a cost function and $\gamma <1$ a discount factor. 

The dynamical model $f$ can be a simple linear (or locally linear) model, a physical model, or a pure data-driven model parameterized by a deep neural network\footnote{Please note that in the RL community, the term \textit{model-based} denotes the presence of a dynamical model $f$, that can either be a pure data-driven model (denoted as \textit{Machine Learning} in this thesis) or a model with a physical prior (denoted as \textit{Model-Based} in this thesis).}. In all cases, the model $f$ is often too simplified to perfectly extrapolate the future trajectories.

A common solution for nonetheless exploiting the incomplete model is to consider short-term rollouts and perform Model Predictive Control (MPC) \cite{nagabandi2018neural,janner2019trust}, which consists in replanning frequently to mitigate the error propagation in the forecasted trajectories.

An interesting future direction would be to explore deep augmented models in this MBRL case. A simplified prior dynamical model of the system could be augmented with a data-driven counterpart and learned together with the APHYNITY framework. This cooperation could improve the accuracy of the predictive model, enabling to perform more truthworthy long-term rollouts, and to replan less frequently.

An other appealing direction  concerns improving the exploration process in Reinforcement Learning with a diversity-promoting mechanism \cite{pathak2017curiosity,eysenbach2018diversity,leurent2020robust}; this mechanism could be implemented with determinantal point processes with adequate kernels to represent structured diversity. 

\subsection{Long-term perspectives}

The field of spatio-temporal forecasting is still a very active area of research in the AI community, and has not reached yet the degree of maturity of deep learning in computer vision or language. Forecasting complex dynamics  remains highly challenging for pure machine learning, due to the relative current scarcity of data for learning complex natural phenomena such as climate. The quantity of training data will likely continue to grow in future years, yet it is not clear at which point it will become sufficient. Relying on this growing data accumulation, the exploration of bigger and bigger models to overcome the underfitting phenomenon is a possible way, which is faced with many computational challenges.

The other way, which was explored in this thesis, is to incorporate external knowledge to regularize machine learning models, in the form of loss functions, model architectures or training strategies. We hope that the contributions of this thesis will open the way towards hybrid and more flexible Machine Learning/Model-Based models for tackling complex real-world applications, \eg in climate science, robotics or reinforcement learning. In particular, the augmentation strategy explored in this thesis - a linear combination - is rather particular. For many incomplete models, there exists high-order interactions between the simplified model and the residual information. Exploring more general augmentations schemes, linked with the growing field of neural architecture search \cite{elsken2019neural}, is an appealing direction for future years.

\clearpage{\pagestyle{empty}\cleardoublepage}





\bibliographystyle{plain} 
\bibliography{refs.bib}
\clearpage{\pagestyle{empty}\cleardoublepage}
\thispagestyle{empty}
\begin{vcenterpage}

\chapter*{Résumé de la thèse}
\addcontentsline{toc}{chapter}{Résumé}
\adjustmtc
\markright{\MakeUppercase{Resume}}

\section{Introduction}

Cette thèse aborde le problème de la prédiction spatio-temporelle par apprentissage profond. Cela correspond à la tâche de prédiction de phénomènes complexes sous forme de séries temporelles ou de vidéos, ce qui nécessite de modéliser des dépendances temporelles complexes avec d’importantes corrélations spatiales. Ce sujet est d’une importance cruciale pour de nombreuses applications, telle que la prévision climatique, le diagnostic médical, l'évolution des marchés financiers, la demande pour des produits en commerce ou la maintenance prédictive dans l'industrie. A Électricité de France (EDF), l’application qui motive cette thèse est la prévision à court-terme de la production photovoltaïque à l’aide d’images fisheye. Cette tâche est habituellement résolue à l'aide d'algorithmes basés sur les prévisions météo et les images satellite. Toutefois ces sources de données ont une résolution spatiale et temporelle insuffisante pour prédire  l'irradiance solaire à très court-terme ($<$ 20min) à l'échelle d'un parc de production photovoltaïque particulier.

Dans cette thèse, nous abordons ces tâches de prédiction avec des méthodes d'intelligence artificielle, en particulier l'apprentissage statistique et l'apprentissage profond. Ces dernières années, l'apprentissage profond a connu un rebond de popularité impressionnant avec le succès du réseau de neurones profond AlexNet \cite{krizhevsky2012imagenet} qui a surpassé toutes les méthodes d'apprentissage machine traditionnel lors de la compétition de classification d'images ImageNet. Depuis, l'apprentissage profond s'est imposé comme le paradigme état de l'art pour de nombreuses tâches liées à la perception, telle que la vision par ordinateur, la reconnaissance vocale ou le traitement du langage naturel. Malgré ces impressionnants succès, les méthodes d’apprentissage entièrement basées sur les données sont limitées pour extrapoler l’évolution de systèmes physiques complexes, particulièrement quand la volumétrie de données est faible et pour des séries temporelles non-stationnaires avec des possibles variations brusques. La tâche d'extrapolation sous-jacente est par nature très différente des tâches de perception pour lesquelles l'apprentissage profond est très efficace, et nécessite de modéliser des dynamiques complexes.

Pour pallier à ces problèmes, nous proposons dans cette thèse d'exploiter de l'information physique a priori en combinaison avec les méthodes d'apprentissage basées données. Il s'agit d'une question très étudiée dans la littérature mais qui reste toujours largement ouverte. Les différents contextes de prévision sont illustrés sur la Figure \ref{fig:physics_data_fr}. D'un côté les méthodes basées modèle (\textit{model-based, MB)} supposent une bonne compréhension mathématique ou physique des phénomènes, souvent formalisée sous forme d'équations différentielles ordinaires ou partielles. A partir de données pour les conditions initiales et aux limites, la prédiction est effectuée par la résolution numérique des équations. C'est le paradigme dominant dans de nombreux domaines scientifiques, par exemple la mécanique des fluides computationnelle. Toutefois ces méthodes sont limitées si la connaissance physique est imparfaite, ce qui est souvent le cas pour des systèmes physiques complexes comme la modélisation du climat.

\begin{figure}
    \centering
    \includegraphics[width=\linewidth]{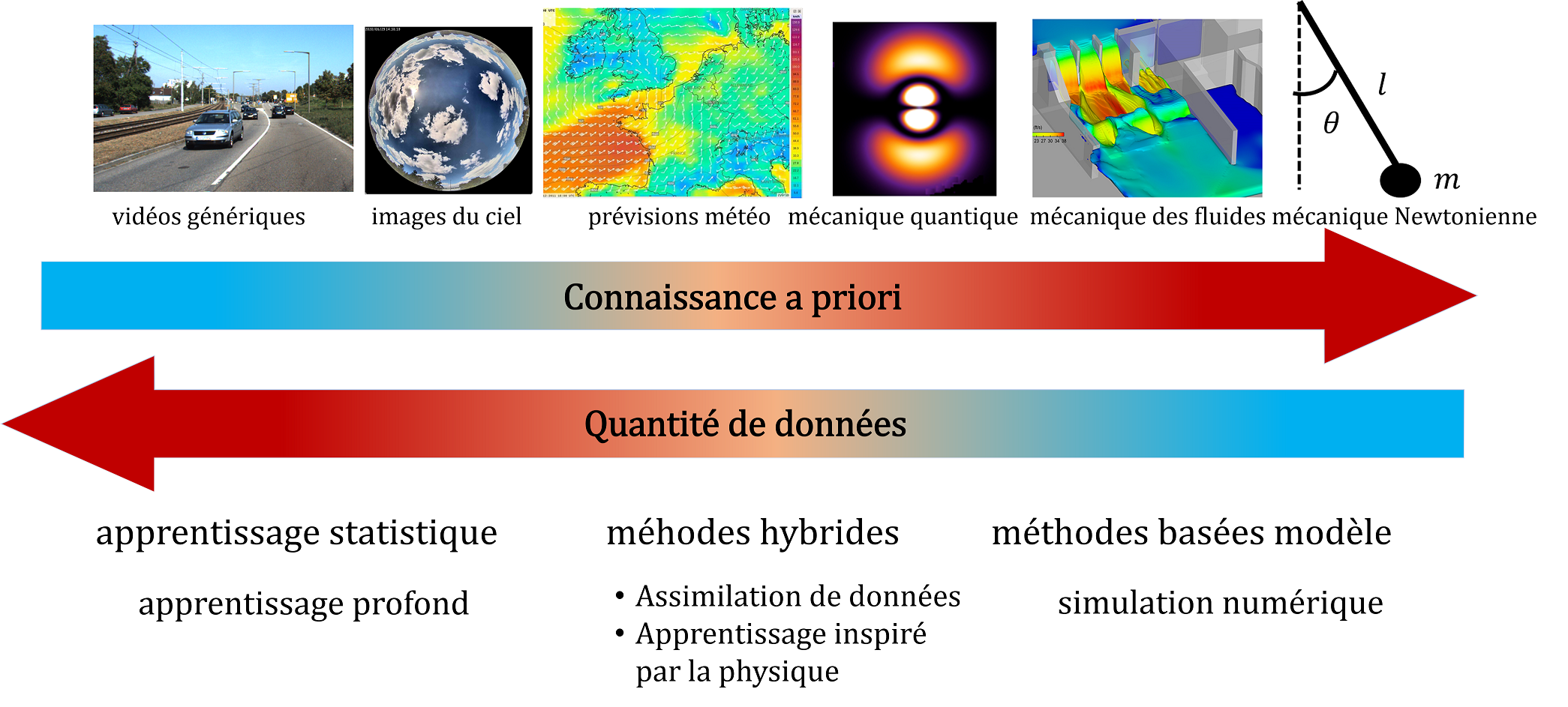}
\caption[Les différents contextes de prédiction.]{\textbf{Les différents contextes de prédiction}.  A gauche, l'apprentissage statistique et profond peuvent extrapoler des systèmes dynamiques sans a priori après apprentissage sur un grand jeu de données. A droite, les méthodes basées modèle supposent une connaissance physique complète du système et prédisent le futur par simulation numérique depuis des conditions aux limites. Entre les deux, les méthodes hybrides utilisant des données et de la connaissance incomplète sont une voie d'exploration très active et prometteuse.}
    \label{fig:physics_data_fr}
\end{figure}

De l'autre côté, les méthodes d'apprentissage machine (\textit{Machine Learning, ML}) sont une alternative agnostique à l'information a priori sur le système. L'apprentissage profond a prouvé sa capacité à apprendre automatiquement des relations complexes à partir de grandes bases de données annotées et est devenu état de l'art pour de nombreuses tâches de prédiction. Toutefois, ces méthodes sont toujours limitées pour modéliser des dynamiques physiques complexes. En outre, elles manquent la plausibilité physique pour interpréter les résultats et extrapoler pour de nouvelles conditions.

Entre les deux, les méthodes hybrides \textit{model-based machine learning} (MB/ML) sont une approche attrayante pour combiner de l'information a priori et des données. Historiquement, les méthodes d'assimilation de données exploitent des données pour corriger les prédictions de modèles physiques en présence d'observations bruitées \cite{bocquet2019data,kalman1960new}. Elles constituent toujours l'état de l'art pour la prévision météorologique.

Revisiter la coopération MB/ML avec l'apprentissage profond moderne est un sujet émergent qui suscite un intérêt majeur pour de nombreuses communautés scientifiques. La physique peut être incorporée dans l'apprentissage de modèles soit sous la forme de contraintes douces dans la fonction de perte \cite{raissi2017physics,sirignano2018dgm}, soit comme des contraintes dures dans les architectures des réseaux \cite{daw2020physics,mohan2020embedding}. Du point de vue apprentissage, ces contraintes physiques permettent de développer des modèles plus interprétables qui se conforment aux lois physiques et qui restent robustes en présence de données bruitées. Cela se traduit typiquement par une plus grande efficacité dans l'utilisation des données et de meilleures performances d'extrapolation au-delà du domaine d'apprentissage. 

Dans cett thèse, nous explorons cette catégorie de méthodes hybrides et nos contributions tâchent de répondre à la question générale suivante:

\begin{center}
    \textit{Comment exploiter de la connaissance physique a priori dans des modèles d'apprentissage statistique?}
\end{center}

Nous nous concentrons sur deux principales directions: incorporation d'information physique a priori dans la fonction d'entraînement des modèles et développement d'architectures augmentées MB/ML dans le cas de connaissance physique incomplète.

\section{Critères différentiable de forme et de temps pour la prédiction déterministe et probabiliste}

Les méthodes traditionnelles de prévision de séries temporelles sont des méthodes statistiques basées modèle qui décrivent des caractéristiques telles que les tendances et la saisonalité. Elles comprennent les méthodes autorégressives comme les modèles ARIMA (\textit{Auto Regressive Integrated Moving Average}) \cite{box2015time}. Ces méthodes font souvent des hypothèses fortes sur les données, par exemple la stationarité, qui ne sont pas vérifiées en pratique.

Avec l'avènement de l'apprentissage profond, les réseaux de neurones profonds sont devenus la méthode état de l'art pour la prédiction de séries temporelles \cite{lai2018modeling,salinas2017deepar,oreshkin2019n,zhou2020informer}, grâce à leur capacité à modéliser des dépendences temporelles complexes à partir d'un corpus d'apprentissage. La plupart des travaux récents se sont concentrés sur l'amélioration des architectures des réseaux de neurones. Le choix de la fonction de perte d'apprentissage, tout aussi important, est quant à lui peu abordé: la plupart des méthodes optimisent l'erreur quadratique moyenne (EQM) ou ses variantes.

L'erreur quadratique moyenne (EQM) est assez peu adaptée pour comparer des séries temporelles à plusieurs pas de temps, comme nous l'illustrons sur la Figure \ref{fig-intro-fr}. L'EQM ne permet pas de modéliser les erreurs de forme ni les décalages temporels entre séries. Pourtant, des critères de forme et de temps sont utilisés dans les applications pour évaluer les prédictions fournies par des algorithmes, par exemple le ramp score \cite{vallance2017towards} pour la forme et le TDI (Temporal Distortion Index) \cite{frias2017assessing} pour le temps. Mais ils ne sont pas utilisés en pratique pour l'entraînement des réseaux de neurones car ils sont la plupart du temps non différentiables.

\begin{figure}
\begin{tabular}{ccc}

\includegraphics[height=4.6cm]{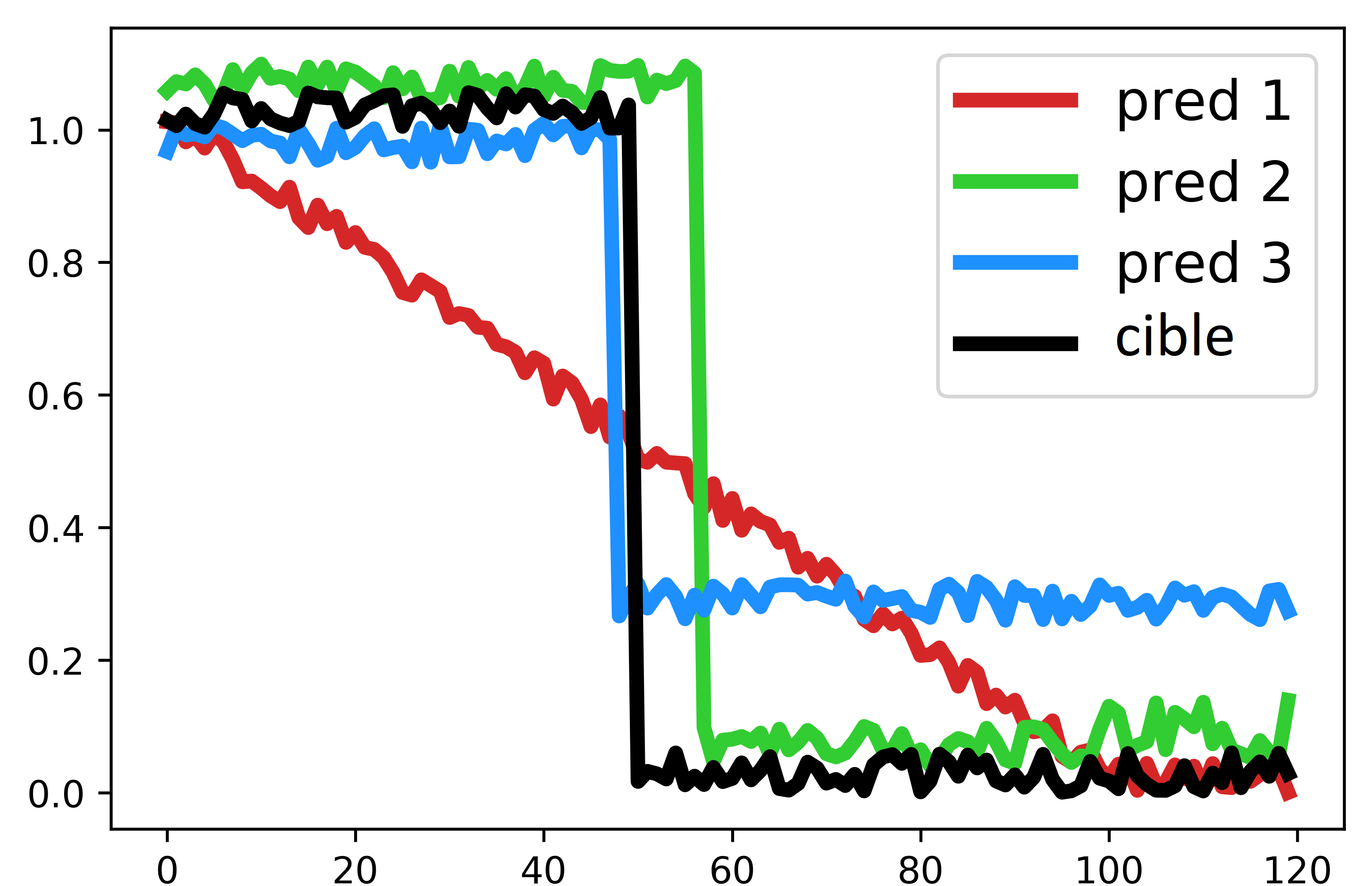}   & \hspace{-0.3cm} 
\includegraphics[height=4.6cm]{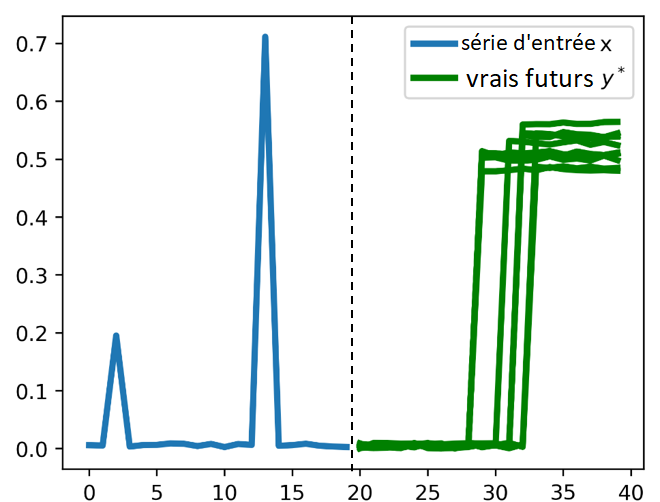}   & 
\hspace{-0.5cm} 
\includegraphics[height=4.6cm]{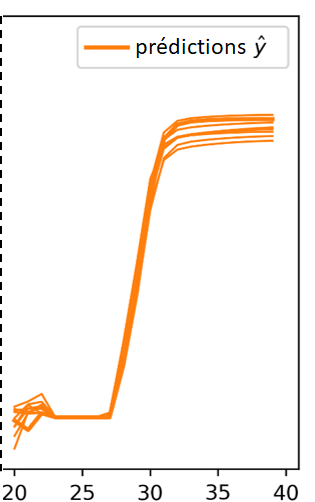}   \\
~ &   \footnotesize{Vraie distribution future}    & \hspace{-0.5cm}  \footnotesize{modèle stoch. \cite{yuan2019diverse}}   \\
 \textbf{(a) Prévision déterministe} & ~ &  \hspace{-5cm} \textbf{(b) Prévision probabiliste}  \\
\end{tabular}{}
    \caption[Limites de l'erreur quadratique moyenne pour la prévision déterministe et probabiliste.]{\textbf{Limites d l'erreur quadratique moyenne pour la prévision déterministe et probabiliste.}
    (a) Pour la prévision déterministe, les trois prédictions (1,2,3) ont la même erreur quadratique moyenne (EQM) par rapport au vrai futur (en noir). Mais on voudrait favoriser la prédiction 2 (bonne forme, léger retard) et 3 (bon positionnement temporel, forme imprécise) sur la prédiction 1 (pas très informative). (b) Pour la prévision probabiliste, les méthodes état de l'art apprises avec l'EQM \cite{yuan2019diverse,rasul2020multi} perdent la capacité à produire des prédictions nettes (en orange) par rapport aux vraies trajectoires futures (en vert).}
    \label{fig-intro-fr}
\end{figure}

Dans cette thèse, nous proposons d'exploiter des critères de forme et de temps pour l'entraînement de réseaux de neurones profonds pour la prédiction de séries temporelles, dans le cas déterministe et probabiliste. Notre objectif est d'aborder des problèmes de prédiction non stationnaires, où les séries temporelles peuvent avoir des variations brutales, comme c'est le cas pour l'irradiance solaire qui chute brutalement lorsqu'un nuage occulte le soleil. Pour cela, nous introduisons des critères différentiables de forme et de temps, que nous formulons à la fois sous la forme de dissimilarités (fonctions de perte) et de similarités (noyaux semi-définis positifs). Les critères de forme sont basés sur une approximation différentiable de l'algorithme du \textit{Dynamic Time Warping (DTW)} \cite{sakoe1990dynamic} et ceux de temps sur le \textit{Temporal Distortion Index} (TDI) \cite{frias2017assessing}.

Nous proposons deux implémentations de ces critères, pour la prévision déterministe et probabiliste.

\subsection{DILATE}

Pour la prévision déterministe de séries temporelles avec des réseaux de neurones profonds, nous introduisons une fonction de perte appelée DILATE (\textit{DIstortion Loss with shApe and TimE}). Conçue comme une alternative à l'EQM, DILATE combine une composante sur la forme des séries temporelles et une composante sur le décalage temporel pour comparer une série prédite $\hat{\y}$ avec le vrai futur $\y^*$:
\begin{align}
\mathcal{L}_{\text{DILATE}}(\hat{\y}, \y^*) &= \alpha~\mathcal{L}_{forme}(\hat{\y}, \y^*) + (1-\alpha)~ \mathcal{L}_{temporelle}(\hat{\y}, \y^*)\\
 &= \alpha ~\text{DTW}^{\mathbf{\Delta}}_{\gamma}(\hat{\y}, \y^*) + (1-\alpha)~ \text{TDI}^{\mathbf{\Delta},\mathbf{\Omega_{dissim}}}_{\gamma}(\hat{\y}, \y^*).
\end{align}

Le principe de DILATE est illustré sur la Figure \ref{fig:dilate_fr}. La perte sur la forme $\mathcal{L}_{forme}$ correspond à la soft-DTW \cite{cuturi2017soft} et la perte temporelle $\mathcal{L}_{temporelle}$ à une relaxation différentiable du TDI \cite{frias2017assessing}. Les deux pertes sont combinées linéairement avec un facteur $\alpha \in [0;1]$ qui est un hyperparamètre de la méthode.

\begin{figure}
    \centering
    \includegraphics[width=\linewidth]{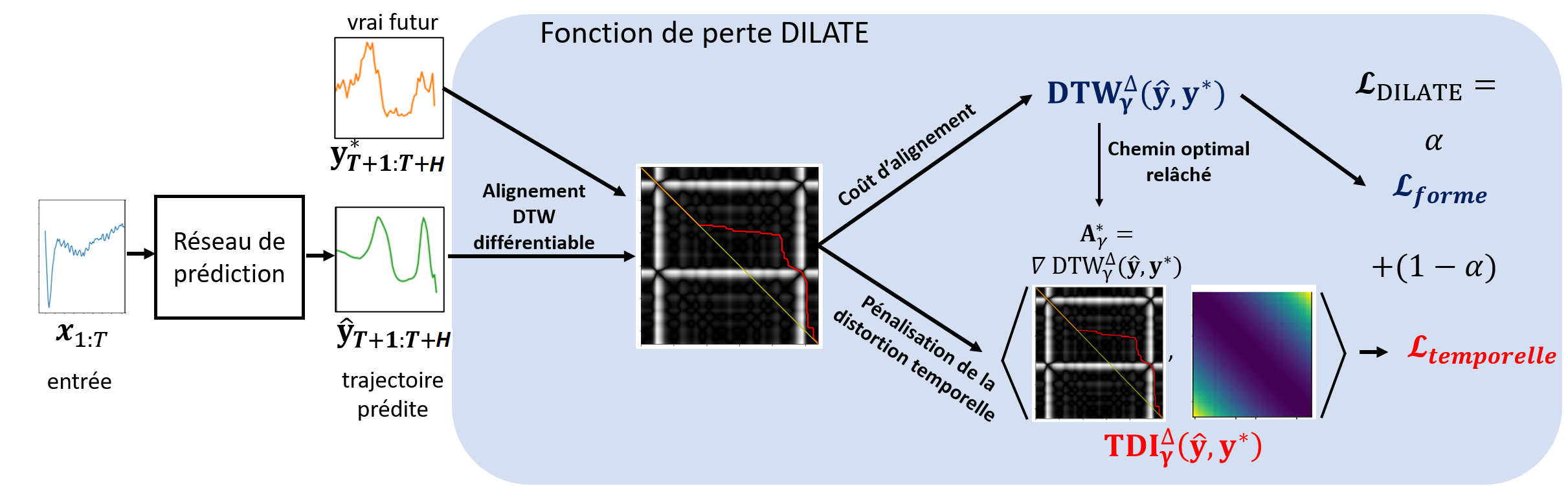}
    \caption{Fonction de perte DILATE pour l'entraînement de réseaux de neurones profonds pour la prédiction déterministe de séries temporelles.}
    \label{fig:dilate_fr}
\end{figure}

Nous conduisons des expériences sur plusieurs jeux de données synthétiques et réels pour évaluer les performances de la perte DILATE. Les résultats révèlent que l'entraînement avec DILATE améliore significativement les performances évaluées sur des critères de forme et de temps, tout en maintenant des performances équivalentes évaluées en EQM. DILATE est agnostique à l'architecture du réseau de neurones et fonctionne aussi bien avec des architecture standard que les dernières architectures état de l'art.

\subsection{STRIPE}

La prévision probabiliste consiste à décrire la loi de probabilité conditionnelle des trajectoires futures sachant une trajectoire d'entrée. Dans cette thèse, notre objectif est de décrire cette loi de probabilité par un petit ensemble (par exemple 10) de trajectoires futures possibles qui représentent bien la variabilité sur l'évolution du phénomène. Ces scénarios doivent être à la fois précis et divers selon des critères de forme et de temps, ce que ne permettent pas les méthodes actuellement état de l'art en prévision probabiliste \cite{salinas2017deepar,rangapuram2018deep}.

Pour cela, nous introduisons un modèle appelé STRIPE (\textit{Shape and Time diverRsIty in Probabilistic forEcasting}). Illustré sur la Figure \ref{fig:stripe-fr}, le modèle STRIPE est une architecture de type encodeur-décodeur qui permet de générer des trajectoires futures à plusieurs pas de temps. Il s'agit d'un modèle génératif où les différents futurs possibles sont générés à partir de l'échantillonnage de variables latentes. Plus précisément, le modèle STRIPE est composé d'un encodeur qui prend la série temporelle d'entrée $\x_{1:T}$ et produit une variable descriptive $h$. On adjoint à cette variable $h$ des variables latentes $z_s$ et $z_t$ qui capturent la variabilité en forme (respectivement en temps). Le décodeur prend en entrée la concaténation $(h,z_s,z_t)$ et produit une trajectoire future $\hat{\y}_{T+1:T+\tau}$.

Pour structurer la diversité des trajectoires prédites, les variables latentes sont générées par des réseaux de neurone appelés STRIPE-forme et STRIPE-temps. La diversité est favorisée par l'ajout d'une fonction de perte de diversité $\mathcal{L}_{diversité}$. Elle est basée sur l'utilisation des processus ponctuels déterminantaux (DPP) \cite{kulesza2012determinantal}, qui sont un outil mathématique élégant pour décrire la diversité d'un ensemble d'éléments. La perte de qualité $\mathcal{L}_{qualité}$ est la perte DILATE pour assurer des prédictions avec à la fois la bonne forme et un faible décalage temporel. Pour assurer le maintien de la qualité des prédictions lors de l'étape de diversification, un réseau postérieur permet d'échantillonner les variables latentes lors de l'entraînement, pour qu'elles correspondent à de réelles trajectoires du jeu de données.

Nous menons des expériences sur un jeu de données synthétique où l'on dispose de l'ensemble des futures trajectoires comme supervision, ainsi que sur des jeux de données réels où l'on a qu'un seul futur disponible. Les résultats montrent que STRIPE parvient à des prédictions avec une bien meilleure diversité mesurée avec des critères de forme et de temps que des mécanismes de diversification concurrents de la littérature \cite{dieng2019prescribed,thiede2019analyzing,elfeki2018gdpp,yuan2019diverse} et que des algorithmes dédiés à la prédiction probabiliste \cite{salinas2017deepar}. De plus, STRIPE maintient une bonne qualité des prédictions obtenues et obtient le meilleur compromis entre qualité et diversité.

\begin{figure}[H]
    \centering
    \includegraphics[width=\linewidth]{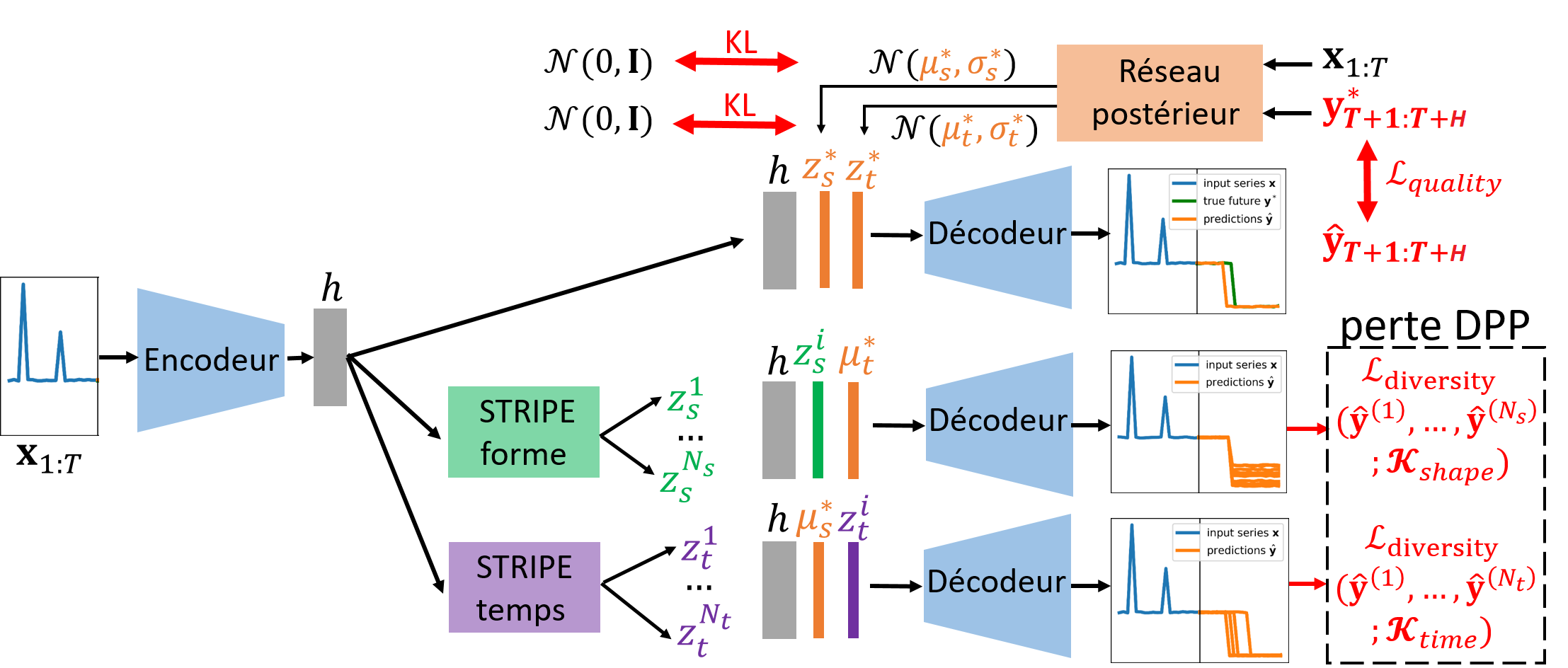}
    \caption[Modèle STRIPE pour la prévision probabiliste.]{Modèle STRIPE pour la prévision probabiliste.}
    \label{fig:stripe-fr}
\end{figure}

\section{Prédiction avec incorporation d'information physique incomplète}

Dans cette partie de la thèse, nous explorons comment incorporer de l'information physique a priori dans les modèles d'apprentissage statistique. En particulier, nous nous intéressons au cas où la connaissance physique est incomplète, ce qui est une question très peu traitée dans la littérature.

\subsection{Modèle PhyDNet pour la prédiction de vidéo}

Nous proposons un modèle d'apprentissage profond dédié à la prédiction de vidéos, dénommé PhyDNet, qui incorpore de l'information physique sous la forme d'une classe d'équations aux dérivées partielles (EDP) linéaires. Toutefois, pour des vidéos génériques, les équations physiques de la dynamique ne s'appliquent pas directement au niveau des pixels. Par exemple, il est nécessaire au préalable de segmenter les objets et de déterminer leur centre de masse avant d'appliquer les lois de Newton. C'est un cas représentatif d'un a priori non observable dans l'espace d'entrée. 

Pour traiter ce problème, nous supposons qu'il existe un espace latent dans lequel le modèle dynamique d'EDP linéaire s'applique. Le modèle PhyDNet est composé d'un encodeur-décodeur pour apprendre automatiquement l'espace latent le plus adapté à partir des données. Dans cet espace latent, nous décomposons la dynamique en deux parties: une partie qui intégre les lois a priori de la physique et une partie qui apprend l'information complémentaire à la physique nécessaire pour avoir une bonne prédiction au niveau des pixels.

Le modèle PhyDNet est un réseau de neurones récurrent, illustré sur la Figure \ref{fig:phydnet_fr} dans sa version pliée (à gauche) et dépliée (à droite). Pour modéliser la partie physique, nous introduisons une cellule récurrente appelée PhyCell qui discrétise une équation aux dérivées partielle linéaire par un schéma d'Euler, pour laquelle les dérivées partielles sont calculées avec des convolutions contraintes \cite{long2018pde}. La deuxième branche modélise le résidu qui n'est pas expliqué par la physique; pour cela nous utilisons un réseau de neurones récurrent assez générique, en l'occurence un ConvLSTM \cite{xingjian2015convolutional}.  Les deux branches sont sommées dans l'espace latent, avant d'être décodées vers une prédiction de l'image future.

\begin{empheq}[left=\empheqlbrace]{alignat=2}
&   \tilde{\mathbf{h}}_{t+1} \!= \mathbf{h}_{t} +  \Phi(\mathbf{h}_{t})   &  \!\!\!\quad \text{\small{\textbf{Prediction}\!}} \label{eq:prediction}\\
&   \mathbf{h}_{t+1} \!= \tilde{\mathbf{h}}_{t+1}  + \mathbf{K}_t \odot \left( \mathbf{E}(\mathbf{u}_t) - \tilde{\mathbf{h}}_{t+1} \right). & \!\!\! \quad \text{\small{\textbf{Correction}\!}} \label{eq:correction}
\end{empheq}

\begin{figure}[H]
    \centering
    \includegraphics[width=\linewidth]{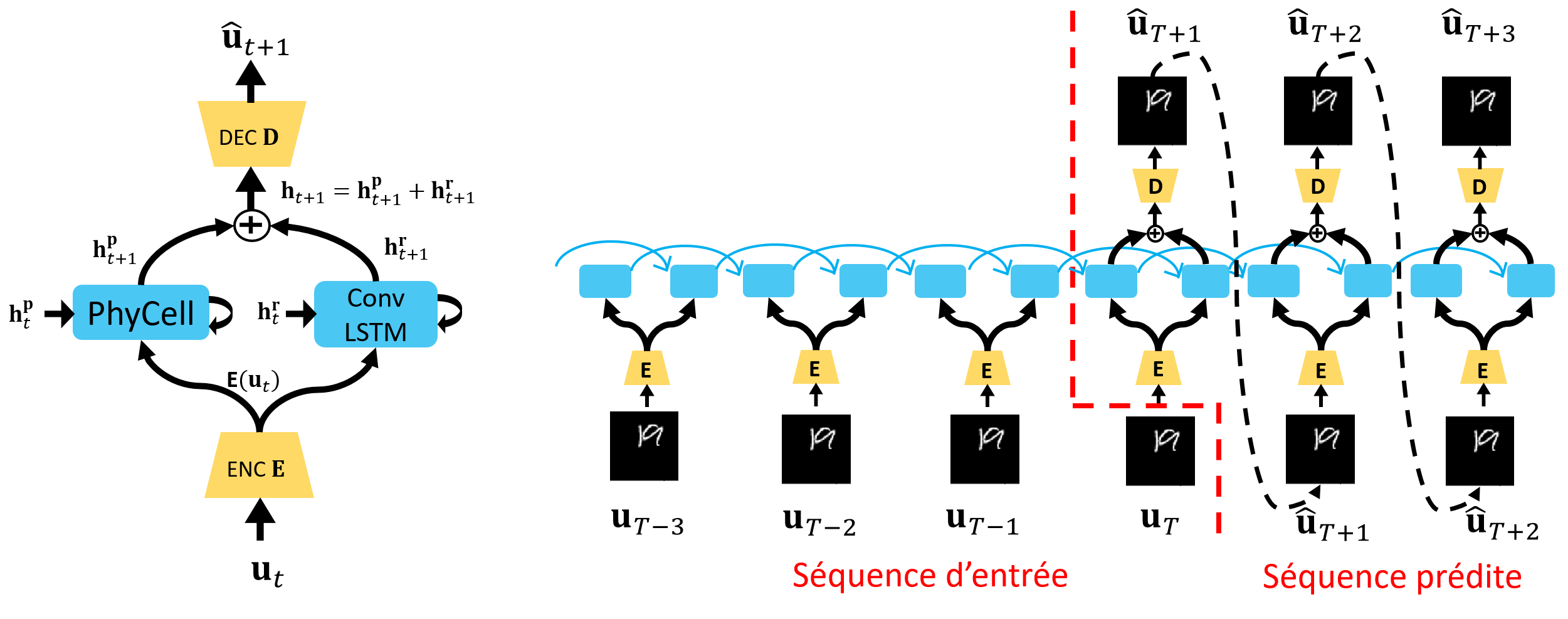}
    \caption[Modèle PhyDNet pour la prévision de vidéo.]{Modèle PhyDNet pour la prédiction de vidéo.}
    \label{fig:phydnet_fr}
\end{figure}

Nous menons des expériences sur des jeux de données avec différents niveaux de connaissance a priori: depuis Moving MNist où la dynamique de déplacement des chiffres est parfaitement connue, jusqu'à des vidéos généralistes de mouvements humains, en passant par des cas où l'on a un a priori physique incomplet sur la dynamique, comme pour le traffic routier ou la température de surface des océans. Dans tous ces cas, nous montrons la supériorité de PhyDNet par rapport à des modèles d'apprentissage profond sans a priori physique.  

\subsection{Modèle APHYNITY pour la coopération optimale entre physique et apprentissage profond}

La prédiction de systèmes dynamiques pour lesquels on a une connaissance partielle de leur dynamique est un problème très courant dans de nombreux champs scientifiques. Par exemple pour la modélisation climatique, il est très compliqué de mettre en équations précisément tous les phénomènes complexes régissant la dynamique de l'atmosphère.

Nous introduisons ici un schéma d'apprentissage, appelé APHYNITY, pour augmenter des modèles physiques simplifiés décrits par des équations aux dérivées partielles, avec des réseaux de neurones profonds. Nous considérons des systèmes dynamiques sous la forme de l'équation différentielle:
\begin{equation}
    \frac{\diff  X_t}{\diff t} = F(X_t).
\end{equation}

Le modèle APHYNITY décompose la fonction de dynamique $F$ en une composante $F_p$ pour laquelle nous avons un a priori physique et une composante d'augmentation $F_a$ qui corrige les erreurs du modèle physique: $F = F_p + F_a$.

Le problème d'apprentissage est formulé de manière à ce que le modèle physique explique la dynamique le plus possible, tandis que le modèle d'augmentation ne capture que l'information qui ne peut pas être capturée par la physique. Inspiré par le principe de moindre action, ce schéma d'apprentissage consiste à minimiser la norme du résidu $F_a$ sous la contrainte de prédiction parfaite du modèle augmenté:
\begin{equation}
\label{eq:aphynity-opt-fr}
\underset{F_p\in\F_p, F_a\in\F}{\min} ~~~\left\Vert  F_a  \right\Vert ~~~ 
\mathrm{subject~to} ~~~~ \forall X\in\D, \forall t, \frac{\diff X_t}{\diff t} =(F_p+F_a)(X_t).
\end{equation}

Sous de faibles hypothèses qui sont vérifiées dans de nombreux cas expérimentaux, il y a existence et unicité du problème d'optimisation APHYNITY, ce qui favorise l'interprétabilité et la généralisation du modèle.

\begin{figure}[H]
    \centering
    \includegraphics[width=\linewidth]{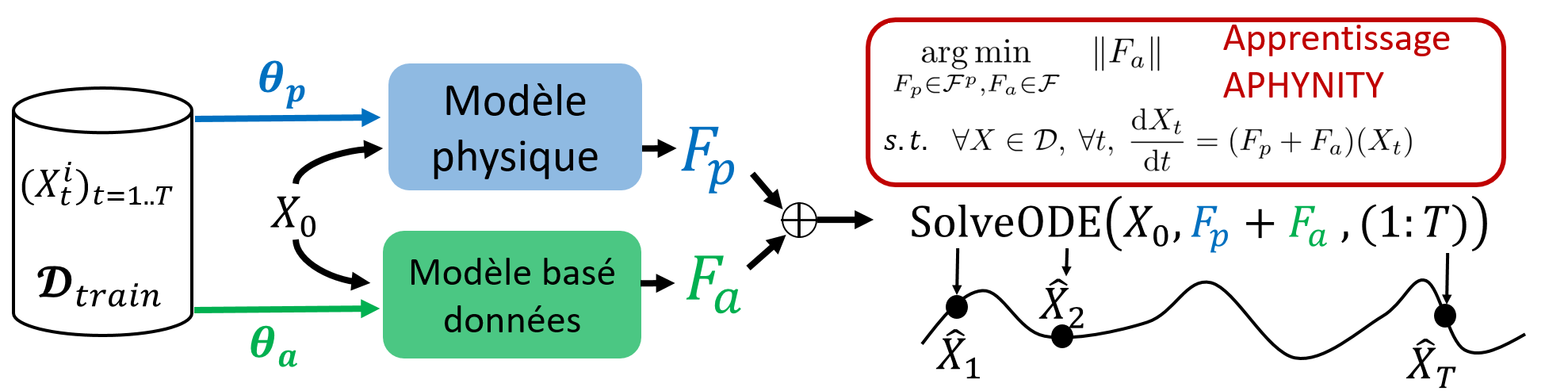}
    \caption[Schéma d'apprentissage APHYNITY.]{Schéma  d'apprentissage APHYNITY pour la coopération optimale entre modèles physiques et modèles d'apprentissage.}
    \label{fig:aphynity_fr}
\end{figure}

Nous proposons une approche trajectoire pour implémenter en pratique le schéma APHYNITY, qui est illustré sur la Figure  \ref{fig:aphynity_fr}. A partir d'une condition initiale $X_0$, un modèle physique paramétré par $\theta_p$ donne la dynamique physique $F_p$, tandis que le modèle d'augmentation basé données paramétrisé par $\theta_a$ fournit la dynamique $F_a$. La dynamique résultante $F=F_p+F_a$ est intégrée dans le temps par un schéma numérique différentiable qui donne les prédictions pour un ensemble de pas de temps futurs. Les paramètres du modèle sont appris par l'optimisation du problème sous contraintes APHYNITY (Eq \ref{eq:aphynity-opt-fr}). Un algorithme d'optimisation sous contraintes adaptatif est utilisé pour résoudre efficacement le problème de l'Eq \ref{eq:aphynity-opt-fr}.

Nous menons des expériences sur trois problèmes representatifs de classes de phénomènes physiques: dynamique Newtonienne (pendule amorti), équations de réaction-diffusion et équations d'ondes. Dans chaque cas, nous considérons des modèles physiques simplifiés (par exemple les équations du pendule sans le terme d'amortissement) et augmentons ces modèles avec le schéma APHYNITY.

Les résultats expérimentaux montrent la supériorité d'APHYNITY sur des modèles basés données uniquement, sur des modèles physiques incomplets et sur des méthodes état de l'art qui combinent données et connaissances. Le gain de performances se voit à la fois sur l'erreur de prédiction et sur l'erreur d'identification des paramètres physiques du modèle. De plus, l'approche APHYNITY est suffisamment flexible pour s'adapter à des niveaux différents de connaissance physique a priori.

\section{Application à la prédiction d'irradiance solaire}

Les énergies renouvelables sont en forte progression dans le monde ces dernières années. Toutefois, leur variabilité spatiale et temporelle reste un défi pour leur intégration à grande échelle dans les réseaux électriques existants, pour lesquels l'équilibre à tout instant entre production et consommation d'électricité est primordial. L'enjeu réside également dans le pilotage indépendant de parcs photovoltaïques ou éoliens qui peuvent être couplés à des moyens de stockage ou de production supplémentaires, notamment dans les systèmes insulaires isolés.

Dans ce contexte, EDF a engagé depuis plusieurs années des travaux sur la prévision de production photovoltaïque, à différents horizons temporels et à l'aide de différentes données d'entrée (modèles météorologiques, images satellites, images au sol, mesures en temps réel). L'amélioration des méthodes de prévision à court terme (de quelques minutes à une heure) est aujourd'hui un enjeu fondamental. La variabilité temporelle à court-terme de la production photovoltaïque est principalement liée à des phénomènes physiques météorologiques, tels que le déplacement des nuages. Les modèles météorologiques et les images satellite ont une résolution spatiale et temporelle insuffisante pour prédire le déplacement des nuages à court-terme au-dessus d'un site de production. Pour cela, l'utilisation de caméras au sol hémisphériques est une piste très prometteuse pour suivre les nuages et anticiper les variations brusques de production à quelques minutes  \cite{gauchet2012surface,chu2013hybrid,chu2016sun,marquez2013intra,schmidt2016evaluating}. EDF dispose de plusieurs sites instrumentés de caméras hémisphériques fisheye et de capteurs de rayonnement solaire (pyranomètres), constituant ainsi une base de données annotées de plusieurs millions d'images du ciel au pas de temps 10s (Figure \ref{fig:fisheye-camera_fr}).

\begin{figure}
    \centering
    \includegraphics[width=14cm]{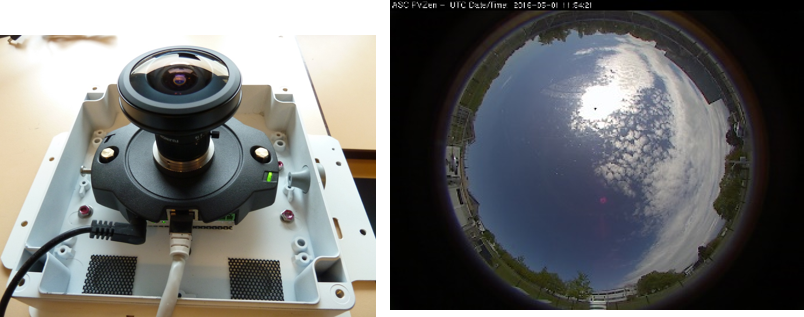}
    \caption{Caméra fisheye et exemple d'image fisheye utilisées pour la prévision à court-terme de l'irradiance solaire.}
    \label{fig:fisheye-camera_fr}
\end{figure}

Les méthodes traditionnelles de prévision par images fisheye reposent sur du traitement d'images classique. La chaîne de traitement typique \cite{gauchet2012surface,chu2013hybrid,chu2016sun,schmidt2016evaluating} se compose des étapes suivantes: calibration de la caméra fisheye, prétraitement de l'image, segmentation de l'image avec des seuillages, calcul du flot optique et propagation du mouvement pour prévoir la future position des nuages et enfin calcul de l'irradiance future avec des algorithmes de régression.

Depuis quelques années, les méthodes d'apprentissage profond se sont révélées être une alternative intéressante pour estimer et prévoir le rayonnement solaire de bout en bout \cite{pothineni2018kloudnet,zhang2018deep,spiess2019learning,sun2019short,nie2020pv,paletta2020temporally,zhen2021ultra}, sans la nécessité de definir des indicateurs  sur les images manuellement. Au début de cette thèse, nous avons exploré de premières architectures de réseaux de neurones profonds pour l'estimation et la prévision du rayonnement \cite{leguen-gretsi}. Pour l'estimation du rayonnement correspondant à l'image courante, nous avons remarqué un gain de performances très important en utilisant des réseaux convolutionnels par rapport aux méthodes traditionelles, ce qui était attendu sachant les succès de l'apprentissage profond pour les tâches de perception. Par contre, la prévision du rayonnement est une tâche beaucoup plus compliquée: notre architecture préliminaire basée sur un ConvLSTM donne de meilleurs résultats que la méthode traditionnelle mais avec une marge plus faible. 

Pour améliorer les prédictions, nous avons appliqué les contributions méthodologiques de cette thèse à ce problème. Nous avons adapté le modèle PhyDNet de prédiction de vidéo à la prédiction jointe des images fisheye et des rayonnements futurs. Illustrée sur la Figure \ref{fig:phydnet_fisheye_fr}, cette architecture prend en entrée une séquence d'images fisheye qui est traitée par le réseau de neurones récurrent PhyDNet. Le réseau est ensuite appliqué récursivement pour décoder les images futures et les rayonnements futurs. 


\begin{figure}[H]
    \centering
    \includegraphics[width=\linewidth]{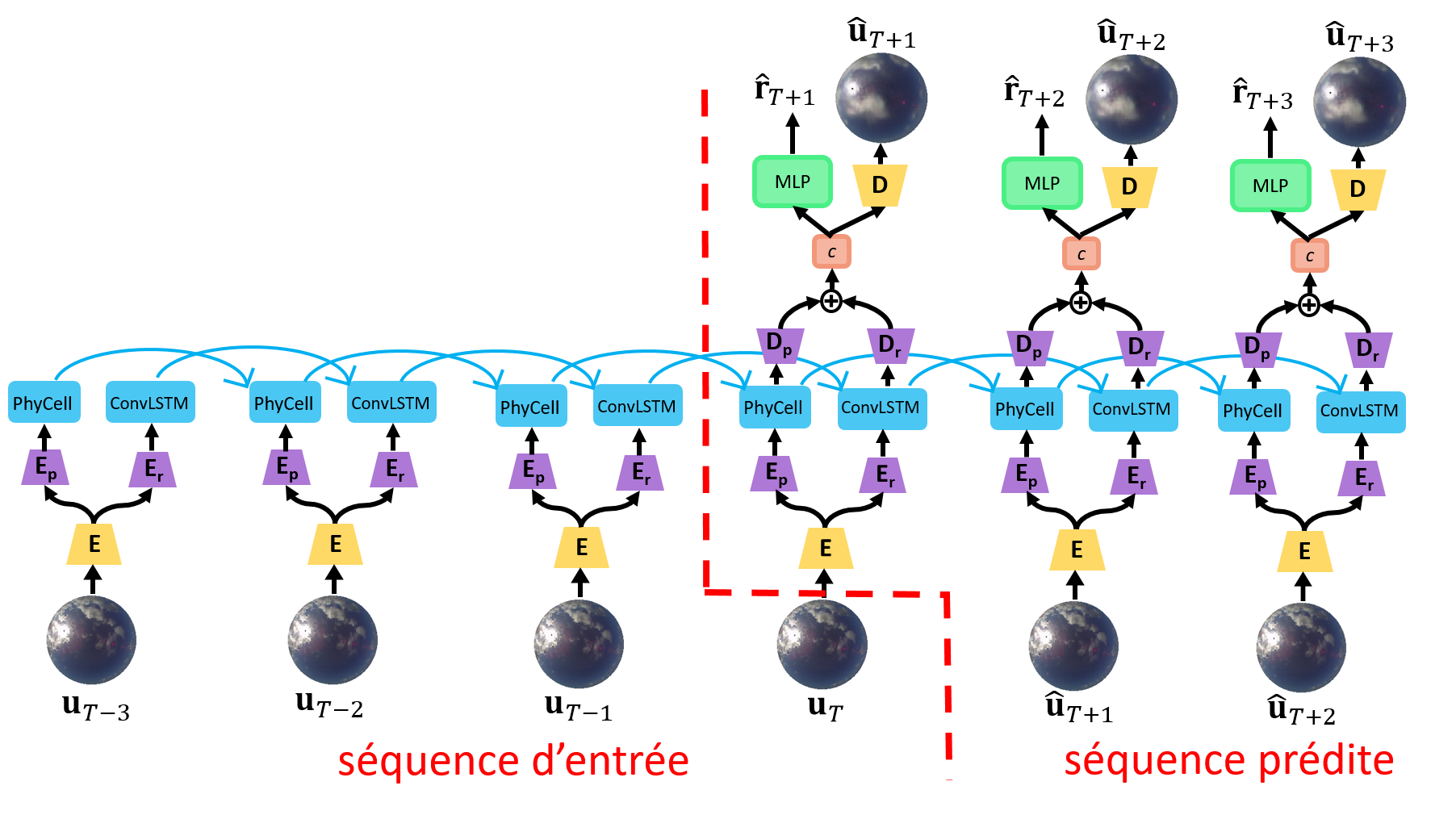}
    \caption{Modèle PhyDNet adapté pour la prévision de l'irradiance solaire.}
    \label{fig:phydnet_fisheye_fr}
\end{figure}

Le modèle PhyDNet a permis un gain de performances important sur les prévisions de l'irradiance solaire à 5min par rapport à notre modèle de base ConvLSTM.

\begin{figure}
    \centering
    \includegraphics[width=15cm]{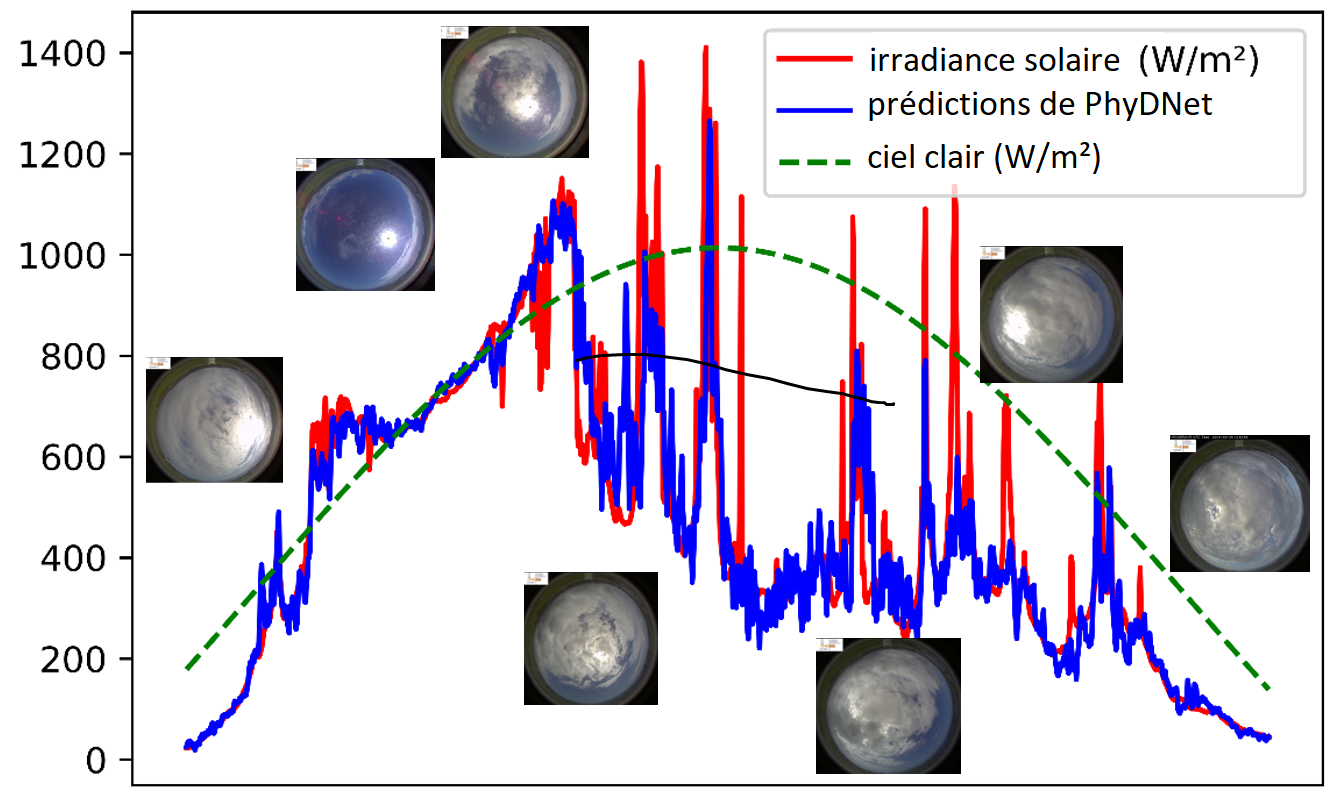}
    \caption[Prévision de l'irradiance solaire à court-terme avec images fisheye.]{Prévisions de l'irradiance à 5min avec des images fisheye. Notre modèle inspiré par la physique prédit correctement les variations brusques de l'irradiance solaire.}
    \label{fig:fisheye-qualitative-fr}
\end{figure}

Nous avons également exploré l'application de la fonction de perte DILATE et du schéma d'apprentissage APHYNITY à ce problème. Ces deux mécanismes permettent d'obtenir un nouveau gain de performances, quoique plus faible que celui apporté par l'architecture inspirée par la physique PhyDNet. Nous en avons analysé les raisons et proposé des pistes d'améliorations futures.

\section{Conclusion et perspectives}

Dans cette thèse, nous avons exploré de manière générale comment incorporer de la connaissance physique a priori dans les modèle d'apprentissage statistique pour améliorer la prévision spatio-temporelle. Plus particulièrement, nous avons abordé deux principales directions de recherche.

 La première concerne le choix de la fonction de perte pour entraîner les modèles. Au lieu de l’erreur quadratique moyenne très majoritairement utilisée, nous proposons d’utiliser des critères de forme et de décalage temporel sur les trajectoires prédites. Nous nous attaquons au contexte de la prévision déterministe avec notre proposition de fonction de perte DILATE, et au contexte probabiliste, où notre objectif est de décrire la distribution prédictive par un faible nombre de scénarios divers et précis, avec notre modèle STRIPE. 
 
 Notre seconde direction de recherche est d’augmenter des modèles physiques incomplets avec des réseaux de neurones profonds basés données. Pour la prédiction de vidéo, nous introduisons le modèle PhyDNet qui sépare une partie de dynamique physique modélisée par des équations aux dérivées partielles, d’une partie résiduelle qui capture l’information complémentaire, comme la texture et les détails, nécessaire à la bonne prédiction. Nous proposons aussi un schéma d’apprentissage, appelé APHYNITY, qui assure une décomposition bien posée et unique entre des modèle physiques incomplets et des réseaux de neurones profonds, sous de faibles hypothèses. 
 
 Nous avons validons les contributions de cette thèse sur de nombreux jeux de données synthétiques et réels, et sur l’application de prévision photovoltaïque à EDF.
 
 Les travaux de cette thèse ouvrent de nombreuses perspectives intéressantes à explorer. A court-terme, les perspectives pour l'améliorations des prédictions d'irradiance comprennent l'utilisation de modèle physiques plus spécifiques à la dynamique de l'atmosphère, l'apprentissage sur des séquences temporelles de plus longue durée, ou encore l'utilisation de réseaux de neurones qui encodent l'invariance par rotation pour le traitement des images fisheye.
 
 A plus long terme, l'étude des modèles physiques augmentés et leur application pour résoudre des problèmes naturels complexes comme la prévision climatique est particulièrement attrayante. Plusieurs applications pourraient directement bénéficier de ces travaux, par exemple l'estimation du flot optique qui est traditionnellement basée sur l'hypothèse simplifiée de la conservation de l'intensité lumineuse, ou l'apprentissage  par renforcement basé modèle qui suppose un modèle de dynamique (souvent simplifié) pour prendre des décisions. 
 
 Par ailleurs, nous avons étudié dans cette thèse des décompositions linéaires entre modèles physiques simplifiés et leur augmentations, ce qui est une hypothèse assez forte. D'autres schémas de décompositions peuvent être envisagés, par exemple entre des modélisations physiques à des échelles spatiales différentes.

Mots-clés : apprentissage profond, prévision spatio-temporelle, prévision photovoltaïque.

\end{vcenterpage}

\clearpage{\pagestyle{empty}\cleardoublepage}

\newpage
\appendix
\addcontentsline{toc}{chapter}{Liste des annexes}

\chapter{Appendix for differentiable shape and temporal criteria  for non-stationary forecasting}
\markright{\MakeUppercase{Annexe A}}

\section{Proof that the temporal kernel is PSD}
\label{app:proof-ktime}

The DTW score between two time series $\y \in \mathbb{R}^{d \times n}$ and $\z \in \mathbb{R}^{d \times m}$ can be written $S(\pi) = \sum_{i=1}^{|\pi|} \mathbf{\Delta}(\y_{\pi_1(i)}, \z_{\pi_2(i)})$ where $\pi=(\pi_1,\pi_2)$ is a valid alignment between both series. Equivalently we can write the DTW score $S(\pi) = S(\mathbf{A}) = \left\langle \mathbf{A}, \mathbf{\Delta(\y,\z)} \right\rangle$, where $\mathbf{A} \subset \left \{  0,1 \right \}  ^{n \times m}$ is the warping path in matrix form ($\mathbf{A}_{ij}=1$ if $\y_i$ is associated to $\z_j$ and 0 otherwise).\\

Let $w: \mathcal{A}_{n,m} \longrightarrow \mathbb{R}_+^*$ be a strictly positive weighting function on alignment paths and let's consider the following kernel:
\begin{align}
\mathcal{K}_w(\y,\z) &=    \sum_{\mathbf{A} \in \mathcal{A}_{n,m}} w(\mathbf{A}) ~~ e^{ -  \frac{S(\mathbf{A})}{\gamma} }  \\
&= \sum_{\mathbf{A} \in \mathcal{A}_{n,m}} w(\mathbf{A}) ~~ e^{ -  \frac{\left\langle \mathbf{A} , \mathbf{\Delta(\y,\z)} \right\rangle}{\gamma} } \\
 &=  \sum_{\pi \in \mathcal{A}_{n,m}} w(\pi)  ~~ e^{ -  \frac{ \sum_{j=1}^{|\pi|} \mathbf{\Delta} \left(  \y_{\pi_1(j)} , \z_{\pi_2(j)} \right) }{\gamma} } \\ 
&=  \sum_{\pi \in \mathcal{A}_{n,m}} w(\pi) \prod_{j=1}^{|\pi|} e^{ - \frac{ \mathbf{\Delta} \left(  \y_{\pi_1(j)} , \z_{\pi_2(j)} \right) }{\gamma} } \\
 &=  \sum_{\pi \in \mathcal{A}_{n,m}} w(\pi) \prod_{j=1}^{|\pi|} k(\y_{\pi_1(j)} , \z_{\pi_2(j)}) ,
\label{eq:kernel_def}
\end{align}
where we denote $k=e^{-\frac{\mathbf{\Delta}}{\gamma}}$. We prove the following result: \\

\begin{prop}
If $k$ is a PSD kernel such that $\frac{k}{1+k}$ is also PSD, the kernel $\mathcal{K}_w$ defined in Eq. \ref{eq:kernel_def} is also PSD.
\end{prop}

\begin{proof}
The proof is adapted from \cite{cuturi2007kernel}. First, for any time series $\y= (\y_1,\dots,\y_n) \in \mathbb{R}^{d \times n}$ of length $n$ and for any sequence $a \in \mathbb{N}^n$, we introduce the notation:
\begin{equation}
    \y_a = (\underset{a_1 \text{~times}}{\underbrace{\y_1,\dots,\y_1}}, \dots,  \underset{a_n \text{~times}}{\underbrace{\y_n,\dots,\y_n}}).
\end{equation}

Let $\chi$ be any PSD kernel defined on $\mathbb{R}^d$ with the following condition $|\chi| < 1$, we introduce the kernel $\kappa$ defined as:
\begin{equation}
    \kappa(\y,\z) = 
    \begin{cases}
    \prod_{i=1}^{|x|} \chi(\y_i, \z_j)  \text{~~if~~} |\y| = |\z| \\
    0 \text{~~~otherwise.}
    \end{cases}
\end{equation}

Then, given a strictly positive weighting function $w(a,b) > 0$, the following kernel $\mathcal{K}_w$ defined in Eq. \ref{eq:Kw}  is PSD by construction:
\begin{equation}
    \mathcal{K}_w(\y,\z) = \sum_{a \in \mathbb{N}^n} \sum_{b \in \mathbb{N}^m} w(a,b) ~ \kappa(\y_a, \z_b).
    \label{eq:Kw}
\end{equation}
where we recall that $n=|\y|$ and $m=|\z|$. We denote $\epsilon_a = (\underset{a_1 \text{~times}}{\underbrace{1,\dots,1}}, \dots,  \underset{a_p \text{~times}}{\underbrace{p,\dots,p}})$ for any $a\in \mathbb{N}^p$. We also write for any sequences $u$ and $v$ of common length $p$: $u \otimes v = ((u_1,v_1),\dots,(u_p,v_p))$. With these notations, we can rewrite $\mathcal{K}_w$ as:
\begin{equation}
    \mathcal{K}_w(\y,\z) = \sum_{ \overset{a \in \mathbb{N}^n, b \in \mathbb{N}^m}{\Vert a \Vert = \Vert b \Vert} } w(a,b)  \prod_{i=1}^{\Vert  a \Vert} \chi((\y,\z)_{\epsilon_a \otimes \epsilon_b(i)}).
    \label{eq:Kw_ab}
\end{equation}

Notice now for each couple $(a,b)$ there exists a unique alignment path $\pi$ and an integral vector $v$ verifying $\pi_v = \epsilon_a \otimes \epsilon_b$. Conversely, for each couple $(\pi,v)$ there exists a unique pair $(a,b)$ verifying  $\pi_v = \epsilon_a \otimes \epsilon_b$. Therefore the kernel $\mathcal{K}_w$ in Eq. \ref{eq:Kw_ab} can be written equivalently with a parameterization on $(\pi,v)$ for $w$:
\begin{equation}
    \mathcal{K}_w(\y,\z) = \sum_{\pi \in \mathcal{A}_{n,m}} \sum_{v \in \mathbb{N}^{|\pi|}} w(\pi,v) \prod_{j=1}^{|\pi|} \chi((\y,\z)_{\pi_v(j)}),
\end{equation}
\label{eq:Kw_piv}
where $\chi_{\pi(j)}$ is a shortcut for $\chi(\y_{\pi_1(j)}, \z_{\pi_2(j)})$.\\

Now we assume that the weighting function $w$ depends only on $\pi$: $w(\pi,v)=w(\pi)$. Then we have:
\begin{align*}
\mathcal{K}_w(\y,\z) &=  \sum_{\pi \in \mathcal{A}_{n,m}} w(\pi) \sum_{v \in \mathbb{N}^{|\pi|}} \prod_{j=1}^{|\pi|} \chi^{v_j}_{\pi(j)} \\
 &= \sum_{\pi \in \mathcal{A}_{n,m}} w(\pi) \prod_{j=1}^{|\pi|} \left( \chi_{\pi(j)} + \chi_{\pi(j)}^2  + \chi_{\pi(j)}^3 + \dots   \right)\\
  &=  \sum_{\pi \in \mathcal{A}_{n,m}} w(\pi) \prod_{j=1}^{|\pi|} \frac{\chi_{\pi(j)}}{1-\chi_{\pi(j)}}.
\end{align*}
By setting now $\chi = \frac{k}{1+k}$ which is PSD by hypothesis and verifies $| \chi | <1$ (recall that $k=e^{- \frac{\mathbf{\Delta}}{\gamma}} $), we get:

\begin{align*}
\mathcal{K}_w(\y,\z) &=  \sum_{\pi \in \mathcal{A}_{n,m}} w(\pi) \prod_{j=1}^{|\pi|} k_{\pi(j)} \\
 &=  \sum_{\pi \in \mathcal{A}_{n,m}} w(\pi) \prod_{j=1}^{|\pi|} k(\y_{\pi_1(j)} , \z_{\pi_2(j)}) ,\\
\end{align*}
which corresponds exactly to the kernel $\mathcal{K}_w$ defined in Eq. \ref{eq:kernel_def}. This proves that $\mathcal{K}_w$ in Eq. \ref{eq:kernel_def} is a well defined PSD kernel. \\

With the particular choice $w(\mathbf{A}) = \left\langle \mathbf{A},\mathbf{\Omega_{sim}} \right\rangle$, we recover: 
\begin{align*}
\mathcal{K}_w(\y,\z) &=   \sum_{\mathbf{A} \in \mathcal{A}} \left\langle \mathbf{A},\mathbf{\Omega_{sim}} \right\rangle  ~~ e^{ -  \frac{\left\langle \mathbf{A} , \mathbf{\Delta(\y,\z)} \right\rangle}{\gamma} } \\
&= Z  \times \text{TDI}^{\mathbf{\Delta,\Omega_{sim}}}_{\gamma}(\y,\z) \\
&=  e^{- \text{DTW}^{\mathbf{\Delta}}_{\gamma}(\y,\z) / \gamma}
  \times \text{TDI}^{\mathbf{\Delta, {\Omega_{sim}}}}_{\gamma} (\y,\z) \\
  &= \mathcal{K}_{time}(\y,\z),
\end{align*}
which finally proves that $\mathcal{K}_{time}$ defined in paper Eq. 9 is a valid PSD kernel.
\end{proof}{}

The particular choice  $k(u,v)= \dfrac{\frac{1}{2} e^{-\Vert u-v \Vert^2_2}} {1-\frac{1}{2} e^{- \Vert u-v \Vert^2_2}}$ fulfills Proposition 1 requirements: $k$ is indeed PSD as the infinite limit of a sequence of PSD kernels $\sum_{i=1}^{\infty} g^i = \frac{g}{1-g} = k$, where $g$ is a halved Gaussian PSD kernel: $g(u,v)=  \frac{1}{2} e^{- \Vert u-v \Vert ^2_2}$. For this choice of $k$, the corresponding pairwise cost matrix writes (it is the half-Gaussian cost defined in Section \ref{sec:shape-kernel}):
\begin{equation}
    \mathbf{\Delta}(\y_i,\z_j) = \gamma \left[\Vert \y_i-\z_j\Vert^2_2  - \log \left( 2 - e^{- \Vert \y_i-\z_j \Vert ^2_2}  \right)  \right] .
\end{equation}

\clearpage{\pagestyle{empty}\cleardoublepage}

\newpage

\mbox{}
\thispagestyle{empty}
\chapter{Appendix for DILATE}
\markright{\MakeUppercase{Annexe B}}

\section{External shape and temporal metrics}
\label{app:dilate_metrics}

We detail here the two external metrics used in our experiments to evaluate the shape and temporal errors.

\paragraph*{Ramp score:} The notion of \textit{ramping event} is a major issue for intermittent renewable energy production that needs to be anticipated for electricity grid management. For assessing the performance of trained forecasting models in presence of ramps, the Ramp Score was proposed in \cite{vallance2017towards}. This score is based on a piecewise linear approximation on both input and target time series by the Swinging Door algorithm \cite{bristol1990swinging,florita2013identifying}. The Ramp Score described in \cite{vallance2017towards} is computed as the integral between the unsigned difference of derivatives of both linear approximated series. For assessing only the shape error component, we apply in our experiments the ramp score on the target and prediction series after alignment by the optimal DTW path.

\paragraph*{Hausdorff distance:} Given a set of change points $\mathcal{T}^*$ in the target signal and change points $\hat{\mathcal{T}}$ in the predicted signal, the Hausdorff distance is defined as:
\begin{equation}
\text{Hausdorff}(\mathcal{T}^*,\hat{\mathcal{T}}) :=  \max (  \underset{\hat{t} \in \mathcal{ \hat{T} }}{\max}  \underset{t^* \in \mathcal{ T^* }}{\min} |\hat{t}-t^* |  ,  \underset{t^* \in \mathcal{ T^* }}{\max}   \underset{\hat{t} \in \mathcal{ \hat{T} }}{\min} |\hat{t}-t^* | ).
\end{equation}{}

It corresponds to the greatest temporal distance between a change point and its prediction. 

We now explain how the change points are computed for each dataset: for Synthetic, we know exactly by construction the positions of the change points in the target signals. For the predictions, we look for a single change point corresponding to the location of the predicted step function. We use the exact segmentation method by dynamic programming described in \cite{truong2018review} with the Python toolbox \url{http://ctruong.perso.math.cnrs.fr/ruptures-docs/build/html/index.html#} .\\

For ECG5000 and Traffic datasets which present sharp peaks, this change point detection algorithm is not suited (detected change points are often located at the inflexion points of peaks and not at the exact peak location). We thus use a simple peak detection algorithm based on first order finite differences. We tune the threshold parameter for outputting a detection and the min distance between detections parameter experimentally for each dataset.

\section{Comparison to DILATE divergence variant \label{app:dilate-div}}

Blondel \etal \cite{blondel2020differentiable} point out two limitations for using $\text{DTW}^{\mathbf{\Delta}}_{\gamma}$ as a loss function: first, it can take negative values and second, $\text{DTW}^{\mathbf{\Delta}}_{\gamma}(\y,\z)$ does not reach its minimum when $\y = \z$. To address these issues, they propose a proper divergence defined as follows \cite{blondel2020differentiable}:
\begin{equation}
        \text{DTW-div}^{\mathbf{\Delta}}_{\gamma}(\y, \z) =  \text{DTW}^{\mathbf{\Delta}}_{\gamma}(\y, \z) \\ - \frac{1}{2} (\text{DTW}^{\mathbf{\Delta}}_{\gamma}(\y, \y) + \text{DTW}^{\mathbf{\Delta}}_{\gamma}(\z, \z)).
\end{equation}
This divergence is non-negative and satisfies $ \text{DTW-div}^{\mathbf{\Delta}}_{\gamma}(\y, \y) = 0$. However, it is still not a distance function since the triangle inequality is not verified (as for the true DTW).\\

These limitations also hold for DILATE. Consequently, we use the same normalization trick to define a proper DILATE-divergence. Forecasting results in Table \ref{tab:dilate-div} show that DILATE-div is equivalent to DILATE with the Seq2Seq and N-Beats \cite{oreshkin2019n} models, and inferior to DILATE with the Informer model \cite{zhou2020informer}. It confirms the good behaviour of the DILATE loss that does not require this renormalization.

\begin{table}[H]
    \caption{Comparison between DILATE and DILATE-div on the synthetic-det dataset.}
    \centering
    \begin{tabular}{ccc}
    \toprule
   Model & MSE  & DILATE    \\
      \midrule
Seq2Seq DILATE &  \textbf{13.1 $\pm$ 1.8} &  \textbf{33.7 $\pm$ 3.1} \\
 Seq2Seq  DILATE-div & \textbf{13.6 $\pm$ 0.9}     & \textbf{33.6 $\pm$ 2.1} \\
   \midrule
  N-Beats \cite{oreshkin2019n} DILATE & \textbf{13.3 $\pm$ 0.7}     & \textbf{37.9 $\pm$ 1.6}  \\
   N-Beats \cite{oreshkin2019n} DILATE-div & \textbf{13.8 $\pm$ 0.9}     & \textbf{38.5 $\pm$ 1.4}  \\
  \midrule
  Informer \cite{zhou2020informer} DILATE & \textbf{11.8 $\pm$ 0.7}    & \textbf{30.1 $\pm$ 1.3} \\
   Informer \cite{zhou2020informer} DILATE-div & 12.9 $\pm$ 0.1   & 31.8 $\pm$ 6.5 \\  
 \bottomrule
    \end{tabular}
    \label{tab:dilate-div}
\end{table}

\section{DILATE additional visualizations}
\label{app:dilate_visus}

\begin{figure*}
\begin{center}
 \includegraphics[width=13cm]{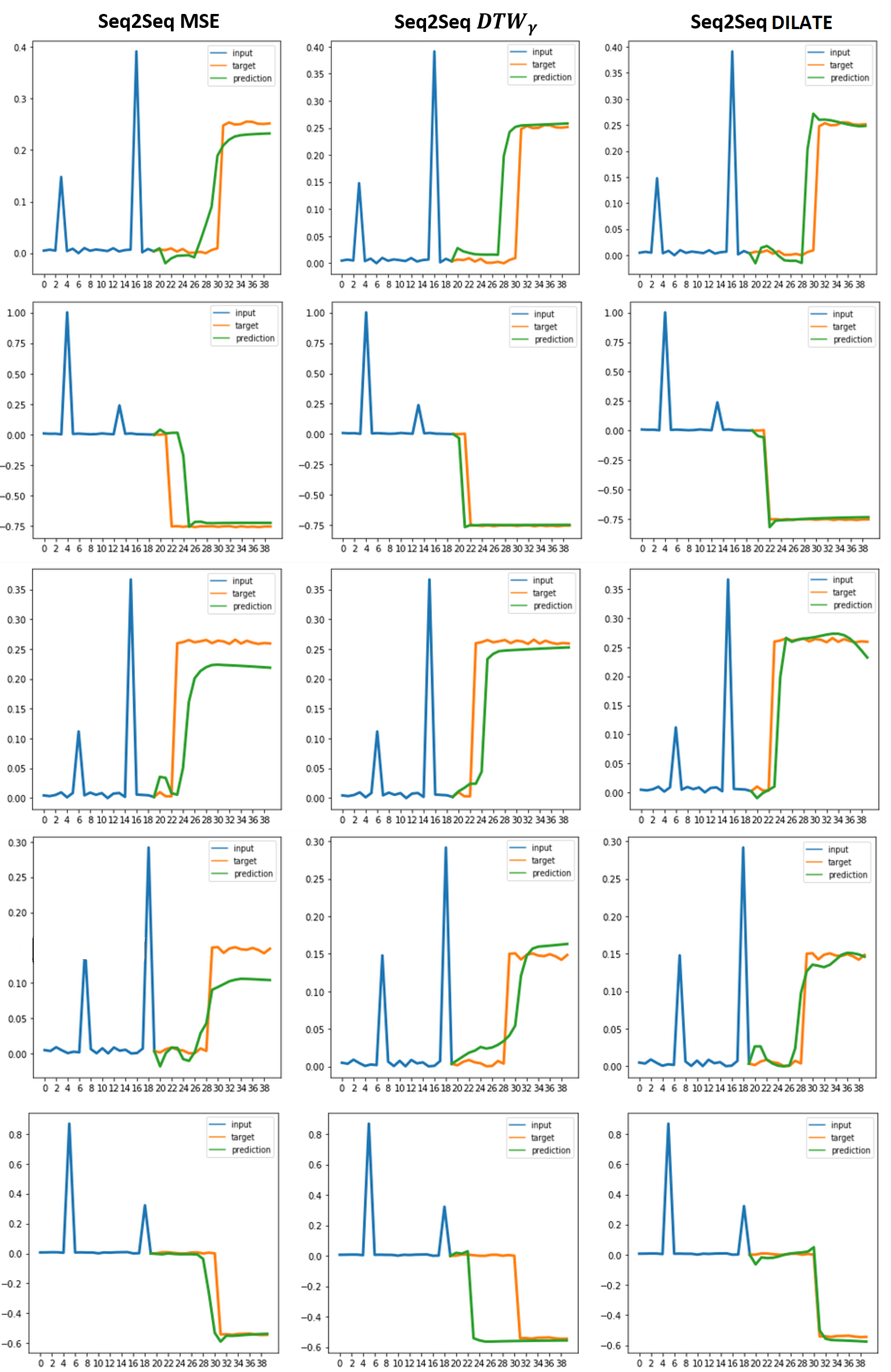}  
\end{center}
   \caption{Qualitative predictions for the \texttt{Synthetic-det} dataset.}
  \label{fig:synth_sup}
\end{figure*}

\begin{figure*}
\begin{center}
 \includegraphics[width=13cm]{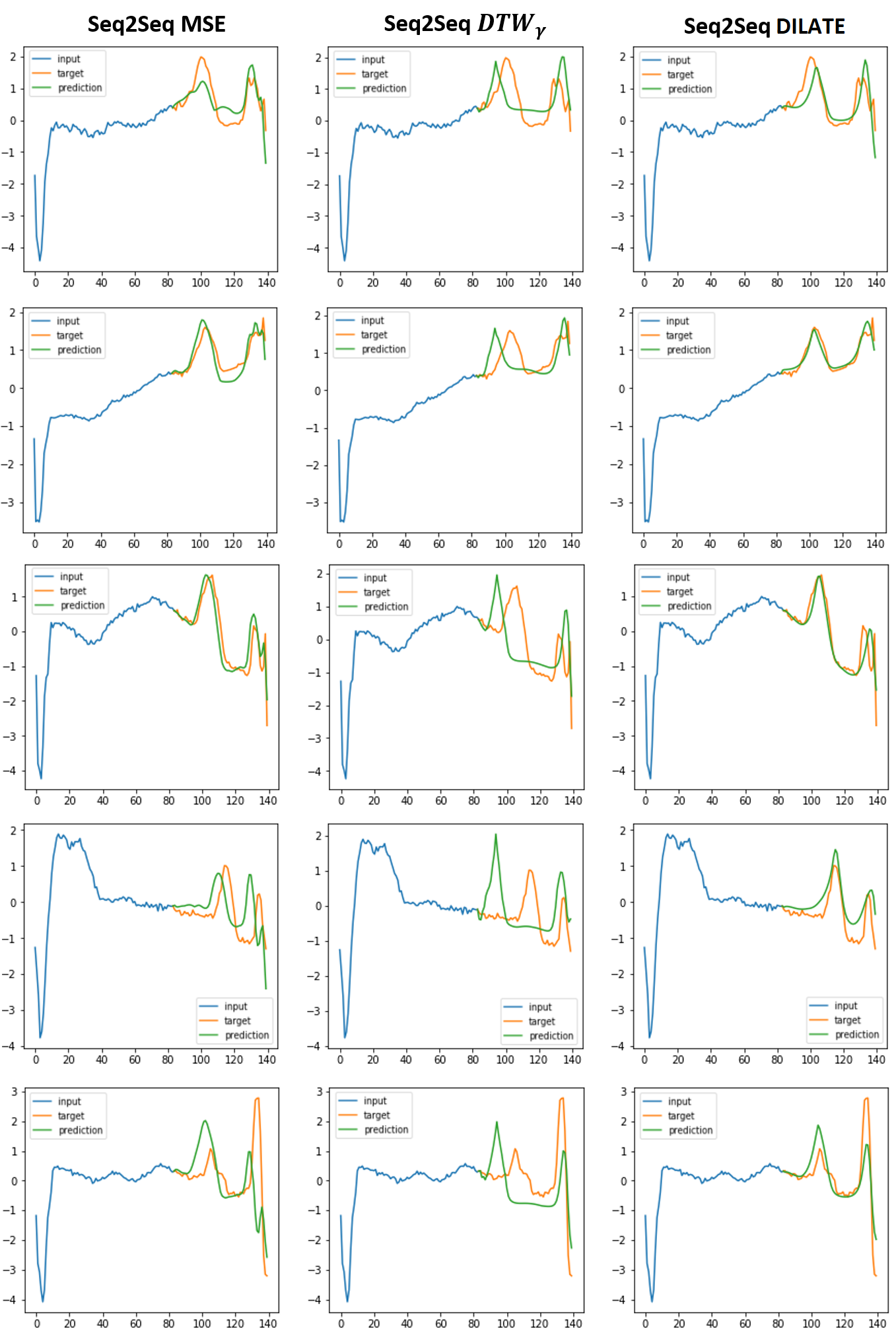}  
\end{center}
   \caption{Qualitative predictions for the \texttt{ECG5000} dataset.}
  \label{fig:ecg_sup}
\end{figure*}

\begin{figure*}
\begin{center}
 \includegraphics[width=13cm]{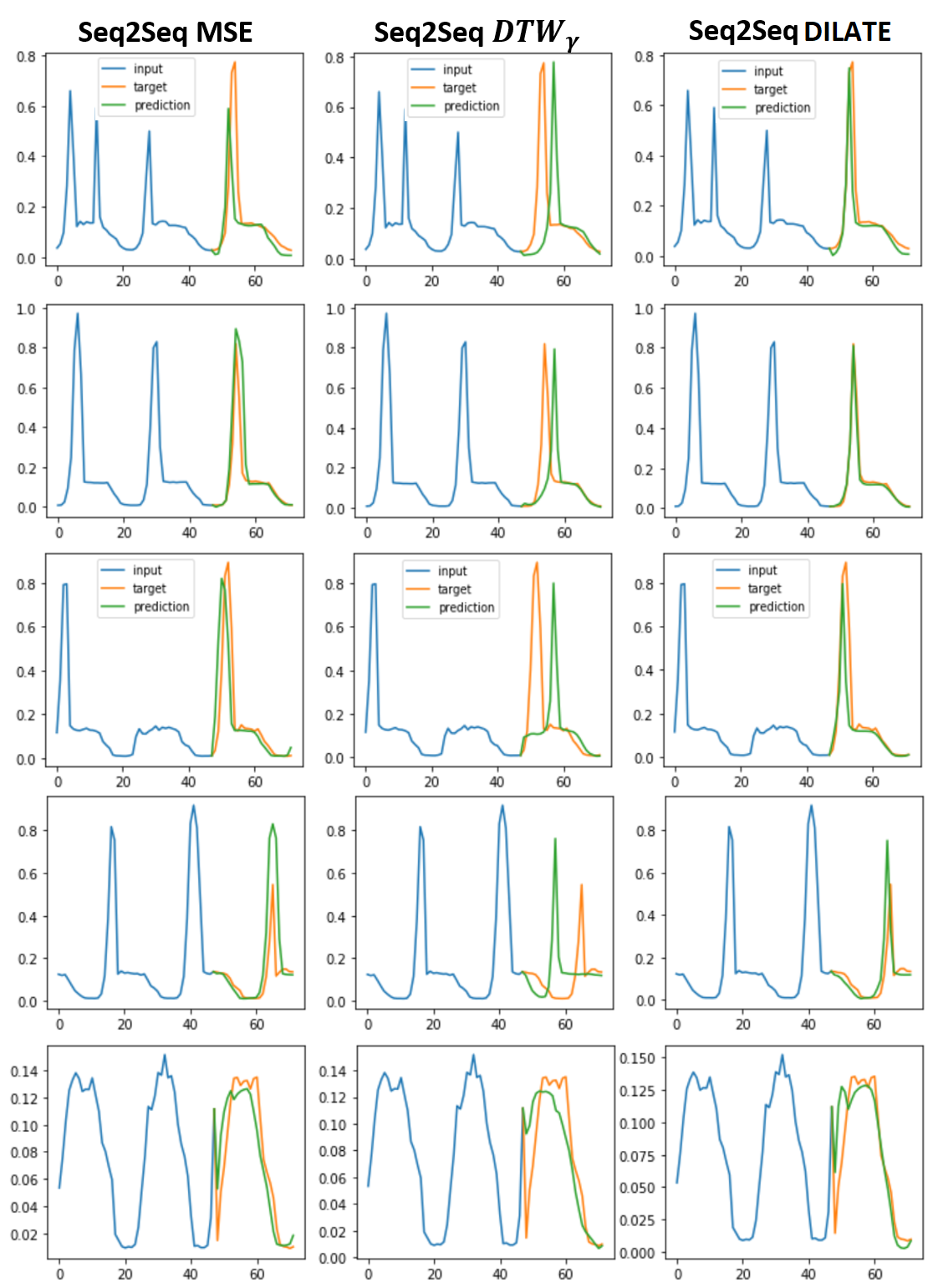}  
\end{center}
   \caption{Qualitative predictions for the \texttt{Traffic} dataset.}
  \label{fig:traffic_sup}
\end{figure*}

We provide additional qualitative predictions with DILATE for the \texttt{Synthetic-det} in Figure \ref{fig:synth_sup}, for \texttt{ECG5000} in Figure \ref{fig:ecg_sup} and for \texttt{Traffic} in Figure \ref{fig:traffic_sup}.

\clearpage{\pagestyle{empty}\cleardoublepage}
\newpage

\mbox{}
\thispagestyle{empty}
\chapter{Appendix for STRIPE}
\markright{\MakeUppercase{Annexe C}}

\section{STRIPE implementation details}
\label{app:stripe}

\textbf{Neural network architectures:} STRIPE++ is composed of a Sequence To Sequence predictive model. The encoder is a recurrent neural network (RNN) with 1 layer of 128 Gated Recurrent Units (GRU) \cite{cho2014learning} units, producing a latent state $h$ of dimension 128. We fixed by cross-validation the dimension of each diversifying variable $z_s$ or $z_t$ to be $k=8$. The decoder is another RNN with $128+8+8=144$ GRU units followed by fully connected layers responsible for producing the future trajectory.\\

The Posterior network has a similar architecture as the encoder: it is a RNN with 1 layer of 128 GRU units that takes as input the full series $(\x_{1:T},\y^*_{T+1:T+H})$, followed by two multi-layer perceptrons (MLP) dedicated to output the parameters $(\mu_s^*,\sigma_s^*)$ and $(\mu_t^*,\sigma_t^*)$ of the Gaussian distribution from which to sample the posterior diversifying variables $z_s^*$ and $z_t^*$.\\

The STRIPE$^{++}_{\text{shape}}$ and STRIPE$^{++}_{\text{time}}$ proposal mechanisms build on top of the encoder (that produces $h$) with a MLP with 3 layers of 512 neurons (with Batch Normalization and LeakyReLU activations) and a final linear layer to produce $N=10$ latent codes of dimension $k=8$ (corresponding to the proposals for $z_s$ or $z_t$).\\

\paragraph{STRIPE hyperparameters:} We cross-validated the relevant hyperparameters of STRIPE:
\begin{itemize}
\setlength{\itemsep}{5pt}
  \setlength{\parskip}{0pt}
  \setlength{\parsep}{0pt}

    \item $k$: dimension of the diversifying latent variables $z$. This dimension should be chosen relatively to the hidden size of the RNN encoders and decoders (128 in our experiments). We fixed $k=8$ in all cases.
    \item $N$: the number of future trajectories to sample. We fixed $N=10$. We performed a sensibility analysis to this parameter in paper Figure 8.
    \item $\mu = 20$: quality constraint hyperparameter in the DPP kernels.
\end{itemize}

\section{STRIPE additional visualizations}

Wee provide additional visualizations for the Traffic and Electricity datasets that confirm that STRIPE predictions are both diverse and sharp.

\subsubsection{Electricity}

\begin{tabular}{cc}
    \includegraphics[width=8cm]{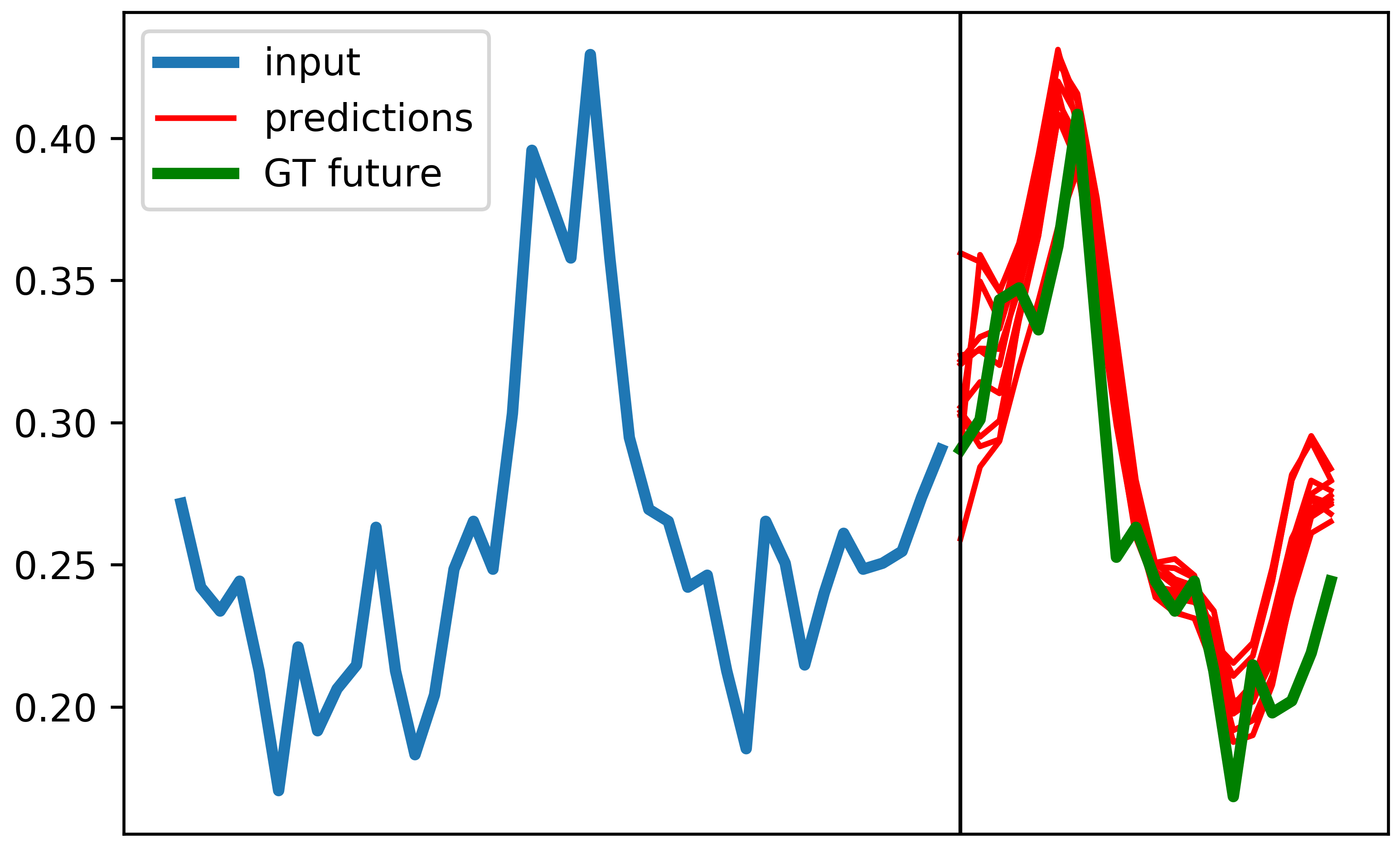} &
     \includegraphics[width=8cm]{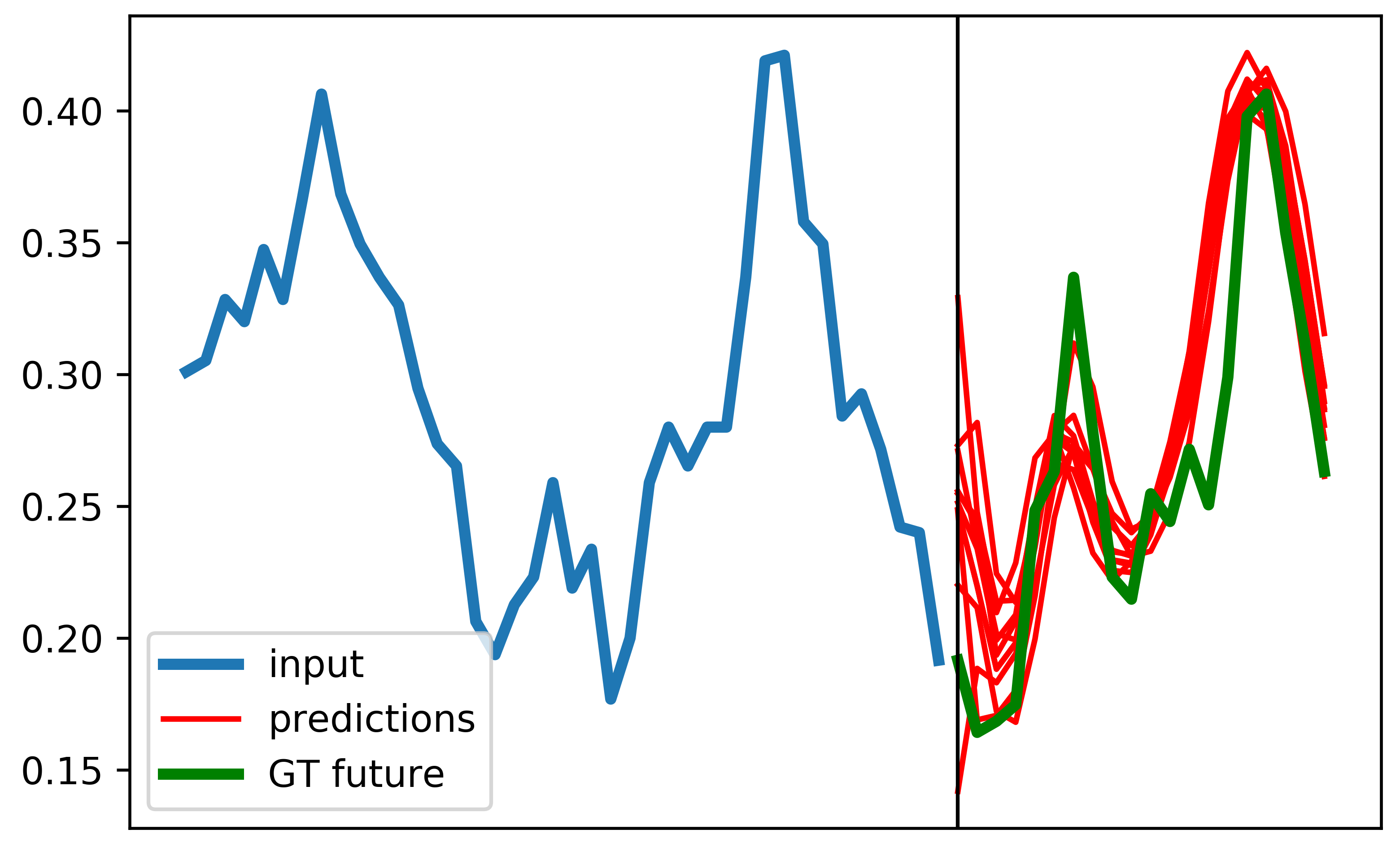}       \\
         \includegraphics[width=8cm]{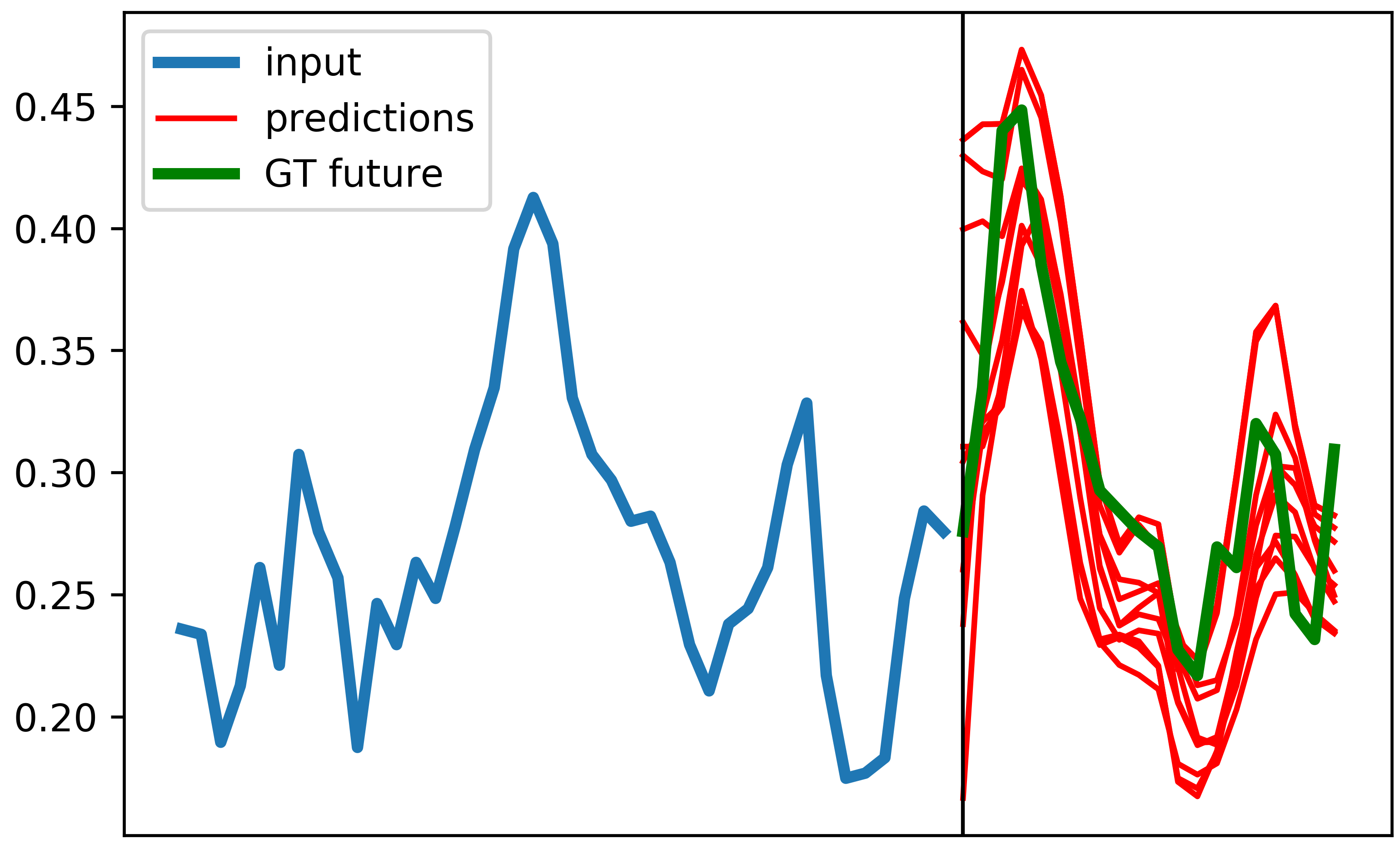}&
     \includegraphics[width=8cm]{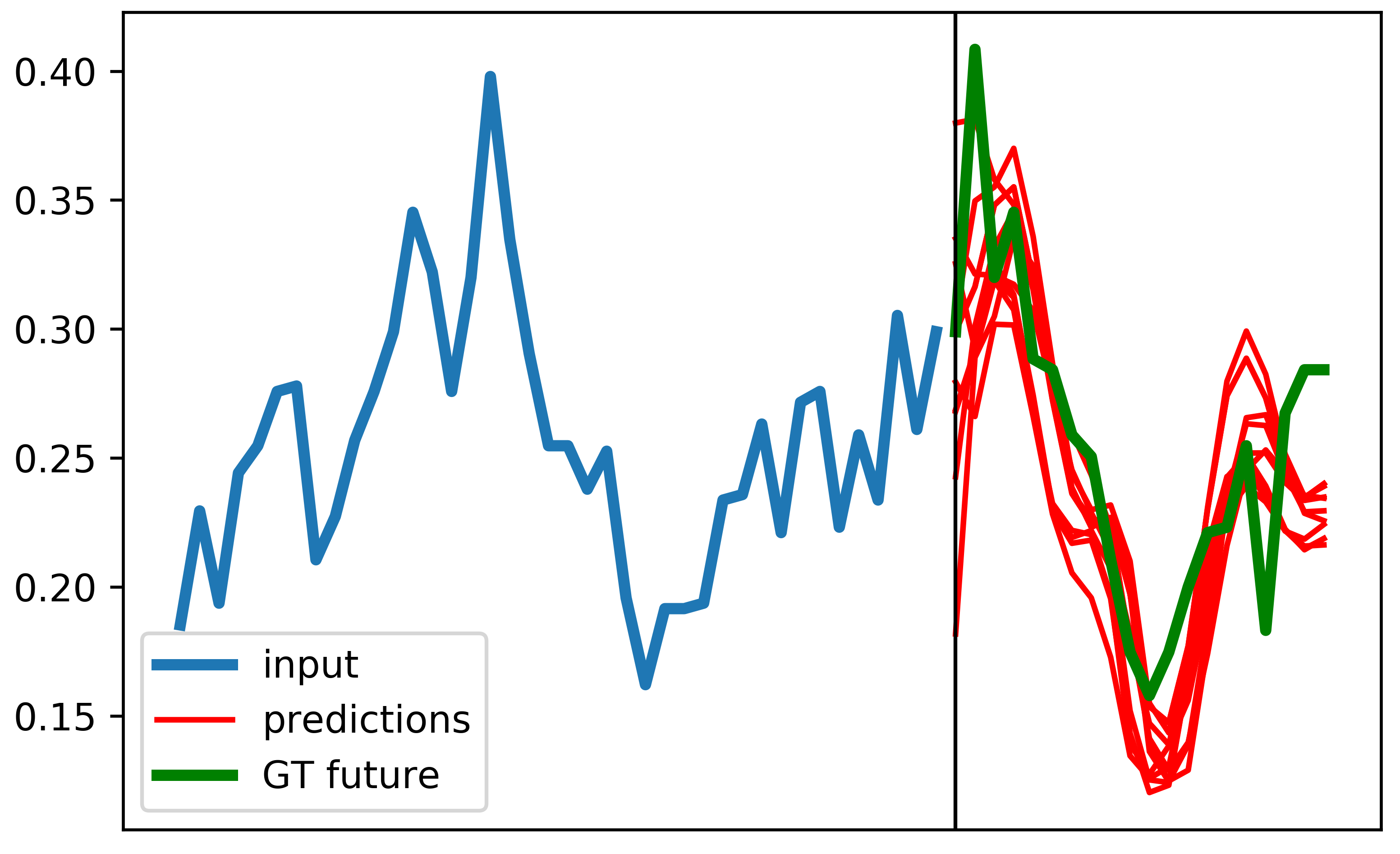}\\
         \includegraphics[width=8cm]{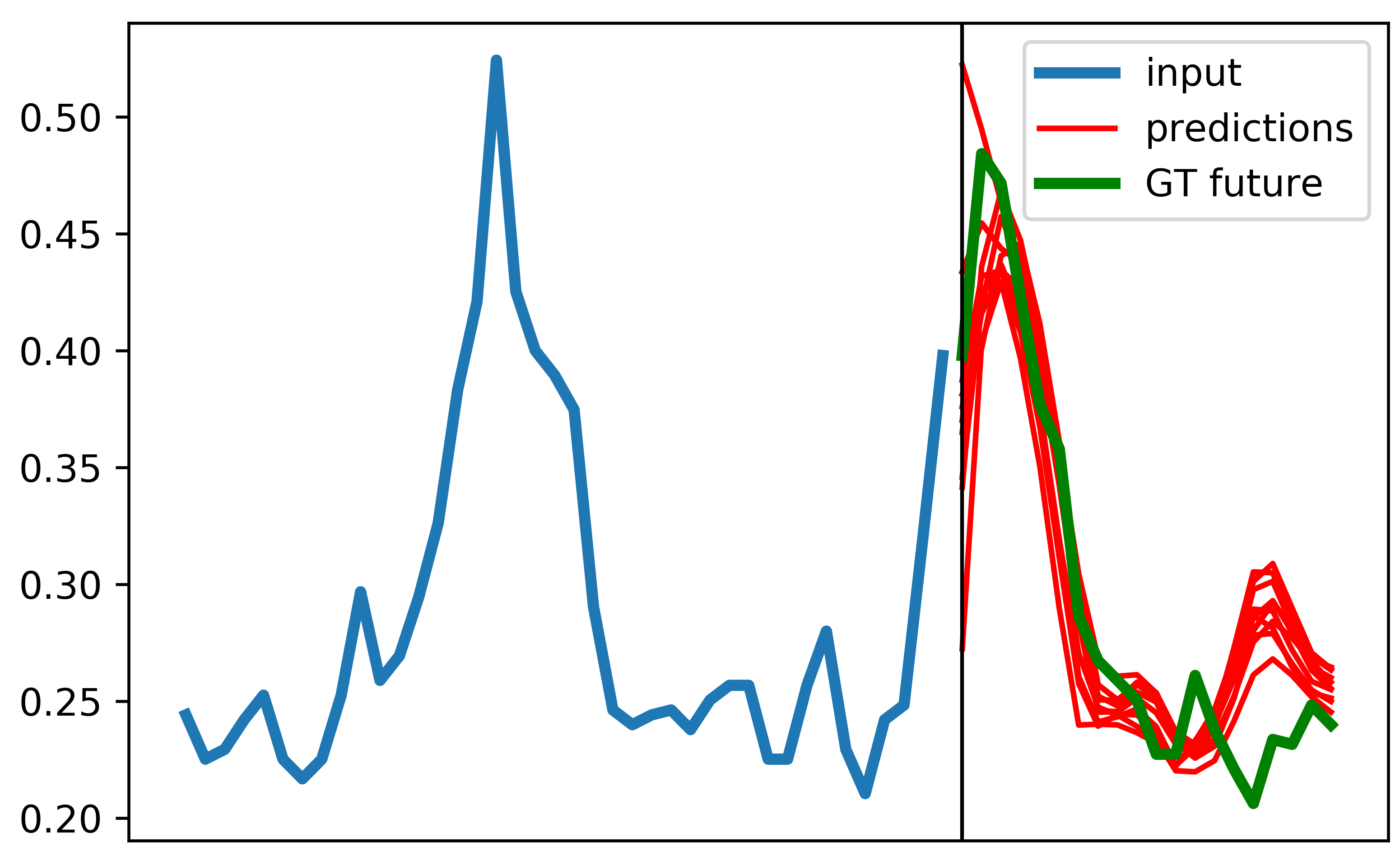}&        \includegraphics[width=8cm]{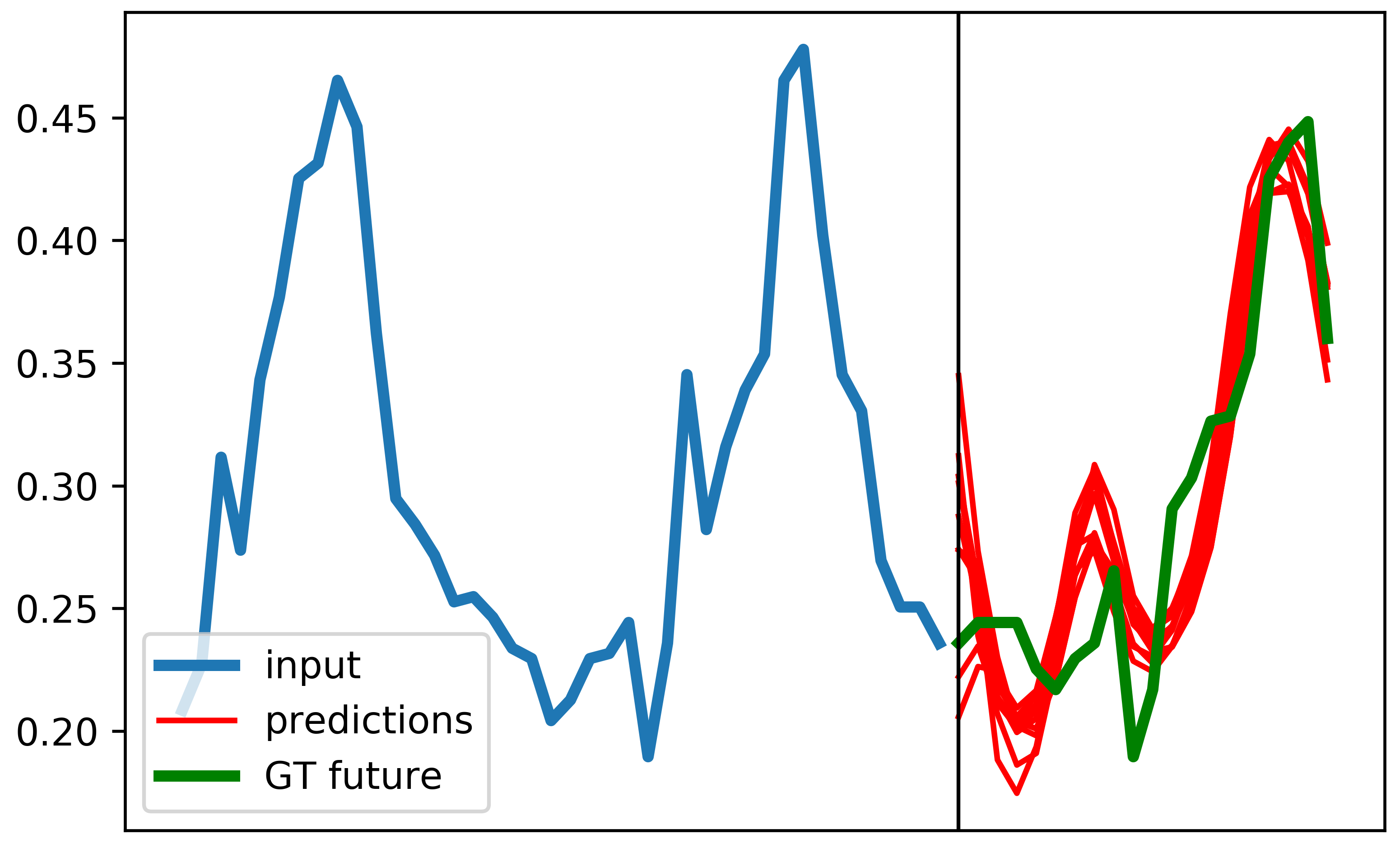}          
   
  \end{tabular}

\subsubsection{Traffic}

\begin{tabular}{cc}
 \includegraphics[width=8cm]{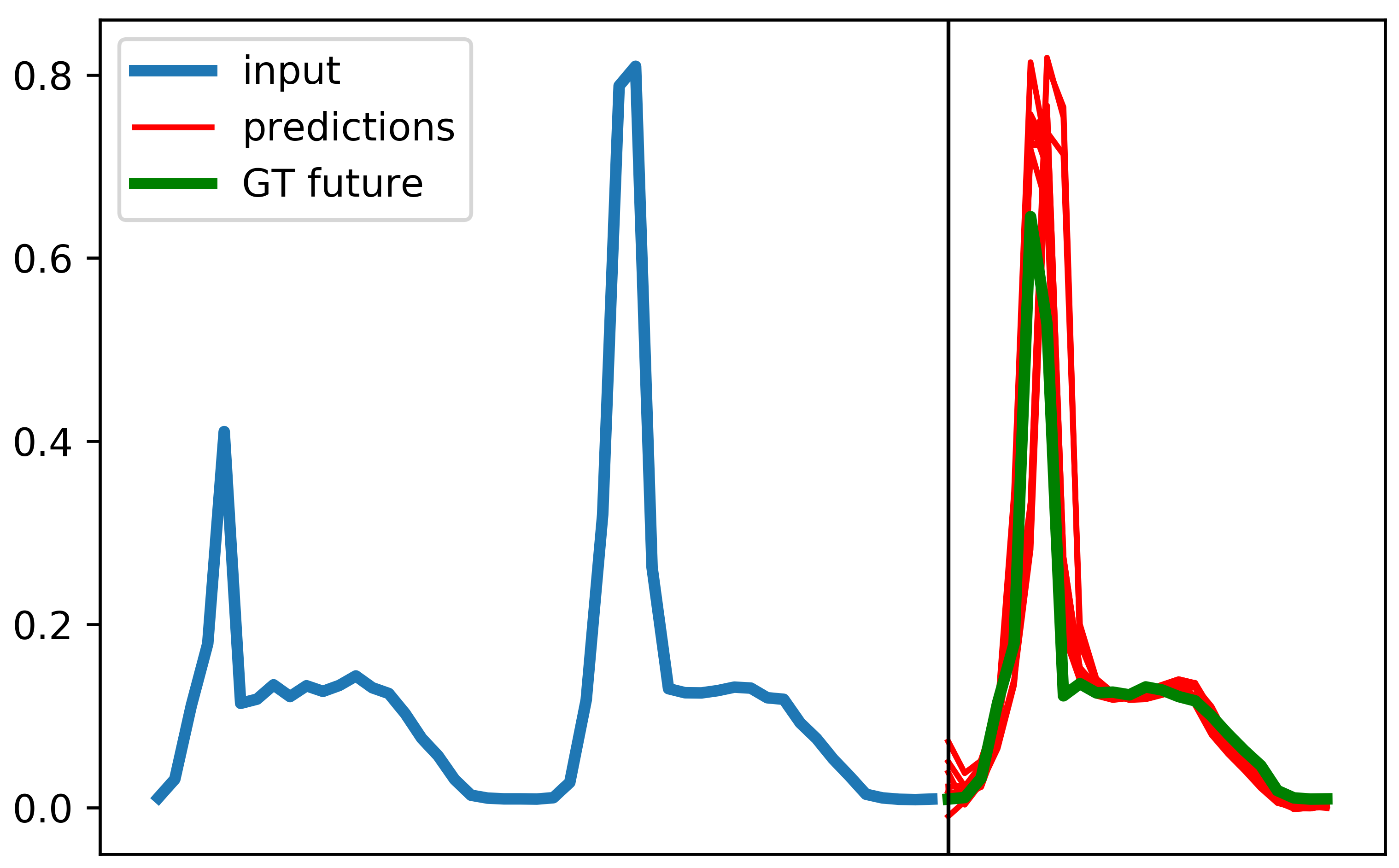}  &\
     \includegraphics[width=8cm]{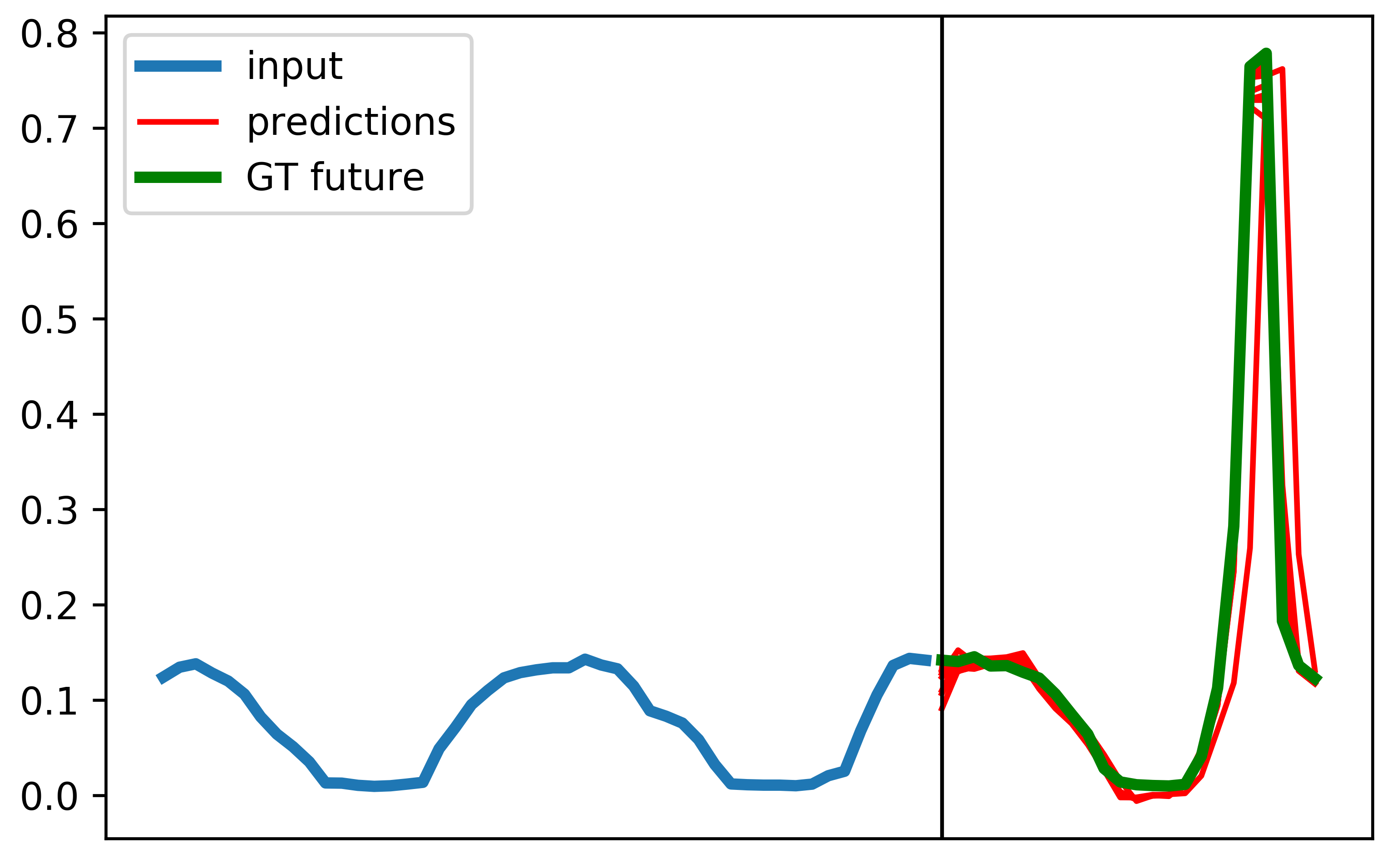}       \\    \includegraphics[width=8cm]{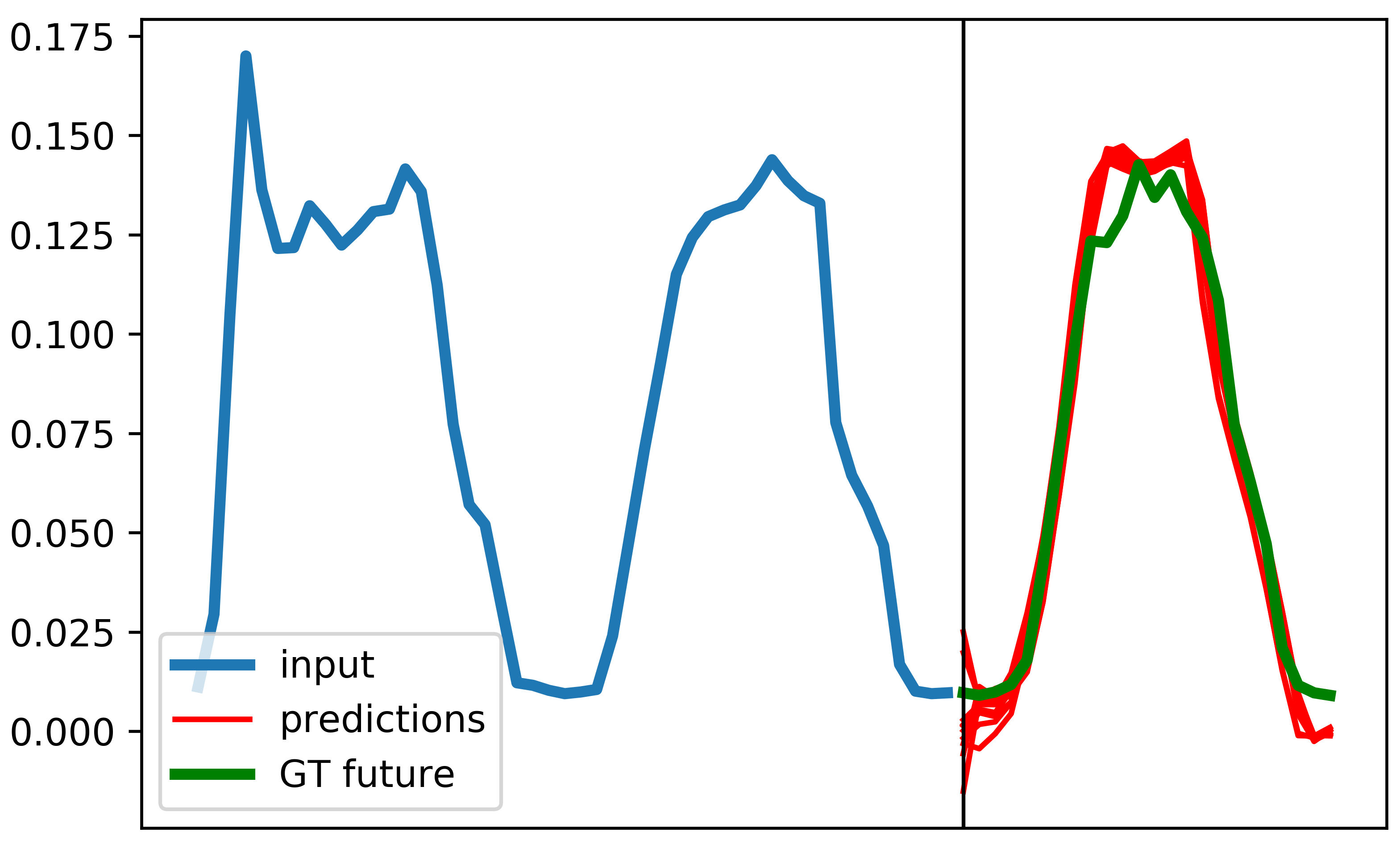}  &
     \includegraphics[width=8cm]{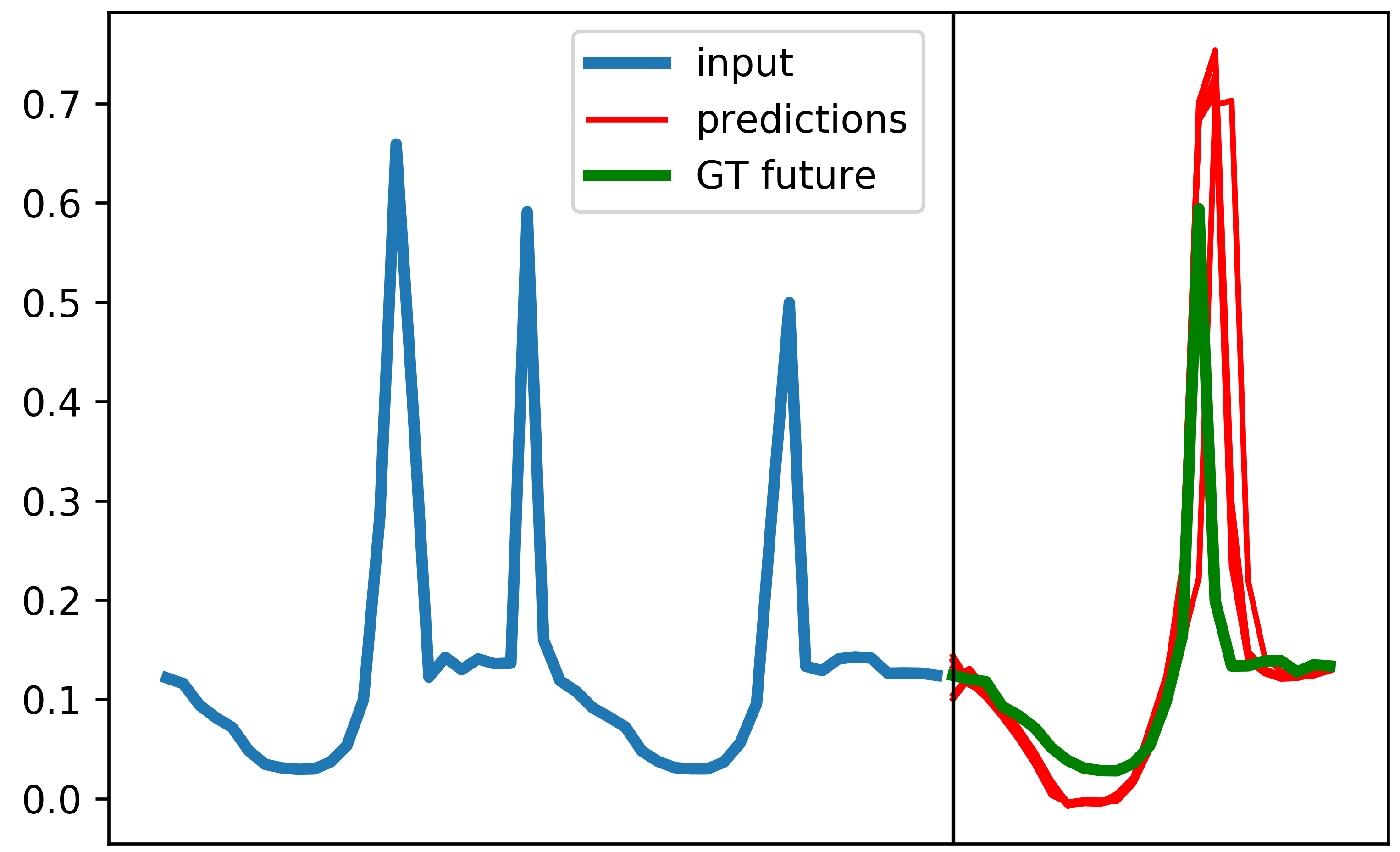}\\
       \includegraphics[width=8cm]{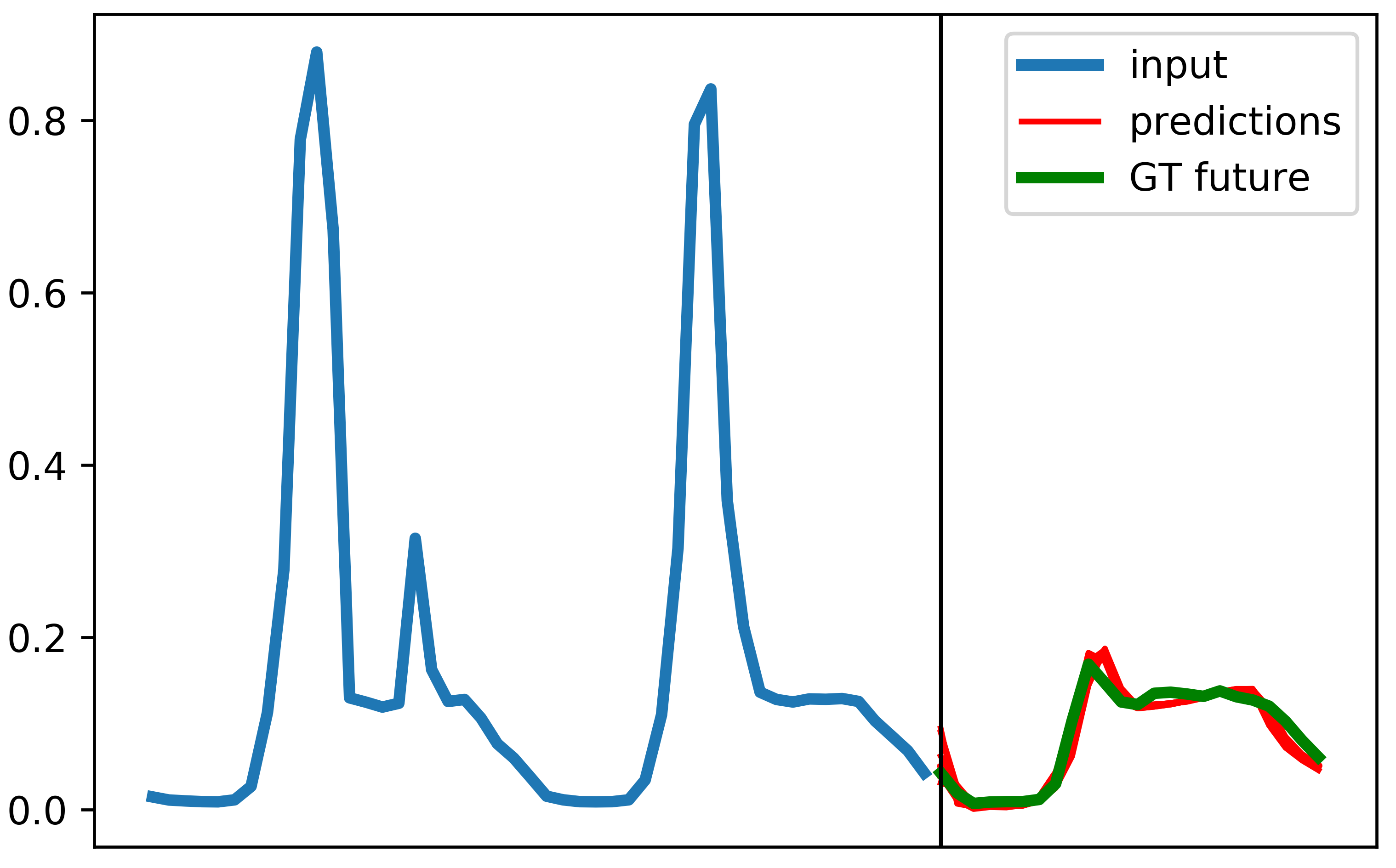} &           
      \includegraphics[width=8cm]{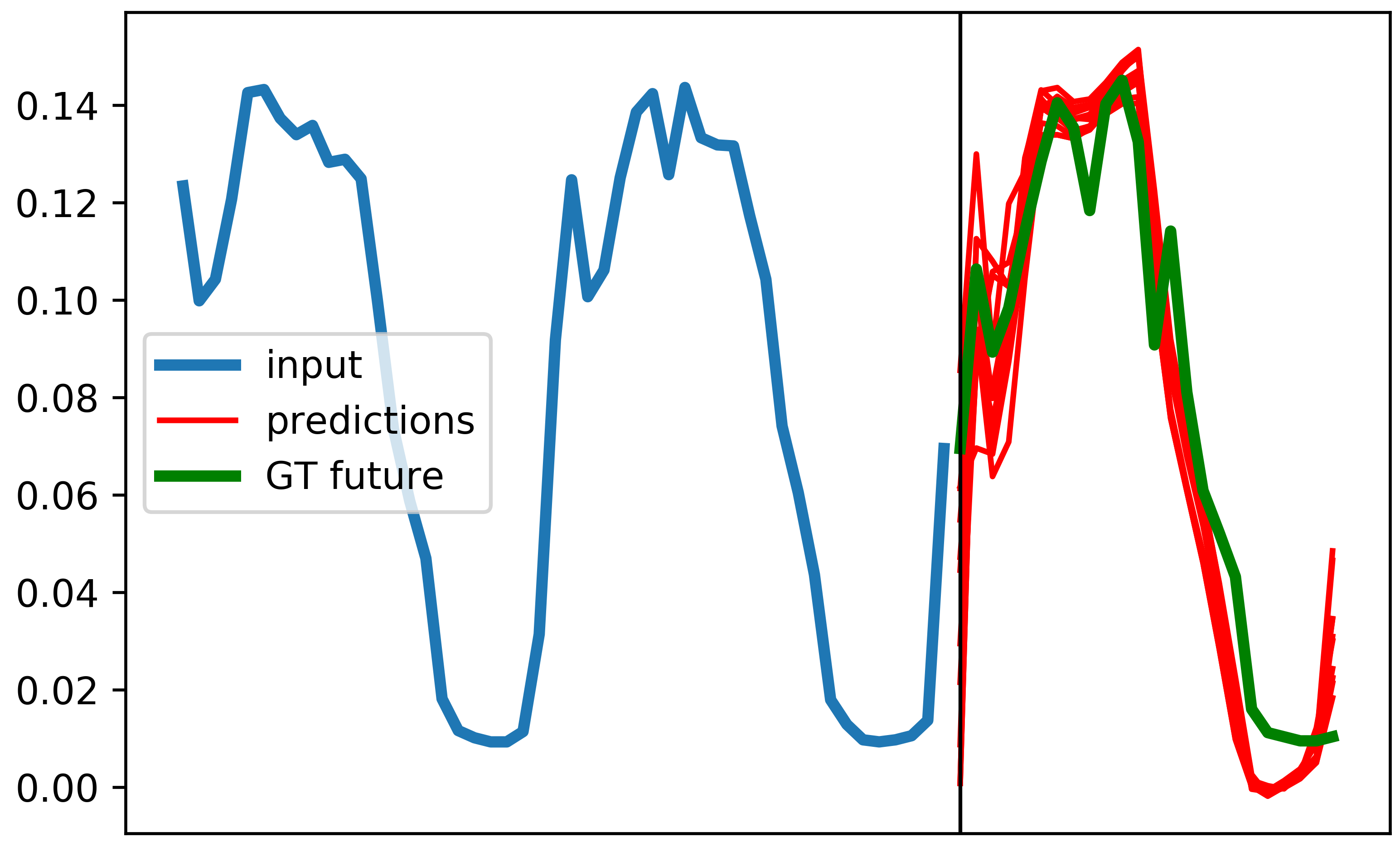}  \\
\end{tabular}

\clearpage{\pagestyle{empty}\cleardoublepage}
\newpage

\mbox{}
\thispagestyle{empty}
\chapter{Appendix for PhyDNet}
\markright{\MakeUppercase{Annexe D}}

\section{PhyDNet model}

\subsection{Discrete PhyCell derivation}
\label{app:phycell-deriv}

\noindent PhyCell dynamics is governed by the PDE:
\begin{align*}
    \dfrac{\partial \mathbf{h}}{\partial t}(t,\mathbf{x}) &= \Phi(\mathbf{h}) + \mathcal{C}(\mathbf{h},\mathbf{u}) \\
     &= \Phi(\mathbf{h}(t,\mathbf{x}) ) + \mathbf{K}(t,\mathbf{x}) \odot  (\mathbf{E}(\mathbf{u}(t,\mathbf{x})) -  (\mathbf{h}(t,\mathbf{x})  + \Phi(\mathbf{h}(t,\mathbf{x}) )).
\end{align*}
By Euler discretization $\frac{\partial \mathbf{h}}{\partial t}= \delta \mathbf{h}_t = \mathbf{h}_{t} - \mathbf{h}_{t-1}$, we get:
\begin{align*}
    \mathbf{h}_{t+1} - \mathbf{h}_t &= \Phi(\mathbf{h}_t) + \mathbf{K}_t \odot (\mathbf{E}(\mathbf{u}_t) -(\mathbf{h}_t+\Phi(\mathbf{h}_t))) \\
    \mathbf{h}_{t+1} &=  \mathbf{h}_t + \Phi(\mathbf{h}_t) + \mathbf{K}_t \odot (\mathbf{E}(\mathbf{u}_t) -(\mathbf{h}_t+\Phi(\mathbf{h}_t)))  \\
\mathbf{h}_{t+1} &= (1-\mathbf{K}_t) \odot (\mathbf{h}_t+\Phi(\mathbf{h}_t)) + \mathbf{K}_t \odot \mathbf{E}(\mathbf{u}_t).
\end{align*}

\subsection{Moment matrix}
\label{app:moment-matrix}

For a filter $\mathbf{w}$ of size $k \times k$, the moment matrix $\mathbf{M(w)}$ is a matrix of size $k \times k$ defined as:
\begin{equation*}
    \mathbf{M}(\mathbf{w})_{i,j} = \frac{1}{i! j!} \sum_{u=-\frac{k-1}{2}}^{\frac{k-1}{2}} \sum_{v=-\frac{k-1}{2}}^{\frac{k-1}{2}} u^i v^j \mathbf{w}[u,v],
\end{equation*}{}
for $i,j=0,...,k-1$.

For any function $h:\mathbb{R}^2 \longrightarrow \mathbb{R}$, we consider the convolution of $h$ with the filter $\mathbf{w}$. Taylor's expansion gives:
\begin{flalign*}
 \sum_{u=-\frac{k-1}{2}}^{\frac{k-1}{2}} \sum_{v=-\frac{k-1}{2}}^{\frac{k-1}{2}} \mathbf{w}[u,v] h(x + \delta x . u,y + \delta y . v) 
    &=  \sum_{u=-\frac{k-1}{2}}^{\frac{k-1}{2}} \sum_{v=-\frac{k-1}{2}}^{\frac{k-1}{2}}  \mathbf{w}[u,v] \sum_{i,j=1}^{k-1} \frac{\partial^{i+j} h}{\partial x^i \partial y^j}(x,y) \frac{u^i v^j}{i! j!} \delta x^i \delta y^j 
  \\  + o(|\delta x|^{k-1} + |\delta y|^{k-1}) \\
    &=   \sum_{i,j=1}^{k-1}  \mathbf{M}(\mathbf{w})_{i,j}  \delta x^i \delta y^j  \frac{\partial^{i+j} h}{\partial x^i \partial y^j}(x,y)   + o(|\delta x|^{k+1} + |\delta y|^{k-1}). \\
\end{flalign*}{}
This equation shows that we can control the differential order approximated by the filter $\mathbf{w}$ by imposing constraints on its moment matrix $\mathbf{M(w)}$.
For example, in order to approximate the differential operator $\frac{\partial^{a+b}}{\partial x^{a} \partial y^{b}} (.)$, it suffices to impose $\mathbf{M(w)}_{i,j} = 0$ for $i \neq a$ and $j \neq b$. By denoting $\mathbf{\Delta}^k_{i,j}$ the Kronecker matrix of size $k \times k$, which equals 1 at position $(i,j)$ and 0 elsewhere, we thus enforce the moment matrix  $\mathbf{M(w)}$ to match the target $\mathbf{\Delta}^k_{a,b}$ with the Frobenius norm. This justifies the choice of our moment loss for enforcing each filter $\mathbf{w}^k_{p,i,j}$ to approximate the corresponding derivative  $\frac{\partial^{i+j}}{\partial x^{i} \partial y^{j}} (.)$:
 \begin{equation*}
   \mathcal{L}_{\text{moment}} = \sum\limits_{i \leq k} \sum\limits_{j \leq k} ||\mathbf{M}(\mathbf{w}^k_{p,i,j}) - \mathbf{\Delta}^k_{i,j} ||_F.
   \label{eq:lmoment}
 \end{equation*}

\subsection{Prediction mode training}
We show in section \ref{sec:pdernn} that the decomposition $\bm{\mathcal{M}}_r(\mathbf{h},\mathbf{u}) = \Phi(\mathbf{h})+ \mathcal{C}(\mathbf{h},\mathbf{u})$ still holds for standard Seq2Seq models (RNN, GRU, LSTM). As mentioned in Chapter \ref{chap:phydnet}, the resulting predictor $\Phi$ is, however, naive and useless for multi-step prediction, \ie $\Phi(\mathbf{h})=-\mathbf{h}$ and  $\mathbf{\tilde{h}}_{t+1}=0$.  

In multi-step prediction, the option followed by standard Seq2seq models is to recursively reinject back predictions as ground truth input for the next time steps. Scheduled Sampling \cite{bengio2015scheduled} is a solution to mitigate error accumulation and train/test discrepancy, that we use in our ConvLSTM branch. This is, however, inferior to the results obtained with our PhyCell trained in the "prediction-only" mode, as shown in Section \ref{sec:expe_prediction}.

\subsubsection{PDE formulation for standard RNNs}
\label{sec:pdernn}
\paragraph{Vanilla RNN}

The equations for the vanilla RNN are:
\begin{equation*}
    \mathbf{h}_t = \tanh(\mathbf{W}_h \mathbf{h}_{t-1} + \mathbf{W}_u \mathbf{u}_t + \mathbf{b} ),
\end{equation*}
with weight matrices $\mathbf{W}_h$, $\mathbf{W}_u$ and bias $\mathbf{b}$. By approximating $\frac{\partial \mathbf{h}}{\partial t}= \delta \mathbf{h}_t = \mathbf{h}_t - \mathbf{h}_{t-1}$, we get the PDE:
\begin{align*}
        \dfrac{\partial \mathbf{h}}{\partial t}(t,\mathbf{x}) &= \bm{\mathcal{M}}(\mathbf{h},\mathbf{u})  \\ &=  
    \tanh(\mathbf{W}_h \mathbf{h}(t) + \mathbf{W}_u \mathbf{u}(t) + \mathbf{b} ) - \mathbf{h}(t).
\end{align*}

A linear decoupling of this PDE is 
\begin{equation*}
\frac{\partial \mathbf{h}}{\partial t}(t,\mathbf{x}) = \Phi(\mathbf{h}) + \mathcal{C}(\mathbf{h},\mathbf{u}),
\end{equation*}
with $\Phi(\mathbf{h}) = -\mathbf{h}(t)$ and $\mathcal{C}(\mathbf{h},\mathbf{u}) =  \tanh(\mathbf{W}_h \mathbf{h}(t) + \mathbf{W}_u \mathbf{u}(t) + \mathbf{b} ) $ which gives in discrete time the prediction-correction scheme:
\begin{empheq}[left=\empheqlbrace]{alignat=2}
&   \tilde{\mathbf{h}}_{t+1}= 0    \label{eq:prediction}\\
&   \mathbf{h}_{t+1} = \tilde{\mathbf{h}}_{t+1} +    \tanh \left(\mathbf{W}_h \mathbf{h}_{t-1} + \mathbf{W}_u \mathbf{u}_t + \mathbf{b} \right). \label{eq:correction}
\end{empheq}
We see that the prior predictor $\Phi$ brings no information and that the correction step drives the whole dynamics.

\paragraph{Gated Recurrent Unit (GRU)}
The equations of the Gated Recurrent Unit \cite{cho2014learning} are:
\begin{align*}
    \mathbf{r}_t &= \sigma(\mathbf{W}_{rh} \mathbf{h}_{t-1} + \mathbf{W}_{ru} \mathbf{u}_t + \mathbf{b}_r) \\
    \mathbf{z}_t &= \sigma(\mathbf{W}_{zh} \mathbf{h}_{t-1} + \mathbf{W}_{zu} \mathbf{u}_t + \mathbf{b}_z) \\    
   \mathbf{g}_t &= \tanh(\mathbf{W}_{gh} (\mathbf{r}_t \odot \mathbf{h}_{t-1}) + \mathbf{W}_{gu} \mathbf{u}_t + \mathbf{b}_g) \\
   \mathbf{h}_t &= \mathbf{z}_t \odot \mathbf{h}_{t-1} + (1-\mathbf{z}_t) \odot \mathbf{g}_t,
\end{align*}
where $\mathbf{r}_t$ is the reset gate, $\mathbf{z}_t$ is the update gate and $\mathbf{g}_t$ is the update vector. By approximating $\frac{\partial \mathbf{h}}{\partial t}= \delta \mathbf{h}_t = \mathbf{h}_t - \mathbf{h}_{t-1}$, we get the PDE:
\begin{align*}
    \dfrac{\partial \mathbf{h}} {\partial t}(t,\mathbf{x}) &=  \bm{\mathcal{M}}(\mathbf{h},\mathbf{u}) \\
    &= \mathbf{z}(t) \odot \mathbf{h}(t) + (1-\mathbf{z}(t)) \odot \mathbf{g}(t) - \mathbf{h}(t).
\end{align*}{}
A linear decoupling of this PDE is 
\begin{equation*}
    \frac{\partial \mathbf{h}}{\partial t}(t,\mathbf{x}) = \Phi(\mathbf{h}) + \mathcal{C}(\mathbf{h},\mathbf{u}) ,
\end{equation*}
with $\Phi(\mathbf{h}) = -\mathbf{h}(t)$ and $\mathcal{C}(\mathbf{h},\mathbf{u}) =  \mathbf{z}(t) \odot \mathbf{h}(t) + (1-\mathbf{z}(t)) \odot \mathbf{g}(t)$ which gives in discrete time the prediction-correction scheme:
\begin{empheq}[left=\empheqlbrace]{alignat=2}
&   \tilde{\mathbf{h}}_{t+1}= 0    \label{eq:prediction}\\
&   \mathbf{h}_{t+1} = \tilde{\mathbf{h}}_{t+1} +  \mathbf{z}_t \odot \mathbf{h}_{t-1} + (1-\mathbf{z}_t) \odot \mathbf{g}_t .  \label{eq:correction}
\end{empheq}
We again see that the prior predictor $\Phi$ brings no information and that the correction step drives the whole dynamics.

\paragraph{Long Short-Term Memory (LSTM)}

We give the formulation for the standard LSTM \cite{Hochreiter:1997:LSM:1246443.1246450} (the ConvLSTM \cite{xingjian2015convolutional} can be immediately deduced by replacing matrix products by convolutions):
\begin{align*}
    \mathbf{i}_t &= \sigma (\mathbf{W}_{ih} \mathbf{h}_{t-1} + \mathbf{W}_{iu} \mathbf{u}_t + \mathbf{b}_i) \\
    \mathbf{f}_t &= \sigma (\mathbf{W}_{fh} \mathbf{h}_{t-1} + \mathbf{W}_{fu} \mathbf{u}_t + \mathbf{b}_f) \\
    \mathbf{g}_t &= \tanh (\mathbf{W}_{gh} \mathbf{h}_{t-1} + \mathbf{W}_{gu} \mathbf{u}_t + \mathbf{b}_g) \\
    \mathbf{c}_t &= \mathbf{f}_t \odot \mathbf{c}_{t-1} + \mathbf{i}_t \odot \mathbf{g}_t \\
    \mathbf{o}_t &= \sigma (\mathbf{W}_{oh} \mathbf{h}_{t-1} + \mathbf{W}_{ou} \mathbf{u}_t + \mathbf{b}_o) \\
    \mathbf{h}_t &= \mathbf{o}_t \odot \tanh(\mathbf{c}_t).
\end{align*}
where $\mathbf{i}_t$ is the input gate, $\mathbf{f}_t$ the forget gate, $\mathbf{g}_t$ the input-modulation gate, $\mathbf{o}_t$ the output gate, $\mathbf{c}_t$ the cell state and $\mathbf{h}_t$ the latent state. We define the LSTM augmented latent state as: 
\begin{equation*}
\Bar{\mathbf{h}} =      \begin{pmatrix}
\mathbf{g}  \\ \mathbf{c} 
\end{pmatrix}.
\end{equation*}
The augmented state $\mathbf{\Bar{h}}$ thus verifies the PDE:
\begin{equation*}
    \dfrac{\partial \Bar{\mathbf{h}}}{\partial t} =   \begin{pmatrix}
\dfrac{\partial \mathbf{h}}{\partial t}  \\ \dfrac{\partial \mathbf{c}}{\partial t}  \end{pmatrix} =  \begin{pmatrix}
\mathbf{o}(t) \odot \tanh(\mathbf{c}(t)) - \mathbf{h}(t))  \\ \mathbf{f}(t) \odot \mathbf{c}(t) + \mathbf{i}(t) \odot \mathbf{g}(t) - \mathbf{c}(t) 
\end{pmatrix}.
\end{equation*}
A linear decoupling of this PDE is 
\begin{equation*}
 \frac{\partial \mathbf{\Bar{h}}}{\partial t}(t,\mathbf{x}) = \Phi(\mathbf{\Bar{h}}) + \mathcal{C}(\mathbf{\Bar{h}},\mathbf{u}) ,   
\end{equation*}
 with $\Phi(\mathbf{\Bar{h}}) = -\mathbf{\Bar{h}}(t)$ and
 \begin{equation*}
 \mathcal{C}(\mathbf{\Bar{h}},\mathbf{u})   =      \begin{pmatrix}
 \mathbf{o}(t) \odot \tanh(\mathbf{c}(t))   \\ \mathbf{f}(t) \odot \mathbf{c}(t) + \mathbf{i}(t) \odot \mathbf{g}(t) 
\end{pmatrix},
 \end{equation*}
which gives in discrete time the prediction-correction scheme:


\begin{empheq}[left=\empheqlbrace]{alignat=2}
&   \tilde{\mathbf{h}}_{t+1} \!= \mathbf{h}_{t} +  \Phi(\mathbf{h}_{t})   &  \!\!\!\quad \text{\small{\textbf{Prediction}\!}} \label{eq:prediction}\\
&   \mathbf{h}_{t+1} \!= \tilde{\mathbf{h}}_{t+1}  + \mathbf{K}_t \odot \left( \mathbf{E}(\mathbf{u}_t) - \tilde{\mathbf{h}}_{t+1} \right). & \!\!\! \quad \text{\small{\textbf{Correction}\!}} \label{eq:correction}
\end{empheq}

We again see that the prior predictor $\Phi$ brings no information and that the correction step drives the whole dynamics.

\section{Experiments}

\subsection{Model architectures and training}
\label{app:phydnet-impl}

\paragraph{Model architectures}
We give here the architecture of the encoder and decoder for all datasets. They share common building blocs, composed of convolutions, GroupNorm activation functions \cite{wu2018group} and LeakyRelu non-linearities. For each of the following architectures, we use skip connections from the encoder to the decoder, as classically done, \eg in \cite{denton2017unsupervised}. We define:
\begin{itemize}
    \item conv-block(input, output, stride) = \{Conv2D + GroupNorm + LeakyRelu(0.2)\}
    \item upconv-block(input,output,stride)=\{TransposedConv2D + GroupNorm + LeakyRelu(0.2) \}
    \item upconv(input,output,stride)=TransposedConv2D(input, output, stride)
\end{itemize}{}

\textbf{Moving MNIST:}
\begin{table}[H]
    \centering
\begin{tabular}{c|c}
\toprule
    Encoder & Decoder  \\ \hline
  conv-block(1,8,1)  &   upconv-block(128,64,1) \\ 
   conv-block(8,16,1) & upconv-block(128,32,2) \\ 
   conv-block(16,32,2) & upconv-block(64,32,1) \\ 
  conv-block(32,32,1) & upconv-block(64,16,2) \\ 
   conv-block(32,64,2) & upconv-block(32,8,1) \\ 
   conv-block(64,64,1) & upconv(16,1,1) \\ 
   \bottomrule
\end{tabular}{}
\end{table}{}

\textbf{Traffic:}
\begin{table}[H]
    \centering
\begin{tabular}{c|c}
\toprule
    Encoder & Decoder  \\ \hline
   conv-block(2,32,1)  &  upconv-block(256,64,1)  \\
   conv-block(32.64,2) &  upconv-block(128,32,2)  \\
   conv-block(64,128,1) & upconv(64,2,1)  \\
   \bottomrule
\end{tabular}{}
\end{table}

\textbf{SST:}
\begin{table}[H]
    \centering
\begin{tabular}{c|c}
\toprule
    Encoder & Decoder  \\ \hline
   conv-block(1,32,1)  &  upconv-block(256,64,1)  \\
   conv-block(32.64,2) &  upconv-block(128,32,2)  \\
   conv-block(64,128,1) & upconv(64,1,1)  \\
   \bottomrule
\end{tabular}{}
\end{table}

\break
\textbf{Human 3.6:}
\begin{table}[H]
    \centering
\begin{tabular}{c|c}
\toprule
    Encoder & Decoder  \\ \hline
  conv-block(3,16,1)  &   upconv-block(256,128,1) \\ 
   conv-block(16,32,1) & upconv-block(256,64,2) \\ 
   conv-block(32,64,2) & upconv-block(128,64,1) \\ 
  conv-block(64,64,1) & upconv-block(128,32,2) \\ 
   conv-block(64,128,2) & upconv-block(64,16,1) \\ 
   conv-block(128,128,1) & upconv(32,3,1) \\ 
   \bottomrule
\end{tabular}{}
\end{table}

\begin{figure}[H]
    \centering
    \includegraphics[width=7cm]{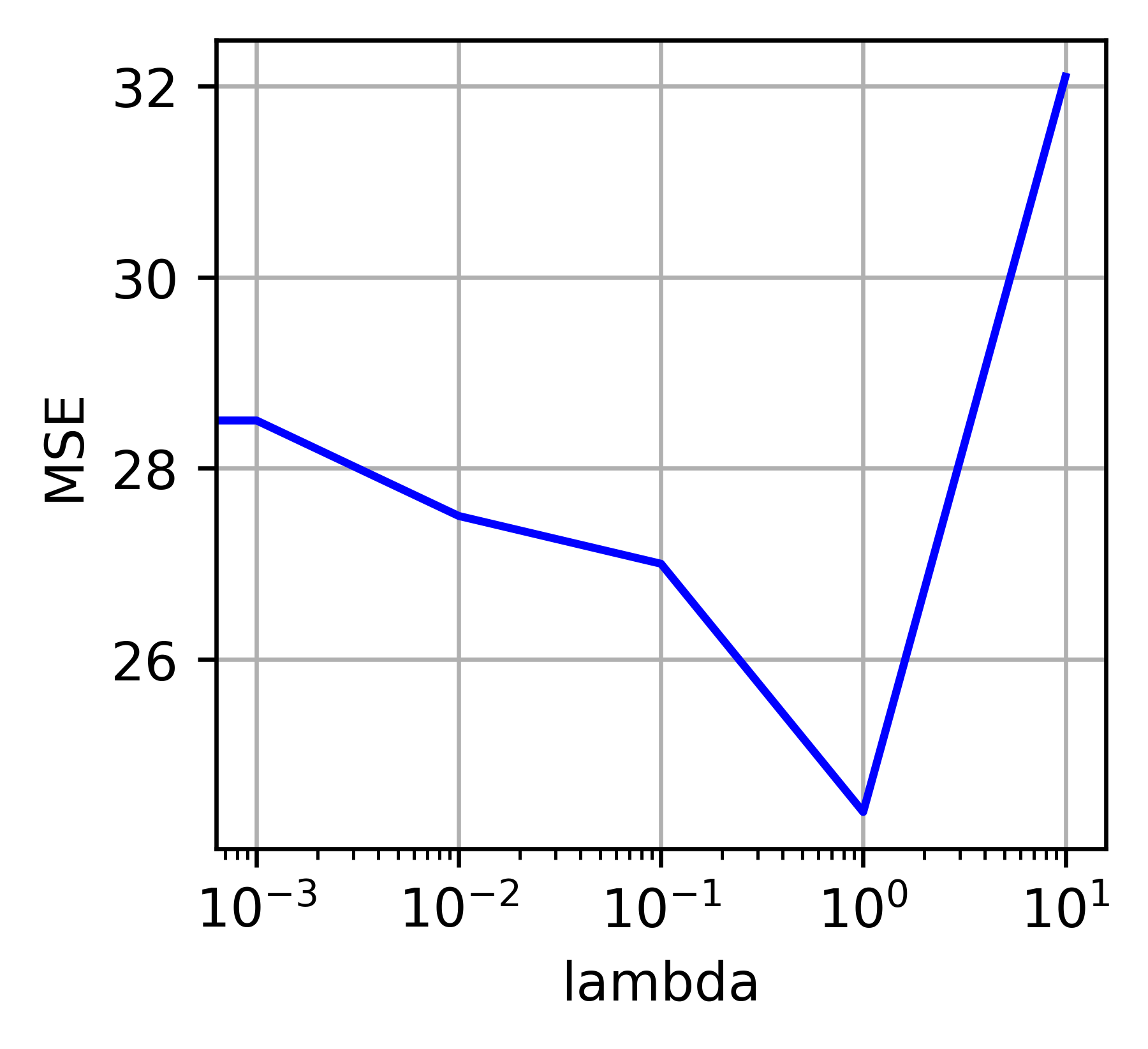}
    \caption{Influence of hyperparameter $\lambda$ when training PhyDNet for Moving MNIST dataset.}
    \label{fig:lambda}
\end{figure}

\paragraph{Influence of $\lambda$} We show in Figure \ref{fig:lambda} the influence of parameter $\lambda$ balancing $\mathcal{L}_{\text{image}}$ and  $\mathcal{L}_{\text{moment}}$ when training PhyDNet for Moving MNIST dataset. When $\lambda$ decreases towards 0, MSE tends towards the unconstrained case at 29. MSE reaches a minimum around $\lambda = 1$. When $\lambda$ further increases, physical regularization is too high and MSE increases above 30. In the paper, we fix $\lambda = 1$ for all datasets.

\subsection{State-of-the art comparison}
\label{app:compa-villegas}

We show here that PhyDNet results are equivalent on Human 3.6 to a recent baseline that explicitly uses additional human pose annotations \cite{villegas2017learning}. In the supplementary of their paper \cite{villegas2017learning}, the authors evaluate their model with Peak Signal over Noise Ratios (PSNR) curves with respect to the forecasting horizon for all deciles of motion in Human 3.6 videos. Regarding prediction horizon up to $H=4$, their method obtains a PSNR always below 21 and around 22 for the $1^{st}$ decile (with the least human motion). In comparison, PhyDNet attains a per-frame MSE of 369, corresponding to a PSNR of 21.2. This shows that PhyDNet performs similarly than \cite{villegas2017learning} for the prediction horizon considered, without requiring additional human pose annotations.

\subsection{Ablation study}

We give in Figure \ref{fig:ablation} additional visualisations completing Figure \ref{fig:ablation}. We qualitatively analyze partial predictions of PhyDNet for the physical branch  $\hat{\mathbf{u}}^{\mathbf{p}}_{t+1} = \mathbf{D}(\mathbf{h}^{\mathbf{p}}_{t+1})$ and residual branch  $\hat{\mathbf{u}}^{\mathbf{r}}_{t+1} = \mathbf{D}(\mathbf{h}^{\mathbf{r}}_{t+1})$. For Moving MNIST (a) and Human 3.6 (d), $\mathbf{h^p}$ captures coarse localisations of objects, while $\mathbf{h^r}$ captures fine-grained details that are not useful for the physical model.  For Traffic BJ, $\mathbf{h^p}$ captures the main patterns of the road network, while $\mathbf{h^r}$ models remaining details. Finally for SST, the visual difference between $\mathbf{h^p}$ and $\mathbf{h^r}$ is slighter, but the cooperation between both branches is crucial, as shown by quantitatives results.

\begin{table*}
    \caption[PhyDNet detailed ablation study.]{A detailed ablation study shows the impact of the physical regularization $\mathcal{L}_{\text{moment}}$ on the performances of PhyCell and PhyDNet for all datasets.}    
    \begin{adjustbox}{max width=\linewidth}
    \begin{tabular}{l|lll|lll|lll|lll}
  \toprule
     \multicolumn{1}{c}{Method} &  \multicolumn{3}{|c|}{\textbf{Moving MNist}} &  \multicolumn{3}{|c|}{\textbf{Traffic BJ}} &  \multicolumn{3}{|c|}{\textbf{Sea Surface Temperature}} &  \multicolumn{3}{|c}{\textbf{Human 3.6}}  \\ 
  \midrule
    ~  & MSE & MAE & SSIM & MSE $\times$ 100 & MAE & SSIM & MSE $\times$ 10 & MAE & SSIM & MSE /10 & MAE /100 & SSIM \\ \midrule
    ConvLSTM  & 103.3  & 182.9  & 0.707  & $48.5^*$ & $17.7^*$  & $0.978^*$ & $45.6^*$ &  $63.1^*$  & $0.949^*$ &  $50.4^*$ & $18.9^*$ &  $0.776^*$  \\ 
    PhyCell  & 50.8 & 129.3  & 0.870  & 48.9  & 17.9 & 0.978  & 38.2  & 60.2 & 0.969 & 42.5  & 18.3 & 0.891  \\ 
    PhyCell  without $\mathcal{L}_{\text{moment}}$  & 43.4  & 112.8  & 0.895  & 43.6  & 16.89  & 0.980 & 35.4 & 56.0 & 0.970 & 39.6  & 17.4 & 0.894  \\ 
    PhyDNet   & \textbf{24.4}  & \textbf{70.3}  & \textbf{0.947}  &  \textbf{41.9} & \textbf{16.2} & \textbf{0.982}  & \textbf{31.9}  & 53.3  & \textbf{0.972}  & 36.9 & 16.2 & 0.901 \\    
    PhyDNet without $\mathcal{L}_{\text{moment}}$ & 29.0  & 81.2 & 0.934  & 43.9  & 16.6  & 0.981  & 32.3 & \textbf{53.1}  & 0.971 & \textbf{36.7}  & \textbf{15.9} &  \textbf{0.904}   \\ 
  \bottomrule
    \end{tabular}
    \end{adjustbox}

        \label{tab:app-ablation}  
\end{table*}

\subsection{Influence of physical regularization}
\label{app:phydnet-influence}

We provide the detailed ablation study for all datasets in Table \ref{tab:app-ablation} that complements Table \ref{tab:ablation}. When we disable $\mathcal{L}_{\text{moment}}$ for training PhyCell, performances improve for all datasets (improvement of 7 MSE points for Moving MNIST, 5 points for Traffic BJ, 3 points for SST and Human 3.6). This again shows that physical constraints alone are too restrictive for learning dynamics in a general  context,  where  other  factors  are  required  for  prediction. When we further include PhyCell in our two-branches disentangling architecture PhyDNet, there is another huge performance gain compared to PhyCell (improvement of 25 MSE points on Moving MNIST, 7 points for Traffic and SST, 5 points for Human 3.6). We also remark that when we disable  $\mathcal{L}_{\text{moment}}$ for training PhyDNet, we get worse performances (drop of 5 MSE points for Moving MNIST and 2 points for Traffic) or equivalent performances (difference below 0.5 MSE point for SST and Human 3.6). This again confirms the relevance of physical constraints.

\subsection{Additional visualisations}
\label{app:phydnet-visu}

\begin{figure*}
    \centering
    \includegraphics[width=17cm]{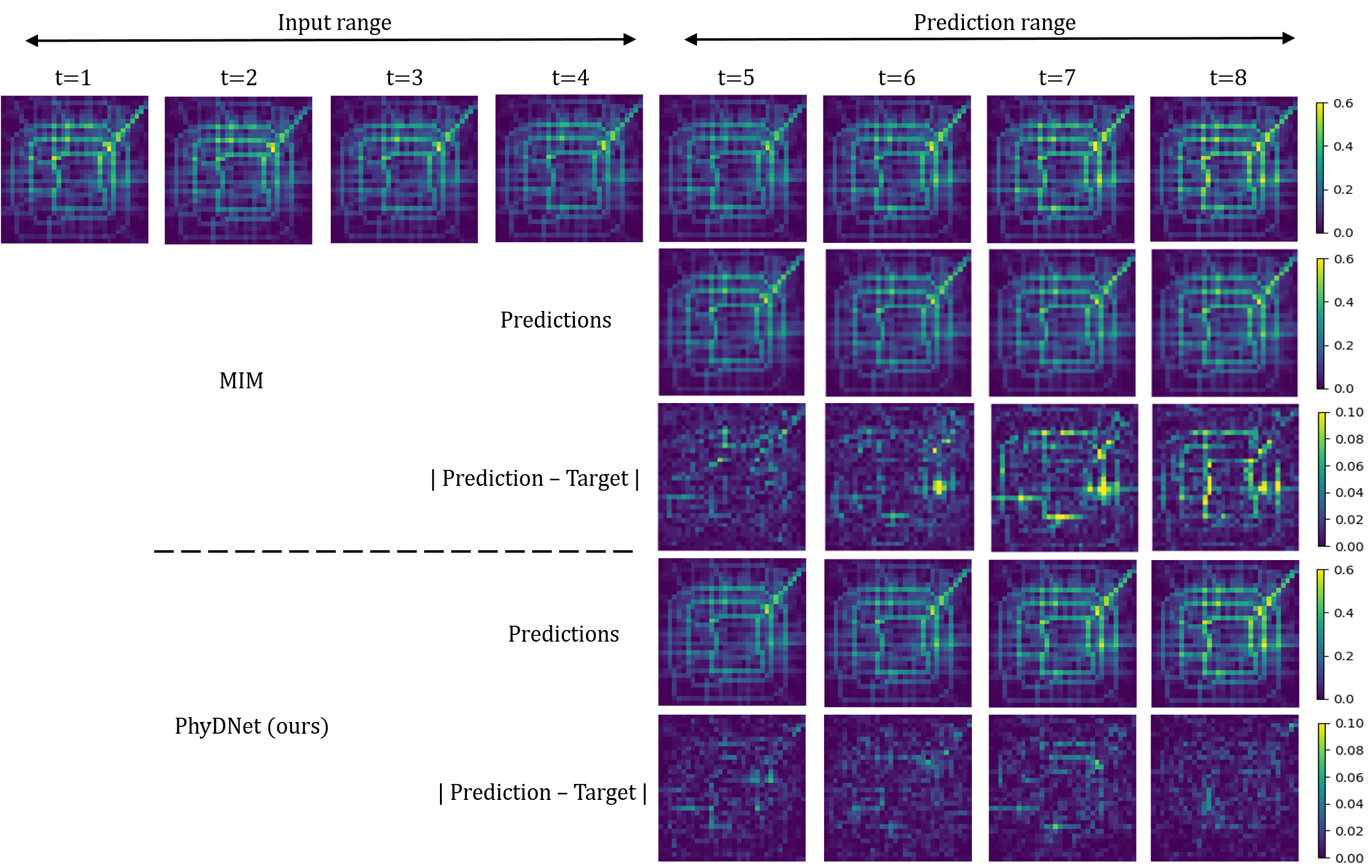}
    \caption[PhyDNet additional qualitative results for Traffic BJ.]{Additional qualitative results for Traffic BJ and comparison to Memory In Memory \cite{wang2019memory}. We see that PhyDNet absolute error are smaller than MIM errors, and independent of the spatial structure of the road network.}
    \label{fig:taxi}
\end{figure*}

We give further qualitative prediction of PhyDNet on Traffic BJ (Figure \ref{fig:taxi}) with a comparison with Memory in Memory \cite{wang2019memory} that is state-of-the-art for this dataset. We see that PhyDNet leads to sharper results and a lower absolute error. Interestingly, PhyDNet absolute errors are approximately spatially independent, whereas MIM errors tend to be higher at a few keys locations of Beijing road network.

\begin{figure*}
    \centering
    \includegraphics[width=17cm]{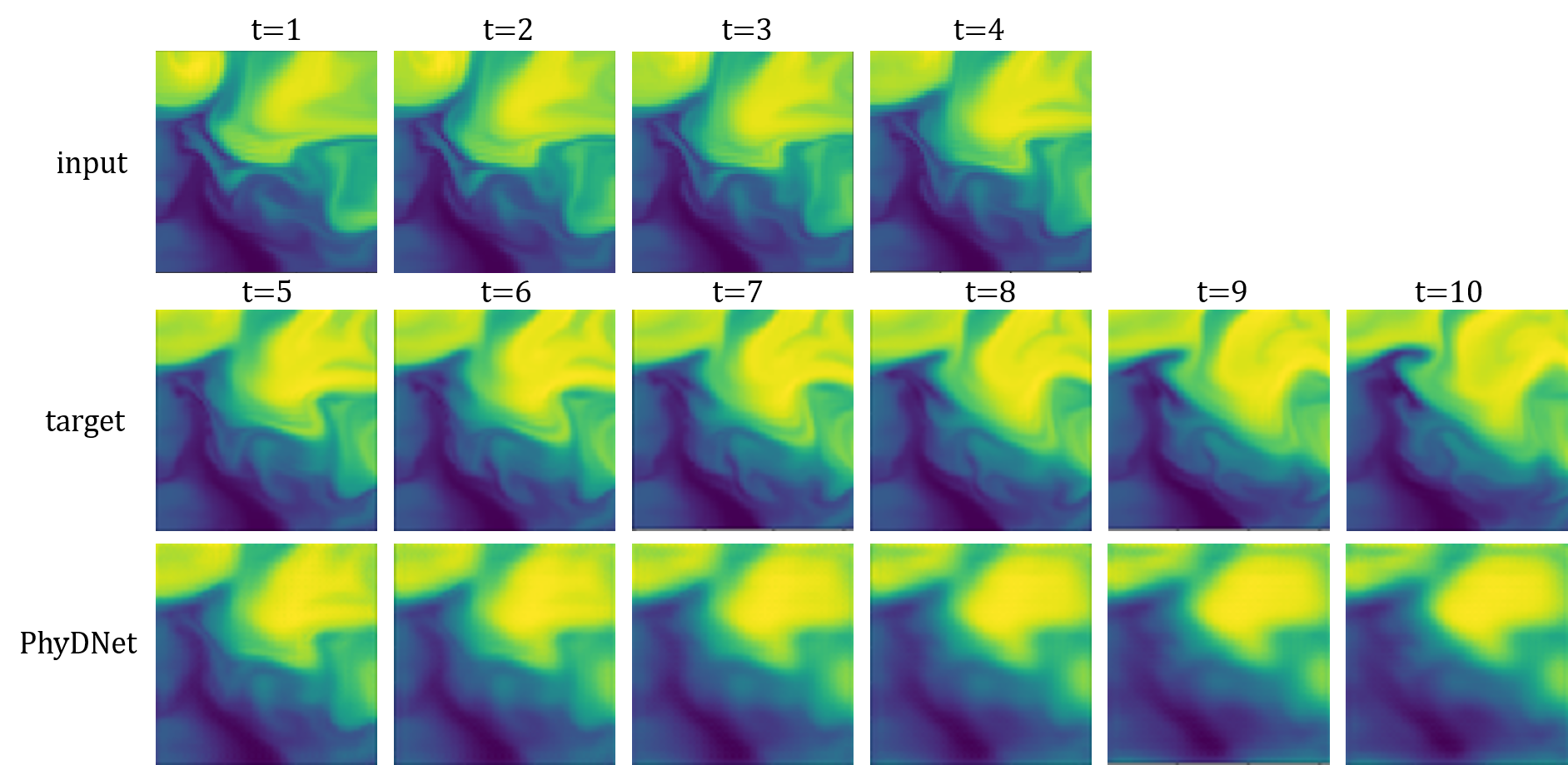}
    \caption{PhyDNet additional qualitative results for Sea Surface Temperature.}
    \label{fig:sst}
\end{figure*}

\begin{figure*}
    \centering
    \includegraphics[width=13cm]{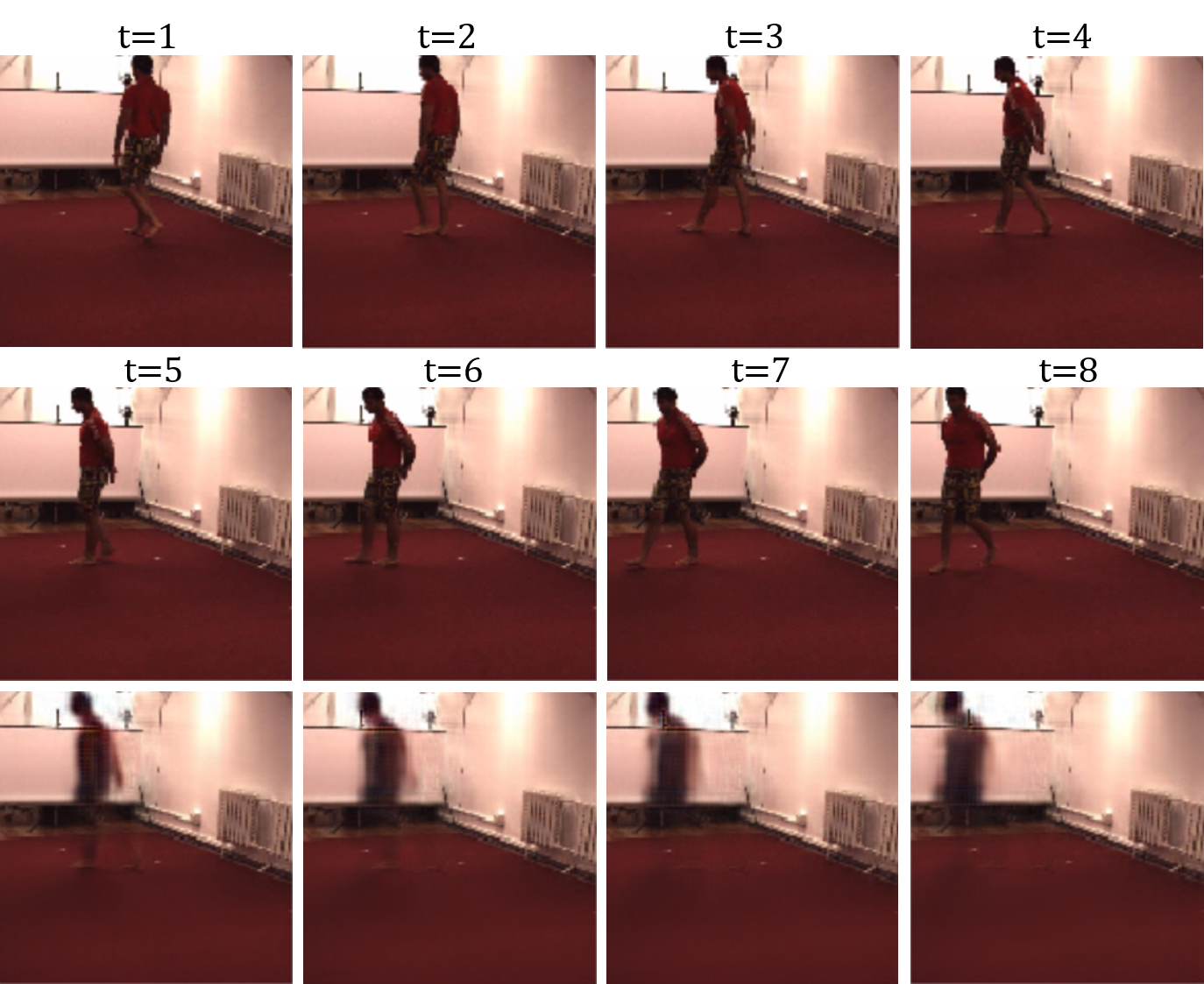}
    \caption{PhyDNet additional qualitative results for Human 3.6.}
    \label{fig:human}
\end{figure*}

We also provide additional prediction visualisations for Sea Surface Temperature (Figure \ref{fig:sst}) and Human 3.6 (Figure \ref{fig:human}) which confirm the good behaviour of PhyDNet. 

\begin{figure*}
    \centering
    \includegraphics[width=17cm]{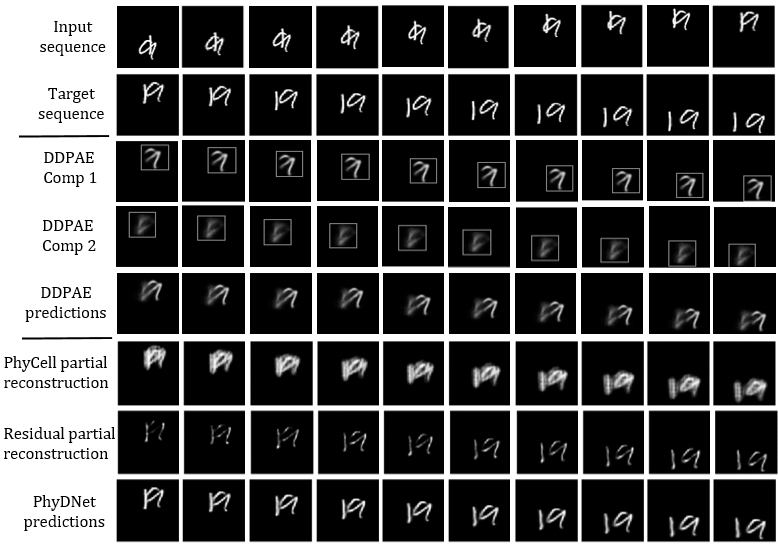}
    \caption{Detailed qualitative comparison to DDPAE \cite{hsieh2018learning} on Moving MNIST dataset.}
    \label{fig:compa-ddpae}
\end{figure*}

We add a detailed qualitative comparison to DDPAE in Figure \ref{fig:compa-ddpae}. DDPAE is a specific disentangling method for Moving MNIST that extracts the positions of the two digits and tracks them with a predictive recurrent neural network. In this example, DDPAE fails to disentangle the two digits (components 1 and 2) in Figure \ref{fig:compa-ddpae} when they overlap in the input sequence, resulting in blurry predictions. In contrast, PhyDNet successfully learns a latent space in which the two digits are disentangled, resulting in far better predictions in terms of sharpness and position of the digits.

{}

{}

{}

{}

\begin{figure*}
    \centering
    \includegraphics[width=\linewidth]{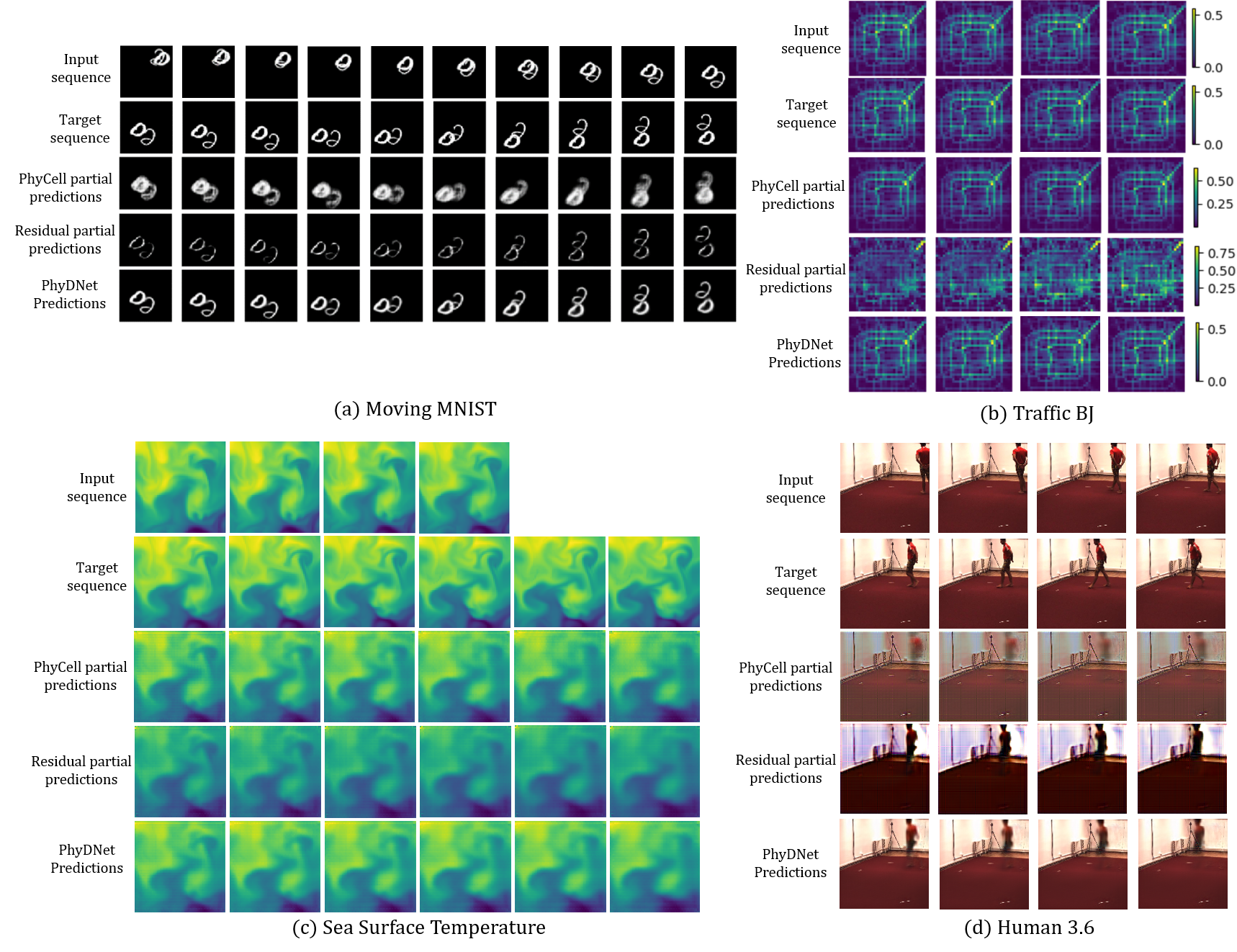}
    \caption{PhyDNet additional ablation visualisations for all datasets.}
    \label{fig:ablation}
\end{figure*}{}

\clearpage{\pagestyle{empty}\cleardoublepage}
\newpage

\mbox{}
\thispagestyle{empty}
\chapter{Appendix for APHYNITY}
\markright{\MakeUppercase{Annexe E}}

\section{\label{app:chebyshev}Reminder on proximinal and Chebyshev sets}

We begin by giving a definition of proximinal and Chebyshev sets, taken from~\cite{chebyshev}:
\begin{definition}
A \textit{proximinal set} of a normed space $(E,\|\cdot\|)$ is a subset $\mathcal{C}\subset E$ such that every $x\in E$ admits at least a nearest point in $\mathcal{C}$.  
\end{definition}

\begin{definition}
A \textit{Chebyshev set} of a normed space $(E,\|\cdot\|)$ is a subset $\mathcal{C}\subset E$ such that every $x\in E$ admits a unique nearest point in $\mathcal{C}$.  
\end{definition}

Proximinality reduces to a compacity condition in finite dimensional spaces. In general, it is a weaker one: boundedly compact sets verify this property for example.

In Euclidean spaces, Chebyshev sets are simply the closed convex subsets. The question of knowing whether it is the case that all Chebyshev sets are closed convex sets in infinite dimensional Hilbert spaces is still an open question. In general, there exists examples of non-convex Chebyshev sets, a famous one being presented in~\cite{chebyshev_counter} for a non-complete inner-product space.

Given the importance of this topic in approximation theory, finding necessary conditions for a set to be Chebyshev and studying the properties of those sets have been the subject of many efforts. Some of those properties are summarized below:
\begin{itemize}
    \item The metric projection on a boundedly compact Chebyshev set is continuous.
    \item If the norm is strict, every closed convex space, in particular any finite dimensional subspace is Chebyshev.
    \item In a Hilbert space, every closed convex set is Chebyshev.
\end{itemize}

\section{\label{app:proof}Proof of Propositions~\ref{prop:exist_unique} and \ref{prop:unique}}

We prove the following result which implies both propositions in the article:
\begin{prop}
The optimization problem:
\begin{equation}
\label{eq:opt_sup}
\underset{F_p\in\F_p, F_a\in\F}{\min} ~~~\left\Vert  F_a  \right\Vert ~~~
\mathrm{subject~to} ~~~~ \forall X\in\D, \forall t, \frac{\diff X_t}{\diff t} =(F_p+F_a)(X_t)
\end{equation}
is equivalent a metric projection onto $\F_p$.

If $\F_p$ is proximinal, Eq \ref{eq:opt_sup} admits a minimizing pair.

If $\F_p$ is Chebyshev, Eq \ref{eq:opt_sup} admits a unique minimizing pair which $F_p$ is the metric projection.
\end{prop}

\begin{proof}
The idea is to reconstruct the full functional from the trajectories of $\D$. By definition, $\A$ is the set of points reached by trajectories in $\D$ so that:
\[
\A = \{x\in\R^d\ |\ \exists X_\cdot\in\D, \exists t,\ X_t = x\}.
\]
Then let us define a function $F^\D$ in the following way: For $a\in \A$, we can find $X_\cdot\in\D$ and $t_0$ such that $X_{t_0} = a$. Differentiating $X$ at $t_0$, which is possible by definition of $\D$, we take:
\[
F^\D(a) = \left.\frac{\diff X_t}{\diff t}\right|_{t=t_0}.
\]

For any $(F_p,F_a)$ satisfying the constraint in Eq \ref{eq:opt_sup}, we then have that $(F_p+F_a)(a) = \nicefrac{\diff X_t}{\diff t}_{|t_0} = F^\D(a)$ for all $a\in\A$. Conversely, any pair such that $(F_p, F_a)\in\F_p\times\F$ and $F_p+F_a = F^\D$, verifies the constraint.

Thus we have the equivalence between Eq \ref{eq:opt_sup} and the metric projection formulated as:
\begin{mini}
    {F_p\in\F_p}{\left\Vert  F^\D - F_p  \right\Vert.} 
    {}{}
\end{mini}

If $\F_p$ is proximinal, the projection problem admits a solution which we denote $F^\star_p$. Taking $F^\star_a = F^\D - F^\star_p$, we have that $F^\star_p+F^\star_a = F^\D$ so that $(F^\star_p, F^\star_a)$ verifies the constraint of Eq \ref{eq:opt_sup}. Moreover, if there is $(F_p,F_a)$ satisfying the constraint of Eq \ref{eq:opt_sup}, we have that $F_p + F_a = F^\D$ by what was shown above and $\|F_a\| = \|F^\D-F_p\|\geq\|F^\D-F^\star_p\|$ by definition of $F^\star_p$. This shows that $(F^\star_p,F^\star_a)$ is minimal.

Moreover, if $\F_p$ is a Chebyshev set, by uniqueness of the projection, if $F_p\not=F^\star_p$ then $\|F_a\|>\|F^\star_a\|$. Thus the minimal pair is unique.
\end{proof}

\section{\label{app:alt_methods}Parameter estimation in incomplete physical models}

Classically, when a set $\F_p\subset\F$ summarizing the most important properties of a system is available, this gives a \textit{simplified model} of the true dynamics and the adopted problem is then to fit the trajectories using this model as well as possible, solving:
\begin{mini}
    {F_p\in\F_p}{\mathbb{E}_{X\sim\D}  L(\widetilde{X}^{X_0},X)} 
    {}{}
    \addConstraint{\forall g\in\I,\ \widetilde{X}_0^g = g\text{ and }\forall t,\ \frac{\diff \widetilde{X}_t^g}{\diff t} = F_p(\widetilde{X}_t^g).}{}
\label{eq:opt_pure_phy}
\end{mini}
where $L$ is a discrepancy measure between trajectories. Recall that $\widetilde{X}^{X_0}$ is the result trajectory of an ODE solver taking $X_0$ as initial condition. In other words, we try to find a function $F_p$ which gives trajectories as close as possible to the ones from the dataset. While estimation of the function becomes easier, there is then a residual part which is left unexplained and this can be a non negligible issue in at least two ways:
\begin{itemize}
    \item When $F\not\in\F_p$, the loss is strictly positive at the minimum. This means that reducing the space of functions $\mathcal{F}_p$ makes us lose in terms of accuracy.\footnote{This is true in theory, although not necessarily in practice when $F$ overfits a small dataset.}
    \item The obtained function $F_p$ might not even be the most meaningful function from $\F_p$ as it would try to capture phenomena which are not explainable with functions in $\F_p$, thus giving the wrong bias to the calculated function. For example, if one is considering a dampened periodic trajectory where only the period can be learned in $\F_p$ but not the dampening, the estimated period will account for the dampening and will thus be biased.
\end{itemize}

This is confirmed in Section \ref{sec:expes}: the incomplete physical models augmented with APHYNITY get different and experimentally better physical identification results than the physical models alone.

Let us compare our approach with this one on the linearized damped pendulum to show how estimates of physical parameters can differ. The equation is the following:
\[
\frac{\diff^2\theta}{\diff t^2} + \omega_0^2\theta + \alpha \frac{\diff \theta}{\diff t} = 0.
\]
We take the same notations as in the article and parametrize the simplified physical models as:
\[
F^{a}_p:X\mapsto (\frac{\diff \theta}{\diff t}, -a\theta),
\]
where $a>0$ corresponds to $\omega_0^2$. The corresponding solution for an initial state $X_0$, which we denote $X^{a}$, can then written explicitly as:
\[
\theta^{a}_t = \theta_0\cos{\sqrt{a} t}.
\]
Let us consider damped pendulum solutions $X$ written as:
\[
\theta_t = \theta_0e^{-t}\cos{t},
\]
which corresponds to:
\[
F : X\mapsto (\frac{\diff \theta}{\diff t}, -2(\theta+\frac{\diff \theta}{\diff t})).
\]
It is then easy to see that the estimate of $a$ with the physical model alone can be obtained by minimizing:
\[
\int_0^T|e^{-t}\cos{t} - \cos{\sqrt{a} t}|^2.
\]
This expression depends on $T$ and thus, depending on the chosen time interval and the way the integral is discretized will almost always give biased estimates. In other words, the estimated value of $a$ will not give us the desired solution $t\mapsto \cos{t}$.

On the other hand, for a given $a$, in the APHYNITY framework, the residual must be equal to:
\[
F^a_r : X\mapsto (0, (a-2)\theta - 2\frac{\diff \theta}{\diff t}).
\]
in order to satisfy the fitting constraint. Here $a$ corresponds to $1+\omega_0^2$ not to $\omega_0^2$ as in the simplified case. Minimizing its norm, we obtain $a=2$ which gives us the desired solution:
\[
\theta_t = \theta_0e^{-t}\cos{t},
\]
with the right period.

\section{\label{app:der_superv}Discussion on supervision over derivatives}

In order to find the appropriate decomposition $(F_p,F_a)$, we use a trajectory-based error by solving:
\begin{mini}
    {F_p\in\F_p, F_a\in\F}{\left\Vert  F_a  \right\Vert} 
    {}{}
    \addConstraint{\forall g\in\I,\ \widetilde{X}_0^g = g\text{ and }\forall t,\ \frac{\diff \widetilde{X}_t^g}{\diff t} = (F_p+F_a)(\widetilde{X}_t^g)}{}
    \addConstraint{\forall X\in\D,\ L(X,\widetilde{X}^{X_0}) = 0.}{}
\label{notre_pbm_int}
\end{mini}

In the continuous setting where the data is available at all times $t$, this problem is in fact equivalent to the following one:
\begin{mini}
    {F_p\in\F_p}{\mathbb{E}_{X\sim\D}  \int \left\Vert \frac{\diff X_t}{\diff t} - F_p(X_t)  \right\Vert.}
    {}{}
\label{der_pbm}
\end{mini}
where the supervision is done directly over derivatives, obtained through finite-difference schemes. This echoes the proof in Section~\ref{app:proof} of the Appendix where $F$ can be reconstructed from the continuous data.

However, in practice, data is only available at discrete times with a certain time resolution. While Eq \ref{der_pbm} is indeed equivalent to Eq \ref{notre_pbm_int} in the continuous setting, in the practical discrete one, the way error propagates is not anymore: For Eq \ref{notre_pbm_int} it is controlled over integrated trajectories while for Eq  \ref{der_pbm} the supervision is over the approximate derivatives of the trajectories from the dataset. We argue that the trajectory-based approach is more flexible and more robust for the following reasons:
\begin{itemize}
    \item In Eq \ref{notre_pbm_int}, if $F_a$ is appropriately parameterized, it is possible to perfectly fit the data trajectories at the sampled points.
    \item The use of finite differences schemes to estimate $F$ as is done in Eq \ref{der_pbm} necessarily induces a non-zero discretization error.
    \item This discretization error is explosive in terms of divergence from the true trajectories.
\end{itemize}

This last point is quite important, especially when time sampling is sparse~(even though we do observe this adverse effect empirically in our experiments with relatively finely time-sampled trajectories). The following gives a heuristical reasoning as to why this is the case. Let $\widetilde{F} = F + \epsilon$ be the function estimated from the sampled points with an error $\epsilon$ such that $\|\epsilon\|_\infty\leq\alpha$. Denoting $\widetilde{X}$ the corresponding trajectory generated by $\widetilde{F}$, we then have, for all $X\in\D$:
\[
\forall t,\ \frac{\diff (X-\widetilde{X})_t}{\diff t} = F(X_t) - F(\widetilde{X}_t) - \epsilon(\widetilde{X}_t).
\]
Integrating over $[0,T]$ and using the triangular inequality as well as the mean value inequality, supposing that $F$ has uniformly bounded spatial derivatives:
\[
\forall t\in[0,T],\ \|(X-\widetilde{X})_t\| \leq \|\nabla F\|_\infty\int_0^t \|X_s-\widetilde{X}_s\| + \alpha t,
\]
which, using a variant of the Grönwall lemma, gives us the inequality:
\[
\forall t\in[0,T],\ \|X_t-\widetilde{X}_t\| \leq  \frac{\alpha}{\|\nabla F\|_\infty}(\exp(\|\nabla F\|_\infty t) -1).
\]

When $\alpha$ tends to $0$, we recover the true trajectories $X$. However, as $\alpha$ is bounded away from $0$ by the available temporal resolution, this inequality gives a rough estimate of the way $\widetilde{X}$ diverges from them, and it can be an equality in many cases. This exponential behaviour explains our choice of a trajectory-based optimization.

\section{Implementation details\label{app:implementation}}

We describe here the three use cases studied in the paper for validating APHYNITY. All experiments are implemented with PyTorch  and the differentiable ODE solvers with the adjoint method implemented in \texttt{torchdiffeq}.\footnote{\url{https://github.com/rtqichen/torchdiffeq}}

\subsection{Damped pendulum} 

We consider the non-linear damped pendulum problem, governed by the ODE \[\frac{\diff ^2 \theta}{\diff t^2} + \omega_0^2 \sin \theta + \alpha \frac{\diff \theta}{\diff t} = 0, \] where $\theta(t)$ is the angle, $\omega_0 = \frac{2 \pi}{T_0}$ is the proper pulsation~($T_0$ being the period) and $\alpha$ is the damping coefficient. With the state $X = (\theta, \frac{\diff\theta}{\diff t})$, the ODE can be written as $\frac{\diff X_t}{\diff t} = F(X_t)$ with
$ F : X \mapsto ( \frac{\diff\theta}{\diff t} ,  - \omega_0^2 \sin \theta  - \alpha \frac{\diff\theta}{\diff t})$.

\paragraph{Dataset} For each train / validation / test split, we simulate a dataset with 25 trajectories of 40 timesteps (time interval $[0,20]$, timestep $\delta t=0.5$) with fixed ODE coefficients $(T_0 = 12, \alpha=0.2)$ and varying initial conditions. The simulation integrator is Dormand-Prince Runge-Kutta method of order (4)5 (DOPRI5, \cite{dormand1980family}). We also add a small amount of white gaussian noise ($\sigma=0.01$) to the state. Note that our pendulum dataset is much more challenging than the ideal frictionless pendulum considered in \cite{greydanus2019hamiltonian}. 

\paragraph{Neural network architectures} We detail in Table \ref{tab:pendulum-nn-archis} the neural architectures used for the damped pendulum experiments. All data-driven augmentations for approximating the mapping $X_t \mapsto F(X_t)$ are implemented by multi-layer perceptrons (MLP) with 3 layers of 200 neurons and ReLU activation functions (except at the last layer: linear activation). The Hamiltonian \cite{greydanus2019hamiltonian,toth2019hamiltonian} is implemented by a MLP that takes the state $X_t$ and outputs a scalar estimation of the Hamiltonian $\mathcal{H}$ of the system: the derivative is then computed by an in-graph gradient of $\mathcal{H}$ with respect to the input: $F(X_t) = \left( \frac{\partial \mathcal{H}}{\partial (\diff\theta/\diff t)}, - \frac{\partial \mathcal{H}}{\diff  \theta} \right)$. 

\begin{table}[H]
    \caption[Neural network architectures for the damped pendulum.]{Neural network architectures for the damped pendulum experiments. n/a corresponds to non-applicable cases.}
    \centering
\begin{adjustbox}{max width=\linewidth}
   \begin{tabular}{lcc}
   \toprule
    Method     & Physical model & Data-driven model  \\
    \midrule
    Neural ODE    &  n/a & MLP(in=2, units=200, layers=3, out=2)  \\
    \midrule
    Hamiltonian  &  MLP(in=2, units=200, layers=3, out=1) & n/a \\
    APHYNITY Hamiltonian & MLP(in=2, units=200, layers=3, out=1)  & MLP(in=2, units=200, layers=3, out=2)  \\
    \midrule
     Param ODE ($\omega_0$) & 1 trainable parameter $\omega_0$ &  n/a \\   
    APHYNITY Param ODE ($\omega_0$) & 1 trainable parameter $\omega_0$ &   MLP(in=2, units=200, layers=3, out=2) \\
    \midrule
     Param ODE ($\omega_0,\alpha$) & 2 trainable parameters $\omega_0, \lambda$ &  n/a \\   
    APHYNITY Param ODE ($\omega_0,\alpha$) & 2 trainable parameters $\omega_0, \lambda$ &   MLP(in=2, units=200, layers=3, out=2) \\    
    \bottomrule
    \end{tabular}
    \end{adjustbox}
   \label{tab:pendulum-nn-archis}
\end{table}

\paragraph{Optimization hyperparameters} The hyperparameters of the APHYNITY optimization algorithm ($Niter,\lambda_0,\tau_1,\tau_2$) were cross-validated on the validation set and are shown in Table \ref{tab:pendulum-hyperparameters}. All models were trained with a maximum number of 5000 steps with early stopping.

\begin{table}[H]
    \caption{Hyperparameters of the damped pendulum experiments.}
    \centering
\begin{adjustbox}{max width=\linewidth}
   \begin{tabular}{ccccc}
   \toprule
    Method     & Niter & $\lambda_0$ & $\tau_1$ & $\tau_2$   \\
    \midrule
    APHYNITY Hamiltonian &  5 & 1 & 1 & 0.1  \\
    APHYNITY ParamODE ($\omega_0$) &  5 & 1 & 1 & 10  \\
    APHYNITY ParamODE ($\omega_0,\lambda$) & 5 & 1000 & 1 & 100  \\    
    \bottomrule
    \end{tabular}
    \end{adjustbox}
   \label{tab:pendulum-hyperparameters}
\end{table}

\subsection{Reaction-diffusion equations}

The system is driven by a FitzHugh-Nagumo type PDE~\cite{klaasen1984fitzhugh}
\begin{align*}
     \frac{\partial u}{\partial t} &= a\Delta u + R_u(u,v; k) \\
      \frac{\partial v}{\partial t} &= b\Delta v + R_v(u,v),
\end{align*}

where $a$ and $b$ are respectively the diffusion coefficients of $u$ and $v$, $\Delta$ is the Laplace operator. The local reaction terms are $R_u(u,v; k) = u - u^3 - k - v, R_v(u,v) = u - v$. 

The state $X=(u,v)$ is defined over a compact rectangular domain $\Omega = [-1,1]^2$ with periodic boundary conditions. $\Omega$ is spatially discretized with a $32\times 32$ 2D uniform square mesh grid. The periodic boundary condition is implemented with circular padding around the borders. $\Delta$ is systematically estimated with a $3\times 3$ discrete Laplace operator. 

\paragraph{Dataset} Starting from a randomly sampled initial state $X_\text{init} \in [0,1]^{2\times 32\times 32}$, we generate states by integrating the true PDE with fixed $a$, $b$, and $k$ in a dataset ($a=1 \times 10^{-3}, b=5 \times 10^{-3},  k=5 \times 10^{-3}$). We firstly simulate high time-resolution ($\delta t_\text{sim} = 0.001$) sequences with explicit finite difference method. We then extract states every $\delta t_\text{data} = 0.1$ to construct our low time-resolution datasets.

We set the time of random initial state to $t=-0.5$ and the time horizon to $t=2.5$. 1920 sequences are generated, with 1600 for training/validation and 320 for test. We take the state at $t=0$ as $X_0$ and predict the sequence until the horizon (equivalent to 25 time steps) in all reaction-diffusion experiments. Note that the sub-sequence with $t<0$ are reserved for the extensive experiments in Appendix~\ref{app:additional_reac_diff}.

\paragraph{Neural network architectures}

Our $F_a$ here is a 3-layer convolution network (ConvNet). The two input channels are $(u, v)$ and two output ones are $(\frac{\partial u}{\partial t}, \frac{\partial v}{\partial t})$. The purely data-driven Neural ODE uses such ConvNet as its $F$. The detailed architecture is provided in Table~\ref{tab:reaction-diffusion-arch}. The estimated physical parameters $\theta_p$ in $F_p$ are simply a trainable vector $(a, b) \in \mathbb{R}_+^2$ or $(a, b, k) \in \mathbb{R}_+^3$.

\begin{table}[H]
    \caption[Model architecture for the reaction-diffusion and wave equations.]{ConvNet architecture in reaction-diffusion and wave equation experiments, used as data-driven derivative operator in APHYNITY and Neural ODE \cite{chen2018neural}.}
    \label{tab:reaction-diffusion-arch}
    \centering
    \begin{tabular}{ll}
    \toprule
        Module & Specification \\
    \midrule
        2D Conv. & $3\times 3$ kernel, 2 input channels, 16 output channels, 1 pixel zero padding \\
        2D Batch Norm. & No average tracking \\
        ReLU activation & --- \\
        2D Conv. & $3\times 3$ kernel, 16 input channels, 16 output channels, 1 pixel zero padding \\
        2D Batch Norm. & No average tracking \\
        ReLU activation & --- \\
        2D Conv. & $3\times 3$ kernel, 16 input channels, 2 output channels, 1 pixel zero padding \\
    \bottomrule
    \end{tabular}
\end{table}

\paragraph{Optimization hyperparameters}
We choose to apply the same hyperparameters for all the reaction-diffusion experiments: $Niter = 1, \lambda_0 = 1, \tau_1 = 1\times 10^{-3}, \tau_2 = 1\times 10^{3}$. 

\subsection{Wave equations}  \label{sup:wave-details}
The damped wave equation is defined by
\[
    \frac{\partial^2 w}{\partial t^2} - c^2 \Delta w + k \frac{\partial w}{\partial t} = 0,
\]
where $c$ is the wave speed and $k$ is the damping coefficient. The state is $X=(w, \frac{\partial w}{\partial t})$. 

We consider a compact spatial domain $\Omega$ represented as a $64\times64$ grid and discretize the Laplacian operator similarly. $\Delta$ is implemented using a $5\times 5$ discrete Laplace operator in simulation whereas in the experiment is a $3\times 3$ Laplace operator. Null Neumann boundary condition are imposed for generation.

\paragraph{Dataset} $\delta t$ was set to $0.001$ to respect Courant number and provide stable integration. The simulation was integrated using a 4th order finite difference Runge-Kutta scheme for 300 steps from an initial Gaussian state, i.e for all sequence at $t=0$, we have:
\begin{equation}
    w(x,y, t=0) = C \times \exp^{\frac{(x-x_0)^2 + (y-y_0)^2}{\sigma^2}}.
\end{equation}
The amplitude $C$ is fixed to $1$, and $(x_0, y_0)=(32,32)$ to make the Gaussian curve centered for all sequences. However, $\sigma$ is different for each sequence and uniformly sampled in $[10, 100]$.
The same $\delta t$ was used for train and test. All initial conditions are Gaussian with varying amplitudes. 250 sequences are generated, 200 are used for training while 50 are reserved as a test set.
In the main paper setting, $c=330$ and $k=50$.
As with the reaction diffusion case, the algorithm takes as input a state $X_{t_0}=(w, \frac{\diff w}{\diff t})(t_0)$ and predicts all states from $t_0+\delta t$ up to $t_0 + 25 \delta t$.

\paragraph{Neural network architectures}
The neural network for $F_a$ is a 3-layer convolution neural network with the same architecture as in Table~\ref{tab:reaction-diffusion-arch}. For $F_p$, the parameter(s) to be estimated is either a scalar $c\in\mathbb{R}_+$ or a vector $(c,k) \in \mathbb{R}_+^2$. Similarly, Neural ODE networks are build as presented in Table~\ref{tab:reaction-diffusion-arch}.


\paragraph{Optimization hyperparameters}
We use the same hyperparameters for the experiments:\\ ${Niter = 3, \lambda_0 = 1, \tau_1 = 1\times 10^{-4}, \tau_2 = 1\times 10^{2}}$.

\section{Ablation study\label{app:ablation}}

We conduct ablation studies to show the effectiveness of APHYNITY's adaptive optimization and trajectory-based learning scheme. 

\subsection{Ablation to vanilla ML/MB cooperation}

In Table~\ref{tab:ablation-nds}, we consider the ablation case with the vanilla augmentation scheme found in \cite{leguen20phydnet,wang2019integrating,neural20}, which does not present any proper decomposition guarantee. We observe that the APHYNITY cooperation scheme outperforms this vanilla scheme in all case, both in terms of forecasting performances (\eg log MSE= -0.35 vs. -3.97 for the Hamiltonian in the pendulum case) and parameter identification (\eg Err Param=8.4\% vs. 2.3 for Param PDE ($a,b$ for reaction-diffusion). It confirms the crucial benefits of APHYNITY's principled decomposition scheme.

\subsection{Detailed ablation study}

We conduct also two other ablations in Table \ref{tab:ablation-others}: 
\begin{itemize}
    \item \textit{derivative supervision}: in which $F_p+F_a$ is trained with supervision over approximated derivatives on ground truth trajectory, as performed in \cite{greydanus2019hamiltonian,cranmer2020lagrangian}. More precisely, APHYNITY's $\mathcal{L}_\text{traj}$ is here replaced with $\mathcal{L}_\text{deriv} = \|\frac{\diff X_t}{\diff t} - F(X_t)\|$ as in \eqref{der_pbm}, where $\frac{\diff X_t}{\diff t}$ is approximated by finite differences on $X_t$.
    \item \textit{non-adaptive optim.}: in which we train APHYNITY by minimizing $\|F_a\|$ without the adaptive optimization of $\lambda$ shown in Algorithm~\ref{alg:optim}. This case is equivalent to $\lambda = 1, \tau_2=0$.
\end{itemize} 

\begin{table}[t]
    \centering
    \caption[Ablation study comparing APHYNITY to the vanilla ML/MB augmentation scheme.]{Ablation study comparing APHYNITY to the vanilla augmentation scheme \cite{wang2019integrating,neural20} for the reaction-diffusion equation, wave equation and damped pendulum.
    \label{tab:ablation-nds}}
\begin{adjustbox}{max width=\linewidth}
    \begin{tabular}{clccc}
    \toprule
    Dataset & Method & $\log$ MSE &
    \%Err Param. & 
    $\|F_a\|^2$ 
    \\ \midrule
    
\multirowcell{8}{\parbox{1cm}{\centering\tiny Damped pendulum}} 
& Hamiltonian with vanilla aug. & -0.35$\pm$0.1  & n/a  & 837$\pm$117 \\
&  APHYNITY Hamiltonian   & \textbf{-3.97$\pm$1.2}  & n/a &  623$\pm$68 \\ \cmidrule{2-5}
& Param ODE ($\omega_0$) with vanilla aug. & -7.02$\pm$1.7 & 4.5   &  148$\pm$49 \\
&  APHYNITY Param ODE ($\omega_0$)  &  \textbf{-7.86$\pm$0.6}  & \textbf{4.0}  &   132$\pm$11  \\ \cmidrule{2-5}
& Param ODE ($\omega_0, \alpha$) with vanilla aug.& -7.60$\pm$0.6  &    4.65     & 35.5$\pm$6.2 \\
& APHYNITY Param ODE ($\omega_0, \alpha$)  & \textbf{-8.31$\pm$0.3}  &  \textbf{0.39}      & 8.5$\pm$2.0   \\
 \cmidrule{2-5}
& Augmented True ODE with vanilla aug. & \textbf{-8.40$\pm$0.2}  & n/a  & 3.4$\pm$0.8  \\
& APHYNITY True ODE   &    \textbf{-8.44$\pm$0.2}  & n/a   & 2.3$\pm$0.4  \\     
\midrule
    
\multirowcell{6}{\parbox{0.7cm}{\centering\tiny Reaction-diffusion}} 
& Param. PDE ($a,b$) with vanilla aug. & -4.56$\pm$0.52 
  & 8.4 &   (7.5$\pm$1.4)e1\\ 
&  APHYNITY Param. PDE ($a,b$) & \textbf{-5.10$\pm$0.21} & \textbf{2.3}  &   (6.7$\pm$0.4)e1 \\
  \cmidrule{2-5}
& Param. PDE ($a,b,k$) with vanilla aug. & -8.04$\pm$0.03 
  & 25.4  & (1.5$\pm$0.2)e-2\\
& APHYNITY Param. PDE ($a,b,k$) & \textbf{-9.35$\pm$0.02} 
  & \textbf{0.096}  & (1.5$\pm$0.4)e-6 \\ 
  \cmidrule{2-5}
& True PDE with vanilla aug. & -8.12$\pm$0.05 
  & n/a  & (6.1$\pm$2.3)e-4\\
& APHYNITY True PDE & \textbf{-9.17$\pm$0.02} 
 & n/a &  (1.4$\pm$0.8)e-7\\
 
  \midrule
\multirowcell{4}{\parbox{1cm}{\centering\tiny Wave equation}} 
& Param PDE ($c$) with vanilla aug. & -3.90 $\pm$  0.27 & 0.51 &   88.66 \\
&  APHYNITY Param PDE ($c$) & \textbf{-4.64$\pm$0.25}& \textbf{0.31} & 71.0 \\
  \cmidrule{2-5}
& Param PDE ($c, k$) with vanilla aug. & -5.96 $\pm$ 0.10 & 0.71  & 25.1 \\
&  APHYNITY Param PDE ($c, k$) & \textbf{-6.09$\pm$0.28} & \textbf{0.70}  & 4.54 \\

 \bottomrule
    \end{tabular}
\end{adjustbox}
\end{table}

\begin{table}[h!]
    \centering
    \caption[Detailed ablation study for APHYNITY.]{Detailed ablation study on supervision and optimization for the reaction-diffusion equation, wave equation and damped pendulum.
    \label{tab:ablation-others}}
\begin{adjustbox}{max width=\linewidth}
    \begin{tabular}{clccc}
    \toprule
    Dataset & Method & $\log$ MSE &
    \%Err Param. & 
    $\|F_a\|^2$ 
    \\ \midrule

    \multirowcell{12}{\parbox{1cm}{\centering\tiny Damped pendulum}} &   Augmented Hamiltonian derivative supervision & -0.83$\pm$0.3   & n/a  & 642$\pm$121 \\
&    Augmented Hamiltonian non-adaptive optim. & -0.49$\pm$0.58  & n/a &  165$\pm$30 \\    
&    APHYNITY Hamiltonian   & \textbf{-3.97$\pm$1.2}  & n/a &  623$\pm$68 \\ \cmidrule{2-5}
&  Augmented Param ODE ($\omega_0$) derivative supervision &  -1.02$\pm$0.04 & 5.8  &  136$\pm$13 \\
&  Augmented Param ODE ($\omega_0$) non-adaptive optim. & -4.30$\pm$1.3  &  4.4   &  90.4$\pm$27 \\  
&  APHYNITY Param ODE ($\omega_0$)  &  \textbf{-7.86$\pm$0.6}  & \textbf{4.0}  &   132$\pm$11  \\ \cmidrule{2-5}
& Augmented Param ODE ($\omega_0, \alpha$) derivative supervision  & -2.61$\pm$0.2  &  5.0  & 3.2$\pm$1.7 \\
&  Augmented Param ODE ($\omega_0, \alpha$) non-adaptive optim. & -7.69$\pm$1.3  &   1.65  & 4.8$\pm$7.7 \\
& APHYNITY Param ODE ($\omega_0, \alpha$)  & \textbf{-8.31$\pm$0.3}  &  \textbf{0.39}      & 8.5$\pm$2.0   \\
 \cmidrule{2-5}
& Augmented True ODE derivative supervision  & -2.14$\pm$0.3 & n/a  & 4.1$\pm$0.6 \\
& Augmented True ODE non-adaptive optim. & \textbf{-8.34$\pm$0.4}  & n/a  &  1.4$\pm$0.3  \\
& APHYNITY True ODE   &    \textbf{-8.44$\pm$0.2}  & n/a   & 2.3$\pm$0.4  \\   
\midrule
    
\multirowcell{9}{\parbox{0.7cm}{\centering\tiny Reaction-diffusion}} & Augmented Param. PDE ($a,b$) derivative supervision  & -4.42$\pm$0.25 

  & 12.6 & (6.8$\pm$0.6)e1\\ 
& Augmented Param. PDE ($a,b$) non-adaptive optim. & -4.55$\pm$0.11 

  & 7.5 &  (7.6$\pm$1.0)e1\\ 
&  APHYNITY Param. PDE ($a,b$) & \textbf{-5.10$\pm$0.21} & \textbf{2.3}  &   (6.7$\pm$0.4)e1 \\
  \cmidrule{2-5}
& Augmented Param. PDE ($a,b,k$) derivative supervision & -4.90$\pm$0.06 

  & 11.7 & (1.9$\pm$0.3)e-1\\
& Augmented Param. PDE ($a,b,k$) non-adaptive optim. & -9.10$\pm$0.02 

  & 0.21  & (5.5$\pm$2.9)e-7\\
&  APHYNITY Param. PDE ($a,b,k$) & \textbf{-9.35$\pm$0.02} 

  & \textbf{0.096}  & (1.5$\pm$0.4)e-6 \\ 
  \cmidrule{2-5}
& Augmented True PDE derivative supervision & -6.03$\pm$0.01 

  & n/a & (3.1$\pm$0.8)e-3\\
& Augmented True PDE non-adaptive optim. & -9.01$\pm$0.01 

  & n/a &  (1.5$\pm$0.8)e-6\\
& APHYNITY True PDE & \textbf{-9.17$\pm$0.02} 

 & n/a &  (1.4$\pm$0.8)e-7\\
  \midrule
\multirowcell{8}{\parbox{1cm}{\centering\tiny Wave equation}} & Augmented Param PDE ($c$) derivative supervision & -1.16$\pm$0.48 & 12.1 &  0.00024 \\
& Augmented Param PDE ($c$) non-adaptive optim. &-2.57$\pm$0.21 & 3.1 &  43.6 \\ 
&  APHYNITY Param PDE ($c$) & \textbf{-4.64$\pm$0.25}& \textbf{0.31} & 71.0 \\
  \cmidrule{2-5}
& Augmented Param PDE ($c, k$) derivative supervision & -4.19$\pm$0.36 &  7.2   & 0.00012 \\
& Augmented  Param PDE ($c, k$) non-adaptive optim. & -4.93$\pm$0.51 & 1.32 & 0.054 \\
&  APHYNITY Param PDE ($c, k$) & \textbf{-6.09$\pm$0.28} & \textbf{0.70}  & 4.54 \\ 
  \cmidrule{2-5}
& Augmented True PDE derivative supervision & -4.42 $\pm$ 0.33 & n/a & 6.02e-5 \\
& Augmented True PDE non-adaptive optim. & -4.97$\pm$0.49 & n/a & 0.23 \\
& APHYNITY True PDE & \textbf{-5.24$\pm$0.45} & n/a  & 0.14 \\

 \bottomrule
    \end{tabular}
\end{adjustbox}
\end{table}


We highlight the importance to use a principled adaptive optimization algorithm (APHYNITY algorithm described in paper) compared to a non-adpative optimization: for example in the reaction-diffusion case, log MSE= -4.55 vs. -5.10 for Param PDE $(a,b)$. Finally, when the supervision occurs on the derivative, both forecasting and parameter identification results are systematically lower than with APHYNITY's trajectory based approach: for example, log MSE=-1.16 vs. -4.64 for Param PDE $(c)$ in the wave equation. It confirms the good properties of the APHYNITY training scheme.

\pagebreak
\section{Additional experiments\label{app:additional}}

\subsection{Reaction-diffusion systems with varying diffusion parameters \label{app:additional_reac_diff}}

We conduct an extensive evaluation on a setting with varying diffusion parameters for reaction-diffusion equations. 
The only varying parameters are diffusion coefficients, \ie individual $a$ and $b$ for each sequence. We randomly sample $a\in [1\times 10^{-3},2\times 10^{-3}]$ and $b \in [3\times 10^{-3},7\times 10^{-3}]$. $k$ is still fixed to $5\times 10^{-3}$ across the dataset.

In order to estimate $a$ and $b$ for each sequence, we use here a ConvNet encoder $E$ to estimate parameters from 5 reserved frames ($t<0$). The architecture of the encoder $E$ is similar to the one in Table~\ref{tab:reaction-diffusion-arch} except that $E$ takes 5 frames (10 channels) as input and $E$ outputs a vector of estimated $(\tilde a,\tilde b)$ after applying a sigmoid activation scaled by $1\times 10^{-2}$ (to avoid possible divergence). For the baseline Neural ODE, we concatenate $a$ and $b$ to each sequence as two channels.

In Table~\ref{tab:reaction-diffusion-supplement}, we observe that combining data-driven and physical components outperforms the pure data-driven one. When applying APHYNITY to Param PDE ($a,b$), the prediction precision is significantly improved ($\log$ MSE: -1.32 vs. -4.32) with $a$ and $b$ respectively reduced from 55.6\% and 54.1\% to 11.8\% and 18.7\%. For complete physics cases, the parameter estimations are also improved for Param PDE ($a,b,k$) by reducing over 60\% of the error of $b$ (3.10 vs. 1.23) and 10\% to 20\% of the errors of $a$ and $k$ (resp. 1.55/0.59 vs. 1.29/0.39). 

The extensive results reflect the same conclusion as shown in the main article: APHYNITY improves the prediction precision and parameter estimation. The same decreasing tendency of $\|F_a\|$ is also confirmed.

\begin{table}[h]
\setlength{\tabcolsep}{4pt}
    \centering
    \caption[APHYNITY results on the reaction-diffusion equations with varying parameters.]{Results of the dataset of reaction-diffusion with varying $(a,b)$. $k=5\times 10^{-3}$ is shared across the dataset. \label{tab:reaction-diffusion-supplement}}
    \begin{tabular}{clccccc}
    \toprule
    & Method & $\log$ MSE 
    & \%Err $a$ & \%Err $b$ & \%Err $k$  & $\|F_a\|^2$ \\
\midrule
\parbox{0.9cm}{\centering\tiny Data-driven}  & Neural ODE \cite{chen2018neural} & -3.61$\pm$0.07 
  & n/a & n/a & n/a & n/a\\
    \midrule
\multirowcell{2}{\parbox{0.9cm}{\centering\tiny Incomplete physics}}   & Param PDE ($a,b$) & -1.32$\pm$0.02 
  & 55.6 & 54.1 &  n/a & n/a\\
  & APHYNITY Param PDE ($a,b$) & \textbf{-4.32$\pm$0.32} 
  & \textbf{11.8} & \textbf{18.7} & n/a &  (4.3$\pm$0.6)e1\\
  \midrule
\multirowcell{4}{\parbox{0.9cm}{\centering\tiny Complete physics}} & Param PDE ($a,b,k$) & \textbf{-5.54$\pm$0.38} 
  &  1.55 & 3.10 & 0.59 & n/a\\
  & APHYNITY Param PDE ($a,b,k$) & \textbf{-5.72$\pm$0.25} 
  & \textbf{1.29} & \textbf{1.23} & \textbf{0.39} & (5.9$\pm$4.3)e-1 \\ 
  \cdashline{2-7}\noalign{\vskip 0.2ex}
  & True PDE &  \textbf{-8.86$\pm$0.02} 
  & n/a & n/a & n/a & n/a\\ 
  & APHYNITY True PDE & \textbf{-8.82$\pm$0.15} 
  & n/a & n/a & n/a & (1.8$\pm$0.6)e-5\\
  \bottomrule
    \end{tabular}
\end{table}

\subsection{Additional results for the wave equation\label{app:additional_wave}}
We conduct an experiment where each sequence is generated with a different wave celerity. This dataset is challenging because both $c$ and the initial conditions vary across the sequences. For each simulated sequence, an initial condition is sampled as described previously, along with a wave celerity $c$ also sampled uniformly in $[300, 400]$. Finally our initial state is integrated with the same Runge-Kutta scheme. $200$ of such sequences are generated for training while $50$ are kept for testing.
 
For this experiment, we also use a ConvNet encoder to estimate the wave speed $c$ from 5 consecutive reserved states $(w, \frac{\partial w}{\partial t})$. The architecture of the encoder $E$ is the same as in Table~\ref{tab:reaction-diffusion-arch} but with 10 input channels.
Here also, $k$ is fixed for all sequences and $k=50$. The hyper-parameters used in these experiments are the same than described in the Section~\ref{sup:wave-details}.

The results when multiple wave speeds $c$ are in the dataset are consistent with the one present when only one is considered. Indeed, while prediction performances are slightly hindered, the parameter estimation remains consistent for both $c$ and $k$. This extension provides elements attesting for the robustness and adaptability of our method to more complex settings. Finally the purely data-driven Neural-ODE fails to cope with the increasing difficulty.

\begin{table}[H]
    \centering
    \setlength{\tabcolsep}{8pt}
    \caption[APHYNITY results on the wave equations with varying parameters.]{Results for the damped wave equation when considering multiple $c$ sampled uniformly in $[300, 400]$ in the dataset, $k$ is shared across all sequences and $k=50$.}
    \begin{tabular}{clcccc}
    \toprule
    &Method & $\log$ MSE &
    \%Error $c$ & \%Error $k$ & $\|F_a\|^2$ 
    \\ \midrule
 \parbox{0.9cm}{\centering\tiny Data-driven} & Neural ODE \cite{chen2018neural}  & 0.056$\pm$0.34& n/a & n/a & n/a \\
  \midrule
\multirowcell{2}{\parbox{0.9cm}{\centering\tiny Incomplete physics}}  &Param PDE ($c$)  &-1.32$\pm$0.27 & 23.9 & n/a & n/a \\ 
  &APHYNITY Param PDE ($c$)  &\textbf{-4.51$\pm$0.38}& 3.2 & n/a & 171 \\ 
  \midrule
\multirowcell{4}{\parbox{0.9cm}{\centering\tiny Complete physics}}  & Param PDE ($c, k$) & -4.25$\pm$0.28 & 3.54 &  1.43& n/a\\
  & APHYNITY Param PDE ($c, k$) &\textbf{-4.84$\pm$0.57} & 2.41 & 0.064 & 3.64 \\ 
  \cdashline{2-6}\noalign{\vskip 0.2ex}
  & True PDE ($c, k$) &\textbf{-4.51$\pm$0.29} & n/a & n/a & n/a \\
  & APHYNITY True PDE ($c, k$) & \textbf{-4.49$\pm$0.22} & n/a &n/a&  0.0005 \\
  \bottomrule
    \end{tabular}
    \label{tab:additional wave}
\end{table}

\clearpage{\pagestyle{empty}\cleardoublepage}
\newpage


\cleardoublepage\phantomsection
\thispagestyle{empty}
\newpage $\ $
\newpage
\thispagestyle{empty}
\setlength{\baselineskip}{11pt}
\setpapersize{A4}
\setmarginsrb{0mm}{0mm}{15mm}{0mm}{0mm}{0mm}{0mm}{0mm}
\begin{center}

\fbox{
\begin{tabular}{p{3cm} |p{10.2cm}|p{3cm}}
		\begin{minipage}{3cm}
			\includegraphics[width=1\linewidth]{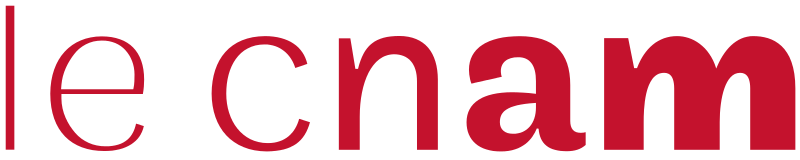}  
		\end{minipage}
	&
		\begin{minipage}{10.2cm}
			\begin{center}
				\textbf{\Large{Vincent LE GUEN}}\\
				\vspace{0.2cm}
				\textbf{\Large{Deep learning for spatio-temporal forecasting - application to solar energy.}}\\
				\vspace{0.2cm}
			\end{center}
		\end{minipage}
		&
		\begin{minipage}{3cm}
			\includegraphics[width=1\linewidth]{images/logo-edf.jpg} 
		\end{minipage}
\end{tabular}
}
\end{center}
\vspace{0.3cm}
\begin{center}

\fcolorbox{redHESAM}{white}{
\begin{minipage}{17.5cm}
\vspace{0.2cm}
\textbf{\large{Résumé :} Cette thèse aborde le problème de la prédiction spatio-temporelle par apprentissage profond, motivée par la prévision à court-terme de la production photovoltaïque à Electricité de France (EDF).  Nous explorons dans cette thèse deux principales directions de recherche. La première concerne le choix de la fonction de perte pour entraîner les modèles: nous proposons d’utiliser des critères de forme et de décalage temporel sur les trajectoires prédites. Nous introduisons la fonction de perte DILATE pour la prévision déterministe et le modèle STRIPE pour la prévision probabiliste. Notre seconde direction de recherche est d’augmenter des modèles physiques incomplets avec des réseaux de neurones profonds. Pour la prédiction de vidéo, nous introduisons le modèle PhyDNet qui sépare une partie de dynamique physique, d’une partie résiduelle qui capture l’information complémentaire, comme la texture et les détails, nécessaire à la bonne prédiction. Nous proposons aussi un schéma d’apprentissage, appelé APHYNITY, qui assure une décomposition bien posée et unique entre des modèles physiques incomplets et des réseaux de neurones profonds, sous de faibles hypothèses.}
\\
\\
\textbf{\large{Mots clés : apprentissage profond, apprentissage statistique, prévision spatio-temporelle, prévision photovoltaïque.}}\\

\vspace{0.2cm}
\end{minipage}
}
\end{center}
\vspace{0.2cm}
\begin{center}
\fcolorbox{redHESAM}{white}{
\begin{minipage}{17.5cm}
\vspace{0.2cm}
\textbf{\large{Abstract :}  This thesis tackles the subject of spatio-temporal forecasting with deep learning. The motivating application at Electricity de France (EDF) is short-term solar energy forecasting with fisheye images. We explore two main research directions for improving deep forecasting methods by injecting external physical knowledge. The first direction concerns the role of the training loss function. We show that differentiable shape and temporal criteria can be leveraged to improve the performances of existing models. We address both the deterministic context with the proposed DILATE loss function and the probabilistic context with the STRIPE model. Our second direction is to augment incomplete physical models with deep data-driven networks for accurate forecasting. For video prediction, we introduce the PhyDNet model that disentangles physical dynamics from residual information necessary for prediction, such as texture or details. We further propose a learning framework (APHYNITY) that ensures a principled and unique linear decomposition between physical and data-driven components under mild assumptions, leading to better forecasting performances and parameter identification.
}
\\
\\
\textbf{\large{Keywords : deep learning, machine learning, spatio-temporal forecasting, solar energy forecasting.}}\\

\end{minipage}
}
\end{center}
\setlength{\voffset}{0pt}

\end{document}